\documentclass[10pt]{article} % For LaTeX2e
\usepackage[accepted]{stylefiles/tmlr}
\usepackage{hyperref}
\usepackage{url}

\usepackage[utf8]{inputenc} % allow utf-8 input
\usepackage[T1]{fontenc}    % use 8-bit T1 fontss
\usepackage{booktabs}       % professional-quality tables
\usepackage{amsfonts}       % blackboard math symbols
\usepackage{nicefrac}       % compact symbols for 1/2, etc.
\usepackage{microtype}      % microtypography
\usepackage{xcolor}         % colors
\usepackage{algorithm}
\usepackage{amsmath}        % for math environments
\usepackage{amssymb}        % for math symbols
\usepackage{graphicx}       % for including graphics
\usepackage{bm}             % proper bold notation in math mode
\usepackage[symbol]{footmisc}
\usepackage{float}
\usepackage{subfigure}
\usepackage{wrapfig}
\usepackage{array}          % to make \newcolumntype work
\usepackage{enumitem}       % can remove spaces from itemize
\usepackage[normalem]{ulem} % for strikeout text
\usepackage{placeins}       % for the /FloatBarrier command that blocks previous floats floating beyond that point

\let\authorAND\AND
\let\AND\relax
\usepackage{algorithmic}

\newcommand{\bc}{\bm{c}}
\newcommand{\bx}{\bm{x}}
\newcommand{\bz}{\bm{z}}
\newcommand{\bu}{\bm{u}}
\newcommand{\bv}{\bm{v}}

\newcommand{\cD}{\ensuremath{\mathcal{D}}}

\newcommand{\NN}[1]{\phi_\theta(#1)}
\newcommand{\mR}{\mathbb R}

\newcommand{\algcom}[1]{\textcolor{gray}{\# {#1}}}

\newcolumntype{x}[1]{>{\centering\arraybackslash\hspace{0pt}}p{#1}}

\title{Exposing Outlier Exposure: What Can Be Learned From Few, One, and Zero Outlier Images}

\author{\name Philipp Liznerski${}^*$ \email liznerski@cs.uni-kl.de \\
      \addr TU Kaiserslautern
      \authorAND
      \name Lukas Ruff${}^*$ \email lukas.ruff@aignostics.com \\
      \addr Aignostics, Berlin
      \authorAND
      \name Robert A. Vandermeulen${}^*$ \email vandermeulen@tu-berlin.de \\
      \addr ML Group TU Berlin, Berlin Institute for the Foundations of Learning and Data
      \authorAND
      \name Billy J. Franks \email franks@cs.uni-kl.de \\
      \addr TU Kaiserslautern
      \authorAND
      \name Klaus-Robert M\"uller \email klaus-robert.mueller@tu-berlin.de  \\
      \addr Google Research, Brain Team, ML Group TU Berlin, MPII, and Korea University
      \authorAND
      \name Marius Kloft \email kloft@cs.uni-kl.de\\
      \addr TU Kaiserslautern}

\def\month{11}  % Insert correct month for camera-ready version
\def\year{2022} % Insert correct year for camera-ready version
\def\openreview{\url{https://openreview.net/forum?id=3v78awEzyB}} % Insert correct link to OpenReview for camera-ready version

\begin{document}

\maketitle

\begin{abstract} 
Due to the intractability of characterizing \emph{everything} that looks unlike the normal data, anomaly detection (AD) is traditionally treated as an unsupervised problem utilizing only normal samples.
However, it has recently been found that unsupervised image AD can be drastically improved through the utilization of huge corpora of random images to represent anomalousness; a technique which is known as \emph{Outlier Exposure}.
In this paper we show that specialized AD learning methods seem unnecessary for state-of-the-art performance, and furthermore one can achieve strong performance with just a small collection of Outlier Exposure data, contradicting common assumptions in the field of AD.
We find that standard classifiers and semi-supervised one-class methods trained to discern between normal samples and relatively few random natural images are able to outperform the current state of the art on an established AD benchmark with ImageNet.
Further experiments reveal that even \emph{one} well-chosen outlier sample is sufficient to achieve decent performance on this benchmark (79.3\% AUC).
We investigate this phenomenon and find that one-class methods are more robust to the choice of training outliers, indicating that there are scenarios where these are still more useful than standard classifiers.
Additionally, we include experiments that delineate the scenarios where our results hold.
Lastly, no training samples are necessary when one uses the representations learned by CLIP, a recent foundation model, which achieves state-of-the-art AD results on CIFAR-10 and ImageNet in a zero-shot setting.
\end{abstract}

\section{Introduction}
\label{sec:intro}
\footnotetext{${}^*$ equal contribution.}
\footnotetext{Our code is available at: \url{https://github.com/liznerski/eoe}.}
\footnotetext{Part of this work has been presented in the ICML 2021 UDL Workshop \citep{ruff2020rethinking}.}

Anomaly detection (AD) \citep{chandola2009anomaly} is the task of determining whether a sample is anomalous compared to a corpus of data. 
Recently there has been a great interest in developing novel deep methods for AD \citep{ruff2021,pang2021}.
Most prior work performs AD in an \emph{unsupervised} way utilizing only an unlabeled corpus of mostly normal data \citep{golan2018deep, hendrycks2019using, bergman2020deep,tack2020}. 
While AD can be interpreted as a classification problem of ``normal vs.~anomalous,'' it is classically treated as an unsupervised problem due to the rather tricky issue of finding or constructing a dataset that captures \emph{everything different} from the normal dataset.

One often has, in addition to normal data, access to some data which is known to be anomalous. 
\citet{hendrycks2019deep} noted that, for an image AD problem, one has access to a virtually limitless amount of random natural images from the internet that are presumably not normal. 
They term the utilization of such auxiliary data \emph{Outlier Exposure}\;(OE).
Many top-performing AD methods on standard image AD benchmarks utilize tens of thousands of OE samples combined with self-supervised learning \citep{hendrycks2019using} or transfer learning \citep{reiss2021panda, deecke2021transfer} to achieve state-of-the-art detection performance.

\begin{figure}[ht] \label{fig:toy_example}
  \begin{center}
    \includegraphics[width=0.92\textwidth]{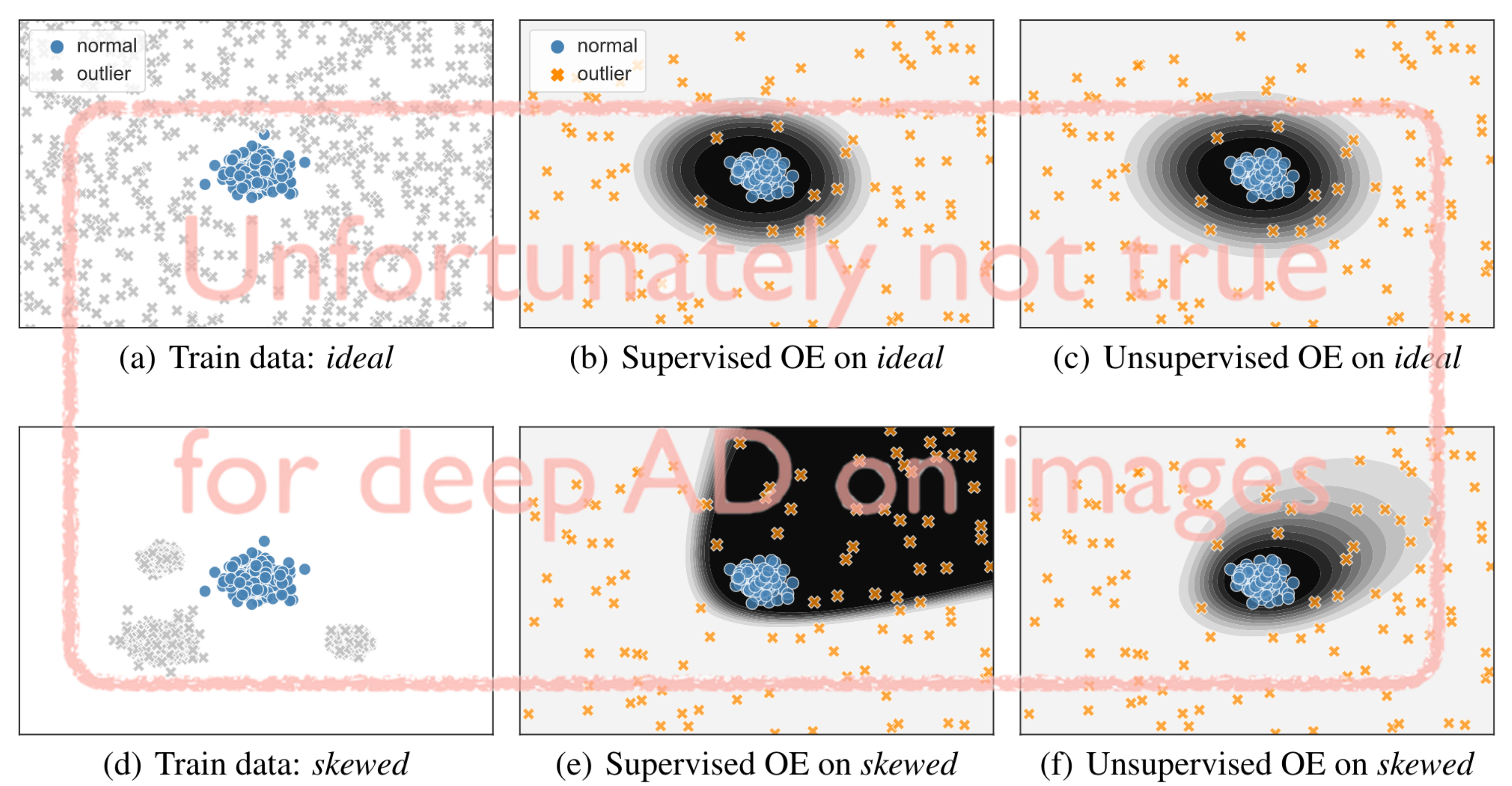}
  \end{center}
  \caption{
  This figure visualizes a deceptively reasonable intuition that one might have for deep AD. 
  It shows the decision boundaries of a supervised OE method and an unsupervised OE method on two toy datasets: \emph{ideal} (a--c) and \emph{skewed} (d--f). 
  The skewed scenario occurs naturally when there are not enough OE samples to cover the ambient space, the data dimensionality is high, or the OE data is clustered. 
  Unsupervised OE (c + f) learns a compact decision region for the normal class that generalizes well in both scenarios. 
  A supervised OE approach (b + e), on the other hand, learns a decision region that generalizes well in the \emph{ideal} case where the outlier training data is fully representative, but not in the \emph{skewed} case. 
  While this intuition is true for shallow AD settings, like in this 2D toy example (a--f), our results suggest that a deep approach does not follow these phenomena for which supervised OE performs remarkably well. }
\end{figure}

For clarity, we here delineate three basic approaches to anomaly detection:
\begin{itemize}[noitemsep,topsep=-8pt,leftmargin=*]
    \item \emph{Unsupervised}: These are methods trained on unlabeled data that is assumed to be mostly normal. This is the classic and most common approach to AD.
    \item \emph{Unsupervised OE}: These are adaptations of unsupervised methods that incorporate auxiliary OE data that is not normal. Elsewhere this is also called ``semi-supervised'' AD \citep{gornitz2013toward,ruff2020}.
    \item \emph{Supervised OE}: This indicates standard classification methods trained to discern between normal data and an auxiliary OE dataset that is not normal.
\end{itemize}\vspace{0.2em}
Using unsupervised OE rather than supervised OE to discern between the normal data and OE samples seems intuitive since the presented anomalies likely do not completely characterize ``anomalousness.'' 
Figure \ref{fig:toy_example} illustrates this classic intuition and highlights the differences between these approaches on a 2D toy dataset.
The benefits of unsupervised OE when incorporating (a few specific) known anomalies has also been observed in previous works \citep{tax2001,gornitz2013toward,ruff2020}.

%%%%%%%%%%%%%%%%%%%%%%%%%%%%%%%%%%%%%%%%%%%%%%%%%%%%%%%%%%%%%%%%%%%%%%%%%%%%%%%%
%%%%%%%%%%%%%%%%%%%%%%%%%%%%%%%%%%%%%%%%%%%%%%%%%%%%%%%%%%%%%%%%%%%%%%%%%%%%%%%%

In this paper we present surprising experimental results that challenge the assumption that deep AD on images needs an unsupervised approach (with or without OE). 
Using the same OE setup as \citet{hendrycks2019using}, which is common in the literature, we find that a standard classifier outperforms current state-of-the-art AD methods on the one vs.~rest AD benchmark with CIFAR-10 and ImageNet. 
The one vs.~rest benchmark has been recommended as a standard evaluation protocol to validate AD methods \citep{emmott13} and is used as a litmus test in virtually all deep AD papers published at top-tier venues; see e.g.~\citep{ruff2018deep,deecke2018image,golan2018deep,samet19,hendrycks2019using,abati19,perera19,wang19,ruff2020,bergman20,kim20,liznerski2021, deecke2021transfer}. 
Further challenging common assumptions, we find that OE does \emph{not} seem to require huge amounts of data to represent ``anomalousness.'' 
A classifier requires only 256 random OE samples to outperform the state of the art on ImageNet and only \emph{one} well-chosen OE sample to score reasonably compared to unsupervised methods and classical AD approaches. 

OE, however, does not solve all types of AD problems, in particular when the normal dataset is highly diverse or when anomalies are very subtle.
For instance, we demonstrate that the methods need more OE samples to achieve top performance on the less studied, more challenging leave-one-class-out AD benchmark \cite{bergman20, ahmed2020, deecke2021transfer} where many classes are combined to form a multimodal normal class.
On the DTD (describable textures) one vs.~rest benchmark and on MVTec-AD, a recent manufacturing dataset, we show that random natural images are not very informative as OE.

We also investigate transfer learning approaches to AD that have recently improved AD on images \citep{reiss2021panda, deecke2021transfer}. 
Using CLIP \citep{radford2021learning}, a recent foundation model, we find that it is possible to set a state of the art on CIFAR-10 and ImageNet \emph{without any additional training data}. 
While transfer learning and standard classification work well, we still find advantages of unsupervised OE over supervised OE.
When there are very few ($<32$) OE samples or the OE samples are not very informative, unsupervised OE approaches outperform classifiers, indicating a certain robustness with respect to the training outliers.
On low-dimensional datasets like Fashion-MNIST and MNIST, where shallow approaches without OE perform well, the intuition described in Figure \ref{fig:toy_example} is still valid: 
Standard classifiers require a lot of OE data to perform well, whereas one-class methods perform competitively even with just one OE sample.

In conclusion, the primary message of this paper is neither that we propose yet another state-of-the-art method nor that one of our investigated methodologies is of general superiority, but that there is a surprisingly strong performance of off-the-shelf classifiers, transfer learning, and just a few OE samples---contradicting widespread common assumptions for deep AD on well-established AD benchmarks. 
Through this work, we want to encourage rethinking how previous AD results extend to deep learning.

\section{Related work}
\label{sec:methods_background}
We here briefly review recent developments in deep AD, including self-supervision, transfer learning, and outlier exposure. We further clarify the differences between out-of-distribution detection and AD and discuss non-natural image AD benchmarks.

\paragraph{Deep anomaly detection}
While there exist many shallow methods for AD, it has been observed that these methods perform poorly on high-dimensional data \citep{huang2006large,kriegel2008angle,erfani2015r1svm,erfani2016high}. 
To address this, deep approaches to AD that scale well with higher dimensions have been proposed \citep{ruff2021,pang2021}.
The most common approaches to deep AD employ autoencoders trained on normal data, where samples not reconstructed well at test time are deemed anomalous \citep{hawkins2002outlier,sakurada2014anomaly,chen2017outlier,zhou2017,nguyen2019,kim20}. 
Deep generative models detect anomalies via a variety of methods \citep{schlegl2017unsupervised,deecke2018image,zenati2018efficient,schlegl2019}, yet their effectiveness has been called into question \citep{nalisnick2019}.

A recent avenue of research uses self-supervision for deep AD on images \citep{gidaris2018unsupervised,golan2018deep,mariusnips19,hendrycks2019using,tack2020,sohn2021}. 
One of the best-performing AD methods is the self-supervised approach from \citet{tack2020}, which combines \citet{hendrycks2019using}'s AD method with contrastive representation learning \citep{chen2020simple}.
\citet{tack2020} train their network on transformed normal data so that it maps similar transformations of a sample close together, while sufficiently distorted transformations and other samples are mapped away.
The network then has to classify each sample's type of transformation as in \citep{hendrycks2019using}. 
For a test sample, both the trained network's certainty on predicting correct transformations and the similarity of the sample with its nearest neighbor in feature space determine its anomaly score:
the larger the certainty and similarity, the smaller the anomaly score.

More recently, transfer learning-based approaches to AD \citep{bergman2020deep, reiss2021panda, deecke2021transfer} that fine-tune supervised classifiers trained on ImageNet have shown to outperform self-supervised methods such as \citet{tack2020} on common benchmarks. 
To the best of our knowledge, \citet{reiss2021panda} is the best performing AD method on CIFAR-10 that does not use OE \citep{hendrycks2019deep}.
Their method fine-tunes a ResNet pre-trained on ImageNet on a deep one-class loss \citep{ruff2018deep} and applies continual learning to avoid a feature collapse. 
Since they use ImageNet pre-trained models, they do not validate their method on the ImageNet one vs.~rest benchmark.

The state of the art on the CIFAR-10 benchmark, however, takes advantage of OE \citep{hendrycks2019deep}, which follows the idea of using a large unstructured corpus of images as ``auxiliary anomalies'' during training.
For example, \citet{hendrycks2019using} use OE to improve their self-supervised method by training the network to predict the uniform distribution for all transforms on OE samples, while leaving training on normal samples unchanged. 
\citet{reiss2021panda} and \citet{deecke2021transfer} combined transfer learning and OE, which yields the currently best performing methods.

\paragraph{Out-of-distribution detection and anomaly detection}
A field of research related to AD is out-of-distribution (OOD) detection, where the aim is to detect anomalous samples that do not belong to any of the given classes of a multi-class classification task \citep{lee2018training}.
One can always apply an AD method to OOD by using it separately from the classifier, treating all training samples of all given classes as normal, ignoring the available class labels. 
However, AD methods typically (and expectedly) perform worse than specialized OOD methods that take advantage of the in-distribution labels and confidence scores of a trained classifier \citep{liang2018enhancing, tack2020, hendrycks2022scaling}.
Such methods define the anomaly score to be large when the maximum of the softmax outputs \citep{liang2018enhancing} or logits \citep{hendrycks2022scaling} is small, i.e.~when the classifier is uncertain about the classification of a sample.
Conversely to AD methods, which are applicable to OOD problems, OOD methods cannot be applied to AD setups due to the absence of in-distribution class labels.
Note that the type of auxiliary supervision via OE we utilize in this paper hence differs from the kind of supervision applied in OOD, as we do not discriminate between different classes of normality but only between normal samples and auxiliary outliers. 

\paragraph{Non-natural image AD}
Recently there has been increasing attention on image AD on ``non-natural'' images (e.g.\ medical images or technical images from manufacturing), where anomalies tend to be more subtle.
For example, the MVTec-AD dataset consists of photos from manufacturing with, for instance, screws being normal and defective screws being anomalous \citep{bergmann2019mvtec}.
In this paper we instead focus on the common and well-established one vs.~rest benchmarks with natural images, aiming to detect images of natural classes that are semantically different from the normal class.
For other types of image data, random natural images from the web are likely not informative as OE.
We show this in Section \ref{sec:exp_var_oe_size} and Appendix \ref{appx:full_results} for the example of MVTec-AD and, less prominently, for DTD.
However, one can see that both transfer learning and OE \emph{can} work well in other settings as many state-of-the-art methods on MVTec-AD rely on one of these. 
For instance, \cite{liznerski2021, schluter2021self, li2021cutpaste} employ OE in the form of synthetically generated anomalies and \cite{defard2021padim,gudovskiy2022cflow,roth2022towards} use transfer learning-based methods.

\section{Methods}
In this section, we introduce the methods that we will use for our experimental evaluation. 
We first motivate why AD typically follows an unsupervised approach and is not viewed as a binary classification problem. Afterwards, we introduce deep one-class classification as well as CLIP \citep{radford2021learning} for zero-shot AD.

\subsection{AD as a classification problem}
Traditionally AD is understood as the problem of estimating the support (or level sets of the support) of the normal data-generating distribution. 
This is known as \emph{density level set estimation} \citep{polonik1995measuring,tsybakov1997}. 
This follows the assumption that normal data is concentrated whereas anomalies are not concentrated \citep{scholkopf2002}. 
\citet{steinwart2005classification} remark that density level set estimation can also be interpreted as a binary classification problem between the normal and an anomalous distribution. 
Many classic AD methods (e.g., kernel density estimation or one-class SVMs) implicitly assume the anomalies to follow a uniform distribution, i.e.~they make an uninformative prior assumption on the anomalous distribution \citep{steinwart2005classification}. 
These methods, as well as a binary classifier trained to discriminate between normal samples and uniform noise, are in fact asymptotically consistent density level set estimators \citep{steinwart2005classification,vert2006}. 
Practically, however, it is preferable to estimate the level set directly rather than classifying against uniform noise. 
Such a classification approach is particularly ineffective and inefficient in high dimensions since it would require massive amounts of noise samples to properly fill the sample space. 
As we show in our experiments, however, we find that this intuition does not seem to extend to deep anomaly detection on images.

\subsection{Deep one-class classification}
\label{sec:hypsphcla}
Deep one-class classification \citep{ruff2018deep} was introduced as a deep learning extension of the one-class classification approach to anomaly detection \citep{scholkopf2001,tax2001}.
Deep SVDD \citep{ruff2018deep} is trained to map normal samples close to a center $\bc$ in feature space, thereby following the concentration assumption \citep{scholkopf2002} mentioned above.
For a neural network $\phi_\theta$ with parameters $\theta$, the Deep SVDD objective is given by
$
  \min_{\theta} \; \frac{1}{n} \sum_{i=1}^n \left\lVert\NN{\bx_i}-\bc\right\rVert^2.
$
\citet{ruff2020} proposed an extension of Deep SVDD that incorporates known anomalies, called \emph{Deep Semi-supervised Anomaly Detection} (Deep SAD). 
Deep SAD trains a network to concentrate normal data near the center $\bc$, while mapping anomalous samples away from that center. 
Hence, this follows an unsupervised OE approach to AD. 
Here, we present a principled modification of Deep SAD based on the cross-entropy loss, which we call \emph{hypersphere classification} (HSC). 
We find that this modification improves performance over Deep SAD and use it in our experiments as a prototypical representative of the unsupervised OE approach to AD.
Potentially, one could further improve the unsupervised OE approach by developing an OE variant of CSI \citep{tack2020}, which we leave to future work and is out of scope of this paper.

Let $\cD = \{(\bx_1,y_1), \ldots, (\bx_n, y_n)\}$ be a dataset with $\bx_i \in \mR^d$ and $y \in \{0, 1\}$ where $y=1$ denotes normal and $y=0$ anomalous instances. Let $\phi_\theta : \mR^d \to \mR^r$ be a neural network and $l:\mR^r \to [0,1]$ a function that maps the output to a probabilistic score. 
Then, the cross-entropy loss is given by
\begin{equation}\label{eqn:cross}
  - \frac{1}{n} \sum_{i=1}^n y_i \log{l(\NN{\bx_i})} + (1{-}y_i) \log{(1{-}l(\NN{\bx_i}))}.
\end{equation}
For a standard binary classifier, $l$ is a linear layer followed by a sigmoid, and the decision region of the mapped samples $\NN{\bx_1},\ldots,\NN{\bx_n}$ is a half-space $S$. 
In this case, the preimage of $S$, $\phi_\theta^{-1}(S)$, is not guaranteed to be compact. 
To encourage the preimage of our normal decision region to be compact, we choose $l$ to be a radial basis function: $l(\bz)=\exp{(-\left\|\bz\right\|^2)}$.
In this case, (\ref{eqn:cross}) becomes
\begin{equation*}\label{eqn:radial}
  \frac{1}{n} \sum_{i=1}^n y_i\left\lVert\NN{\bx_i}\right\rVert^2 - (1{-}y_i)\log{(1{-}\exp{({-}\left\lVert\NN{\bx_i}\right\rVert^2)})}.
\end{equation*}
If there are no anomalies, the HSC loss simplifies to $\frac{1}{n}\sum_{i=1}^n\left\lVert\NN{\bx_i}\right\rVert^2$. 
For $\bc = 0$, we thus recover Deep SVDD as a special case.
Similar to Deep SVDD/SAD, we define our anomaly score as $s(\bx) := \left\lVert\NN{\bx}\right\rVert^2$.
Motivated by robust statistics \citep{hampel2005,huber2009}, we also considered replacing $l$ with radial functions that replace the squared-norm with more robust alternatives. 
Here, we found the pseudo-Huber loss \citep{charbonnier1997deterministic} to consistently yield the best results. 
We refer to Appendix \ref{appx:losses} for a detailed analysis.

\subsection{Contrastive language-image pre-training} \label{sec:clip}
To challenge the assumption that transfer learning approaches require OE for state-of-the-art detection performance, we consider a zero-shot approach to AD using the features of the contrastive language-image pre-training (CLIP) model \citep{radford2021learning}.
CLIP is trained on a massive dataset of 400 million (image, text) pairs with an objective to align corresponding pairs in feature space while keeping other pairs apart. 
Let $(\bx_u, \bx_v)$ denote an (image, text) pair, $\bu = f_u(\bx_u)$ and $\bv = f_v(\bx_v)$ the corresponding representations obtained by networks $f_u$ and $f_v$, and consequently $\bu_i$ and $\bv_i$ the representations of the $i$-th data pair.
CLIP uses the following losses: the text-to-image loss $l_i^{(v \rightarrow u)}$ and the image-to-text loss $l_i^{(u \rightarrow v)}$
\begin{equation}
    l_i^{(v \rightarrow u)} =   - \log \frac{\exp(\left\langle \bv_i, \bu_i \right\rangle  e^\tau)}{\sum_{k=1}^N \exp(\left\langle \bv_i, \bu_k \right\rangle e^\tau)},
    \quad
    l_i^{(u \rightarrow v)} =   - \log \frac{\exp(\left\langle \bu_i, \bv_i \right\rangle e^\tau)}{\sum_{k=1}^N \exp(\left\langle \bu_i, \bv_k \right\rangle e^\tau)},
\end{equation}
where $\left\langle \cdot, \cdot \right\rangle$ denotes the cosine similarity and $\tau$ is a temperature parameter. 
CLIP's final objective is
\begin{equation}
     \min_{f_u, f_v, \tau} \frac{1}{N} \sum_{i=1}^N \left( l_i^{(v \rightarrow u)} + l_i^{(u \rightarrow v)}  \right) / 2.
\end{equation}

\citet{radford2021learning} report that, without any fine-tuning on the downstream task, CLIP is able to outperform a fully supervised linear classifier with ResNet-50 features on several classification benchmarks, including ImageNet. 
For this, they use the names of the dataset classes as potential text candidates and predict the class whose text has the largest alignment with a given image. 
For out-of-distribution detection, \citet{fort2021exploring} explored using CLIP by taking the in-distribution and out-of-distribution text labels as candidates.
We use CLIP in a similar way to perform zero-shot AD, where we use the text pair $(v_1, v_2) =$ (``a photo of a \{NORMAL\_CLASS\}'', ``a photo of something'').
For a test image $\bx$, we compute its anomaly score as
\begin{equation} \label{eq:clip_ad}
    s(\bx) = \frac{\exp(\left\langle f_u(\bx), f_v(v_2) \right\rangle \cdot 100)}{\sum_{k=1}^2 \exp(\left\langle f_u(\bx), f_v(v_k) \right\rangle \cdot 100)}.
\end{equation}
Fine-tuning CLIP for AD with OE also is straightforward.
We simply minimize the score (\ref{eq:clip_ad}) for normal samples and maximize it for OE samples.
Since this corresponds to a binary cross-entropy loss, this is an instance of supervised OE, which we term ``BCE with CLIP'' (or just BCE-CL). 

\paragraph{On the legitimacy of transfer learning for AD} 
The use of transfer learning improved the performance of deep AD approaches significantly.
Yet, it seems at least questionable whether the use of pre-trained models is experimentally sound since there may be a semantic overlap between the pre-training data and the anomalies seen at test time. 
While it is technically true that the AD model is still unsupervised and does not exploit knowledge of test samples, the reported performance on typical image AD benchmarks might be spurious as the model may not generalize well to other data.
\citet{radford2021learning}, however, have investigated the overlap of data used for pre-training CLIP, where they remove all the data that overlaps with the downstream tasks and observe only an insignificant drop in performance on average. 
This suggests that our experiments and results based on CLIP reasonably explore generalization performance.
We still want to raise awareness for this somewhat problematic trend in deep AD, however, for which we propose future research below.

\section{Experimental setups}
\label{sec:experimental_setups}
Before we present our results, we explain the experimental setup. 
In particular, we introduce the common one vs.~rest benchmark, the CIFAR-10, ImageNet, CUB, DTD, Fashion-MNIST, and MNIST datasets, and the state-of-the-art AD methods we consider in our experiments.

\paragraph{One vs.~rest benchmark} The one vs.~rest evaluation protocol is a ubiquitous benchmark in the deep AD literature  \citep{ruff2018deep, golan2018deep, hendrycks2019deep, hendrycks2019using, ruff2020,sohn2021, deecke2021transfer, liznerski2021, reiss2021panda}.
This benchmark constructs AD settings from classification datasets (e.g., CIFAR-10) by considering the ``one'' class (e.g., ``airplane'') as being normal and the ``rest'' classes (e.g., ``automobile'', ``bird'', ...) as being anomalous at test time.
In each experiment, we train a model using only the training set of the normal class and samples from an OE set that are not contained in the anomaly classes of the benchmark. 
We use the same OE auxiliary datasets as suggested in previous works \citep{hendrycks2019deep,hendrycks2019using,liznerski2021}.
To evaluate detection performance, we consider the commonly used Area Under the ROC curve (AUC) on the one vs.~rest test sets. 
This is repeated over classes and multiple random seeds. 

\paragraph{Datasets}
For our experiments we focus on the well-established one vs.~rest versions of CIFAR-10 and ImageNet. We also consider less common datasets, for which there are yet no AD results in the literature. If not mentioned otherwise, we use all available classes as our one vs.~rest classes.
\begin{itemize}[noitemsep,topsep=-8pt,leftmargin=*]
    \item CIFAR-10: For CIFAR-10 \citep{krizhevsky2009learning}, we use 80 Million Tiny Images (80MTI)\citep{torralba200880} as OE, with CIFAR-10 and CIFAR-100 images removed. 
    This follows the experimental setup in \citet{hendrycks2019using}. 
    \item ImageNet-30: For ImageNet \citep{imagenet}, we use a subset of 30 classes as the one vs.~rest classes, which are the same classes used in \citet{hendrycks2019using}. 
    For OE, we use ImageNet-22K with ImageNet-1K removed, again following \citet{hendrycks2019using}. 
    \item CUB: CUB (Caltech-UCSD Birds-200-2011) \citep{cub} is a more challenging dataset where each of the 200 classes consists of images of a specific bird type (e.g., ``black footed albatross'', ``blue jay'', ...). We again use ImageNet-22k as OE.
    \item DTD: DTD (Describable Textures Dataset) \citep{dtd} is a dataset containing 47 different classes of images of textures like ``cracked'' and ``braided''. We use ImageNet-22k as OE.
    \item Fashion-MNIST: For Fashion-MNIST \citep{fmnist}, we consider a grayscale version of CIFAR-100 as OE, as suggested in \citet{liznerski2021}.
    \item MNIST: For MNIST \citep{mnist}, we use EMNIST \citep{emnist} as OE.
\end{itemize}

\paragraph{End-to-end methods} We present results from end-to-end methods (without transfer learning) including all methods that achieve state-of-the-art performance on the CIFAR-10 and ImageNet-30 one vs.~rest benchmarks.
\begin{itemize}[noitemsep,topsep=-8pt,leftmargin=*]
    \item Unsupervised: Shorthands for unsupervised methods are DSVDD \citep{ruff2018deep}, GT \citep{golan2018deep}, GT+ \citep{hendrycks2019using}, and CSI \citep{tack2020}.
    \item Unsupervised OE: We implement HSC from Section \ref{sec:hypsphcla} and DSAD \citep{ruff2020} as unsupervised OE methods and also report the results from the unsupervised OE variant of GT+ \citep{hendrycks2019using}.
    \item Supervised OE: BCE denotes a standard binary cross-entropy classifier.
    We also implement the Focal loss classifier with $\gamma=2$ \citep{lin17}, a BCE variant for imbalanced classes that was also presented in \citet{hendrycks2019using}.
    Results from \citet{hendrycks2019using} are marked with an asterisk as Focal*.
\end{itemize}

\paragraph{Transfer learning-based methods} 
We consider the following transfer learning-based methods.
\begin{itemize}[noitemsep,topsep=-8pt,leftmargin=*]
    \item Unsupervised: We implement a zero-shot anomaly detector using CLIP's feature space as described in Section \ref{sec:clip} and use DN2 \citep{bergman2020deep} and PANDA \citep{reiss2021panda} as shorthands respectively for these unsupervised methods from the literature. 
    \item Supervised OE: We consider a fine-tuned version of CLIP with a binary cross-entropy classifier, denoted as BCE-CL.
    When available, we also report the results of ADIP \citep{deecke2021transfer} and the supervised OE variant of PANDA \citep{reiss2021panda}.
\end{itemize} \vspace{1em}

We provide all network architecture and optimization details in Appendix \ref{appx:architectures} and investigate the impact of $\gamma$ for the Focal loss in Appendix \ref{appx:focal}.
We report the mean AUC performance over all classes and seeds in the main paper. 
Individual results per class and method are given in Appendix \ref{appx:full_results}. 
Each experiment is averaged over 10 seeds if not mentioned otherwise.

\section{On the usefulness of samples in deep AD} \label{sec:meth_experiments}
Traditionally, AD methods utilize an unsupervised approach due to the assumption that it is impossible to characterize everything that is not normal. 
As mentioned above, later works introduced the idea of including a large collection of random images (OE) during training that serve as auxiliary examples of anomalousness to improve unsupervised AD methods \citep{hendrycks2019deep}. 
Models are trained with these OE samples using a loss that is essentially inverting a given unsupervised AD loss. 
We here look into whether an unsupervised OE approach, instead of a straightforward supervised OE approach using standard binary cross-entropy, is really necessary.

\subsection{Supervised OE achieves state-of-the-art results} \label{sec:exp_sota_without_transfer}
The first work on OE \citep{hendrycks2019deep} applied an unsupervised OE method to two experimental setups: CIFAR-10 with 80MTI as OE and ImageNet-30 with ImageNet-22K as OE. 
Here we consider the same experimental setups with the basic HSC and BCE classifiers. 
We also include experiments on the mostly unexplored one vs.~rest versions of CUB, DTD, Fashion-MNIST, and MNIST.
The results are shown in Table \ref{tab:aucs_without_transfer}.
In this section we only consider end-to-end methods (i.e., no transfer learning).

%%%%%%%%%%%%%%%%%%%%%%%%%%%%%%%%%%%%%%%%%%%%%%%%%%%%%%%%%%%%%%%%%%%%%%%%%%%%%%%%
\begin{table}[hb]
  \caption{Mean AUC detection performance in \% for end-to-end methods on the CIFAR-10, ImageNet-30, CUB, DTD, Fashion-MNIST, and MNIST one vs.~rest benchmarks.}
  \label{tab:aucs_without_transfer}
  \begin{center}
    \small \begin{tabular}{lccccccccc} 
\toprule 
& \multicolumn{3}{c|}{Unsupervised} & \multicolumn{3}{c|}{Unsupervised OE} & \multicolumn{3}{c}{Supervised OE} \\ 
 & DSVDD & GT+* & \multicolumn{1}{c|}{CSI*} & GT+* & DSAD & \multicolumn{1}{c|}{HSC} & Focal* & Focal & \multicolumn{1}{c}{BCE} \\ 
\midrule 
CIFAR-10 & 64.8* & 90.1 & \multicolumn{1}{c|}{94.3} & 95.6 & 94.5 & \multicolumn{1}{c|}{95.9} & 87.3 & 95.8 & \multicolumn{1}{c}{\bf 96.1} \\ 
ImageNet-30 & 61.1 & 84.8 & \multicolumn{1}{c|}{91.6} & 85.7 & 96.7 & \multicolumn{1}{c|}{97.3} & 56.1 & 97.5 & \bf 97.7  \\
\midrule
CUB & 59.0 & $\times$ & \multicolumn{1}{c|}{$\times$} & $\times$ & 81.1 & \multicolumn{1}{c|}{83.2} & $\times$ & 83.5 & \bf 84.1  \\
DTD & 56.8 & $\times$ & \multicolumn{1}{c|}{$\times$} & $\times$ & 72.4 & \multicolumn{1}{c|}{72.7} & $\times$ & 73.2 & \bf 73.3  \\
Fashion-MNIST & 86.3 & $\times$ & \multicolumn{1}{c|}{$\times$} & $\times$ & 86.4 & \multicolumn{1}{c|}{87.3} & $\times$ & \bf 87.7 & 86.4   \\
MNIST & 97.6 & $\times$ & \multicolumn{1}{c|}{$\times$} & $\times$ & 97.6 & \multicolumn{1}{c|}{98.2} & $\times$ & \bf 98.4 & \bf 98.4   \\
\bottomrule 
\end{tabular} 
  \end{center}
\end{table}
%%%%%%%%%%%%%%%%%%%%%%%%%%%%%%%%%%%%%%%%%%%%%%%%%%%%%%%%%%%%%%%%%%%%%%%%%%%%%%%%

\paragraph{Discussion} 
Surprisingly, we find that the choice of AD method has little impact on performance.
On CIFAR-10, all unsupervised OE methods yield a comparable detection performance, while the supervised methods (Focal and BCE) show state-of-the-art performance, with BCE attaining the overall best mean AUC.
On ImageNet, Deep SAD, HSC, Focal, and BCE all outperform the current state of the art (CSI*) by a significant margin.
We are unsure as to why the Focal* performs so poorly in \citet{hendrycks2019using} since their experimental code is not publicly available. 
On CUB and DTD, we observe a similar behavior, but the methods perform slightly worse, most likely due to the more fine-grained class categorization in CUB and the more subtle anomalies in DTD.
These benchmarks, as they show room for improvement, thus present themselves useful for future AD evaluation. 
On the rather low-dimensional Fashion-MNIST and MNIST datasets, OE is not able to improve significantly over unsupervised approaches, and the different OE approaches again perform similarly.
Overall, our results show that a vanilla classifier using OE outperforms all previous deep AD approaches.

Our experiments seem to suggest that the inclusion of OE does not just improve AD performance, it also changes the problem into a typical supervised classification problem that does not require a compact decision boundary (cf. Figure \ref{fig:toy_example}).
This stands in contrast to previous observations in shallow AD \citep{tax2001, gornitz2013toward}.
While there is an abundance of OE data for the sorts of AD problems we investigate here, we want to highlight that this data is likely not very helpful for AD problems where anomalies are more subtle. 
For example, in the realm of manufacturing, pictures of cats and trucks are not as useful for OE as industrial images, which are not so widely available.
Nonetheless, one may still have a few anomalous examples to incorporate during training.
As one removes OE data, we would expect the behavior of an unsupervised OE approach to transition to the behavior of an unsupervised method that outperforms supervised OE.
Our next experiment investigates this transition.

\subsection{Few OE samples are sufficient}  \label{sec:exp_var_oe_size}
\begin{figure}[hbt] % OVR CIFAR-10, CUB, ImageNet-30
  \begin{center} 
      \subfigure[CIFAR-10 (over 10 classes $\times$ 10 seeds)]{\label{fig:cifarvstiny}\includegraphics[height=0.142\textheight]{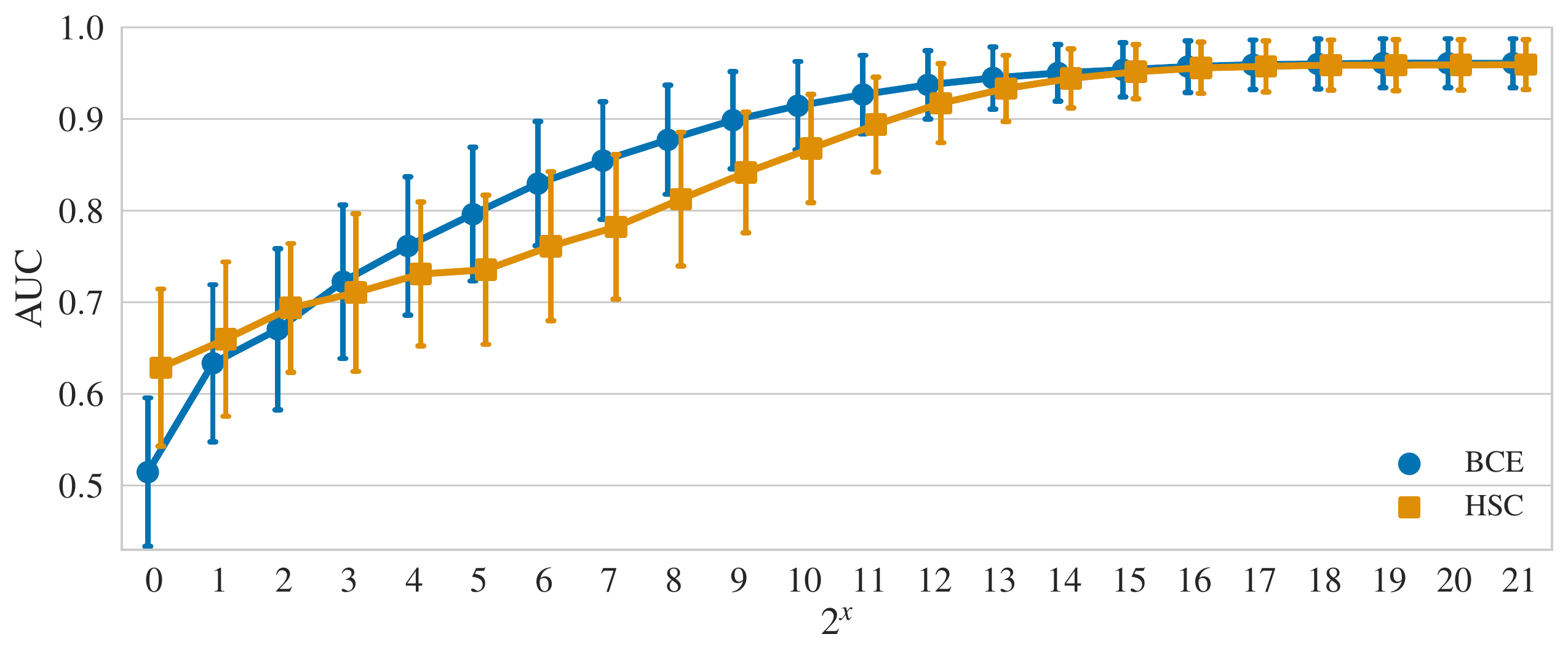}}
      \subfigure[CUB (over 200 classes $\times$ 2 seeds)]{\label{fig:cub_varoe}\includegraphics[height=0.142\textheight]{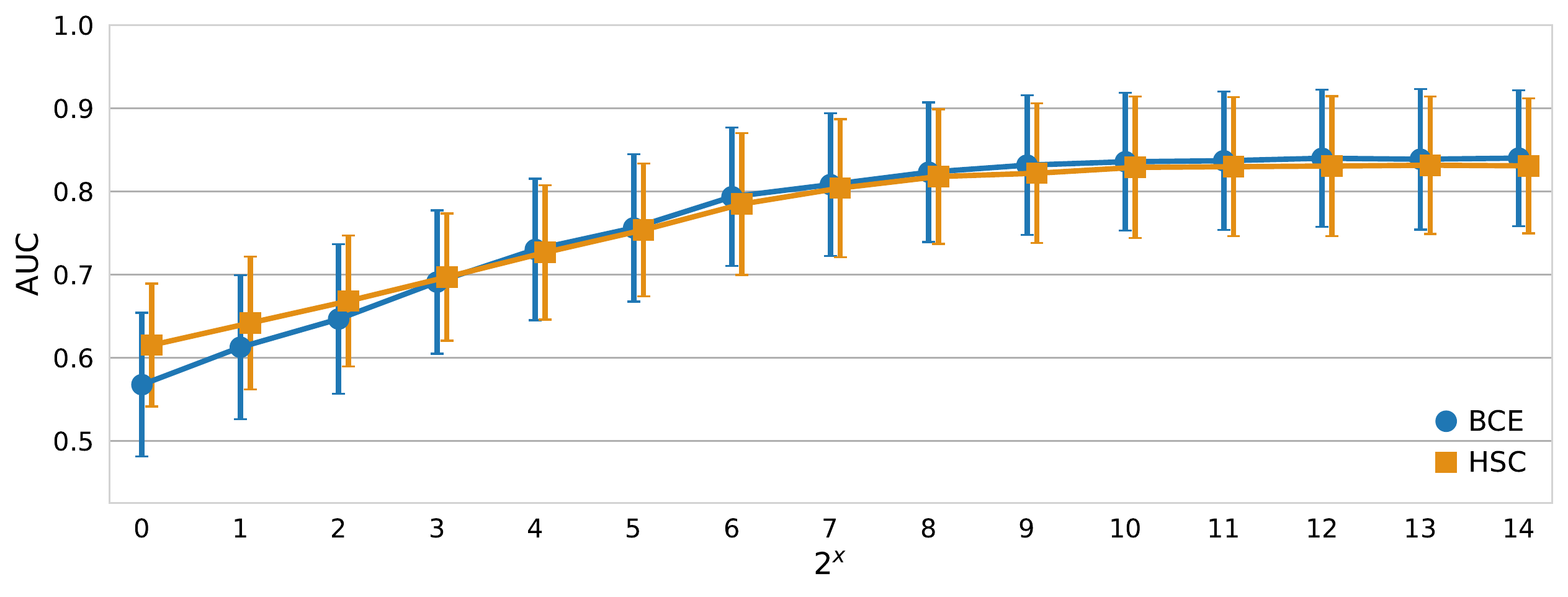}}
      \subfigure[ImageNet-30 (over 30 classes $\times$ 5 seeds)]{\label{fig:imagenet1kvs22k}\includegraphics[width=0.65\textwidth]{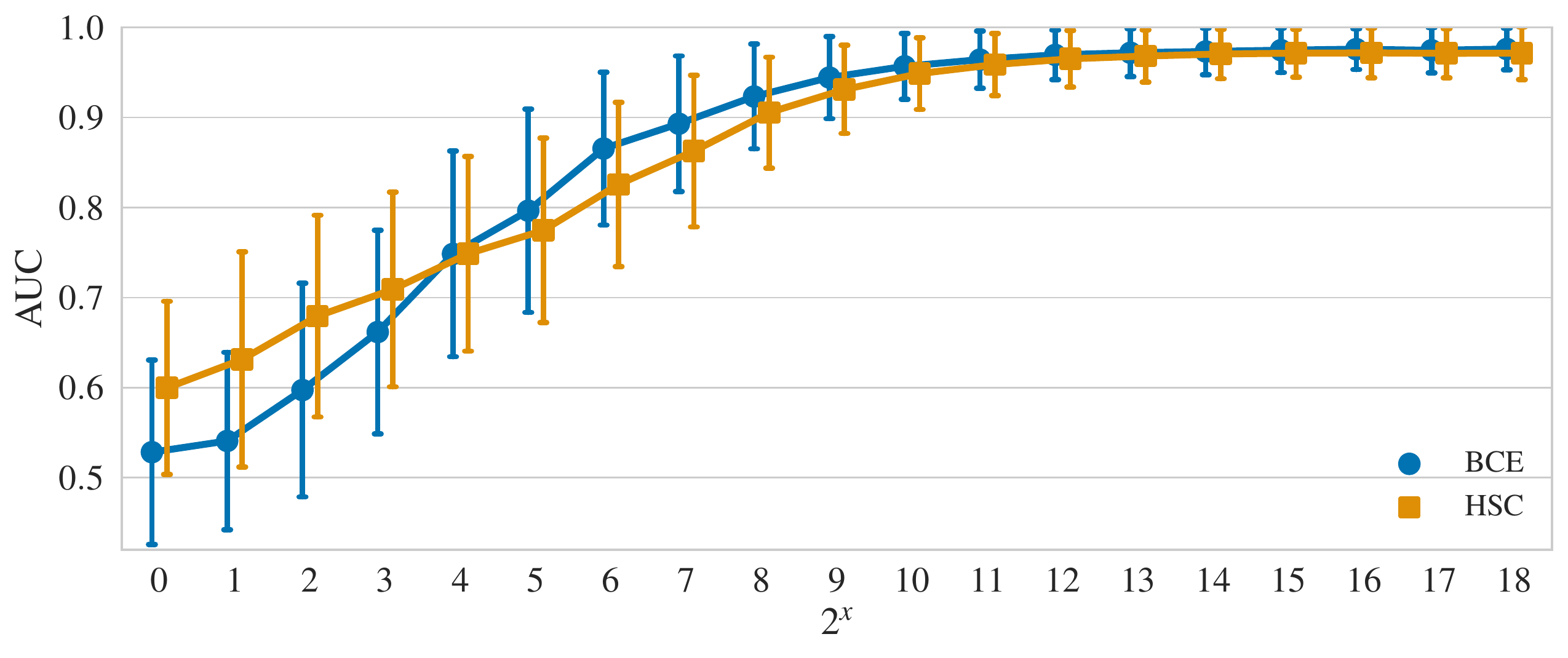}}
      \caption{Mean AUC detection performance in \% on the CIFAR-10, CUB, and ImageNet-30 one vs.~rest benchmarks when varying the number of 80MTI and ImageNet-22K OE samples, respectively.}
      \label{fig:var_oe_size}
  \end{center}
\end{figure}

To investigate the effect of the size of the OE corpus, we perform experiments varying the OE training set size from just \emph{one} sample ($2^0=1$) to using the maximal amount of OE data.
Our results for the high-dimensional datasets CIFAR-10, ImageNet-30, and CUB are shown in Figure \ref{fig:var_oe_size}. 
With sufficiently few samples, HSC outperforms BCE, which seems to indicate a regime where the unsupervised OE approach is advantageous. 
Interestingly, this regime seems to be quite small, with supervised OE needing only 8 samples to outperform unsupervised OE on CIFAR-10 and only $\sim32$ samples for such a transition on ImageNet and CUB.
Remarkably, BCE classification outperforms previous state-of-the-art methods on ImageNet with only using $\sim256$ OE samples.
A training set with so few outliers and high-dimensional data represents an instance of skewed data (see Figure \ref{fig:toy_example} and caption), where the common intuition suggests that supervised OE should \emph{not} generalize well.
As before, we see that the choice of the specific method seems negligible as soon as sufficient OE data is available.

%%%%%%%%%%%%%%%%%%%%%%%%%%%%%%%%%%%%%%%%%%%%%%%%%%%%%%%%%%%%%%%%%%%%%%%%%%%%%%%%
%%%%%%%%%%%%%%%%%%%%%%%%%%%%%%%%%%%%%%%%%%%%%%%%%%%%%%%%%%%%%%%%%%%%%%%%%%%%%%%%

\begin{figure}[hbt] % OVR FMNIST, MNIST
  \begin{center} 
      \subfigure[Fashion-MNIST (over 10 classes $\times$ 5 seeds)]{\label{fig:fmnist_varoe}\includegraphics[width=0.465\textwidth]{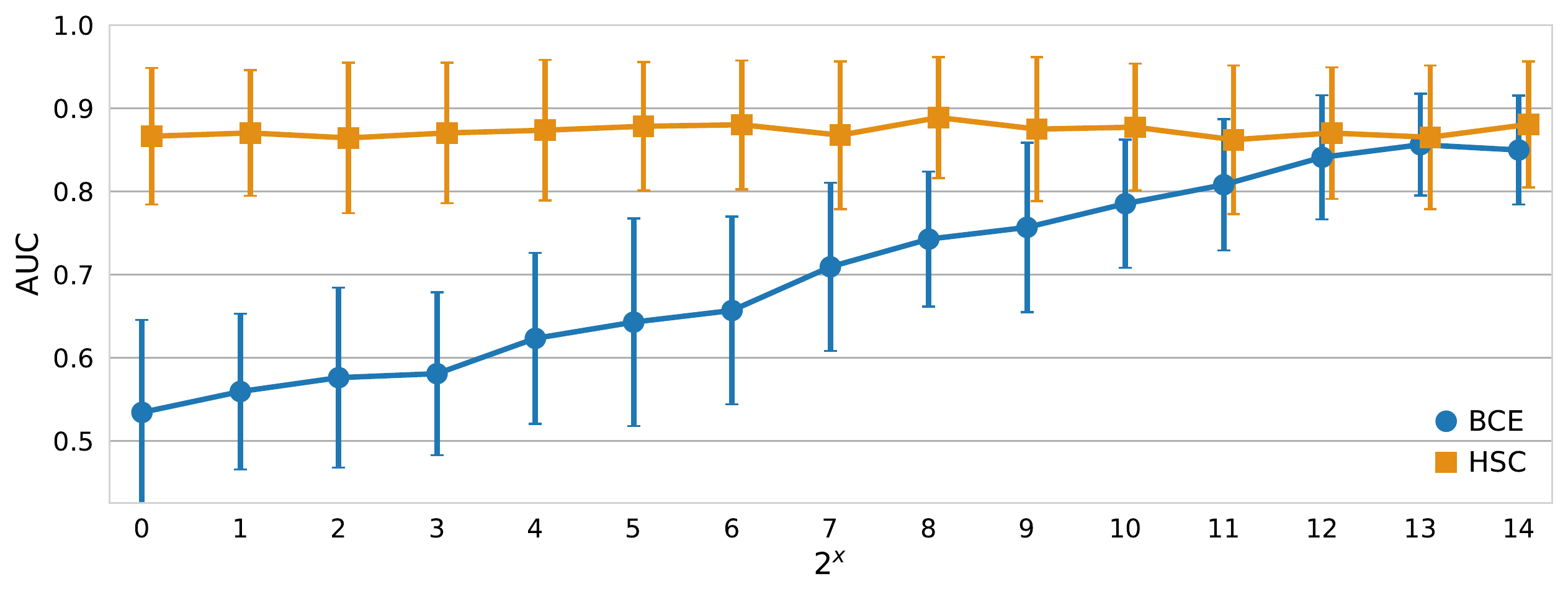}}
      \subfigure[MNIST (over 10 classes $\times$ 5 seeds)]{\label{fig:mnist_varoe}\includegraphics[width=0.465\textwidth]{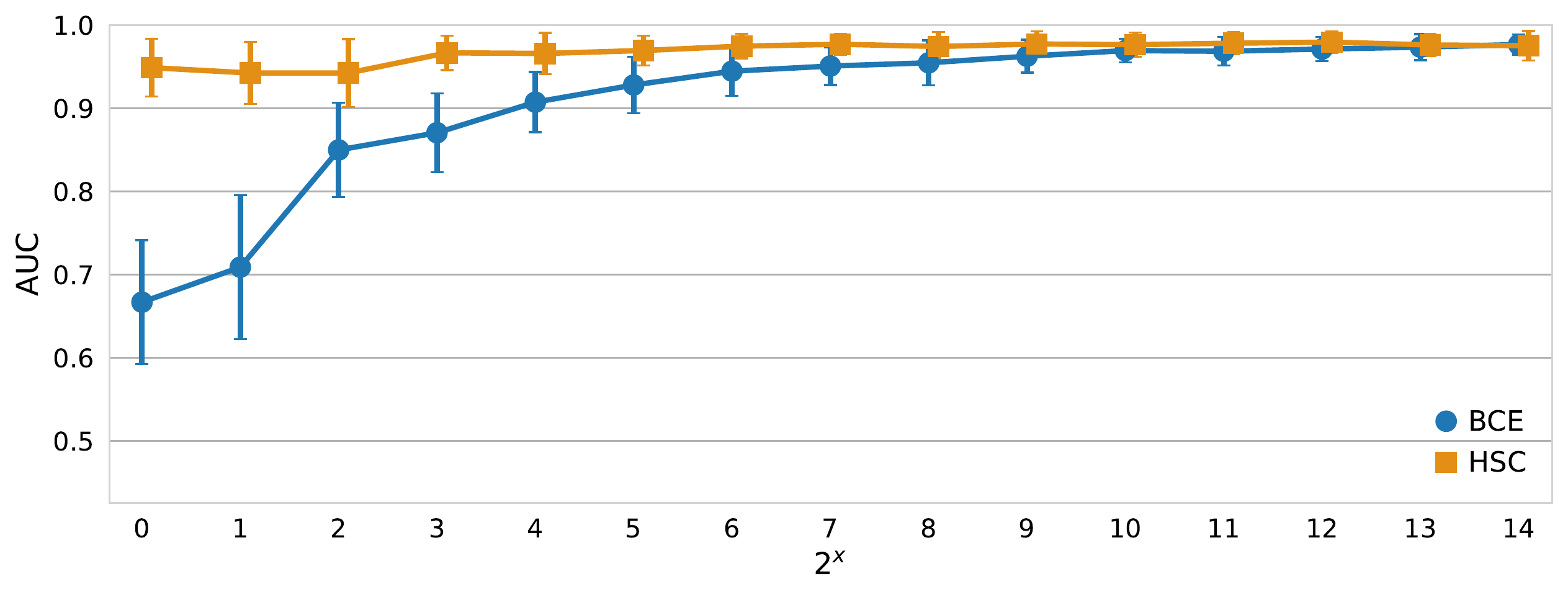}}
      \caption{Mean AUC detection performance in \% on the Fashion-MNIST and MNIST one vs.~rest benchmarks when varying the number of grayscale CIFAR-100 and EMNIST OE samples, respectively.}
      \label{fig:var_oe_size_mnist}
  \end{center}
\end{figure}

Results with varied OE set size for Fashion-MNIST and MNIST are presented in Figure \ref{fig:var_oe_size_mnist}. 
We find that, in contrast to the experiments on high-dimensional datasets, the amount of OE has little impact on performance for the well-performing HSC, whereas BCE requires a lot of OE to perform competitively.
It seems that the intuition of Figure \ref{fig:toy_example} does still hold for low-dimensional datasets where shallow methods perform well.

%%%%%%%%%%%%%%%%%%%%%%%%%%%%%%%%%%%%%%%%%%%%%%%%%%%%%%%%%%%%%%%%%%%%%%%%%%%%%%%%
%%%%%%%%%%%%%%%%%%%%%%%%%%%%%%%%%%%%%%%%%%%%%%%%%%%%%%%%%%%%%%%%%%%%%%%%%%%%%%%%

\begin{figure}[hbt]  % LOCO CIFAR-10, ImageNet-30
  \begin{center} 
      \subfigure[CIFAR-10 (over 10 classes $\times$ 2 seeds)]{\label{fig:cifarvstiny_loco}\includegraphics[width=0.495\textwidth]{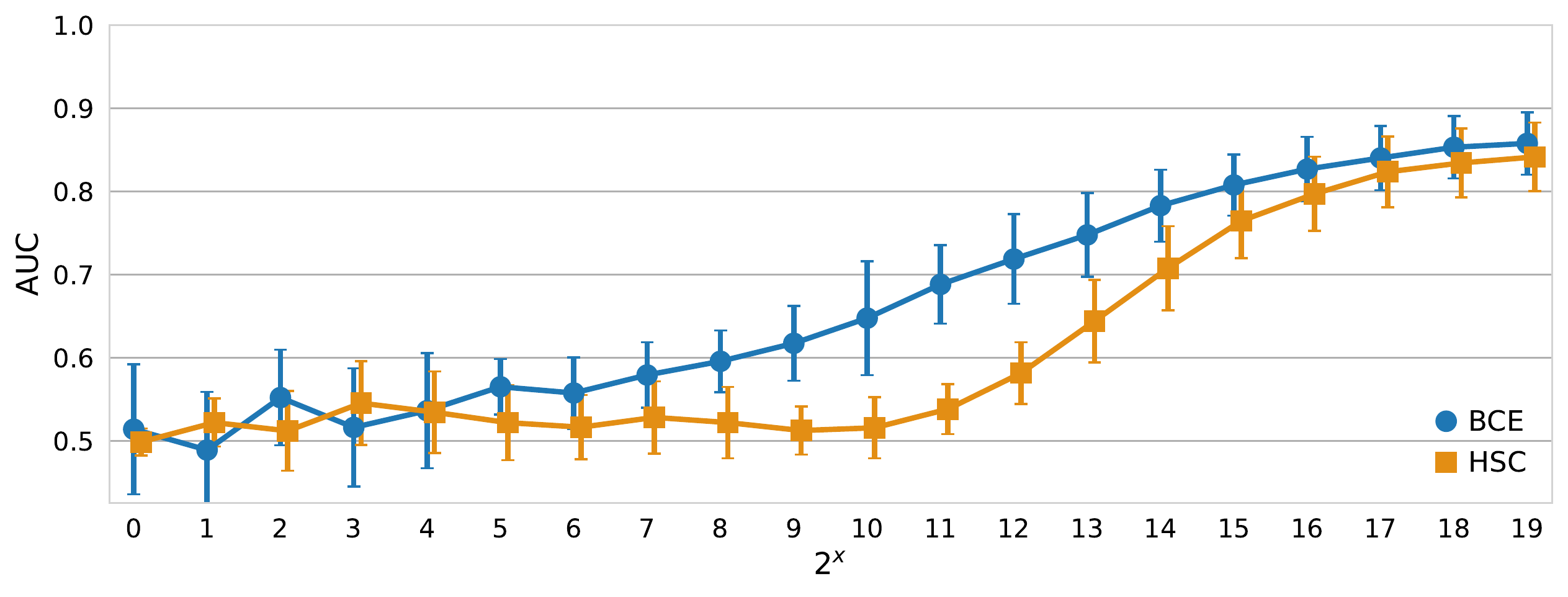}}
      \subfigure[ImageNet-30 (over 30 classes $\times$ 2 seeds)]{\label{fig:imagenet1kvs22k_loco}\includegraphics[width=0.495\textwidth]{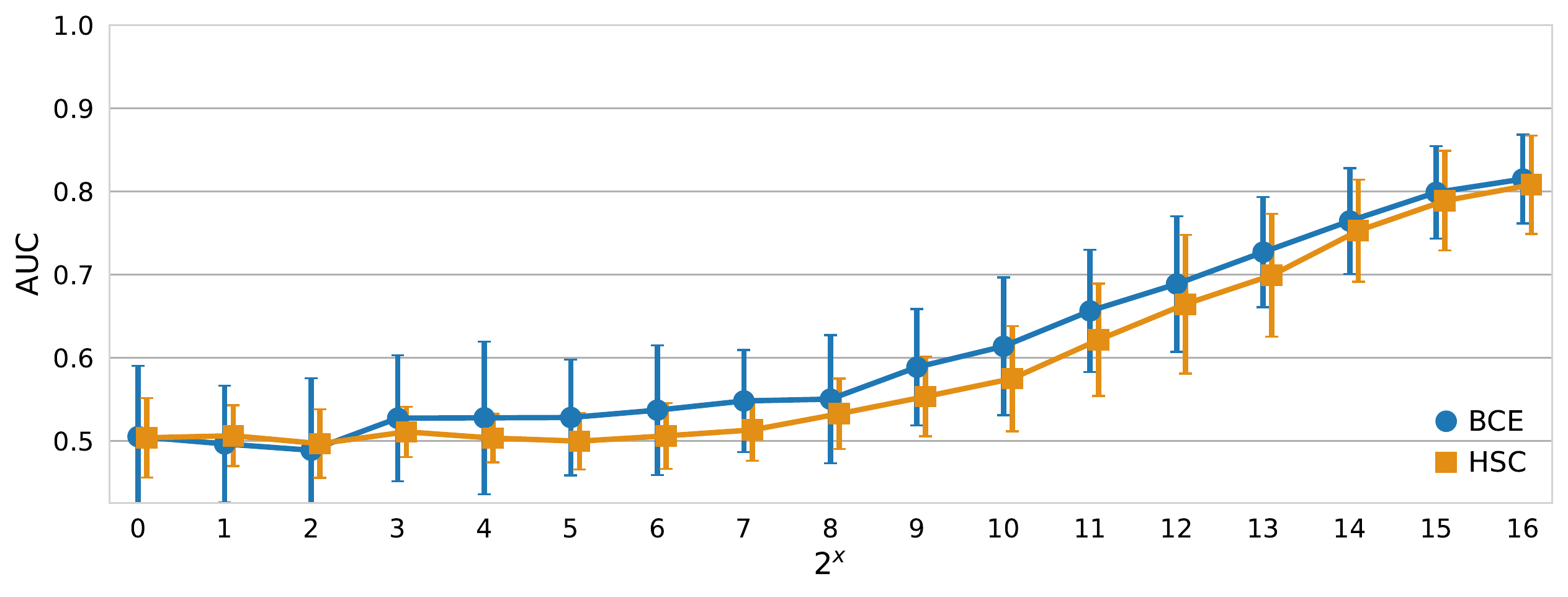}}
      \caption{Mean AUC detection performance in \% on the CIFAR-10 and ImageNet-30 leave-one-class-out benchmarks when varying the number of 80MTI and ImageNet-22K OE samples, respectively.}
      \label{fig:var_oe_size_loco}
  \end{center}
\end{figure}

\paragraph{There are settings that require more OE data}
The one vs.~rest image-AD benchmark represents a certain type of typical AD problem where the normal data distribution is roughly unimodal since the normal samples are drawn from just one class. 
However, one can also create a more challenging, and perhaps less typical, AD benchmark by considering the ``rest'' classes (e.g., ``automobile'', ``bird'', ...) as being normal and the ``one'' class (e.g., ``airplane'') as being anomalous at test time.  
The distribution of normal samples in this leave-one-class-out approach becomes multimodal. 
Figure \ref{fig:var_oe_size_loco} shows the AD performance of HSC and BCE on the CIFAR-10 and ImageNet-30 leave-one-class-out image-AD benchmark, when one varies the size of the OE training set.

%%%%%%%%%%%%%%%%%%%%%%%%%%%%%%%%%%%%%%%%%%%%%%%%%%%%%%%%%%%%%%%%%%%%%%%%%%%%%%%%
%%%%%%%%%%%%%%%%%%%%%%%%%%%%%%%%%%%%%%%%%%%%%%%%%%%%%%%%%%%%%%%%%%%%%%%%%%%%%%%%

Compared to previous experiments it seems that more OE data is required to achieve strong performance. 
This indicates that more OE samples are necessary when the normal class is not concentrated.
We also report results when one uses the full OE dataset in Appendix \ref{appx:section_loco_results}, where we see that the various methods perform similarly with respect to one another as in the one vs.~rest tasks, albeit overall slightly worse due to the more difficult problem setting. Notably BCE is the best performing method, further supporting its effectiveness at AD with OE.

\begin{figure}[hbt]  % LOCO CIFAR-10, ImageNet-30
  \begin{center} 
      \subfigure[DTD (over 47 classes $\times$ 2 seeds)]{\label{fig:dtd_varoe}\includegraphics[width=0.495\textwidth]{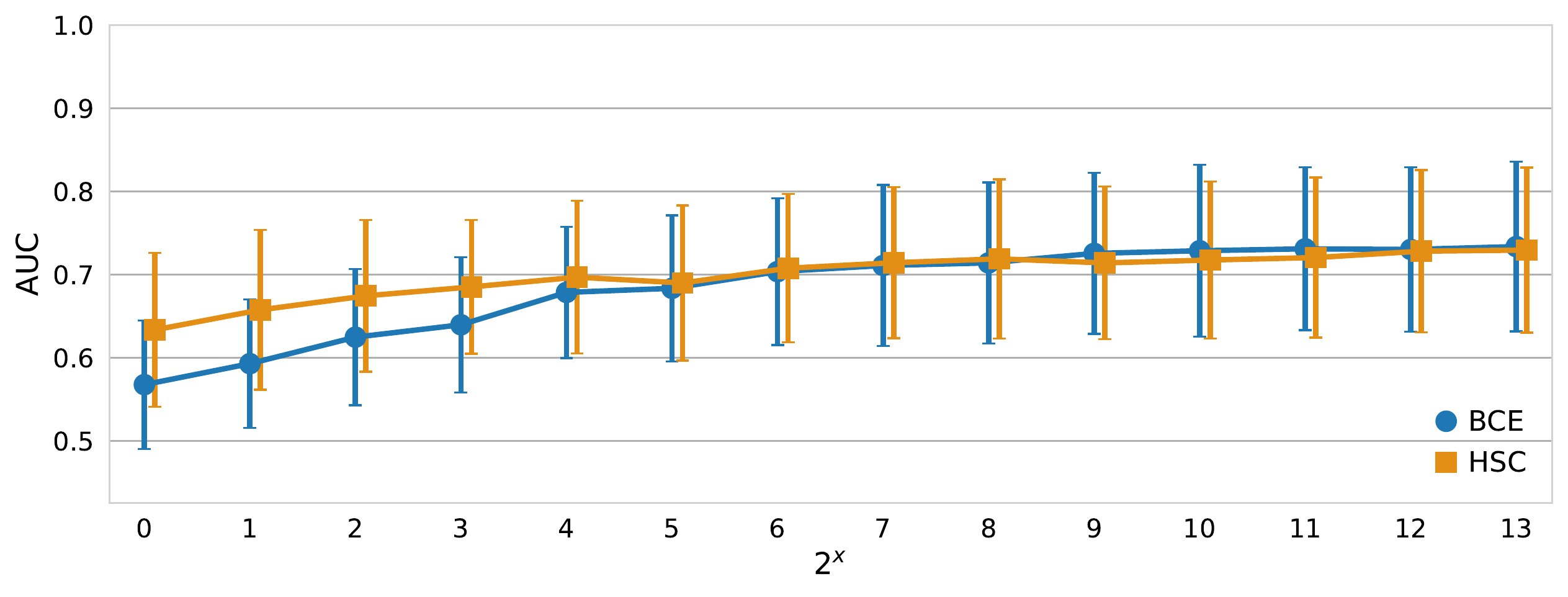}}
      \subfigure[MVTec-AD (over 15 classes $\times$ 2 seeds)]{\label{fig:mvtec_varoe}\includegraphics[width=0.495\textwidth]{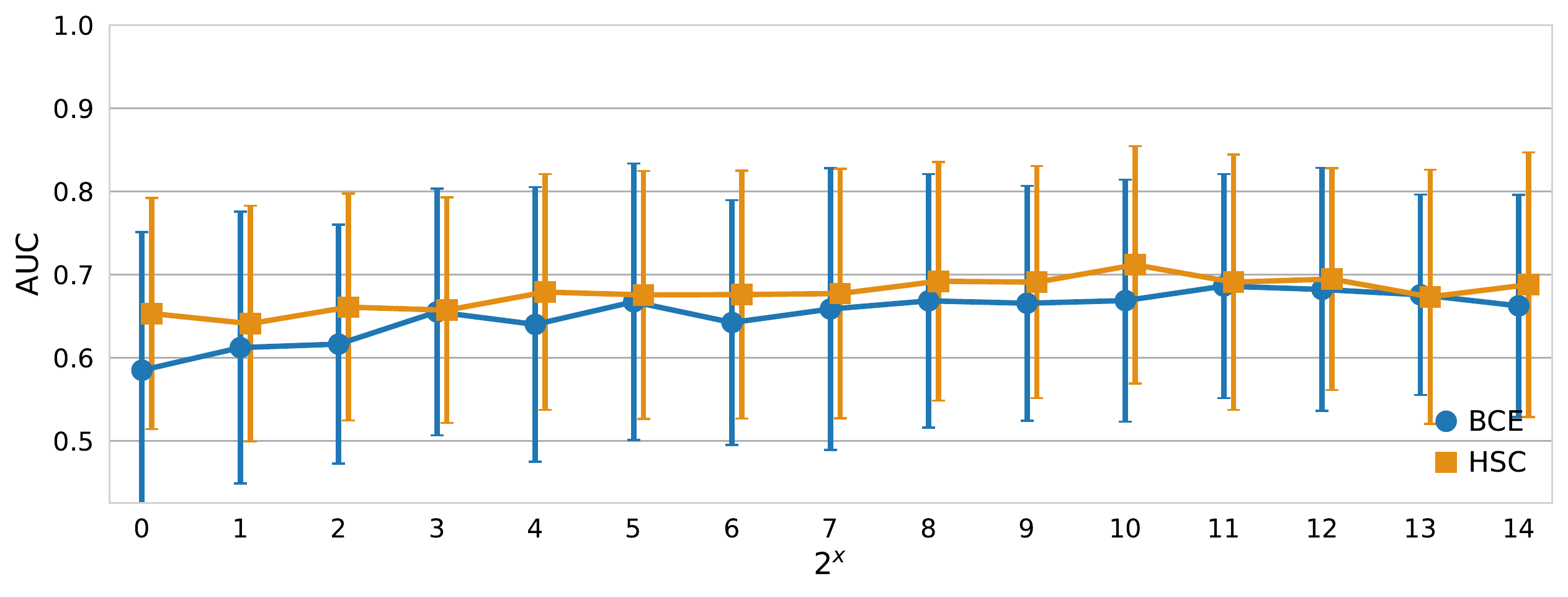}}
      \caption{Mean AUC detection performance in \% on the DTD one vs.~rest and on the MVTec-AD benchmarks when varying the number of ImageNet-22K OE samples.}
      \label{fig:var_oe_size_mvtec}
  \end{center}
\end{figure}

\paragraph{There are settings where random images are not informative as OE}
As mentioned in Section \ref{sec:methods_background}, this paper focuses on natural images because random natural images are likely not informative when used as OE in other settings. 
To demonstrate this, we apply our methods to the one vs.~rest DTD benchmark and the MVTec-AD dataset. 
DTD contains different classes of textures.
MVTec-AD contains image-AD scenarios for detecting manufacturing defects for a variety of object types (e.g., screws, bottles, wires, or sections of carpet).
For example, one scenario consists of a training set containing images of normal screws and a test set containing images of normal screws along with images of screws with defects, which serve as anomalies.
Figure \ref{fig:var_oe_size_mvtec} shows results when varying the number of ImageNet-22k samples used as OE for training. 
We find that the OE training set size has little impact on the AD performance, especially for MVTec-AD.

%%%%%%%%%%%%%%%%%%%%%%%%%%%%%%%%%%%%%%%%%%%%%%%%%%%%%%%%%%%%%%%%%%%%%%%%%%%%%%%%
%%%%%%%%%%%%%%%%%%%%%%%%%%%%%%%%%%%%%%%%%%%%%%%%%%%%%%%%%%%%%%%%%%%%%%%%%%%%%%%%

We also show results using the full ImageNet-22k dataset as OE for various methods for MVTec-AD in Table \ref{tab:appx_mvtec} in Appendix \ref{appx:section_mvtec_results}. 
As expected, OE performs more poorly on MVTec-AD than state-of-the-art methods: HSC scores around 70\% while \citet{roth2022towards} score around 99\% AUC. 
Contrary to the experiments with CIFAR-10, CUB, and ImageNet-30, we observe that HSC is a bit stronger than BCE (66\%). There seems to be some evidence that HSC performs better when the OE data is not very informative, either when there is a dearth of OE data like in Section \ref{sec:exp_var_oe_size}, or when the OE data simply isn't relevant to the AD scenario. We explore this a bit further in Section \ref{sec:exp_robustness}.

\paragraph{Diversity of OE data is important}
To measure the impact of data diversity we also experiment with varying the number of OE classes.
We defer these results to Appendix \ref{appx:oe_diversity}.
In summary, we find that performance overall increases with OE data diversity, but already using just one OE class still performs relatively well.

While our results show that surprisingly few samples are needed to achieve competitive performance for end-to-end models on the standard one vs.~rest AD benchmarks, more recent methods in deep AD tend to use transfer learning. 
Interestingly, we find that one can achieve state-of-the-art AD performance using pre-trained models with \emph{no training samples} (i.e., in a zero-shot setting).

\subsection{Transfer learning enables zero-shot AD with state-of-the-art performance} \label{sec:exp_sota_with_transfer}
More recent progress in deep AD has been achieved through transfer learning, which has further improved the state of the art on standard deep AD benchmarks.
These methods use a network pre-trained on large datasets to provide rich representations as a starting basis for deep AD. 
Again, we look into how useful auxiliary data is for these algorithms. 
Here we investigate the effect of using OE data (vs.~no OE data at all) and the implicit use of an extraneous dataset via pre-training. 
We also consider the situation where \emph{no normal data} is used and find that CLIP outperforms all previous end-to-end methods on virtually all datasets.
Table \ref{tab:aucs_transfer} shows the results for transfer learning-based methods on the one vs.~rest benchmarks. 
Note that, since DN2, PANDA, and ADIP employ ImageNet pre-trained networks, they cannot be compared on the ImageNet AD benchmark in a fair way.

%%%%%%%%%%%%%%%%%%%%%%%%%%%%%%%%%%%%%%%%%%%%%%%%%%%%%%%%%%%%%%%%%%%%%%%%%%%%%%%%
\begin{wraptable}[12]{r}{0.6\textwidth} 
  \vspace{-2.3em}
  \caption{Mean AUC detection performance in \% for methods with transfer learning on the CIFAR-10, ImageNet-30, CUB, DTD, Fashion-MNIST, and MNIST one vs.~rest benchmark.}
  \label{tab:aucs_transfer}
  \centering 
  \vspace{0.5em}
  \resizebox{0.6\textwidth}{!}{% \begin{tabular}{lccccc} 
% \toprule 
% & \multicolumn{2}{c|}{Unsupervised} & \multicolumn{1}{c|}{Unsupervised OE} & \multicolumn{2}{c}{Supervised OE} \\ 
% & PANDA* & \multicolumn{1}{c|}{CLIP} & \multicolumn{1}{c|}{PANDA*} & ADIP* & \multicolumn{1}{c}{BCE} \\ 
% \midrule 
% CIFAR-10 & 96.2 & \multicolumn{1}{c|}{98.5} & \multicolumn{1}{c|}{98.9} & 99.1 & \multicolumn{1}{c}{\bf 99.6} \\ 
% ImageNet & $\times$ & \multicolumn{1}{c|}{99.88} & \multicolumn{1}{c|}{$\times$} & $\times$ & \multicolumn{1}{c}{\bf 99.90} \\ 
% \bottomrule 
% \end{tabular} 
\begin{tabular}{lcccccc} 
\toprule 
& \multicolumn{3}{c|}{Unsupervised} & \multicolumn{3}{c}{Supervised OE} \\ 
& DN2* & PANDA* & \multicolumn{1}{c|}{CLIP} & PANDA* & ADIP* & \multicolumn{1}{c}{BCE-CL} \\ 
\midrule 
CIFAR-10 & 92.5 & 96.2 & \multicolumn{1}{c|}{98.5} & 98.9 & 99.1 & \multicolumn{1}{c}{\bf 99.6} \\ 
ImageNet-30 & $\times$ & $\times$ & \multicolumn{1}{c|}{99.88} & $\times$ & $\times$ & \multicolumn{1}{c}{\bf 99.90} \\ 
\midrule
CUB & $\times$ & $\times$ & \multicolumn{1}{c|}{97.1} & $\times$ & $\times$ & \multicolumn{1}{c}{\bf 97.5} \\ 
DTD & $\times$ & $\times$ & \multicolumn{1}{c|}{90.2} & $\times$ & $\times$ & \multicolumn{1}{c}{\bf 94.6} \\ 
Fashion-MNIST & $\times$ & $\times$ & \multicolumn{1}{c|}{89.0} & $\times$ & $\times$ & \multicolumn{1}{c}{\bf 94.7} \\ 
MNIST & $\times$ & $\times$ & \multicolumn{1}{c|}{59.0} & $\times$ & $\times$ & \multicolumn{1}{c}{\bf 96.0} \\ 
\bottomrule 
\end{tabular} 
}
\end{wraptable}
%%%%%%%%%%%%%%%%%%%%%%%%%%%%%%%%%%%%%%%%%%%%%%%%%%%%%%%%%%%%%%%%%%%%%%%%%%%%%%%%

\paragraph{Discussion}
These results highlight the remarkable efficacy of pre-training. 
Disregarding the CLIP results for now, we see that pre-training improves over previous unsupervised deep results on CIFAR-10. 
The results of PANDA indicates that OE is still useful for deep AD: OE provides additional information that is not learned in the pre-training task. 
Undoubtedly, the most interesting result here is the observation that one can outperform all previous state-of-the-art methods using no additional training from the benchmarks.
In a zero-shot setting, CLIP (unsupervised) outperforms all previous end-to-end (see Table \ref{tab:aucs_without_transfer}) and unsupervised methods on all datasets apart from MNIST. 
Fine-tuning CLIP with OE along with normal samples (BCE-CL), further improves CLIP's results setting a new state of the art on the CIFAR-10 and ImageNet benchmarks.
Remarkably, CLIP performs similarly on DTD that consists of textures instead of natural images and, on CUB, outperforms end-to-end methods by a large margin (13.4\% AUC), essentially solving this quite challenging benchmark.

Similar to results in other areas of deep learning \citep{bommasani2021opportunities}, the use of large pre-trained networks offers an effective and convenient way to improve performance.
Though transfer learning seems to be a natural endpoint for a certain class of deep AD problems, it still leaves a more general question about the difference between supervised and unsupervised approaches to deep AD in settings where transfer learning is not appropriate. 
This may happen when there are very subtle semantic novelties \citep{vaze2022openset}, when one simply cannot use a pre-trained network; for example, when one must train a network from scratch due to architectural considerations (e.g., when one requires a smaller architecture due to hardware constraints), or when there are security concerns regarding the white-box nature of pre-trained networks \citep{samek2021explaining}.
Additionally, some AD techniques offer no obvious way to utilize a pre-trained network, like the recently introduced explainable one-class variant \citep{liznerski2021} or methods based on probabilistic models.

\subsection{On the robustness of HSC vs.~BCE} \label{sec:exp_robustness}
Our previous experiments (Section \ref{sec:exp_var_oe_size}) have shown that, though end-to-end BCE overall outperforms HSC, an unsupervised OE approach is more effective when only very few ($<32$) OE samples are available or when the OE data is not very informative.
This indicates a certain degree of robustness to the anomalous training samples for HSC. 
This robustness with regards to OE is likely inherited from the learning objectives of unsupervised and semi-supervised AD, which encourage the normal representations to be concentrated thereby avoiding the issue with skewed OE data (see Figure \ref{fig:toy_example}).
So, even on high-dimensional datasets, there seems to be a regime, albeit a very small one, where the intuition with skewed data in Figure \ref{fig:toy_example} \emph{does} hold.
\vspace{-0.5em}
%%%%%%%%%%%%%%%%%%%%%%%%%%%%%%%%%%%%%%%%%%%%%%%%%%%%%%%%%%%%%%%%%%%%%%%%%%%%%%%%
\begin{wraptable}[12]{r}{0.44\textwidth} 
  \vspace{-0.45em}
  \caption{Mean AUC detection performance in \% for the best and worst single OE samples on the CIFAR-10 AD benchmark with 80MTI as OE and on the ImageNet-10 AD benchmark with ImageNet-22K (without the 1K classes) as OE.}
  \label{tab:evolve}
  \vspace{0.5em}
  \centering
  \resizebox{0.4\textwidth}{!}{\begin{tabular}{ccccc} 
\toprule 
& \multicolumn{2}{c|}{CIFAR-10} & \multicolumn{2}{c}{ImageNet-10} \\ 
& HSC & \multicolumn{1}{c|}{BCE} & HSC & BCE \\ 
\midrule 
Best OE & 77.7 & \multicolumn{1}{c|}{69.9} & 79.3 & 75.5 \\ 
Worst OE & 43.3 & \multicolumn{1}{c|}{31.6} & 39.2 & 26.3 \\
\bottomrule 
\end{tabular} 
}
\end{wraptable}
%%%%%%%%%%%%%%%%%%%%%%%%%%%%%%%%%%%%%%%%%%%%%%%%%%%%%%%%%%%%%%%%%%%%%%%%%%%%%%%%

To demonstrate the robustness of HSC when one has few OE samples we investigate the extreme case where an OE dataset consists of only \emph{one} sample in more detail. 
How much can a single sample help or hinder the different approaches to deep AD?
This is analogous to the experiments in Section \ref{sec:exp_var_oe_size} with the number of OE samples fixed to one. 
To investigate robustness, we search for the OE sample that gives the worst test AUC for each class and report the average AUC over all classes. 
While it is desirable for a method to be unaffected by detrimental OE examples, one would still want an AD method to exploit beneficial OE examples. 
To measure this trade-off, we perform an analogous experiment where we search for the best performing OE sample and again report the average test AUC over all classes.

\begin{figure*}[ht]
    \begin{center}
        \small
        \begin{minipage}[t]{1mm}
            \rotatebox{90}{
                \scriptsize
                \begin{tabular}{p{9mm}p{11mm}p{7mm}} \;worst & \quad\; best & \;\;norm  \end{tabular}
            }
        \end{minipage}
        \subfigure[
            ``ship'' is normal 
        ]{
            \includegraphics[height=0.175\textheight]{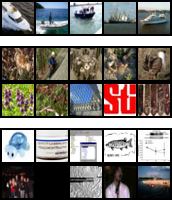}
        }
        \subfigure[
            ``cat'' is normal 
        ]{
            \includegraphics[height=0.175\textheight]{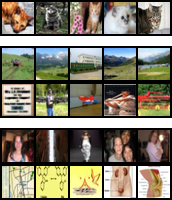}
        }
        \begin{minipage}[t]{1mm}
            \rotatebox{90}{
                \scriptsize
                \begin{tabular}{p{9mm}p{11mm}p{7mm}} \;worst & \quad\; best & \;\;norm  \end{tabular}
            }
        \end{minipage}
        \subfigure[
            ``airplane'' is normal 
        ]{
            \includegraphics[height=0.175\textheight]{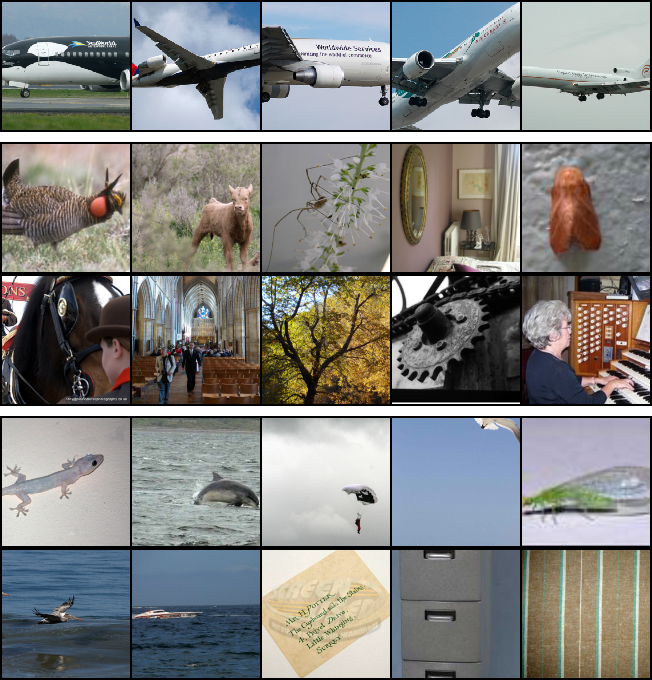}
        }
        \subfigure[
            ``dragonfly'' is normal 
        ]{
            \includegraphics[height=0.175\textheight]{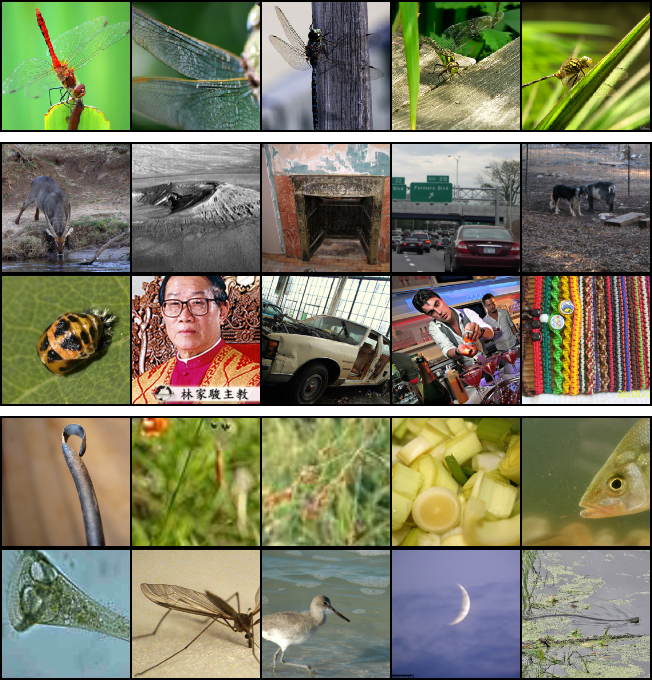}
        }
    \end{center}
    \caption{OE samples for CIFAR-10 with 80MTI as OE (a-b) and for ImageNet-10 with ImageNet-22k as OE (c-d). The first row shows normal samples, the next two rows show the best samples for HSC (top) and BCE (bottom), and the last two rows show the worst samples for HSC (top) and BCE (bottom).}
    \label{fig:ev_cifar_imagenet}
\end{figure*}

As it is computationally prohibitive to test every possible OE sample, we utilize an evolutionary algorithm to attempt to minimize or maximize the class AUC.
A detailed description of the algorithm can be found in Appendix \ref{appx:evol_algo}.
To ensure that this optimization scheme does not find poor local minima, we also evaluate the AUC of many randomly chosen OE samples (see Appendix \ref{appx:rand_search}). 
We find that the evolutionary algorithm almost always finds better optima.
We report our results for CIFAR-10 and for the first 10 classes of ImageNet-30 in Table \ref{tab:evolve}.
Figure \ref{fig:ev_cifar_imagenet} shows optimal OE examples on CIFAR-10 for the normal classes ``ship'' and ``cat,'' and on ImageNet-10 for ``airplane'' and ``dragonfly.'' 
%%%%%%%%%%%%%%%%%%%%%%%%%%%%%%%%%%%%%%%%%%%%%%%%%%%%%%%%%%%%%%%%%%%%%%%%%%%%%%%%
%%%%%%%%%%%%%%%%%%%%%%%%%%%%%%%%%%%%%%%%%%%%%%%%%%%%%%%%%%%%%%%%%%%%%%%%%%%%%%%%

\paragraph{Discussion}
On both datasets, we observe that HSC performs better than BCE when using both the best and the worst OE samples. 
Looking at the samples chosen for the optima, there also appears to be more consistency within a setting (class, dataset, best or worst) for HSC. 
For example on ``best'' with ``ship,'' HSC's images are all brownish outdoor photos, whereas the BCE samples vary from greenery to stylized text to an image containing mostly sky. 
This is likely due to the fact that HSC has, in some sense, an initial notion of anomalousness due to its unsupervised term.
For instance, the most useful OE samples are those not already contained in this notion of anomalousness, resulting in HSC having a stable region for selecting OE samples that yield the greatest improvement.
BCE lacks this notion so it can benefit from a large variety of OE samples.
With HSC, choosing one optimal sample achieves roughly the same performance as using 32 random ones.
Interestingly, it seems that near-distribution outliers are less useful as OE samples since samples with similar color patterns as the normal ones occur more often as the ``worst'' samples. 
Previous works have found that near-distribution OE samples are useful for OE \citep{lee2018training,goyal2020drocc}, however, our results suggest that this may not hold when little OE data is available.

\paragraph{HSC focuses on low frequency features}
To gain further insight into the difference between BCE and HSC, we extend the previous experiments to include frequency-domain corruptions. 
This sort of analysis has been insightful in other works on deep learning \citep{yin2019fourier}. 
We find that HSC is again more robust than BCE since it's generally less affected by the frequency corruptions and tends to focus on low frequency signals in the input. 
We defer the results and discussion to Appendix \ref{appx:exp_freq_anal} and also show some more examples for the best and worst single OE samples in Appendix \ref{appx:full_results}.

\section{Broader impact} \label{sec:ethics}
Anomaly detection methods on images may be deployed on tasks which have societal implications such as screening images or automated surveillance, and it is thus imperative that these tasks are done in a fair and transparent way. 
The use of OE is potentially harmful as there may be OE images biasing the model towards detecting certain entities as anomalous. 
Our paper aids in this since it demonstrates that no huge corpora are necessary, which enables a controlled selection of OE samples. 
Further, we have shown that HSC is more robust, and the fact that it chooses optimal OE samples that coincide with human intuition suggests that it is more interpretable than BCE, where the rationale for optimal OE samples is opaque. 
This makes HSC more suitable for critical applications. 
We trained some of our models with the 80MTI dataset, which is known to contain problematic data such as offensive labels, but were required to do so to be comparable with the previous line of research.

\section{Conclusion}
\label{sec:disc}
We presented surprising results that challenge common assumptions in AD. 
Neither does deep AD on natural images seem to require specialized AD methodologies nor huge amounts of Outlier Exposure (OE).
A standard classifier outperforms all end-to-end methods on the common one vs.~rest benchmark, for which it only needs 256 random OE samples on ImageNet and only one well-chosen OE sample for competitive performance. 
We showed some limitations of the few-OE approaches when applied to settings where the normal dataset is multimodal or the anomalies are more subtle. 
Using transfer learning, standard classifiers set a new state of the art on CIFAR-10 and, in a zero-shot setting \emph{without using any additional training data,} on ImageNet. 
Despite the overall strong performance of standard classifiers, we find that semi-supervised one-class methods are more robust to the choice of OE when only few OE examples are available.
Our results provide insights about deep AD that are useful for future research.

\section{Acknowledgements}
PL, BJF, and MK acknowledge support by the Carl-Zeiss Foundation, the German Research Foundation (DFG) awards KL 2698/2-1 and KL 2698/5-1, and the German Federal Ministry of Education and Research (BMBF) awards 01|S18051A, 03|B0770E, and 01|S21010C.
RV acknowledges support by the Federal Ministry of Education and Research (BMBF) for the Berlin Institute for the Foundations of Learning and Data (BIFOLD) (01IS18037A).
This work was supported in part by the German Ministry for Education and Research under Grant Nos. 01IS14013A-E, 01GQ1115, 01GQ0850, 01IS18025A, 031L0207D, and 01IS18037A. 
KRM was partly supported by the Institute of Information \& Communications Technology Planning \& Evaluation (IITP) grants funded by the Korea government(MSIT) (No. 2019-0-00079, Artificial Intelligence Graduate School Program, Korea University and No. 2022-0-00984, Development of Artificial Intelligence Technology for Personalized Plug-and-Play Explanation and Verification of Explanation).

\bibliography{references}

\begin{thebibliography}{83}
\providecommand{\natexlab}[1]{#1}
\providecommand{\url}[1]{\texttt{#1}}
\expandafter\ifx\csname urlstyle\endcsname\relax
  \providecommand{\doi}[1]{doi: #1}\else
  \providecommand{\doi}{doi: \begingroup \urlstyle{rm}\Url}\fi

\bibitem[Abati et~al.(2019)Abati, Porrello, Calderara, and Cucchiara]{abati19}
Davide Abati, Angelo Porrello, Simone Calderara, and Rita Cucchiara.
\newblock Latent space autoregression for novelty detection.
\newblock In \emph{{IEEE/CVF} Conference on Computer Vision and Pattern
  Recognition}, pp.\  481--490, 2019.

\bibitem[Ahmed \& Courville(2020)Ahmed and Courville]{ahmed2020}
Faruk Ahmed and Aaron Courville.
\newblock Detecting semantic anomalies.
\newblock In \emph{{AAAI} Conference on Artificial Intelligence}, pp.\
  3154--3162, 2020.

\bibitem[Akcay et~al.(2018)Akcay, Atapour-Abarghouei, and Breckon]{samet19}
Samet Akcay, Amir Atapour-Abarghouei, and Toby~P Breckon.
\newblock {GAN}omaly: Semi-supervised anomaly detection via adversarial
  training.
\newblock In \emph{Asian Conference on Computer Vision}, pp.\  622--637, 2018.

\bibitem[Bergman \& Hoshen(2020)Bergman and Hoshen]{bergman20}
Liron Bergman and Yedid Hoshen.
\newblock Classification-based anomaly detection for general data.
\newblock In \emph{International Conference on Learning Representations}, 2020.

\bibitem[Bergman et~al.(2020)Bergman, Cohen, and Hoshen]{bergman2020deep}
Liron Bergman, Niv Cohen, and Yedid Hoshen.
\newblock Deep nearest neighbor anomaly detection.
\newblock \emph{arXiv preprint arXiv:2002.10445}, 2020.

\bibitem[Bergmann et~al.(2019)Bergmann, Fauser, Sattlegger, and
  Steger]{bergmann2019mvtec}
Paul Bergmann, Michael Fauser, David Sattlegger, and Carsten Steger.
\newblock Mvtec ad--a comprehensive real-world dataset for unsupervised anomaly
  detection.
\newblock In \emph{Proceedings of the IEEE/CVF conference on computer vision
  and pattern recognition}, pp.\  9592--9600, 2019.

\bibitem[Blickle \& Thiele(1995)Blickle and Thiele]{blickle1995mathematical}
Tobias Blickle and Lothar Thiele.
\newblock A mathematical analysis of tournament selection.
\newblock In \emph{ICGA}, volume~95, pp.\  9--15. Citeseer, 1995.

\bibitem[Bommasani et~al.(2021)Bommasani, Hudson, Adeli, Altman, Arora, von
  Arx, Bernstein, Bohg, Bosselut, Brunskill,
  et~al.]{bommasani2021opportunities}
Rishi Bommasani, Drew~A Hudson, Ehsan Adeli, Russ Altman, Simran Arora, Sydney
  von Arx, Michael~S Bernstein, Jeannette Bohg, Antoine Bosselut, Emma
  Brunskill, et~al.
\newblock On the opportunities and risks of foundation models.
\newblock \emph{arXiv preprint arXiv:2108.07258}, 2021.

\bibitem[Chandola et~al.(2009)Chandola, Banerjee, and
  Kumar]{chandola2009anomaly}
Varun Chandola, Arindam Banerjee, and Vipin Kumar.
\newblock Anomaly detection: A survey.
\newblock \emph{ACM Computing Surveys}, 41\penalty0 (3):\penalty0 1--58, 2009.

\bibitem[Charbonnier et~al.(1997)Charbonnier, Blanc-F{\'e}raud, Aubert, and
  Barlaud]{charbonnier1997deterministic}
Pierre Charbonnier, Laure Blanc-F{\'e}raud, Gilles Aubert, and Michel Barlaud.
\newblock Deterministic edge-preserving regularization in computed imaging.
\newblock \emph{IEEE Transactions on Image Processing}, 6\penalty0
  (2):\penalty0 298--311, 1997.

\bibitem[Chen et~al.(2017)Chen, Sathe, Aggarwal, and Turaga]{chen2017outlier}
Jinghui Chen, Saket Sathe, Charu~C Aggarwal, and Deepak~S Turaga.
\newblock Outlier detection with autoencoder ensembles.
\newblock In \emph{{SIAM} International Conference on Data Mining}, pp.\
  90--98, 2017.

\bibitem[Chen et~al.(2020)Chen, Kornblith, Norouzi, and Hinton]{chen2020simple}
Ting Chen, Simon Kornblith, Mohammad Norouzi, and Geoffrey Hinton.
\newblock A simple framework for contrastive learning of visual
  representations.
\newblock In \emph{International conference on machine learning}, pp.\
  1597--1607. PMLR, 2020.

\bibitem[Cimpoi et~al.(2014)Cimpoi, Maji, Kokkinos, Mohamed, , and
  Vedaldi]{dtd}
M.~Cimpoi, S.~Maji, I.~Kokkinos, S.~Mohamed, , and A.~Vedaldi.
\newblock Describing textures in the wild.
\newblock In \emph{Proceedings of the {IEEE} Conf. on Computer Vision and
  Pattern Recognition ({CVPR})}, 2014.

\bibitem[Cohen et~al.(2017)Cohen, Afshar, Tapson, and Van~Schaik]{emnist}
Gregory Cohen, Saeed Afshar, Jonathan Tapson, and Andre Van~Schaik.
\newblock {EMNIST}: {Extending} {MNIST} to handwritten letters.
\newblock In \emph{International Joint Conference on Neural Networks}, pp.\
  2921--2926, 2017.

\bibitem[Deecke et~al.(2018)Deecke, Vandermeulen, Ruff, Mandt, and
  Kloft]{deecke2018image}
Lucas Deecke, Robert~A Vandermeulen, Lukas Ruff, Stephan Mandt, and Marius
  Kloft.
\newblock Image anomaly detection with generative adversarial networks.
\newblock In \emph{European Conference on Machine Learning and Principles and
  Practice of Knowledge Discovery in Databases}, pp.\  3--17, 2018.

\bibitem[Deecke et~al.(2021)Deecke, Ruff, Vandermeulen, and
  Bilen]{deecke2021transfer}
Lucas Deecke, Lukas Ruff, Robert~A Vandermeulen, and Hakan Bilen.
\newblock Transfer-based semantic anomaly detection.
\newblock In \emph{International Conference on Machine Learning}, pp.\
  2546--2558. PMLR, 2021.

\bibitem[Defard et~al.(2021)Defard, Setkov, Loesch, and
  Audigier]{defard2021padim}
Thomas Defard, Aleksandr Setkov, Angelique Loesch, and Romaric Audigier.
\newblock Padim: a patch distribution modeling framework for anomaly detection
  and localization.
\newblock In \emph{International Conference on Pattern Recognition}, pp.\
  475--489. Springer, 2021.

\bibitem[Deng et~al.(2009)Deng, Dong, Socher, Li, Li, and Fei-Fei]{imagenet}
Jia Deng, Wei Dong, Richard Socher, Li-Jia Li, Kai Li, and Li~Fei-Fei.
\newblock {ImageNet}: A large-scale hierarchical image database.
\newblock In \emph{{IEEE/CVF} Conference on Computer Vision and Pattern
  Recognition}, pp.\  248--255, 2009.

\bibitem[Deng(2012)]{mnist}
Li~Deng.
\newblock The mnist database of handwritten digit images for machine learning
  research.
\newblock \emph{IEEE Signal Processing Magazine}, 29\penalty0 (6):\penalty0
  141--142, 2012.

\bibitem[Emmott et~al.(2013)Emmott, Das, Dietterich, Fern, and Wong]{emmott13}
Andrew~F Emmott, Shubhomoy Das, Thomas Dietterich, Alan Fern, and Weng-Keen
  Wong.
\newblock Systematic construction of anomaly detection benchmarks from real
  data.
\newblock In \emph{KDD 2013 Workshop on Outlier Detection and Description},
  pp.\  16--21, 2013.

\bibitem[Erfani et~al.(2015)Erfani, Baktashmotlagh, Rajasegarar, Karunasekera,
  and Leckie]{erfani2015r1svm}
Sarah Erfani, Mahsa Baktashmotlagh, Sutharshan Rajasegarar, Shanika
  Karunasekera, and Chris Leckie.
\newblock {R1SVM}: {A} randomised nonlinear approach to large-acale anomaly
  detection.
\newblock In \emph{{AAAI} Conference on Artificial Intelligence}, pp.\
  432--438, 2015.

\bibitem[Erfani et~al.(2016)Erfani, Rajasegarar, Karunasekera, and
  Leckie]{erfani2016high}
Sarah~M Erfani, Sutharshan Rajasegarar, Shanika Karunasekera, and Christopher
  Leckie.
\newblock High-dimensional and large-scale anomaly detection using a linear
  one-class {SVM} with deep learning.
\newblock \emph{Pattern Recognition}, 58:\penalty0 121--134, 2016.

\bibitem[Fort et~al.(2021)Fort, Ren, and Lakshminarayanan]{fort2021exploring}
Stanislav Fort, Jie Ren, and Balaji Lakshminarayanan.
\newblock Exploring the limits of out-of-distribution detection.
\newblock \emph{Advances in Neural Information Processing Systems}, 34, 2021.

\bibitem[Fortin et~al.(2012)Fortin, De~Rainville, Gardner, Parizeau, and
  Gagn{\'e}]{fortin2012deap}
F{\'e}lix-Antoine Fortin, Fran{\c{c}}ois-Michel De~Rainville,
  Marc-Andr{\'e}~Gardner Gardner, Marc Parizeau, and Christian Gagn{\'e}.
\newblock Deap: Evolutionary algorithms made easy.
\newblock \emph{The Journal of Machine Learning Research}, 13\penalty0
  (1):\penalty0 2171--2175, 2012.

\bibitem[Gidaris et~al.(2018)Gidaris, Singh, and
  Komodakis]{gidaris2018unsupervised}
Spyros Gidaris, Praveer Singh, and Nikos Komodakis.
\newblock Unsupervised representation learning by predicting image rotations.
\newblock In \emph{International Conference on Learning Representations}, 2018.

\bibitem[Golan \& El-Yaniv(2018)Golan and El-Yaniv]{golan2018deep}
Izhak Golan and Ran El-Yaniv.
\newblock Deep anomaly detection using geometric transformations.
\newblock In \emph{Advances in Neural Information Processing Systems}, pp.\
  9758--9769, 2018.

\bibitem[G{\"o}rnitz et~al.(2013)G{\"o}rnitz, Kloft, Rieck, and
  Brefeld]{gornitz2013toward}
Nico G{\"o}rnitz, Marius Kloft, Konrad Rieck, and Ulf Brefeld.
\newblock Toward supervised anomaly detection.
\newblock \emph{Journal of Artificial Intelligence Research}, 46:\penalty0
  235--262, 2013.

\bibitem[Goyal et~al.(2020)Goyal, Raghunathan, Jain, Simhadri, and
  Jain]{goyal2020drocc}
Sachin Goyal, Aditi Raghunathan, Moksh Jain, Harsha~Vardhan Simhadri, and
  Prateek Jain.
\newblock Drocc: Deep robust one-class classification.
\newblock In \emph{International Conference on Machine Learning}, pp.\
  3711--3721. PMLR, 2020.

\bibitem[Gudovskiy et~al.(2022)Gudovskiy, Ishizaka, and
  Kozuka]{gudovskiy2022cflow}
Denis Gudovskiy, Shun Ishizaka, and Kazuki Kozuka.
\newblock Cflow-ad: Real-time unsupervised anomaly detection with localization
  via conditional normalizing flows.
\newblock In \emph{Proceedings of the IEEE/CVF Winter Conference on
  Applications of Computer Vision}, pp.\  98--107, 2022.

\bibitem[Hampel et~al.(2005)Hampel, Ronchetti, Rousseeuw, and
  Stahel]{hampel2005}
Frank~R Hampel, Elvezio~M Ronchetti, Peter~J Rousseeuw, and Werner~A Stahel.
\newblock \emph{Robust Statistics: The Approach Based on Influence Functions}.
\newblock John Wiley \& Sons, 2005.

\bibitem[Hawkins et~al.(2002)Hawkins, He, Williams, and
  Baxter]{hawkins2002outlier}
Simon Hawkins, Hongxing He, Graham Williams, and Rohan Baxter.
\newblock Outlier detection using replicator neural networks.
\newblock In \emph{International Conference on Data Warehousing and Knowledge
  Discovery}, volume 2454, pp.\  170--180, 2002.

\bibitem[Hendrycks et~al.(2019{\natexlab{a}})Hendrycks, Mazeika, and
  Dietterich]{hendrycks2019deep}
Dan Hendrycks, Mantas Mazeika, and Thomas~G Dietterich.
\newblock Deep anomaly detection with outlier exposure.
\newblock In \emph{International Conference on Learning Representations},
  2019{\natexlab{a}}.

\bibitem[Hendrycks et~al.(2019{\natexlab{b}})Hendrycks, Mazeika, Kadavath, and
  Song]{hendrycks2019using}
Dan Hendrycks, Mantas Mazeika, Saurav Kadavath, and Dawn Song.
\newblock Using self-supervised learning can improve model robustness and
  uncertainty.
\newblock In \emph{Advances in Neural Information Processing Systems}, pp.\
  15637--15648, 2019{\natexlab{b}}.

\bibitem[Hendrycks et~al.(2022)Hendrycks, Basart, Mazeika, Zou, Kwon,
  Mostajabi, Steinhardt, and Song]{hendrycks2022scaling}
Dan Hendrycks, Steven Basart, Mantas Mazeika, Andy Zou, Joseph Kwon,
  Mohammadreza Mostajabi, Jacob Steinhardt, and Dawn Song.
\newblock Scaling out-of-distribution detection for real-world settings.
\newblock In \emph{International Conference on Machine Learning}, pp.\
  8759--8773. {PMLR}, 2022.

\bibitem[Huang \& LeCun(2006)Huang and LeCun]{huang2006large}
Fu~Jie Huang and Yann LeCun.
\newblock Large-scale learning with {SVM} and convolutional nets for generic
  object categorization.
\newblock In \emph{{IEEE/CVF} Conference on Computer Vision and Pattern
  Recognition}, pp.\  284--291, 2006.

\bibitem[Huber \& Ronchetti(2009)Huber and Ronchetti]{huber2009}
Peter~J Huber and Elvezio~M Ronchetti.
\newblock \emph{Robust Statistics}.
\newblock John Wiley \& Sons, 2nd edition, 2009.

\bibitem[Ioffe \& Szegedy(2015)Ioffe and Szegedy]{ioffe2015}
Sergey Ioffe and Christian Szegedy.
\newblock Batch {N}ormalization: {A}ccelerating {D}eep {N}etwork {T}raining by
  {R}educing {I}nternal {C}ovariate {S}hift.
\newblock In \emph{International Conference on Machine Learning}, volume~37,
  pp.\  448--456, 2015.

\bibitem[Kim et~al.(2020)Kim, Shim, Lim, Jeon, Choi, Kim, and Yoon]{kim20}
Ki~Hyun Kim, Sangwoo Shim, Yongsub Lim, Jongseob Jeon, Jeongwoo Choi, Byungchan
  Kim, and Andre~S. Yoon.
\newblock {RaPP}: Novelty detection with reconstruction along projection
  pathway.
\newblock In \emph{International Conference on Learning Representations}, 2020.

\bibitem[Kingma \& Ba(2015)Kingma and Ba]{kingma2014}
Diederik~P Kingma and Jimmy Ba.
\newblock Adam: {A} method for stochastic optimization.
\newblock In \emph{International Conference on Learning Representations}, 2015.

\bibitem[Kriegel et~al.(2008)Kriegel, Schubert, and Zimek]{kriegel2008angle}
Hans-Peter Kriegel, Matthias Schubert, and Arthur Zimek.
\newblock Angle-based outlier detection in high-dimensional data.
\newblock In \emph{International Conference on Knowledge Discovery \& Data
  Mining}, pp.\  444--452, 2008.

\bibitem[Krizhevsky et~al.(2009)Krizhevsky, Hinton,
  et~al.]{krizhevsky2009learning}
Alex Krizhevsky, Geoffrey Hinton, et~al.
\newblock Learning multiple layers of features from tiny images.
\newblock Technical report, Citeseer, 2009.

\bibitem[LeCun et~al.(1990)LeCun, Boser, Denker, Henderson, Howard, Hubbard,
  and Jackel]{lecun1990handwritten}
Yann LeCun, Bernhard~E Boser, John~S Denker, Donnie Henderson, Richard~E
  Howard, Wayne~E Hubbard, and Lawrence~D Jackel.
\newblock Handwritten digit recognition with a back-propagation network.
\newblock In \emph{Advances in Neural Information Processing Systems}, pp.\
  396--404, 1990.

\bibitem[Lee et~al.(2018)Lee, Lee, Lee, and Shin]{lee2018training}
Kimin Lee, Honglak Lee, Kibok Lee, and Jinwoo Shin.
\newblock Training confidence-calibrated classifiers for detecting
  out-of-distribution samples.
\newblock In \emph{International Conference on Learning Representations}, 2018.

\bibitem[Li et~al.(2021)Li, Sohn, Yoon, and Pfister]{li2021cutpaste}
Chun-Liang Li, Kihyuk Sohn, Jinsung Yoon, and Tomas Pfister.
\newblock Cutpaste: Self-supervised learning for anomaly detection and
  localization.
\newblock In \emph{Proceedings of the IEEE/CVF Conference on Computer Vision
  and Pattern Recognition}, pp.\  9664--9674, 2021.

\bibitem[Liang et~al.(2018)Liang, Li, and Srikant]{liang2018enhancing}
Shiyu Liang, Yixuan Li, and R~Srikant.
\newblock Enhancing the reliability of out-of-distribution image detection in
  neural networks.
\newblock In \emph{International Conference on Learning Representations}, 2018.

\bibitem[Lin et~al.(2017)Lin, Goyal, Girshick, He, and Doll{\'a}r]{lin17}
Tsung-Yi Lin, Priya Goyal, Ross Girshick, Kaiming He, and Piotr Doll{\'a}r.
\newblock Focal loss for dense object detection.
\newblock In \emph{International Conference on Computer Vision}, pp.\
  2980--2988, 2017.

\bibitem[Liznerski et~al.(2021)Liznerski, Ruff, Vandermeulen, Franks, Kloft,
  and M{\"u}ller]{liznerski2021}
Philipp Liznerski, Lukas Ruff, Robert~A. Vandermeulen, Billy~Joe Franks, Marius
  Kloft, and Klaus-Robert M{\"u}ller.
\newblock Explainable deep one-class classification.
\newblock In \emph{International Conference on Learning Representations}, 2021.

\bibitem[Nalisnick et~al.(2019)Nalisnick, Matsukawa, Teh, Gorur, and
  Lakshminarayanan]{nalisnick2019}
Eric Nalisnick, Akihiro Matsukawa, Yee~Whye Teh, Dilan Gorur, and Balaji
  Lakshminarayanan.
\newblock Do deep generative models know what they don't know?
\newblock In \emph{International Conference on Learning Representations}, 2019.

\bibitem[Nguyen et~al.(2019)Nguyen, Lou, Klar, and Brox]{nguyen2019}
Duc~Tam Nguyen, Zhongyu Lou, Michael Klar, and Thomas Brox.
\newblock Anomaly detection with multiple-hypotheses predictions.
\newblock In \emph{International Conference on Machine Learning}, volume~97,
  pp.\  4800--4809, 2019.

\bibitem[Pang et~al.(2021)Pang, Shen, Cao, and Hengel]{pang2021}
Guansong Pang, Chunhua Shen, Longbing Cao, and Anton Van~Den Hengel.
\newblock Deep learning for anomaly detection: A review.
\newblock \emph{ACM Computing Surveys}, 54\penalty0 (2), 2021.

\bibitem[Perera et~al.(2019)Perera, Nallapati, and Xiang]{perera19}
Pramuditha Perera, Ramesh Nallapati, and Bing Xiang.
\newblock {OCGAN}: One-class novelty detection using {GAN}s with constrained
  latent representations.
\newblock In \emph{{IEEE/CVF} Conference on Computer Vision and Pattern
  Recognition}, pp.\  2898--2906, 2019.

\bibitem[Polonik(1995)]{polonik1995measuring}
Wolfgang Polonik.
\newblock Measuring mass concentrations and estimating density contour
  clusters-an excess mass approach.
\newblock \emph{The Annals of Statistics}, 23\penalty0 (3):\penalty0 855--881,
  1995.

\bibitem[Radford et~al.(2021)Radford, Kim, Hallacy, Ramesh, Goh, Agarwal,
  Sastry, Askell, Mishkin, Clark, Krueger, and Sutskever]{radford2021learning}
Alec Radford, Jong~Wook Kim, Chris Hallacy, Aditya Ramesh, Gabriel Goh,
  Sandhini Agarwal, Girish Sastry, Amanda Askell, Pamela Mishkin, Jack Clark,
  Gretchen Krueger, and Ilya Sutskever.
\newblock Learning transferable visual models from natural language
  supervision.
\newblock In \emph{International Conference on Machine Learning}, volume 139,
  pp.\  8748--8763, 2021.

\bibitem[Reiss et~al.(2021)Reiss, Cohen, Bergman, and Hoshen]{reiss2021panda}
Tal Reiss, Niv Cohen, Liron Bergman, and Yedid Hoshen.
\newblock Panda: Adapting pretrained features for anomaly detection and
  segmentation.
\newblock In \emph{Proceedings of the IEEE/CVF Conference on Computer Vision
  and Pattern Recognition}, pp.\  2806--2814, 2021.

\bibitem[Roth et~al.(2022)Roth, Pemula, Zepeda, Sch{\"o}lkopf, Brox, and
  Gehler]{roth2022towards}
Karsten Roth, Latha Pemula, Joaquin Zepeda, Bernhard Sch{\"o}lkopf, Thomas
  Brox, and Peter Gehler.
\newblock Towards total recall in industrial anomaly detection.
\newblock In \emph{Proceedings of the IEEE/CVF Conference on Computer Vision
  and Pattern Recognition}, pp.\  14318--14328, 2022.

\bibitem[Ruff et~al.(2018)Ruff, Vandermeulen, G{\"o}rnitz, Deecke, Siddiqui,
  Binder, M{\"u}ller, and Kloft]{ruff2018deep}
Lukas Ruff, Robert~A Vandermeulen, Nico G{\"o}rnitz, Lucas Deecke, Shoaib~A.
  Siddiqui, Alexander Binder, Emmanuel M{\"u}ller, and Marius Kloft.
\newblock Deep one-class classification.
\newblock In \emph{International Conference on Machine Learning}, volume~80,
  pp.\  4390--4399, 2018.

\bibitem[Ruff et~al.(2020)Ruff, Vandermeulen, G{\"o}rnitz, Binder, M{\"u}ller,
  M{\"u}ller, and Kloft]{ruff2020}
Lukas Ruff, Robert~A Vandermeulen, Nico G{\"o}rnitz, Alexander Binder, Emmanuel
  M{\"u}ller, Klaus-Robert M{\"u}ller, and Marius Kloft.
\newblock Deep semi-supervised anomaly detection.
\newblock In \emph{International Conference on Learning Representations}, 2020.

\bibitem[Ruff et~al.(2021{\natexlab{a}})Ruff, Kauffmann, Vandermeulen,
  Montavon, Samek, Kloft, Dietterich, and M{\"u}ller]{ruff2021}
Lukas Ruff, Jacob~R Kauffmann, Robert~A Vandermeulen, Gr{\'e}goire Montavon,
  Wojciech Samek, Marius Kloft, Thomas~G Dietterich, and Klaus-Robert
  M{\"u}ller.
\newblock A unifying review of deep and shallow anomaly detection.
\newblock \emph{Proceedings of the IEEE}, 109\penalty0 (5):\penalty0 756--795,
  2021{\natexlab{a}}.

\bibitem[Ruff et~al.(2021{\natexlab{b}})Ruff, Vandermeulen, Franks, M{\"u}ller,
  and Kloft]{ruff2020rethinking}
Lukas Ruff, Robert~A Vandermeulen, Billy~Joe Franks, Klaus-Robert M{\"u}ller,
  and Marius Kloft.
\newblock Rethinking assumptions in deep anomaly detection.
\newblock In \emph{ICML 2021 Workshop on Uncertainty \& Robustness in Deep
  Learning}, 2021{\natexlab{b}}.

\bibitem[Sakurada \& Yairi(2014)Sakurada and Yairi]{sakurada2014anomaly}
Mayu Sakurada and Takehisa Yairi.
\newblock Anomaly detection using autoencoders with nonlinear dimensionality
  reduction.
\newblock In \emph{2nd Workshop on Machine Learning for Sensory Data Analysis
  (MLSDA 2014)}, pp.\  4--11, 2014.

\bibitem[Samek et~al.(2021)Samek, Montavon, Lapuschkin, Anders, and
  M{\"u}ller]{samek2021explaining}
Wojciech Samek, Gr{\'e}goire Montavon, Sebastian Lapuschkin, Christopher~J
  Anders, and Klaus-Robert M{\"u}ller.
\newblock Explaining deep neural networks and beyond: A review of methods and
  applications.
\newblock \emph{Proceedings of the IEEE}, 109\penalty0 (3):\penalty0 247--278,
  2021.

\bibitem[Schlegl et~al.(2017)Schlegl, Seeb{\"o}ck, Waldstein, Schmidt-Erfurth,
  and Langs]{schlegl2017unsupervised}
Thomas Schlegl, Philipp Seeb{\"o}ck, Sebastian~M Waldstein, Ursula
  Schmidt-Erfurth, and Georg Langs.
\newblock Unsupervised anomaly detection with generative adversarial networks
  to guide marker discovery.
\newblock In \emph{International Conference on Information Processing in
  Medical Imaging}, pp.\  146--157, 2017.

\bibitem[Schlegl et~al.(2019)Schlegl, Seeb{\"o}ck, Waldstein, Langs, and
  Schmidt-Erfurth]{schlegl2019}
Thomas Schlegl, Philipp Seeb{\"o}ck, Sebastian~M Waldstein, Georg Langs, and
  Ursula Schmidt-Erfurth.
\newblock f-{AnoGAN}: Fast unsupervised anomaly detection with generative
  adversarial networks.
\newblock \emph{Medical Image Analysis}, 54:\penalty0 30--44, 2019.

\bibitem[Schl{\"u}ter et~al.(2021)Schl{\"u}ter, Tan, Hou, and
  Kainz]{schluter2021self}
Hannah~M Schl{\"u}ter, Jeremy Tan, Benjamin Hou, and Bernhard Kainz.
\newblock Self-supervised out-of-distribution detection and localization with
  natural synthetic anomalies (nsa).
\newblock \emph{arXiv preprint arXiv:2109.15222}, 2021.

\bibitem[Sch{\"o}lkopf \& Smola(2002)Sch{\"o}lkopf and Smola]{scholkopf2002}
Bernhard Sch{\"o}lkopf and Alex~J Smola.
\newblock \emph{Learning with Kernels}.
\newblock MIT press, 2002.

\bibitem[Sch{\"o}lkopf et~al.(2001)Sch{\"o}lkopf, Platt, Shawe-Taylor, Smola,
  and Williamson]{scholkopf2001}
Bernhard Sch{\"o}lkopf, John~C Platt, John Shawe-Taylor, Alex~J Smola, and
  Robert~C Williamson.
\newblock Estimating the support of a high-dimensional distribution.
\newblock \emph{Neural Computation}, 13\penalty0 (7):\penalty0 1443--1471,
  2001.

\bibitem[Sohn et~al.(2021)Sohn, Li, Yoon, Jin, and Pfister]{sohn2021}
Kihyuk Sohn, Chun-Liang Li, Jinsung Yoon, Minho Jin, and Tomas Pfister.
\newblock Learning and evaluating representations for deep one-class
  classification.
\newblock In \emph{International Conference on Learning Representations}, 2021.

\bibitem[Steinwart et~al.(2005)Steinwart, Hush, and
  Scovel]{steinwart2005classification}
Ingo Steinwart, Don Hush, and Clint Scovel.
\newblock A classification framework for anomaly detection.
\newblock \emph{Journal of Machine Learning Research}, 6\penalty0
  (Feb):\penalty0 211--232, 2005.

\bibitem[Tack et~al.(2020)Tack, Mo, Jeong, and Shin]{tack2020}
Jihoon Tack, Sangwoo Mo, Jongheon Jeong, and Jinwoo Shin.
\newblock {CSI}: Novelty detection via contrastive learning on distributionally
  shifted instances.
\newblock In \emph{Advances in Neural Information Processing Systems},
  volume~33, pp.\  11839--11852, 2020.

\bibitem[Tax(2001)]{tax2001}
David Martinus~Johannes Tax.
\newblock \emph{One-Class Classification}.
\newblock PhD thesis, Delft University of Technology, 2001.

\bibitem[Torralba et~al.(2008)Torralba, Fergus, and Freeman]{torralba200880}
Antonio Torralba, Rob Fergus, and William~T Freeman.
\newblock 80 million tiny images: {A} large data set for nonparametric object
  and scene recognition.
\newblock \emph{{IEEE} Transactions on Pattern Analysis and Machine
  Intelligence}, 30\penalty0 (11):\penalty0 1958--1970, 2008.

\bibitem[Tsybakov(1997)]{tsybakov1997}
Alexandre~B Tsybakov.
\newblock On nonparametric estimation of density level sets.
\newblock \emph{The Annals of Statistics}, 25\penalty0 (3):\penalty0 948--969,
  1997.

\bibitem[Vaze et~al.(2022)Vaze, Han, Vedaldi, and Zisserman]{vaze2022openset}
Sagar Vaze, Kai Han, Andrea Vedaldi, and Andrew Zisserman.
\newblock Open-set recognition: A good closed-set classifier is all you need.
\newblock In \emph{International Conference on Learning Representations}, 2022.

\bibitem[Vert \& Vert(2006)Vert and Vert]{vert2006}
R{\'e}gis Vert and Jean-Philippe Vert.
\newblock Consistency and convergence rates of one-class {SVM}s and related
  algorithms.
\newblock \emph{Journal of Machine Learning Research}, 7\penalty0
  (May):\penalty0 817--854, 2006.

\bibitem[Wah et~al.(2011)Wah, Branson, Welinder, Perona, and Belongie]{cub}
C.~Wah, S.~Branson, P.~Welinder, P.~Perona, and S.~Belongie.
\newblock The caltech-ucsd birds-200-2011 dataset.
\newblock Technical Report CNS-TR-2011-001, California Institute of Technology,
  2011.

\bibitem[Wang et~al.(2019{\natexlab{a}})Wang, Sun, and Yu]{wang19}
Jingjing Wang, Sun Sun, and Yaoliang Yu.
\newblock Multivariate triangular quantile maps for novelty detection.
\newblock In \emph{Advances in Neural Information Processing Systems}, pp.\
  5061--5072, 2019{\natexlab{a}}.

\bibitem[Wang et~al.(2019{\natexlab{b}})Wang, Zeng, Liu, Zhu, Yin, Xu, and
  Kloft]{mariusnips19}
Siqi Wang, Yijie Zeng, Xinwang Liu, En~Zhu, Jianping Yin, Chuanfu Xu, and
  Marius Kloft.
\newblock Effective end-to-end unsupervised outlier detection via inlier
  priority of discriminative network.
\newblock In \emph{Advances in Neural Information Processing Systems}, pp.\
  5960--5973, 2019{\natexlab{b}}.

\bibitem[Xiao et~al.(2017)Xiao, Rasul, and Vollgraf]{fmnist}
Han Xiao, Kashif Rasul, and Roland Vollgraf.
\newblock Fashion-mnist: a novel image dataset for benchmarking machine
  learning algorithms.
\newblock 2017.

\bibitem[Yin et~al.(2019)Yin, Lopes, Shlens, Cubuk, and Gilmer]{yin2019fourier}
Dong Yin, Raphael~Gontijo Lopes, Jon Shlens, Ekin~Dogus Cubuk, and Justin
  Gilmer.
\newblock A fourier perspective on model robustness in computer vision.
\newblock In \emph{Advances in Neural Information Processing Systems}, pp.\
  13255--13265, 2019.

\bibitem[Yu \& Gen(2010)Yu and Gen]{yu2010introduction}
Xinjie Yu and Mitsuo Gen.
\newblock \emph{Introduction to evolutionary algorithms}.
\newblock Springer Science \& Business Media, 2010.

\bibitem[Zagoruyko \& Komodakis(2016)Zagoruyko and
  Komodakis]{zagoruyko2016wide}
Sergey Zagoruyko and Nikos Komodakis.
\newblock Wide residual networks.
\newblock In \emph{British Machine Vision Conference}, 2016.

\bibitem[Zenati et~al.(2018)Zenati, Romain, Foo, Lecouat, and
  Chandrasekhar]{zenati2018efficient}
Houssam Zenati, Manon Romain, Chuan-Sheng Foo, Bruno Lecouat, and Vijay
  Chandrasekhar.
\newblock Adversarially learned anomaly detection.
\newblock In \emph{{IEEE} International Conference on Data Mining}, pp.\
  727--736, 2018.

\bibitem[Zhou \& Paffenroth(2017)Zhou and Paffenroth]{zhou2017}
Chong Zhou and Randy~C Paffenroth.
\newblock Anomaly detection with robust deep autoencoders.
\newblock In \emph{International Conference on Knowledge Discovery \& Data
  Mining}, pp.\  665--674, 2017.

\end{thebibliography}
\bibliographystyle{stylefiles/tmlr}

\clearpage
\appendix

% --- don't center figures on otherwise empty pages
\makeatletter
\setlength{\@fptop}{0pt}
\makeatother

\section{Diversity of the Outlier Exposure data}
\label{appx:oe_diversity}
Here we evaluate how data diversity influences detection performance for unsupervised and supervised OE, again comparing HSC to BCE.
For this purpose, instead of 80MTI, we now use CIFAR-100 as OE varying the number of anomaly classes available for the CIFAR-10 benchmark. 
The OE data is varied by choosing $k$ classes at random for each random seed and using the union of these classes as the OE dataset.

%%%%%%%%%%%%%%%%%%%%%%%%%%%%%%%%%%%%%%%%%%%%%%%%%%%%%%%%%%%%%%%%%%%%%%%%%%%%%%%%
\begin{figure}[htb]
  \centering \small
  \includegraphics[width=0.495\columnwidth]{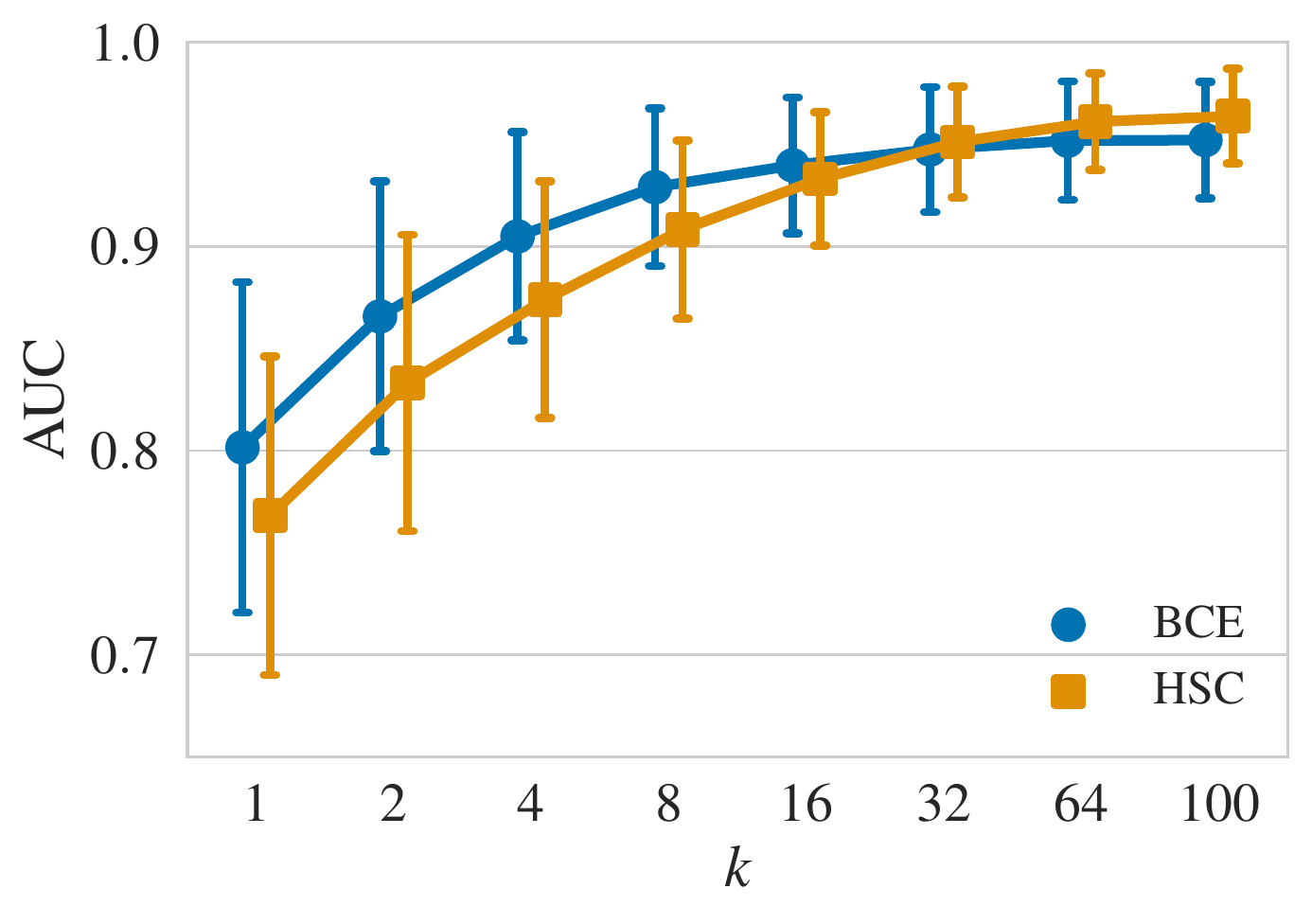}
  \caption{Mean AUC detection performance in \% (over 10 classes with 10 seeds per class) on the CIFAR-10 with CIFAR-100 OE one vs.~rest benchmark when varying the number of $k$ classes that comprise the OE dataset.}
  \label{fig:oe_diversity}
\end{figure}
%%%%%%%%%%%%%%%%%%%%%%%%%%%%%%%%%%%%%%%%%%%%%%%%%%%%%%%%%%%%%%%%%%%%%%%%%%%%%%%%

The results are presented in Figure \ref{fig:oe_diversity}. 
As expected, the performance increases with the diversity of the OE dataset.
Interestingly, drawing OE samples from just $k=1$ class, i.e.~binary classification between the normal class and a single OE class (which is not present as an anomaly class at test time!) already yields good detection performance on the CIFAR-10 benchmark. For example, training a standard classification network to discern between automobiles and beavers performs competitively as an automobile anomaly detector, even when no beavers are present as anomalies during test time.

Compared to Section \ref{sec:exp_var_oe_size} in the main paper, where we see a transition to BCE outperforming HSC at $2^3 = 8$ OE samples, we here see that with even one OE class we have passed this transition: there are already enough samples in a single class that BCE outperforms HSC. The takeaway is that OE sample diversity is not as important as one may expect, simply having many OE smaples suffices to enter the regime where BCE outperforms HSC. Again, with many samples, BCE and HSC's performances are comparable.

\section{HSC focuses on low frequency features}  \label{appx:exp_freq_anal}
\begin{figure*}[htb]
    \centering \small
    \begin{minipage}[t]{1mm}
        \rotatebox{90}{
            \scriptsize
            \begin{tabular}{p{9mm}p{11mm}p{6mm}} \;worst & \quad\; best & \;norm  \end{tabular}
        }
    \end{minipage}
    \subfigure[
        ``ship'' is normal 
    ]{
        \includegraphics[height=0.17\textheight]{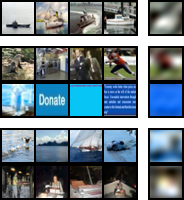}
    }
    \subfigure[
        ``cat'' is normal 
    ]{
        \includegraphics[height=0.17\textheight]{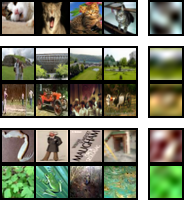}
    }
    \begin{minipage}[t]{1mm}
        \rotatebox{90}{
            \scriptsize
            \begin{tabular}{p{9mm}p{11mm}p{6mm}} \;worst & \quad\; best & \;norm  \end{tabular}
        }
    \end{minipage}
    \subfigure[
        ``airplane'' is normal 
    ]{
        \includegraphics[height=0.17\textheight]{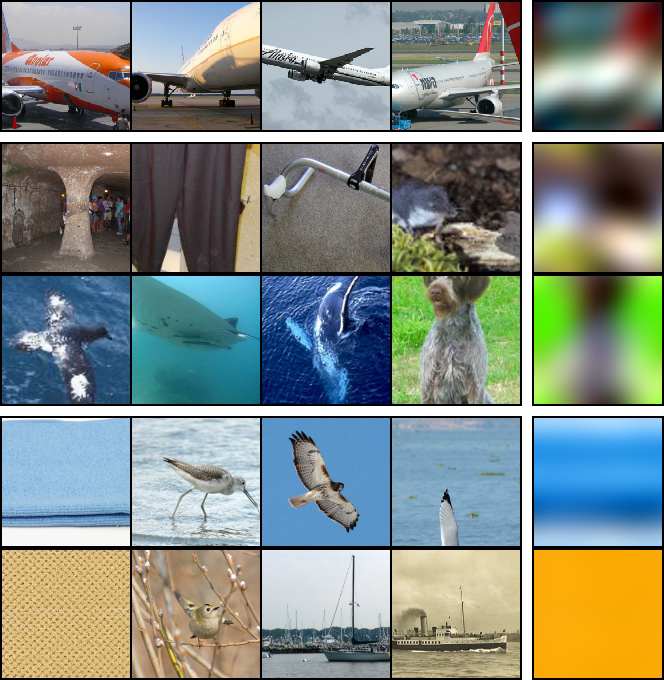}
    }
    \subfigure[
        ``dragonfly'' is normal 
    ]{
        \includegraphics[height=0.17\textheight]{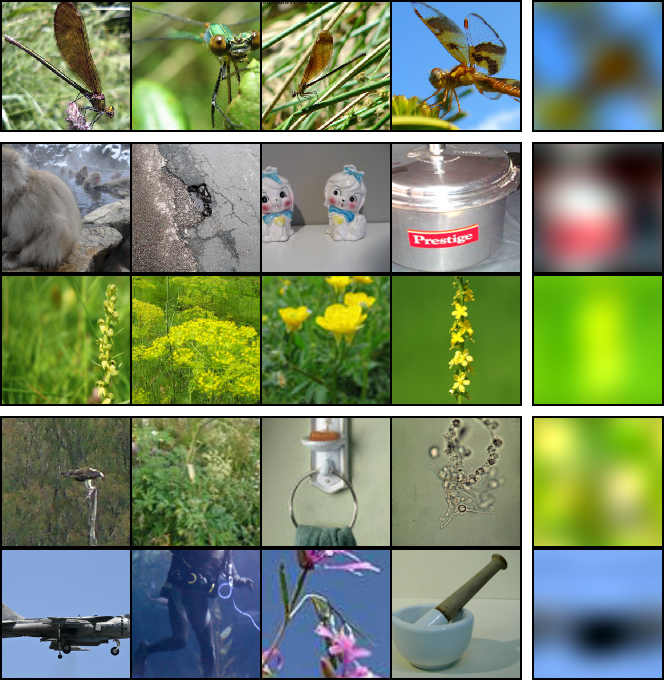}
    }
    \begin{minipage}[t]{1mm}
        \rotatebox{90}{
            \scriptsize
            \begin{tabular}{p{9mm}p{11mm}p{6mm}} \;worst & \quad\; best & \;norm  \end{tabular}
        }
    \end{minipage}
    \subfigure[
        ``ship'' is normal 
    ]{
        \includegraphics[height=0.17\textheight]{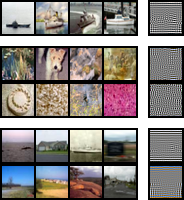}
    }
    \subfigure[
        ``cat'' is normal 
    ]{
        \includegraphics[height=0.17\textheight]{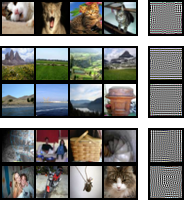}
    }
    \begin{minipage}[t]{1mm}
        \rotatebox{90}{
            \scriptsize
            \begin{tabular}{p{9mm}p{11mm}p{6mm}} \;worst & \quad\; best & \;norm  \end{tabular}
        }
    \end{minipage}
    \subfigure[
        ``airplane'' is normal 
    ]{
        \includegraphics[height=0.17\textheight]{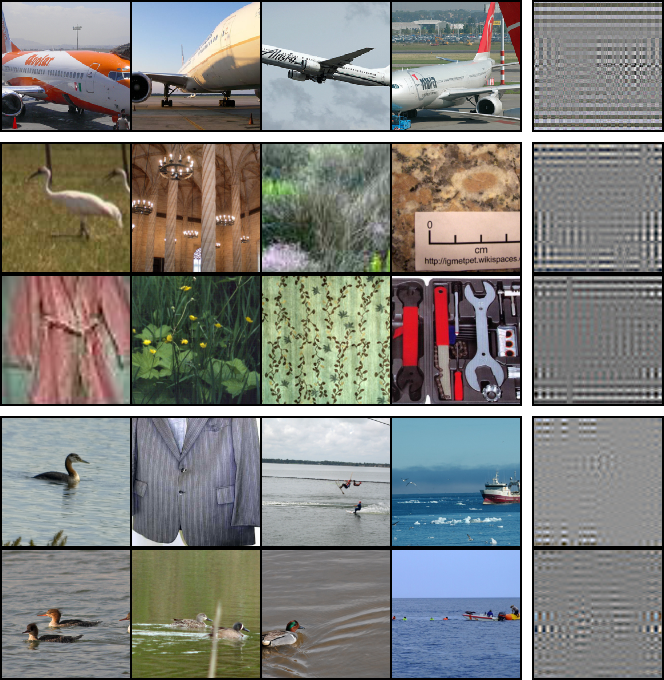}
    }
    \subfigure[
        ``dragonfly'' is normal 
    ]{
        \includegraphics[height=0.17\textheight]{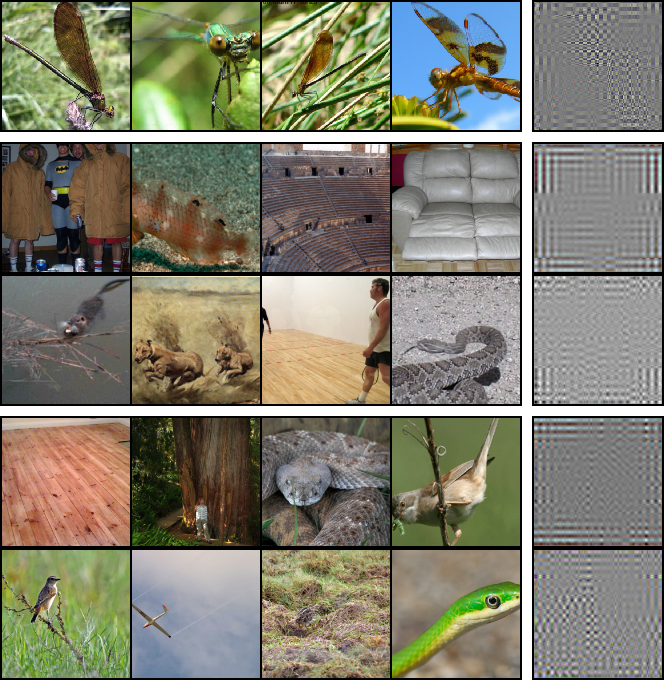}
    }
    \caption{OE samples for low-pass-filtered (a-d) and high-pass-filtered (e-h) versions of CIFAR-10 with 80MTI as OE (a,b,e,f) and ImageNet-10 with ImageNet-22k as OE (c,d,g,h). In each figure, the first row shows normal samples, the next two rows the best samples for HSC (top) and BCE (bottom), and the last two rows the worst samples for HSC (top) and BCE (bottom). The last column shows the filtered version of the images, which is what the network sees during training and testing. }
    \label{fig:ev_cifar_imagenet_lpf_hpf}
\end{figure*}

To gain further insight into the difference between BCE and HSC and why the best (and worst) found OE samples are quite different for these methods, we repeat the experiment from Section \ref{sec:exp_robustness} in the main paper with frequency-domain corruptions. 
%This sort of analysis has been insightful in other works on deep learning \citep{yin2019fourier}. 
%and the cognitive sciences \citep{marr2010vision} conjecture that human vision considers multiple levels of detail in natural images. 
That is, we low-pass-filter (LPF) or high-pass-filter (HPF) the \emph{entire} dataset (training, testing, and OE) and then proceed exactly as we did in Section \ref{sec:exp_robustness}. 
An LPF removes all higher frequencies and preserves only global information such as a scenery's color. This roughly corresponds to blurring images.
An HPF removes all lower frequencies and roughly corresponds to edge detection.
The AUC scores for the CIFAR-10 and ImageNet-10 one vs.~rest benchmarks are in Table \ref{tab:evolve_lpf_hpf}.
Figure \ref{fig:ev_cifar_imagenet_lpf_hpf} contains examples of the best and worst OE samples and also shows filtered examples. 
We provide details on the filter implementation in Appendix \ref{appx:freq_sens_anal}, where we evaluate the general performance of BCE and HSC for varying filter magnitudes. 

HSC seems to be more robust than BCE since it's generally less affected by the frequency corruptions. 
Focusing on CIFAR-10, we see that for HSC frequency corruption makes little change to performance except for the Best OE HPF experiment, where HSC's performance is significantly lower than in the LPF variant, causing HSC to behave similarly to BCE. 
This may have some implication that when useful signal is contained in the OE sample, corresponding to the ``Best OE'' experiments, it is concentrated in the low frequency spectrum. 
It seems as though BCE is not capable of exploiting this data. 
Interestingly it was also found in \citet{yin2019fourier} that low frequency features tend do be more robust (``low frequency bias results in improved robustness to corruptions'').
On ImageNet we observe that HSC again performs better on the LPF experiments.
%%%%%%%%%%%%%%%%%%%%%%%%%%%%%%%%%%%%%%%%%%%%%%%%%%%%%%%%%%%%%%%%%%%%%%%%%%%%%%%%
\begin{table}[h]
  \caption{Mean AUC detection performance in \% for the best and worst single OE samples on CIFAR-10 using 80MTI as OE and on ImageNet-10 using ImageNet-22K (with the 1K classes removed) as OE. All images have been either high-pass-filtered (HPF) or low-pass-filtered (LPF) both during training and testing. Arrows indicate the change compared to Table \ref{tab:evolve}.}
  \label{tab:evolve_lpf_hpf}
  \begin{center}
    \small
    % \begin{tabular}{ccccc} 
% \toprule 
% & \multicolumn{2}{c|}{CIFAR-10} & \multicolumn{2}{c}{ImageNet} \\ 
% & HSC & \multicolumn{1}{c|}{BCE} & HSC & BCE \\ 
% \midrule 
% Best OE LPF & 77.5 & \multicolumn{1}{c|}{68.5} & 77.1 & 73.8 \\ 
% Worst OE LPF & 44.1 & \multicolumn{1}{c|}{31.1} & 44.6 & 26.1 \\
% Best OE HPF & 68.8 & \multicolumn{1}{c|}{66.4} & 75.0 & 77.3 \\ 
% Worst OE HPF & 43.6 & \multicolumn{1}{c|}{38.0} & 44.1 & 27.9 \\
% \bottomrule 
% \end{tabular} 
\begin{tabular}{ccccc} 
\toprule 
& \multicolumn{2}{c|}{CIFAR-10} & \multicolumn{2}{c}{ImageNet} \\ 
& HSC & \multicolumn{1}{c|}{BCE} & HSC & BCE \\ 
\midrule 
Best OE LPF & 77.5$\rightarrow$ & \multicolumn{1}{c|}{68.5$\rightarrow$} & 77.2$\searrow$ & 73.8$\rightarrow$ \\ 
Worst OE LPF & 44.1$\rightarrow$ & \multicolumn{1}{c|}{31.1$\rightarrow$} & 44.6$\; \; \uparrow$ & 26.1$\rightarrow$ \\
Best OE HPF & 68.8$\; \; \downarrow$& \multicolumn{1}{c|}{66.4$\searrow$} & 75.0$\searrow$ & 77.3$\rightarrow$ \\ 
Worst OE HPF & 43.6$\rightarrow$ & \multicolumn{1}{c|}{38.0$\; \; \uparrow$} & 44.1$\; \; \uparrow$ & 27.9$\rightarrow$ \\
\bottomrule 
\end{tabular} 

% -0.2, -1.4, -2.2, -1.7
% +0.8, -0.5, +5.4, -0.2
% -8.9, -3.5, -4.3, +1.8
% +0.3, +6.4, +4.9, +1.6

  \end{center}
\end{table}
%%%%%%%%%%%%%%%%%%%%%%%%%%%%%%%%%%%%%%%%%%%%%%%%%%%%%%%%%%%%%%%%%%%%%%%%%%%%%%%%

%%%%%%%%%%%%%%%%%%%%%%%%%%%%%%%%%%%%%%%%%%%%%%%%%%%%%%%%%%%%%%%%%%%%%%%%%%%%%%%%
%%%%%%%%%%%%%%%%%%%%%%%%%%%%%%%%%%%%%%%%%%%%%%%%%%%%%%%%%%%%%%%%%%%%%%%%%%%%%%%%

\section{Frequency sensitivity analysis} \label{appx:freq_sens_anal}
\begin{figure*}[ht]
    \centering \small
    \subfigure[
        HSC
    ]{
        \includegraphics[width=0.48\textwidth]{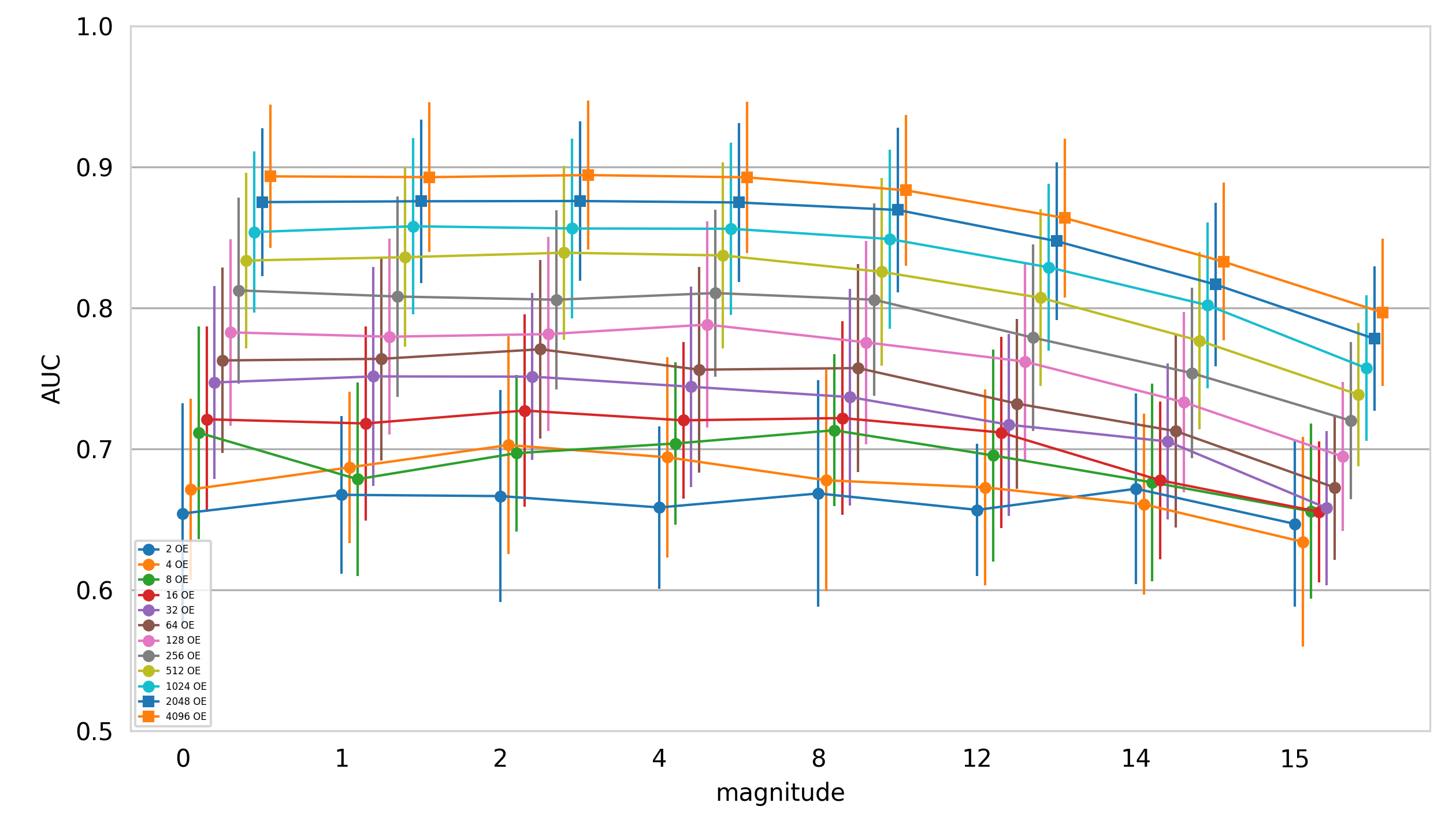}
    }
    \subfigure[
        BCE
    ]{
        \includegraphics[width=0.48\textwidth]{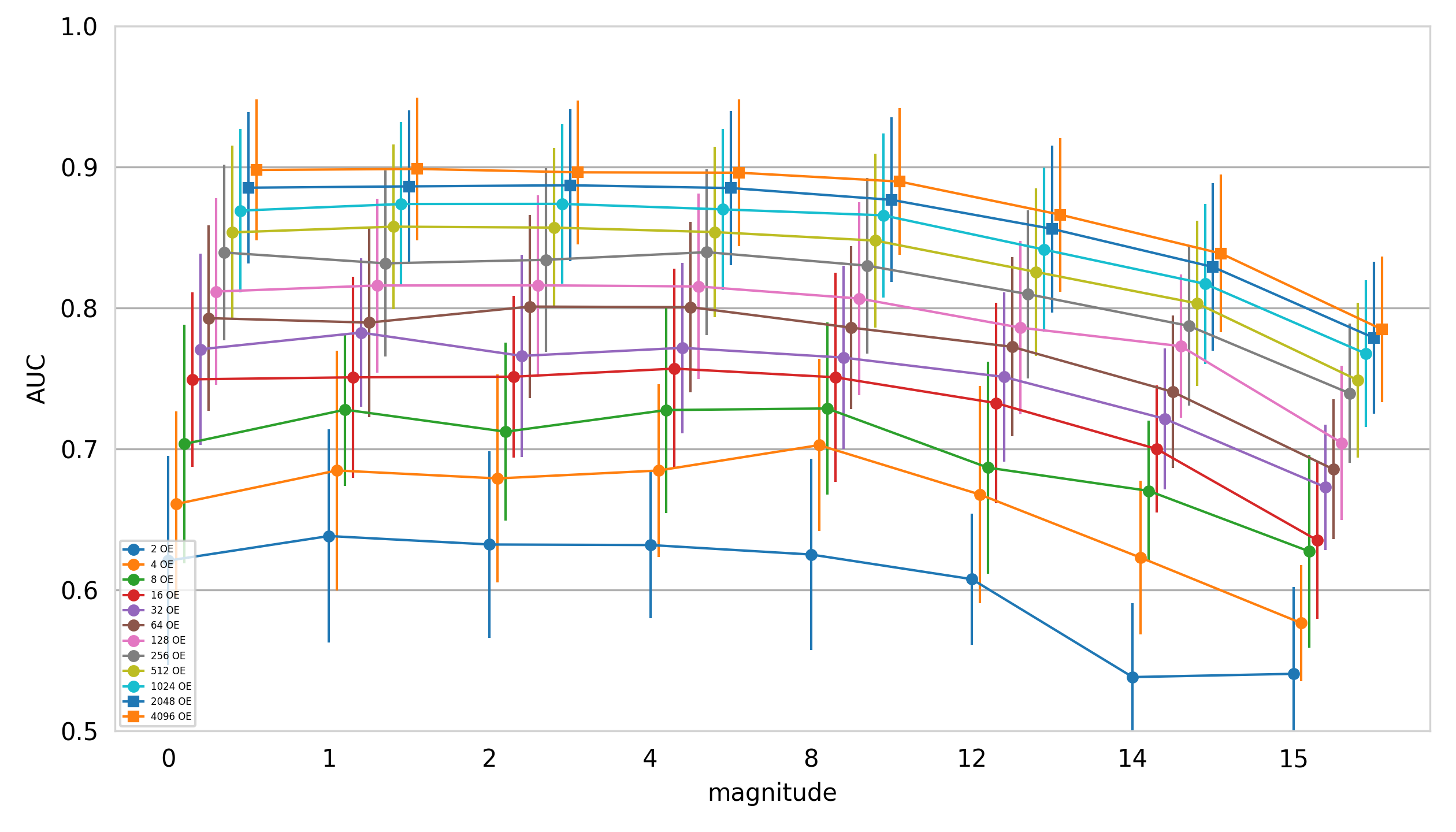}
    }
    \subfigure{
        \hspace{0.040\textwidth} 
        \includegraphics[width=0.415\textwidth]{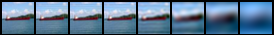}
        \hspace{0.010\textwidth}
    }
    \subfigure{
        \hspace{0.040\textwidth} 
        \includegraphics[width=0.415\textwidth]{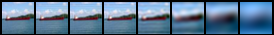}
        \hspace{0.010\textwidth}
    }
    \caption{LPF CIFAR-10 with 80MTI OE AD benchmark. }
    \label{fig:multiscale_cifar_lpf}
\end{figure*}

\begin{figure*}[ht]
    \centering \small
    \subfigure[
        HSC
    ]{
        \includegraphics[width=0.48\textwidth]{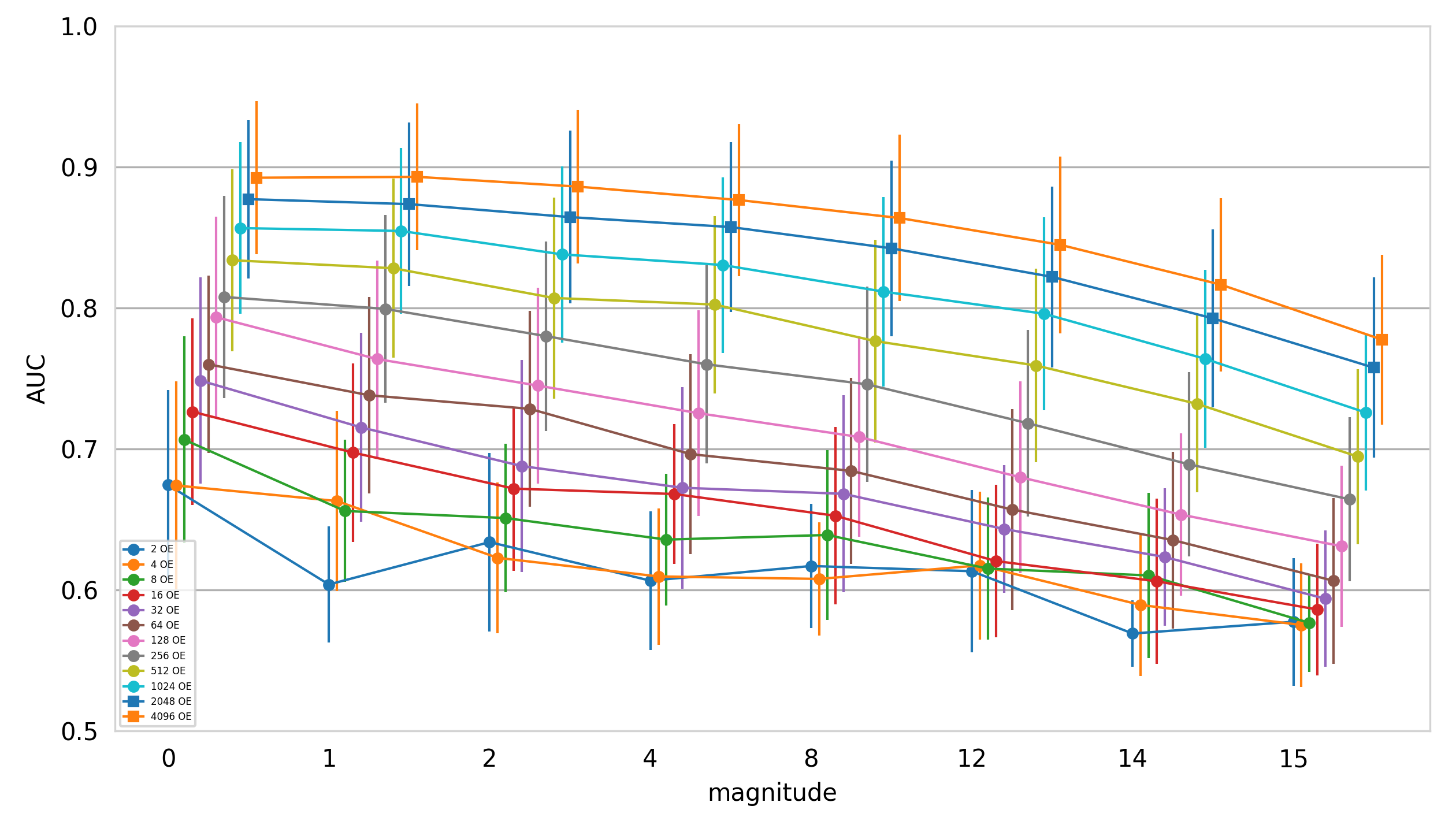}
    }
    \subfigure[
        BCE
    ]{
        \includegraphics[width=0.48\textwidth]{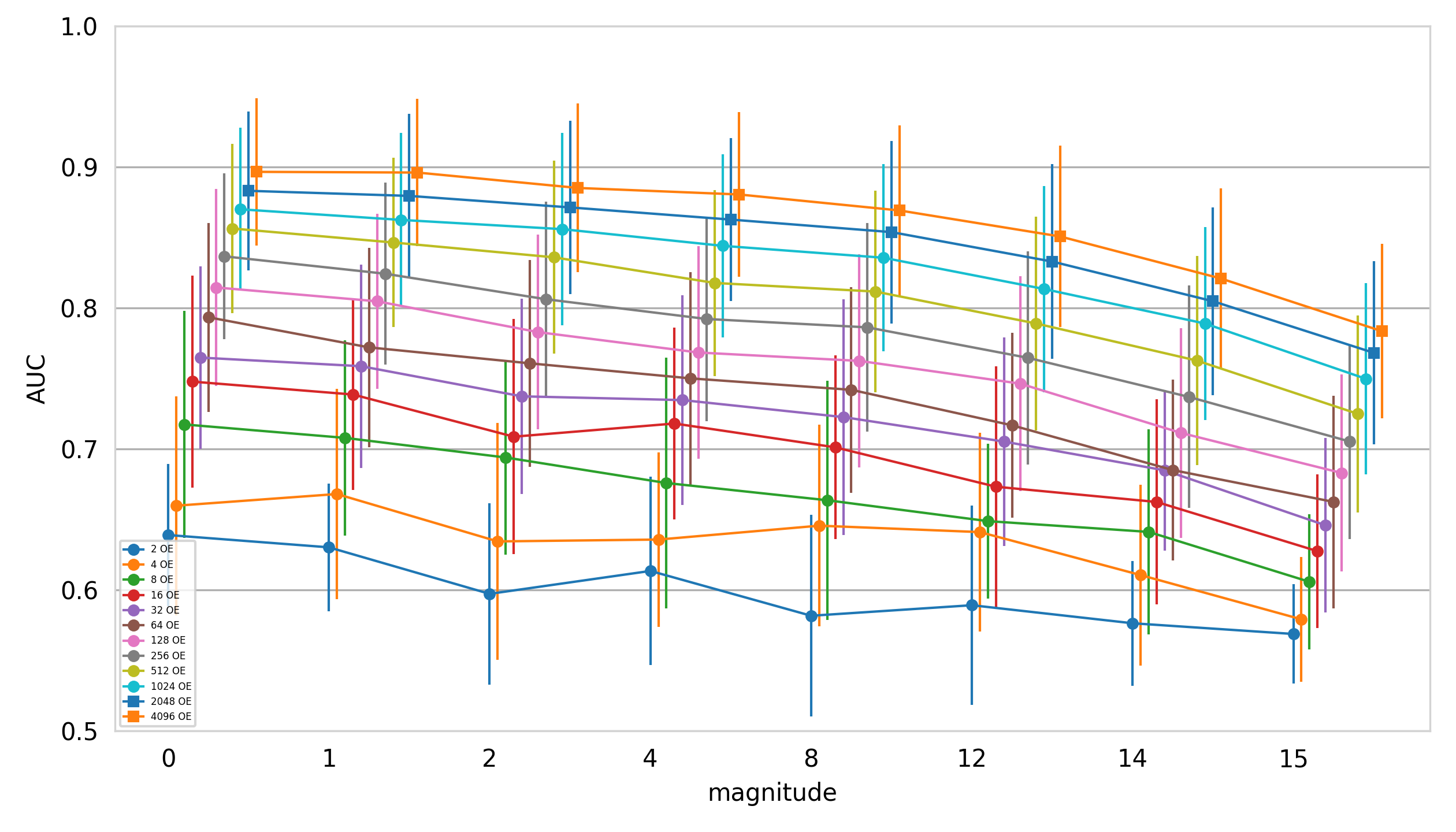}
    }
    \subfigure{
        \hspace{0.040\textwidth} 
        \includegraphics[width=0.415\textwidth]{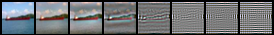}
        \hspace{0.010\textwidth}
    }
    \subfigure{
        \hspace{0.040\textwidth} 
        \includegraphics[width=0.415\textwidth]{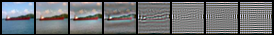}
        \hspace{0.010\textwidth}
    }
    \caption{
    HPF CIFAR-10 with 80MTI OE AD benchmark. 
    }
    \label{fig:multiscale_cifar_hpf}
\end{figure*}

\begin{figure*}[ht]
    \centering \small
    \subfigure[
        HSC
    ]{
        \includegraphics[width=0.48\textwidth]{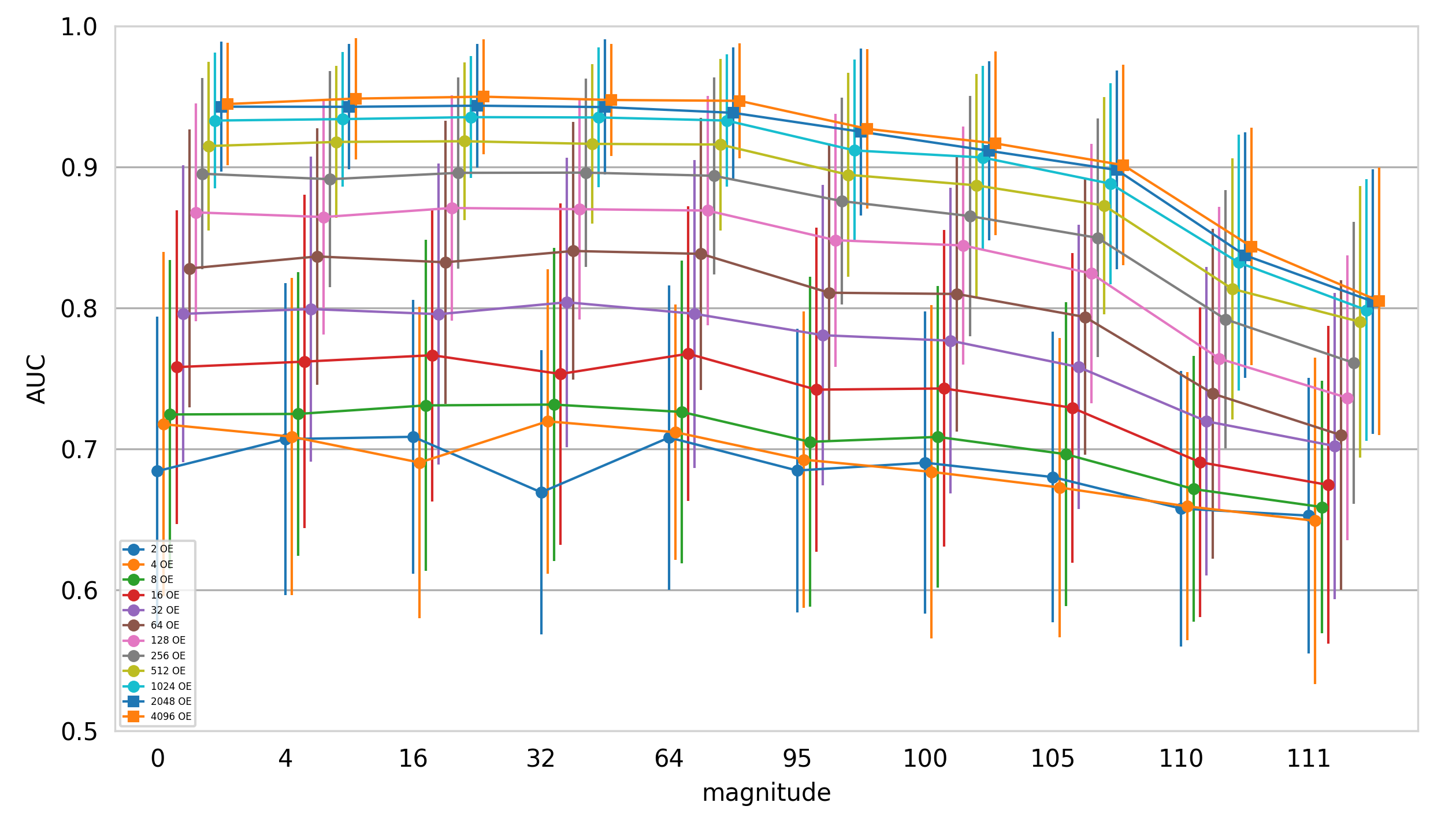}
    }
    \subfigure[
        BCE
    ]{
        \includegraphics[width=0.48\textwidth]{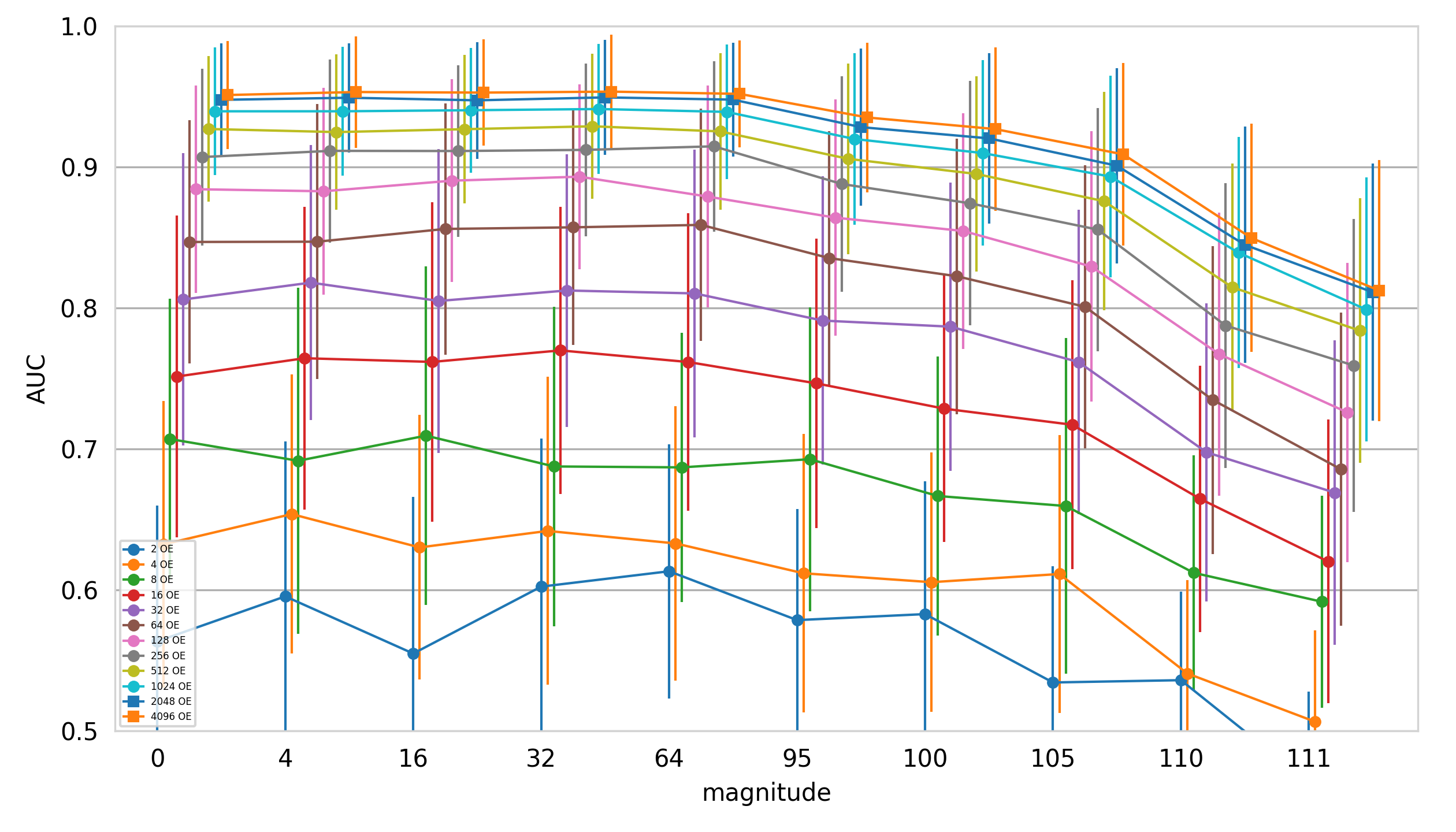}
    }
    \subfigure{
        \hspace{0.040\textwidth} 
        \includegraphics[width=0.415\textwidth]{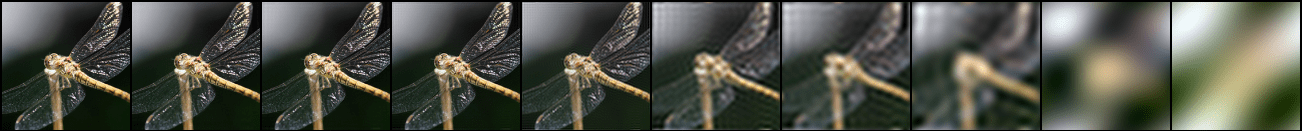}
        \hspace{0.010\textwidth}
    }
    \subfigure{
        \hspace{0.040\textwidth} 
        \includegraphics[width=0.415\textwidth]{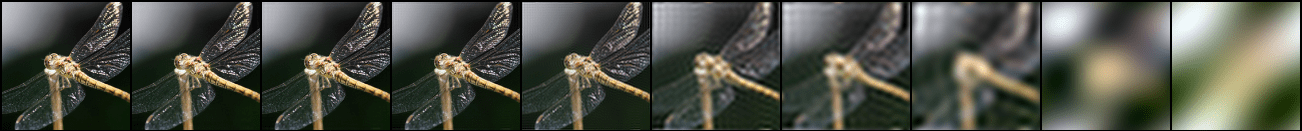}
        \hspace{0.010\textwidth}
    }
    \caption{
    LPF ImageNet-30 with ImageNet-22k (with ImageNet-30 removed) OE AD benchmark. 
    }
    \label{fig:multiscale_imagenet_lpf}
\end{figure*}

\begin{figure*}[ht]
    \centering \small
    \subfigure[
        HSC
    ]{
        \includegraphics[width=0.48\textwidth]{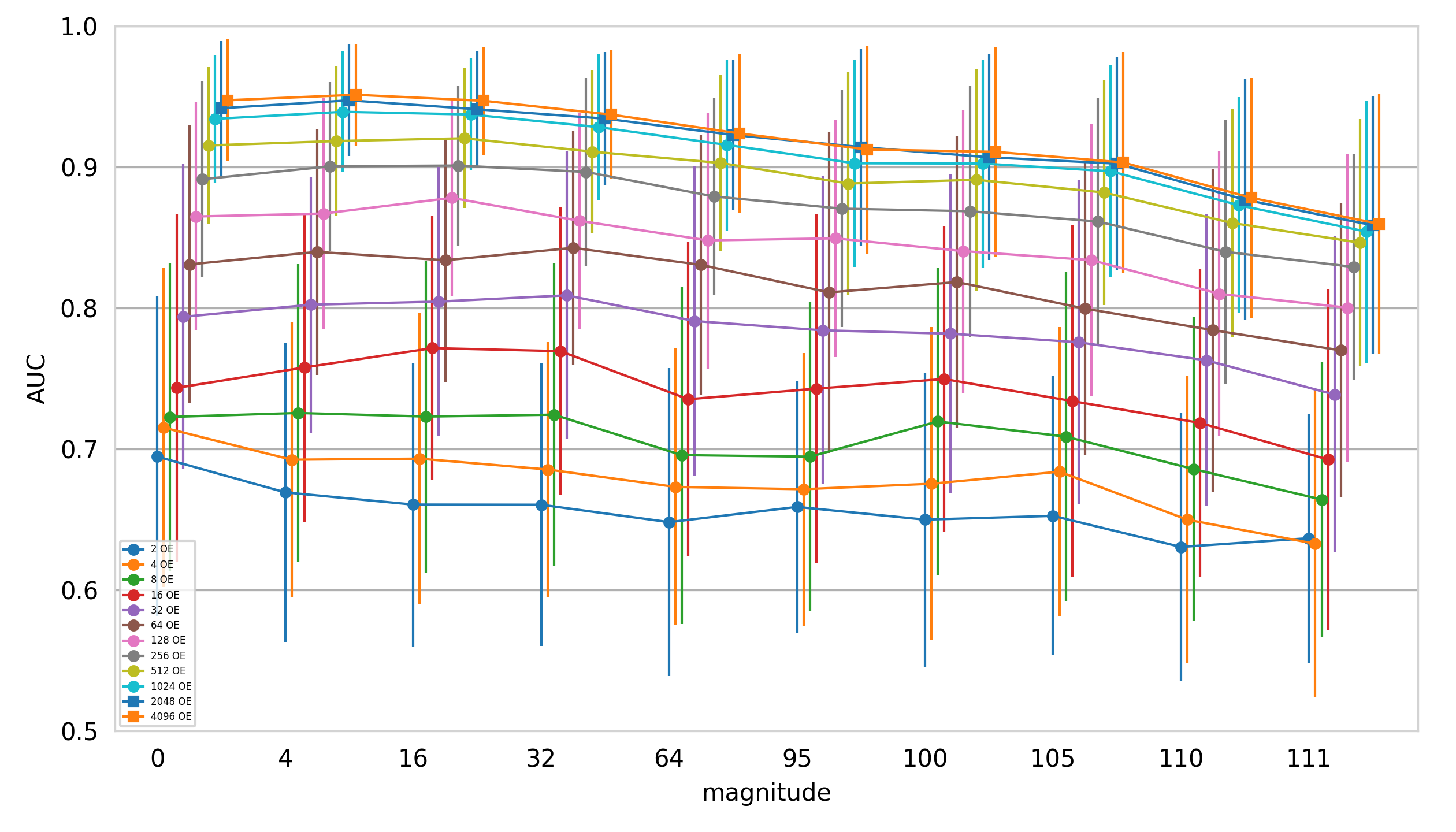}
    }
    \subfigure[
        BCE
    ]{
        \includegraphics[width=0.48\textwidth]{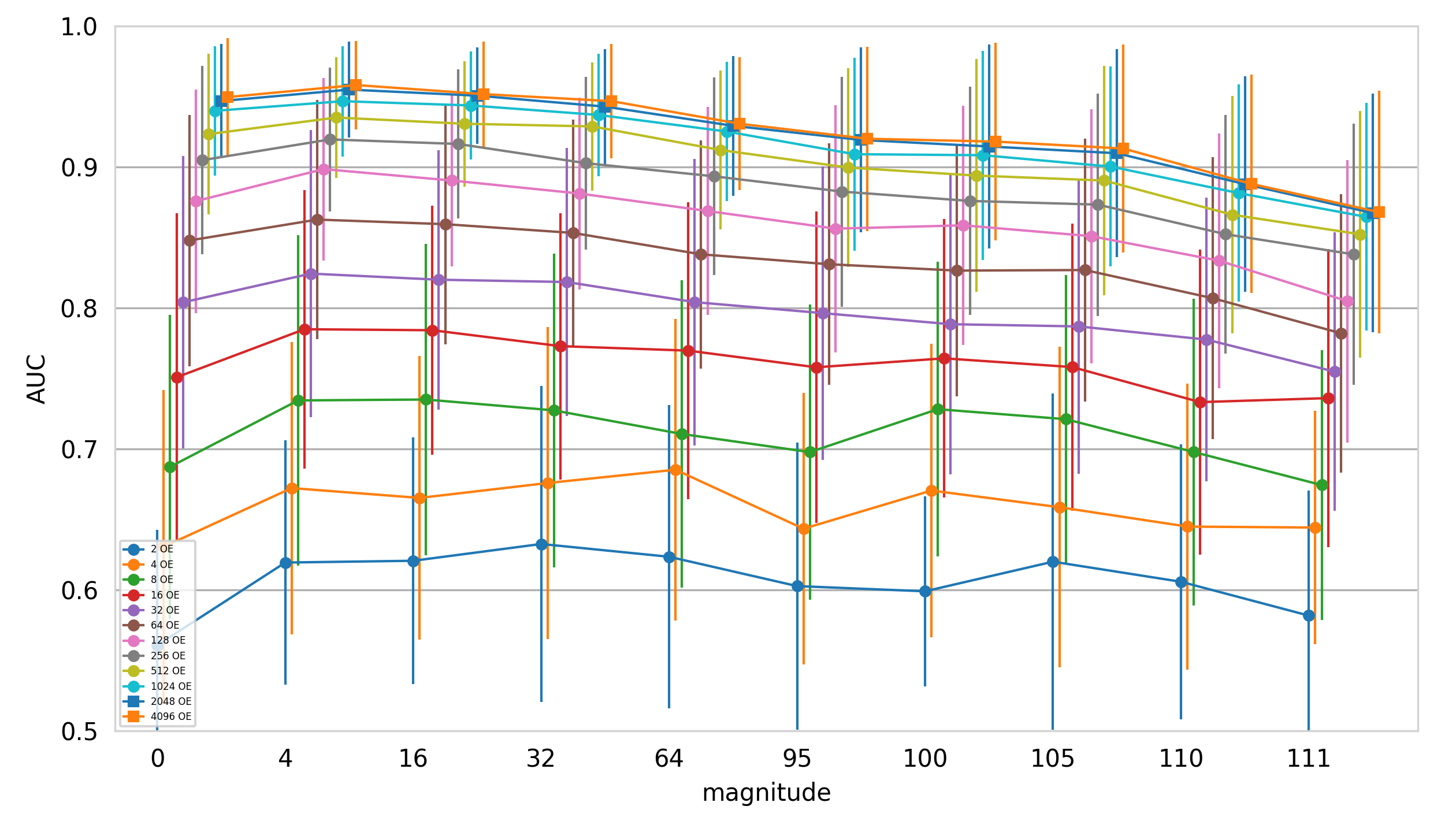}
    }
    \subfigure{
        \hspace{0.040\textwidth} 
        \includegraphics[width=0.415\textwidth]{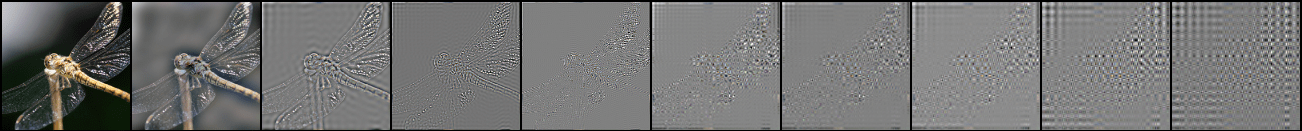}
        \hspace{0.010\textwidth}
    }
    \subfigure{
        \hspace{0.040\textwidth} 
        \includegraphics[width=0.415\textwidth]{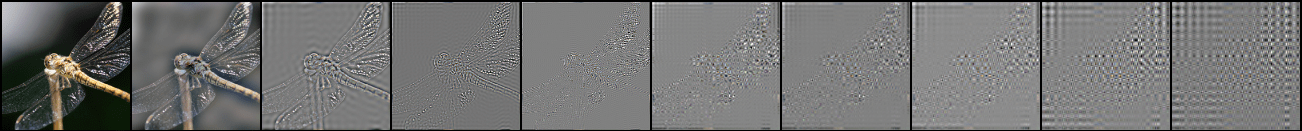}
        \hspace{0.010\textwidth}
    }
    \caption{
    HPF ImageNet-30 with ImageNet-22k (with ImageNet-1k removed) OE AD benchmark. 
    }
    \label{fig:multiscale_imagenet_hpf}
\end{figure*}

To understand why so few OE samples are that effective, we investigate the general detection performance of HSC and BCE for images with limited frequency spectra.
Analog to Appendix \ref{appx:exp_freq_anal}, we either low-pass (LPF) or high-pass-filter (HPF) all images, both during training and testing, both normal and anomalous samples. 
We train and evaluate either HSC or BCE for varying OE dataset sizes and different magnitudes of filters.
Note that, due to computational constraints, we decrease the number of epochs and restrict the augmentations for the frequency experiments. 
Figures \ref{fig:multiscale_cifar_lpf}, \ref{fig:multiscale_cifar_hpf}, \ref{fig:multiscale_imagenet_lpf}, and \ref{fig:multiscale_imagenet_hpf} show the results on LPF CIFAR-10, HPF CIFAR-10, LPF ImageNet-30, and HPF ImageNet-30, respectively.
Each point in the plots corresponds to the mean AUC detection performance over all classes and 2 seeds per class.
Different colors/markers correspond to different amounts of random OE samples used.
The magnitudes shown on the horizontal axes correspond to the number of rows and columns removed in the frequency domain. 
For example, an LPF with a magnitude of $m$ sets the first and last $m$ rows and the first and last $m$ columns of the Fourier-transformed image to zero, before applying the inverse Fourier transformation. 
A magnitude of $0$ corresponds to unfiltered images, a magnitude of $15$ on CIFAR-10 images (which have a resolution of $32 \times 32$) corresponds to filtered images where just 4 ``pixels'' in the center remain in the frequency domain. 
Similarly, an HPF with a magnitude of $m$ sets the center of size $m \times m$ to zero. 
The extended OE robustness experiments in Appendix \ref{appx:exp_freq_anal} and Appendix \ref{appx:rand_search} use a magnitude of 14 for CIFAR-10 and a magnitude of 110 for ImageNet.

%%%%%%%%%%%%%%%%%%%%%%%%%%%%%%%%%%%%%%%%%%%%%%%%%%%%%%%%%%%%%%%%%%%%%%%%%%%%%%%%
%%%%%%%%%%%%%%%%%%%%%%%%%%%%%%%%%%%%%%%%%%%%%%%%%%%%%%%%%%%%%%%%%%%%%%%%%%%%%%%%
%%%%%%%%%%%%%%%%%%%%%%%%%%%%%%%%%%%%%%%%%%%%%%%%%%%%%%%%%%%%%%%%%%%%%%%%%%%%%%%%
%%%%%%%%%%%%%%%%%%%%%%%%%%%%%%%%%%%%%%%%%%%%%%%%%%%%%%%%%%%%%%%%%%%%%%%%%%%%%%%%
%%%%%%%%%%%%%%%%%%%%%%%%%%%%%%%%%%%%%%%%%%%%%%%%%%%%%%%%%%%%%%%%%%%%%%%%%%%%%%%%
%%%%%%%%%%%%%%%%%%%%%%%%%%%%%%%%%%%%%%%%%%%%%%%%%%%%%%%%%%%%%%%%%%%%%%%%%%%%%%%%
%%%%%%%%%%%%%%%%%%%%%%%%%%%%%%%%%%%%%%%%%%%%%%%%%%%%%%%%%%%%%%%%%%%%%%%%%%%%%%%%
%%%%%%%%%%%%%%%%%%%%%%%%%%%%%%%%%%%%%%%%%%%%%%%%%%%%%%%%%%%%%%%%%%%%%%%%%%%%%%%%

\section{OE robustness results with random search} \label{appx:rand_search}
In Section \ref{sec:exp_robustness} we use an evolutionary algorithm to find the best and worst single OE samples for AD. 
The main difference between an evolutionary algorithm and a pure random search is that the former ``fine-tunes'' samples by picking the next ones depending on the best-performing previous ones. 
However, the algorithm also explores completely new samples.
To show that this is the case and the algorithm doesn't quickly converge to a local optimum, we here present results using a pure random search for the best single OE samples and compare those to the ones in the main paper. 
%%%%%%%%%%%%%%%%%%%%%%%%%%%%%%%%%%%%%%%%%%%%%%%%%%%%%%%%%%%%%%%%%%%%%%%%%%%%%%%%
\begin{table}[ht]
  \caption{Mean AUC detection performance in \% for the best single OE samples on the CIFAR-10 AD benchmark using 80MTI as OE and on the ImageNet-10 AD benchmark using ImageNet-22K (with the 1K classes removed) as OE. All images have been either unfiltered, high-pass-filtered (HPF), or low-pass-filtered (LPF). Here we use pure random search instead of an evolutionary algorithm. }
  \label{tab:random_oe}
  \begin{center}
    \small
     \begin{tabular}{ccccc} 
\toprule 
& \multicolumn{2}{c|}{CIFAR-10} & \multicolumn{2}{c}{ImageNet} \\ 
& HSC & \multicolumn{1}{c|}{BCE} & HSC & BCE \\ 
\midrule 
Best OE & 76.5 & \multicolumn{1}{c|}{68.7} & 79.5 & 77.8 \\
Best OE LPF & 77.0 & \multicolumn{1}{c|}{66.6} & 76.7 & 73.3 \\ 
Best OE HPF & 68.1& \multicolumn{1}{c|}{65.9} & 75.0 & 77.2 \\ 
\bottomrule 
\end{tabular}

  \end{center}
\end{table}
%%%%%%%%%%%%%%%%%%%%%%%%%%%%%%%%%%%%%%%%%%%%%%%%%%%%%%%%%%%%%%%%%%%%%%%%%%%%%%%%

\begin{figure*}[ht]
\centering
\subfigure[``ship'' is normal]{\includegraphics[height=0.159\linewidth]{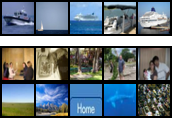}}
\subfigure[``cat'' is normal]{\includegraphics[height=0.159\linewidth]{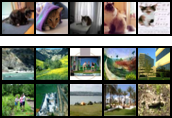}} 
\subfigure[``airplane'' is normal]{\includegraphics[height=0.159\linewidth]{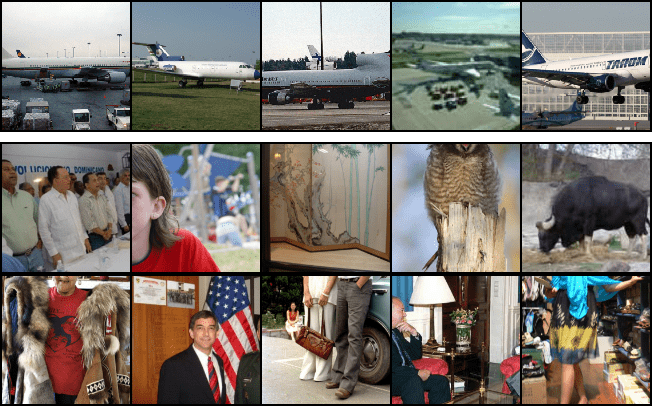}}
\subfigure[``dragonfly'' is normal]{\includegraphics[height=0.159\linewidth]{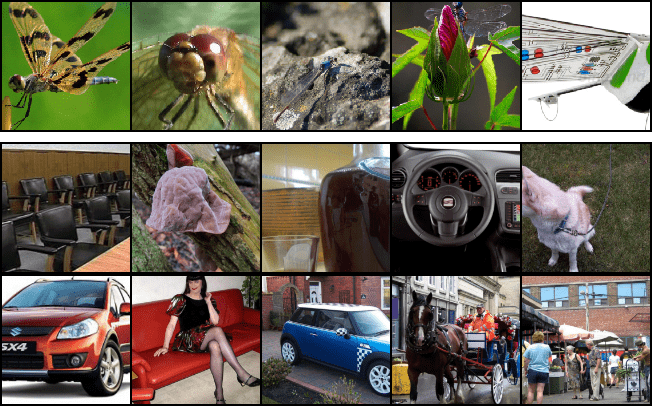}}
\subfigure[``ship'' is normal]{\includegraphics[height=0.152\linewidth]{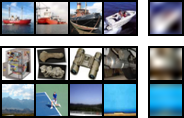}}
\subfigure[``cat'' is normal]{\includegraphics[height=0.152\linewidth]{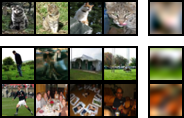}} 
\subfigure[``airplane'' is normal]{\includegraphics[height=0.152\linewidth]{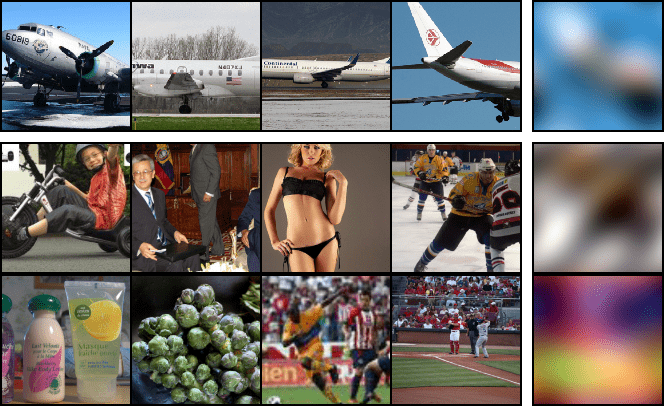}}
\subfigure[``dragonfly'' is normal]{\includegraphics[height=0.152\linewidth]{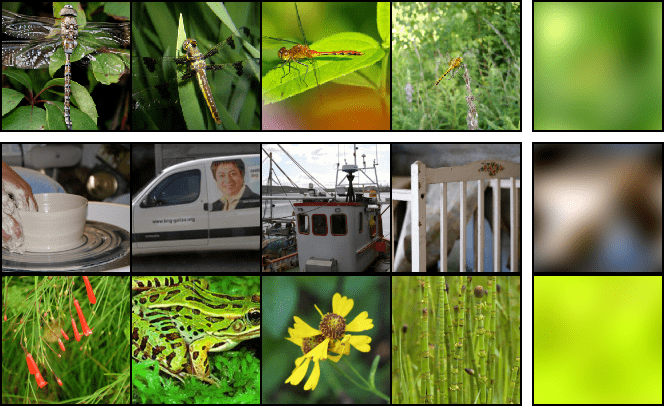}}\\
\subfigure[``ship'' is normal]{\includegraphics[height=0.152\linewidth]{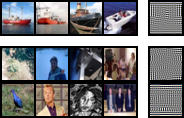}}
\subfigure[``cat'' is normal]{\includegraphics[height=0.152\linewidth]{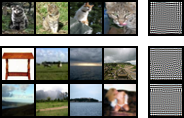}}
\subfigure[``airplane'' is normal]{\includegraphics[height=0.152\linewidth]{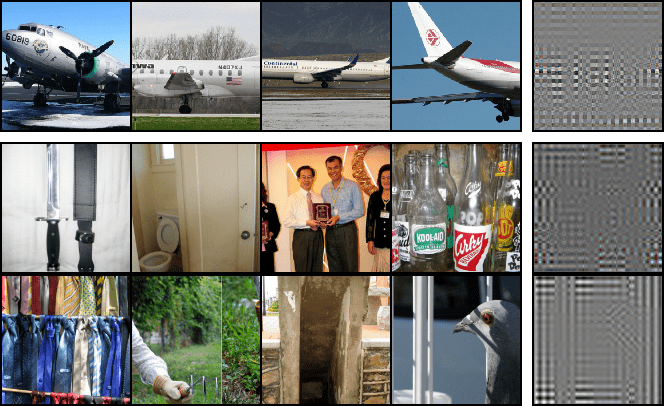}}
\subfigure[``dragonfly'' is normal]{\includegraphics[height=0.152\linewidth]{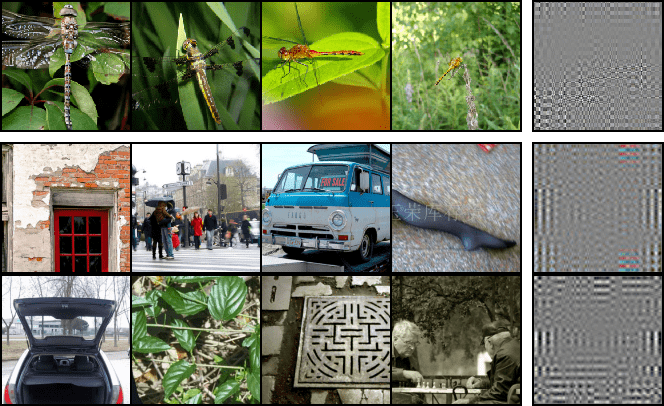}}
\caption{Best OE samples for CIFAR-10 with 80MTI as OE (a-b, e-f, i-j) and ImageNet-10 with ImageNet-22k as OE (c-d, g-h, k-l). In each figure, the first row shows normal samples, and the next two rows the best samples found via HSC (top) and BCE (bottom). In Figures (e-h), all samples (train, test, anomalous, and normal) are low-pass-filtered, in Figures (i-l) they are high-pass-filtered. These figures also contain a separate last column showing the filtered version of the images, which is what the network sees during training or testing. Here we use pure random search instead of an evolutionary algorithm.}
\label{fig:random_oe}
\end{figure*}

Table \ref{tab:random_oe} shows the mean AUC over ten classes (analogue to Table \ref{tab:evolve} in the main paper and Table \ref{tab:evolve_lpf_hpf} in Appendix \ref{appx:exp_freq_anal}). 
We see that pure random search consistently yields slightly worse samples on CIFAR-10 and mostly very similar performing samples on ImageNet-10.
There is a minor exception for unfiltered images with BCE, where random search found on average 2\% better performing single OE samples, resulting in the same conclusion, though.
Figure \ref{fig:random_oe} shows examples for the best samples found via a random search for unfiltered and filtered images. 
We see that the results are quite similar to those found via the evolutionary algorithm.
%%%%%%%%%%%%%%%%%%%%%%%%%%%%%%%%%%%%%%%%%%%%%%%%%%%%%%%%%%%%%%%%%%%%%%%%%%%%%%%%
%%%%%%%%%%%%%%%%%%%%%%%%%%%%%%%%%%%%%%%%%%%%%%%%%%%%%%%%%%%%%%%%%%%%%%%%%%%%%%%%

\section{Hypersphere Classifier sensitivity analysis}
\label{appx:losses}
In this section, we show results for the Hypersphere Classifier (HSC) (Section \ref{sec:hypsphcla}) when varying the radial function $l(\bz) = \exp\left(-h(\bz) \right)$.
For this, we run the CIFAR-10 one vs.~rest benchmark with 80MTI as OE, as presented in Table \ref{tab:aucs_without_transfer} in the main paper, for different functions $h:\mR^r \to [0, \infty), \bz \mapsto h(\bz)$.
We also alter training to be with or without data augmentation in these experiments.
The results are presented in Table \ref{tab:loss_sensitivity}.
We see that data augmentation leads to an improvement in performance even in the case where we have the full 80MTI dataset as OE.
HSC shows the overall best performance with data augmentation and using the robust Pseudo-Huber loss $h(\bz) = \sqrt{\left\lVert\bz\right\rVert^2+1}-1$.

%%%%%%%%%%%%%%%%%%%%%%%%%%%%%%%%%%%%%%%%%%%%%%%%%%%%%%%%%%%%
\begin{table}[ht]
  \caption{Mean AUC detection performance in \% (over 10 seeds) on the CIFAR-10 one vs.~rest benchmark using 80MTI as OE for different choices of $h(\bz)$ in the radial function $l$ of HSC.}
  \label{tab:loss_sensitivity}
  \begin{center}
    \small
    \begin{tabular}{lcccccccc}
    \toprule
    Data augment.\       & $\left\|\bz\right\|_1$ & $\left\|\bz\right\|_2$ & $\left\|\bz\right\|_2^2$ & $\sqrt{\left\lVert\bz\right\rVert^2+1}-1$ \\
    \midrule
    w/o                     & 90.6 & 92.3 & 89.1 & 91.8 \\
    w/                      & 92.5 & 94.1 & 94.5 & 96.1 \\
    \bottomrule
\end{tabular}
  \end{center}
\end{table}
%%%%%%%%%%%%%%%%%%%%%%%%%%%%%%%%%%%%%%%%%%%%%%%%%%%%%%%%%%%%

\section{Focal Loss with varying \texorpdfstring{$\gamma$}{TEXT}} 
\label{appx:focal}
\begin{figure}[htb]
  \centering \small
  \subfigure[CIFAR-10]{\includegraphics[width=0.495\columnwidth]{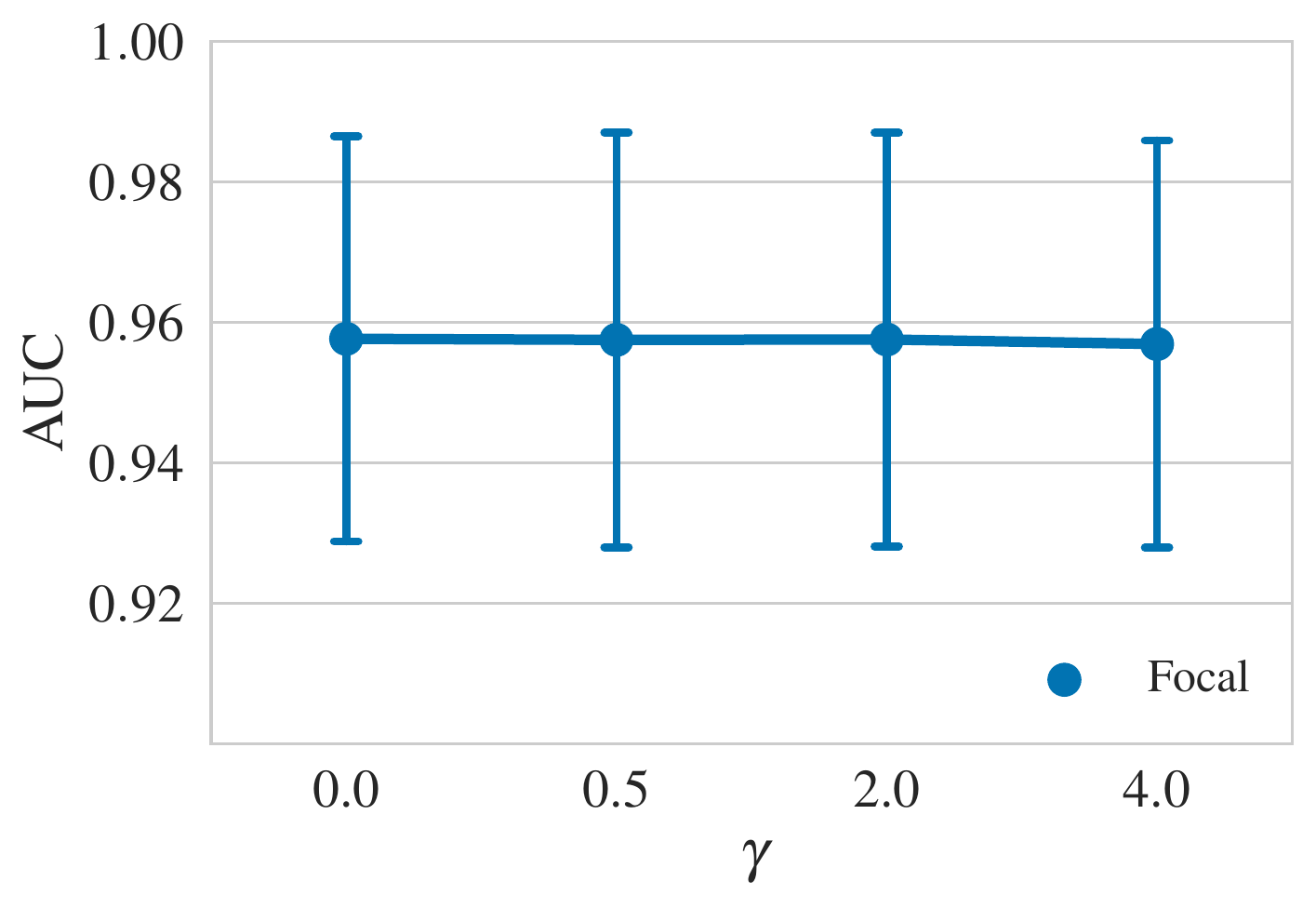}}
  \subfigure[ImageNet-30]{\includegraphics[width=0.495\columnwidth]{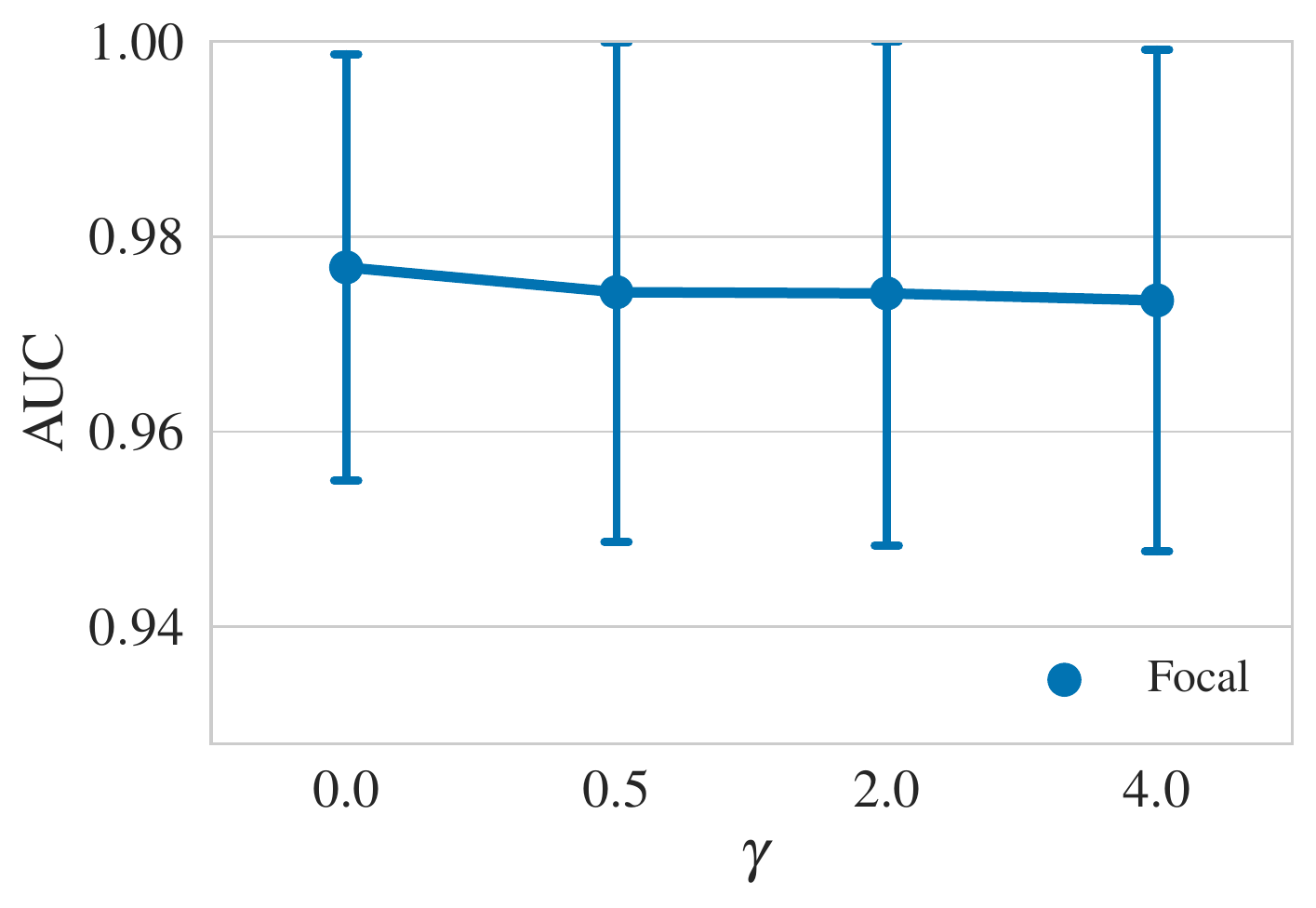}}
  \caption{Focal loss detection performance in mean AUC in \% when varying $\gamma$ on the CIFAR-10 with 80MTI OE (a) and ImageNet-30 with ImageNet-22K OE (b) one vs.~rest benchmarks.}
  \label{fig:focal_sensitivity}
\end{figure}

Here we include results showing how mean AUC detection performance changes with $\gamma$ on the Focal loss.
Since we balance every batch to contain 128 normal and 128 OE samples during training, we set the weighting factor $\alpha$ to be $\alpha=0.5$ \citep{lin17}.
Again, note that $\gamma = 0$ corresponds to standard binary cross entropy.
Figure \ref{fig:focal_sensitivity} shows that mean AUC performance is insensitive to the choice of $\gamma$ on the CIFAR-10 and ImageNet-30 one vs.~rest benchmarks.

%%%%%%%%%%%%%%%%%%%%%%%%%%%%%%%%%%%%%%%%%%%%%%%%%%%%%%%%%%%%%%%%%%%%%%%%%%%%%%%%
%%%%%%%%%%%%%%%%%%%%%%%%%%%%%%%%%%%%%%%%%%%%%%%%%%%%%%%%%%%%%%%%%%%%%%%%%%%%%%%%

\section{Network architectures and optimization}
\label{appx:architectures}
We provide details on the network architectures and optimization below, where we distinguish between end-to-end methods and the ones that use transfer learning (CLIP-based). 

\subsection{End-to-end methods}
We always use the same underlying network $\phi_\theta$ in each experimental setting for our HSC, Deep SAD, Focal, and BCE implementations to control architectural effects. 
For Focal and BCE, the output of the network $\phi_\theta$ is followed by a linear layer with sigmoid activation. 
For the experiments on CIFAR-10 and (Fashion-)MNIST we use standard LeNet-style networks \citep{lecun1990handwritten} having three convolutional layers followed by two fully connected layers. 
We use batch normalization \citep{ioffe2015} and (leaky) ReLU activations in these networks. 
For our experiments on ImageNet, CUB, DTD, and MVTec-AD we use the same WideResNet \citep{zagoruyko2016wide} as \citet{hendrycks2019using}, which has ResNet-18 as its architectural backbone. 
We use Adam \citep{kingma2014} for optimization and balance every batch to contain 128 normal and 128 OE samples during training. 
For data augmentation, we use standard color jitter, random cropping, horizontal flipping, and Gaussian pixel noise.
Due to computational constraints, there is an exception for the OE robustness experiments (Section \ref{sec:exp_robustness}), the frequency analysis (Appendices \ref{appx:exp_freq_anal} and \ref{appx:freq_sens_anal}), and the leave-one-class-out experiment with varying OE sizes (Figure \ref{fig:var_oe_size_loco}), where we use no augmentations for CIFAR-10 and only random cropping for ImageNet. 
We provide further dataset-specific details below.

\paragraph{CIFAR-10}
On CIFAR-10, we use LeNet-style networks having three convolutional layers and two fully connected layers. Each convolutional layer is followed by batch normalization, a leaky ReLU activation, and max-pooling. The first fully connected layer is followed by batch normalization, and a leaky ReLU activation, while the last layer is just a linear transformation. The number of kernels in the convolutional layers are, from first to last: 32-64-128. The fully connected layers have 512-256 units respectively.
We use Adam \citep{kingma2014} for optimization and balance every batch to contain
128 normal and 128 OE samples during training.
We train for 200 epochs starting with a learning rate of $\eta=0.001$ and have learning rate milestones at 100 and 150 epochs. The learning rate is reduced by a factor of 10 at every milestone.
For the OE robustness experiments (Section \ref{sec:exp_robustness}), the frequency analysis (Appendices \ref{appx:exp_freq_anal} and \ref{appx:freq_sens_anal}), and the leave-one-class-out experiment with varying OE sizes (Figure \ref{fig:var_oe_size_loco}), we trained for 30 epochs with a milestone at 25 instead. 

\paragraph{ImageNet}
On ImageNet, we use exactly the same WideResNet \citep{zagoruyko2016wide} as was used in \citet{hendrycks2019using}, which has a ResNet-18 as architectural backbone.
We use Adam \citep{kingma2014} for optimization and balance every batch to contain
128 normal and 128 OE samples during training.
We train for 150 epochs starting with a learning rate of $\eta=0.001$ and milestones at epochs 100 and 125. The learning rate is reduced by a factor of 10 at every milestone.
For the OE robustness experiments (Section \ref{sec:exp_robustness}), the frequency analysis (Appendices \ref{appx:exp_freq_anal} and \ref{appx:freq_sens_anal}), and the leave-one-class-out experiment with varying OE sizes (Figure \ref{fig:var_oe_size_loco}) we trained for 30 epochs with a milestone at 25 instead.

\paragraph{CUB}
On CUB, we use the same setup as we used on ImageNet.
However, we balance every batch to contain 30 normal and 30 OE samples during training since there are only 30 training samples per class.
Also, we used a subset of ImageNet-22k where we sampled 2 images per ImageNet-22k class as OE. 
This speeds up data loading but should not have an impact on the performance because with 150 epochs and 30 training samples the trainer only sees up to 4500 OE samples.

\paragraph{DTD}
On DTD, similar to CUB, we use the same setup as we used on ImageNet but balance every batch to contain 40 normal and 40 OE samples during training since there are only 40 training samples per class.
Again, we use the subset of ImageNet-22k as OE.

\paragraph{MVTec-AD}
On MVTec-AD, we use almost the same setup as we used on ImageNet.
However, we train for 300 epochs starting with a learning rate of $\eta=0.001$ and milestones at epochs 200 and 250. 

\paragraph{Fashion-MNIST}
On Fashion-MNIST, we use a similar setup as we used on CIFAR-10. Here the network consists of only two convolutional layers with 16 and 32 kernels, respectively.

\paragraph{MNIST}
On MNIST, we use the same setup as on Fashion-MNIST but don't use any data augmentation.

\subsection{CLIP}
Apart from the following changes, we use the same setup for our CLIP-based implementations as for the end-to-end methods.
One of the changes is that we use the pre-trained CLIP network architecture \citep{radford2021learning}.
For fine-tuning CLIP with a BEC classifier, we use SGD with Nesterov momentum instead of ADAM, train for 80 epochs starting with a learning rate of $\eta=0.0001$, and have learning rate milestones at 50, 60, 70, and 75 epochs. 
The learning rate is reduced by a factor of 10 at every milestone.
This applies completely to ImageNet and MVTec-AD.
For CIFAR-10, CUB, DTD, Fashion-MNIST, and MNIST, we do the same but start with a learning rate of $\eta=0.00002$ instead.

\section{The evolutionary algorithm for finding OE samples} \label{appx:evol_algo}

In Section \ref{sec:exp_robustness} we search for a sample in the OE dataset that is either particularly harmful or useful for the AD model. 
A common approach to perform a discrete search in a large set is to use evolutionary algorithms \citep{yu2010introduction, fortin2012deap}.
We here stick to a simple version that uses tournament selection \citep{blickle1995mathematical} with three competitors.
In our scenario, the individuals are images of the OE dataset and randomly initialized.

\begin{algorithm}[H]
    \footnotesize
    \caption{Evolve OE samples}
    \label{alg:evolve}
    
    \textbf{Input:} AD model $\phi$, OE dataset $D$, normal train set $X$, full test set $X_{test}$. \hfill \hphantom{1em}
    
    \textbf{Output:} A collection of optimal single training outliers (individuals) $D'$. \hfill \hphantom{1em}
    
    \textbf{Define:} 
    \begin{algorithmic}
        \STATE COIN($p$): returns TRUE with chance $p$, else FALSE.
        \STATE RND($D, k$): selects $k$ samples in set $D$ randomly. 
        \STATE TRAIN($\phi$, $d$): randomly initializes $\phi$'s weights, then trains $\phi$ with normal dataset $X$ and the single training outlier $d$ as OE.
        \STATE EVAL($\phi$): computes and returns the test AUC of $\phi$ using the one vs.~rest approach on $X_{test}$.
    \end{algorithmic} 
    \vspace{1em}

    \textbf{Algorithm:} 
    \begin{algorithmic}
        \STATE \algcom{Initialize:}
        \STATE  $D' \gets$ RND($D, 64$)
        \FORALL{$d' \in D'$}
            \STATE TRAIN($\phi$, $d'$)
            \STATE $a \gets$ EVAL($\phi$)
            \STATE Set fitness of $d'$ to $a$
        \ENDFOR
        \vspace{0.5em}
        
        \FOR{50 iterations}
        \vspace{0.5em}
        
            \STATE \algcom{Select:}
            \STATE $D_{temp} \gets D'$
            \FORALL{$d_{temp} \in D_{temp}$}
                \STATE Replace $d_{temp}$ in $D_{temp}$ with the best in RND($D', 3$). \quad [best according to fitness]
            \ENDFOR
            \STATE $D' \gets D_{temp}$
            \vspace{0.5em}
            
            \STATE \algcom{Mate:}
            \FORALL{$i \in [0, 2, 4, ..., |D'|]$}
                \IF{COIN(0.05)}
                    \STATE $d'_1 \gets D'[i]$
                    \STATE $d'_2 \gets D'[i+1]$ 
                    \FORALL{$j \in [0, 1]$}
                        \STATE $P \gets$ RND($D, 10000$)
                        \STATE sort($P$) according to $(\lVert p - d'_1 \rVert^2 + \lVert p - d'_2 \rVert^2)$ for every $p \in P$.
                        \STATE $P \gets P[:50]$, get the 50 samples with least average distance to $d'_1$ and $d'_2$.
                        \STATE Replace $D'[i+j]$ with RND($P, 1$).
                    \ENDFOR
                \ENDIF
            \ENDFOR
            \vspace{0.5em}
            
            \STATE \algcom{Mutate:}
            \FORALL{$d' \in D'$}
                \IF{COIN(0.55)}
                    \STATE $P \gets$ RND($D, 10000$)
                    \STATE sort($P$) according to $\lVert p - d' \rVert^2$ for every $p \in P$.
                    \STATE $P \gets P[:50]$, get the 50 samples with least distance to $d'$.
                    \STATE Replace $d'$ in $D'$ with RND($P, 1$).
                \ENDIF
            \ENDFOR
            \vspace{0.5em}
    
            \STATE \algcom{Evaluate:}
            \FORALL{$d' \in D'$}
                \STATE TRAIN($\phi$, $d'$)
                \STATE $a \gets$ EVAL($\phi$)
                \STATE Set fitness of $d'$ to $a$
            \ENDFOR
            \vspace{0.5em}
            
        \ENDFOR
    
    \end{algorithmic}
\end{algorithm}

Mutating an individual works by first sampling random images from the OE set and then randomly picking one among those $50$ in that subset that have the least $L^2$ distance to the individual. 
Mating works similarly, it picks one with the least $L^2$ distance to both parents.  
The fitness of an individual is the test AUC of an AD model trained with the full normal dataset and just the individual as the only training outlier. 
The algorithm's objective is to either maximize (``best'' OE samples) or minimize (``worst'' OE samples) the average fitness of its final generation.

This experimental setup is exceptionally computationally expensive, with the evaluation of a single individual's fitness requiring a full training of a neural network. 
Because of this we decrease the number of training epochs, restrict the augmentations, and average over just two different random seeds to evaluate an individual's fitness (see Appendix \ref{appx:architectures}).
On ImageNet, we only consider the first 10 classes (see Appendix \ref{appx:full_results}) out of the 30 that are used in \citet{hendrycks2019deep}, but evaluate these with the full 30-classes one vs.~rest benchmark.

Algorithm \ref{alg:evolve} provides a detailed explanation of the evolutionary algorithm. 
For the sake of readability, we fix the evolutionary hyperparameters to the values used in our experiments.
We set the generation size to $64$, the number of generations to $50$, the tournament size to $3$, the mating chance to $0.05$, the mutation chance to $0.55$, the initial candidate pool size to $10000$, and the final candidate pool size to $50$.
We apply this algorithm to each class separately, thereby changing the normal set $X$ and test set $X_{test}$.
For finding the worst single OE samples instead of the best ones, we change the selection procedure to simply replace $d_{temp}$ with the worst in RND($D', 3$) instead of the best.

\clearpage

\section{Results on individual classes}
\label{appx:full_results}
We report results averaged over all classes in the main paper.
Here we provide full class-wise results for all experiments of Sections \ref{sec:exp_sota_without_transfer}, \ref{sec:exp_sota_with_transfer}, \ref{sec:exp_robustness}, and Appendix \ref{appx:exp_freq_anal}. We also include figures for each class for the varying amount of OE experiment (Section \ref{sec:exp_var_oe_size}) for CIFAR-10 and ImageNet-30.

\subsection{Image AD with end-to-end methods on the one vs.~rest benchmark}
\begin{figure}[hbt] 
  \begin{center} 
      \includegraphics[width=0.91\textwidth]{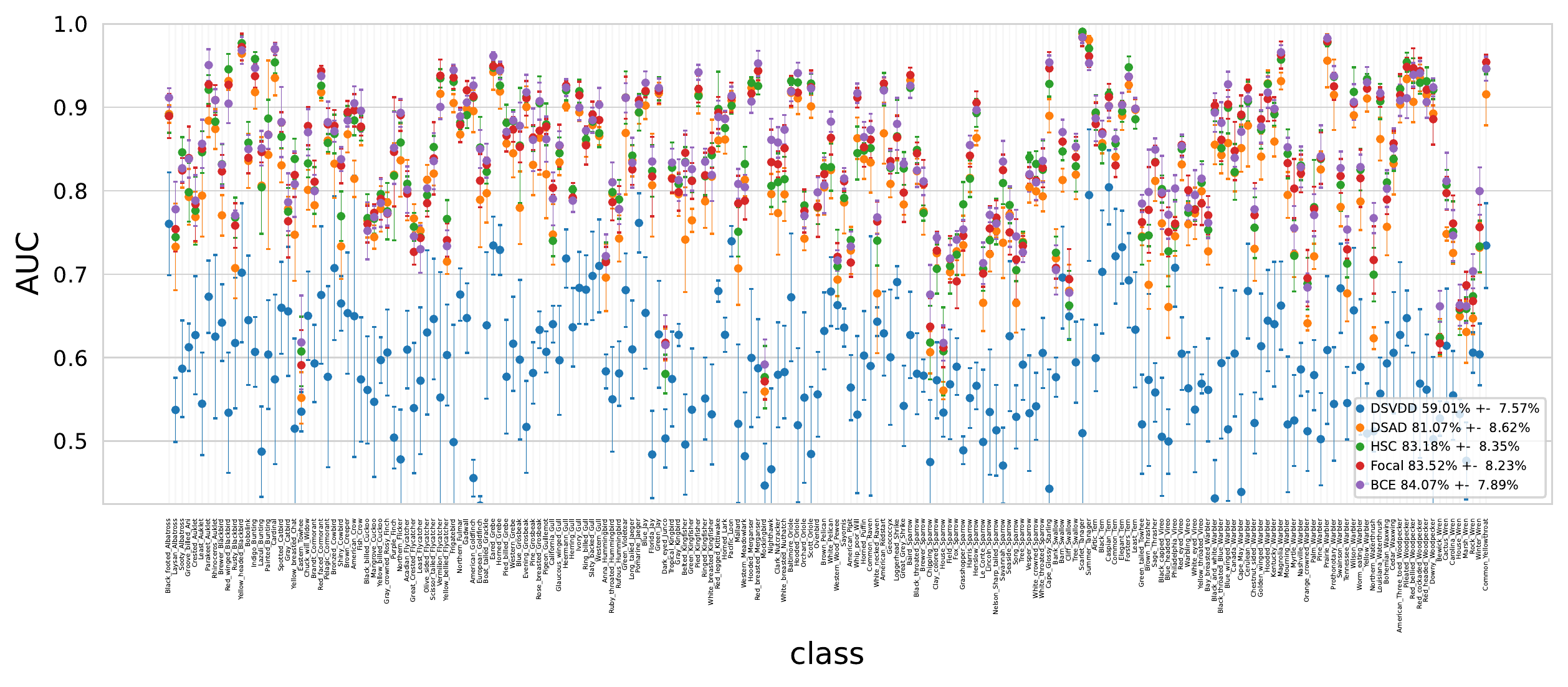}
      \caption{Mean AUC detection performance in \% (over 10 seeds) for all individual classes for our end-to-end implementations on the CUB one vs.~rest benchmark with ImageNet-22k OE from Section \ref{sec:exp_sota_without_transfer}.}
      \label{fig:auc_summary_cub_wo_transfer}
  \end{center}
\end{figure}

\begin{figure}[hbt] 
  \begin{center} 
      \includegraphics[width=0.91\textwidth]{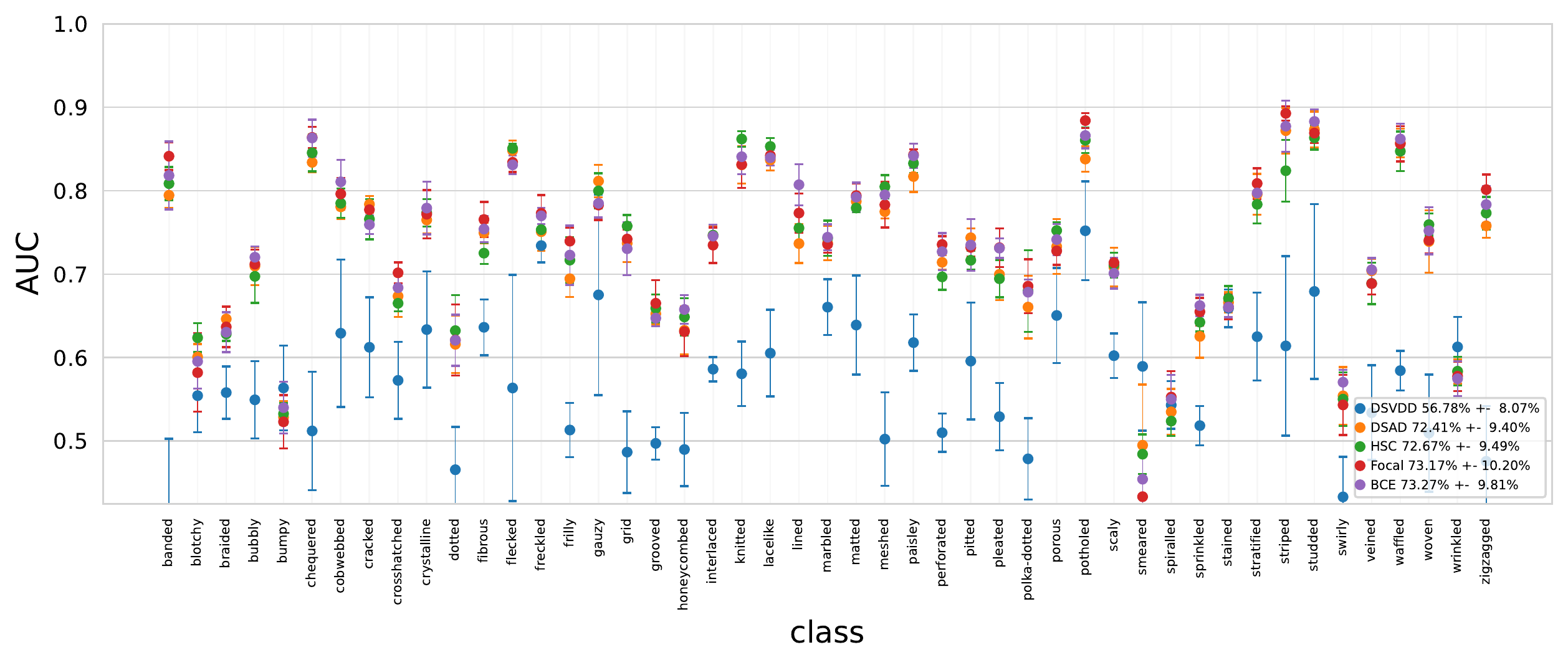}
      \caption{Mean AUC detection performance in \% (over 10 seeds) for all individual classes for our end-to-end implementations on the DTD one vs.~rest benchmark with ImageNet-22k OE from Section \ref{sec:exp_sota_without_transfer}.}
      \label{fig:auc_summary_dtd_wo_transfer}
  \end{center}
\end{figure}

Here, we report results for the one vs.~rest benchmark experiments from Section \ref{sec:exp_sota_without_transfer} for all individual classes and \emph{end-to-end} methods. 
For CIFAR-10, we report results in Table \ref{tab:cifar10_classes_wo_transfer}, including results from the literature. 
For CIFAR-10, ImageNet-30, Fashion-MNIST, and MNIST, we report results with standard deviations for our implementations in Tables \ref{tab:cifar10_classes_}, \ref{tab:imagenet_one_vs_rest_wo_transfer}, \ref{tab:fmnist_one_vs_rest_wo_transfer}, and \ref{tab:mnist_one_vs_rest_wo_transfer}, respectively. 
For CUB and DTD, we report class-wise results instead in form of a plot in Figures \ref{fig:auc_summary_cub_wo_transfer} and \ref{fig:auc_summary_dtd_wo_transfer}, as there are too many classes (47 for DTD and 200 for CUB) to report concisely in a Table. 

%%%%%%%%%%%%%%%%%%%%%%%%%%%%%%%%%%%%%%%%%%%%%%%%%%%%%%%%%%%%
\begin{table*}[th]
    \caption{Mean AUC detection performance in \% (over 10 seeds) for all individual classes and end-to-end methods on the CIFAR-10 one vs.~rest benchmark with 80MTI OE from Section \ref{sec:exp_sota_without_transfer}.}
    \label{tab:cifar10_classes_wo_transfer}
    \vspace{0.5em}
    \centering\small
    \begin{tabular}{lccccccccccc} 
\toprule 
 & \multicolumn{5}{c|}{Unsupervised} & \multicolumn{3}{c|}{Unsupervised OE} & \multicolumn{3}{c}{Supervised OE} \\ 
Class & SVDD* & DSVDD* & GT* & GT+* & \multicolumn{1}{c|}{CSI*} & GT+* & DSAD & \multicolumn{1}{c|}{HSC} & Focal* & Focal & \multicolumn{1}{c}{BCE} \\ 
\midrule 
Airplane & 65.6 & 61.7 & 74.7 & 77.5 & \multicolumn{1}{c|}{89.9} & 90.4 & 94.2 & \multicolumn{1}{c|}{96.3} & 87.6 & 95.9 & \multicolumn{1}{c}{\bf 96.4} \\ 
Automobile & 40.9 & 65.9 & 95.7 & 96.9 & \multicolumn{1}{c|}{99.1} & \bf 99.3 & 98.1 & \multicolumn{1}{c|}{98.7} & 93.9 & 98.7 & \multicolumn{1}{c}{98.8} \\ 
Bird & 65.3 & 50.8 & 78.1 & 87.3 & \multicolumn{1}{c|}{93.1} & \bf 93.7 & 89.8 & \multicolumn{1}{c|}{92.7} & 78.6 & 92.3 & \multicolumn{1}{c}{93.0} \\ 
Cat & 50.1 & 59.1 & 72.4 & 80.9 & \multicolumn{1}{c|}{86.4} & 88.1 & 87.4 & \multicolumn{1}{c|}{89.8} & 79.9 & 88.8 & \multicolumn{1}{c}{\bf 90.0} \\ 
Deer & 75.2 & 60.9 & 87.8 & 92.7 & \multicolumn{1}{c|}{93.9} & \bf 97.4 & 95.0 & \multicolumn{1}{c|}{96.6} & 81.7 & 96.6 & \multicolumn{1}{c}{97.1} \\ 
Dog & 51.2 & 65.7 & 87.8 & 90.2 & \multicolumn{1}{c|}{93.2} & \bf 94.3 & 93.0 & \multicolumn{1}{c|}{94.2} & 85.6 & 94.1 & \multicolumn{1}{c}{94.2} \\ 
Frog & 71.8 & 67.7 & 83.4 & 90.9 & \multicolumn{1}{c|}{95.1} & 97.1 & 96.9 & \multicolumn{1}{c|}{97.9} & 93.3 & 97.8 & \multicolumn{1}{c}{\bf 98.0} \\ 
Horse & 51.2 & 67.3 & 95.5 & 96.5 & \multicolumn{1}{c|}{98.7} & \bf 98.8 & 96.8 & \multicolumn{1}{c|}{97.6} & 87.9 & 97.6 & \multicolumn{1}{c}{97.6} \\ 
Ship & 67.9 & 75.9 & 93.3 & 95.2 & \multicolumn{1}{c|}{97.9} & \bf 98.7 & 97.1 & \multicolumn{1}{c|}{98.2} & 92.6 & 98.0 & \multicolumn{1}{c}{98.1} \\ 
Truck & 48.5 & 73.1 & 91.3 & 93.3 & \multicolumn{1}{c|}{95.5} & \bf 98.5 & 96.2 & \multicolumn{1}{c|}{97.4} & 92.1 & 97.5 & \multicolumn{1}{c}{97.7} \\ 
\midrule 
Mean AUC & 58.8 & 64.8 & 86.0 & 90.1 & \multicolumn{1}{c|}{94.3} & 95.6 & 94.5 & \multicolumn{1}{c|}{95.9} & 87.3 & 95.8 & \multicolumn{1}{c}{\bf 96.1} \\ 
\bottomrule 
\end{tabular} 
\end{table*}
%%%%%%%%%%%%%%%%%%%%%%%%%%%%%%%%%%%%%%%%%%%%%%%%%%%%%%%%%%%%
%%%%%%%%%%%%%%%%%%%%%%%%%%%%%%%%%%%%%%%%%%%%%%%%%%%%%%%%%%%%
\begin{table*}[th]
    \caption{Mean AUC detection performance in \% (over 10 seeds) \emph{with standard deviations} for all individual classes for our end-to-end implementations on the CIFAR-10 one vs.~rest benchmark with 80MTI OE from Section \ref{sec:exp_sota_without_transfer}. }
    \label{tab:cifar10_classes_}
    \vspace{0.5em}
    \centering\small
    \begin{tabular}{lcccc}
    \toprule
                & \multicolumn{2}{c|}{Unsupervised OE} & \multicolumn{2}{c}{Supervised OE}\\
    Class       & DSAD & \multicolumn{1}{c|}{HSC} & Focal & BCE \\
    \midrule
    Airplane    & 94.2 $\pm$ 0.34 & 96.3 $\pm$ 0.13 & 95.9 $\pm$ 0.11 & 96.4 $\pm$ 0.17\\
    Automobile  & 98.1 $\pm$ 0.19 & 98.7 $\pm$ 0.07 & 98.7 $\pm$ 0.09 & 98.8 $\pm$ 0.06\\
    Bird        & 89.8 $\pm$ 0.54 & 92.7 $\pm$ 0.27 & 92.3 $\pm$ 0.32 & 93.0 $\pm$ 0.14\\
    Cat         & 87.4 $\pm$ 0.38 & 89.8 $\pm$ 0.27 & 88.8 $\pm$ 0.33 & 90.0 $\pm$ 0.27\\
    Deer        & 95.0 $\pm$ 0.22 & 96.6 $\pm$ 0.17 & 96.6 $\pm$ 0.10 & 97.1 $\pm$ 0.10\\
    Dog         & 93.0 $\pm$ 0.30 & 94.2 $\pm$ 0.13 & 94.1 $\pm$ 0.21 & 94.2 $\pm$ 0.12\\
    Frog        & 96.9 $\pm$ 0.22 & 97.9 $\pm$ 0.08 & 97.8 $\pm$ 0.07 & 98.0 $\pm$ 0.09\\
    Horse       & 96.8 $\pm$ 0.15 & 97.6 $\pm$ 0.10 & 97.6 $\pm$ 0.16 & 97.6 $\pm$ 0.09\\
    Ship        & 97.1 $\pm$ 0.21 & 98.2 $\pm$ 0.09 & 98.0 $\pm$ 0.11 & 98.1 $\pm$ 0.08\\
    Truck       & 96.2 $\pm$ 0.22 & 97.4 $\pm$ 0.13 & 97.5 $\pm$ 0.12 & 97.7 $\pm$ 0.16\\
    \midrule
    Mean AUC    & 94.5 $\pm$ 3.30 & 95.9 $\pm$ 2.68 & 95.8 $\pm$ 2.97 & 96.1 $\pm$ 2.71\\
    \bottomrule
\end{tabular}
\end{table*}
%%%%%%%%%%%%%%%%%%%%%%%%%%%%%%%%%%%%%%%%%%%%%%%%%%%%%%%%%%%%
%%%%%%%%%%%%%%%%%%%%%%%%%%%%%%%%%%%%%%%%%%%%%%%%%%%%%%%%%%%%
\begin{table*}[th]
    \caption{Mean AUC detection performance in \% (over 10 seeds) for all individual classes for our end-to-end implementations on the ImageNet-30 one vs.~rest benchmark with ImageNet-22K OE from Section \ref{sec:exp_sota_without_transfer}. Note that for GT+*, Focal*, and CSI* as reported in Table \ref{tab:aucs_without_transfer} in the main paper, \citet{hendrycks2019using} and \citet{tack2020} do not provide results on a per class basis. }
    \label{tab:imagenet_one_vs_rest_wo_transfer}
    \vspace{0.5em}
    \centering\small
    
\begin{tabular}{lccccc}
\toprule
& \multicolumn{1}{c|}{Unsupervised} & \multicolumn{2}{c|}{Unsupervised OE} & \multicolumn{2}{c}{Supervised OE} \\
Class &   \multicolumn{1}{c|}{DSVDD} & DSAD & \multicolumn{1}{c|}{HSC} & Focal & \multicolumn{1}{c}{BCE} \\
\midrule
Acorn & \multicolumn{1}{c|}{62.7 $\pm$ 2.94} & 98.5 $\pm$ 0.28 & \multicolumn{1}{c|}{98.8 $\pm$ 0.42} & 99.0 $\pm$ 0.15 & \multicolumn{1}{c}{99.0 $\pm$ 0.19} \\
Airliner & \multicolumn{1}{c|}{55.8 $\pm$ 1.76} & 99.6 $\pm$ 0.16 & \multicolumn{1}{c|}{99.8 $\pm$ 0.10} & 99.9 $\pm$ 0.02 & \multicolumn{1}{c}{99.8 $\pm$ 0.04} \\
Ambulance & \multicolumn{1}{c|}{47.3 $\pm$ 3.58} & 99.0 $\pm$ 0.13 & \multicolumn{1}{c|}{99.8 $\pm$ 0.13} & 99.2 $\pm$ 0.14 & \multicolumn{1}{c}{99.9 $\pm$ 0.07} \\
American alligator & \multicolumn{1}{c|}{73.0 $\pm$ 0.65} & 92.9 $\pm$ 1.06 & \multicolumn{1}{c|}{98.0 $\pm$ 0.32} & 94.7 $\pm$ 0.67 & \multicolumn{1}{c}{98.2 $\pm$ 0.27} \\
Banjo & \multicolumn{1}{c|}{56.8 $\pm$ 2.22} & 97.0 $\pm$ 0.51 & \multicolumn{1}{c|}{98.2 $\pm$ 0.41} & 97.0 $\pm$ 0.33 & \multicolumn{1}{c}{98.7 $\pm$ 0.22} \\
Barn & \multicolumn{1}{c|}{67.6 $\pm$ 1.32} & 98.5 $\pm$ 0.29 & \multicolumn{1}{c|}{99.8 $\pm$ 0.05} & 98.7 $\pm$ 0.24 & \multicolumn{1}{c}{99.8 $\pm$ 0.08} \\
Bikini & \multicolumn{1}{c|}{59.7 $\pm$ 2.81} & 96.5 $\pm$ 0.84 & \multicolumn{1}{c|}{98.6 $\pm$ 0.57} & 97.2 $\pm$ 0.89 & \multicolumn{1}{c}{99.1 $\pm$ 0.30} \\
Digital clock & \multicolumn{1}{c|}{61.2 $\pm$ 2.08} & 99.4 $\pm$ 0.33 & \multicolumn{1}{c|}{96.8 $\pm$ 0.79} & 99.8 $\pm$ 0.03 & \multicolumn{1}{c}{97.2 $\pm$ 0.29} \\
Dragonfly & \multicolumn{1}{c|}{61.9 $\pm$ 3.02} & 98.8 $\pm$ 0.28 & \multicolumn{1}{c|}{98.4 $\pm$ 0.16} & 99.1 $\pm$ 0.21 & \multicolumn{1}{c}{98.3 $\pm$ 0.04} \\
Dumbbell & \multicolumn{1}{c|}{51.3 $\pm$ 1.09} & 93.0 $\pm$ 0.53 & \multicolumn{1}{c|}{91.6 $\pm$ 0.88} & 94.0 $\pm$ 0.04 & \multicolumn{1}{c}{92.6 $\pm$ 0.97} \\
Forklift & \multicolumn{1}{c|}{48.0 $\pm$ 5.00} & 90.6 $\pm$ 1.43 & \multicolumn{1}{c|}{99.1 $\pm$ 0.33} & 94.2 $\pm$ 0.90 & \multicolumn{1}{c}{99.5 $\pm$ 0.09} \\
Goblet & \multicolumn{1}{c|}{63.3 $\pm$ 0.25} & 92.4 $\pm$ 1.05 & \multicolumn{1}{c|}{93.8 $\pm$ 0.38} & 93.8 $\pm$ 0.27 & \multicolumn{1}{c}{94.7 $\pm$ 1.43} \\
Grand piano & \multicolumn{1}{c|}{58.4 $\pm$ 0.16} & 99.7 $\pm$ 0.06 & \multicolumn{1}{c|}{97.4 $\pm$ 0.37} & 99.9 $\pm$ 0.04 & \multicolumn{1}{c}{97.6 $\pm$ 0.34} \\
Hotdog & \multicolumn{1}{c|}{62.0 $\pm$ 2.46} & 95.9 $\pm$ 2.01 & \multicolumn{1}{c|}{98.5 $\pm$ 0.34} & 97.2 $\pm$ 0.05 & \multicolumn{1}{c}{98.8 $\pm$ 0.34} \\
Hourglass & \multicolumn{1}{c|}{60.2 $\pm$ 2.93} & 96.3 $\pm$ 0.37 & \multicolumn{1}{c|}{96.9 $\pm$ 0.26} & 97.5 $\pm$ 0.17 & \multicolumn{1}{c}{97.6 $\pm$ 0.48} \\
Manhole cover & \multicolumn{1}{c|}{63.9 $\pm$ 0.56} & 98.5 $\pm$ 0.29 & \multicolumn{1}{c|}{99.6 $\pm$ 0.34} & 99.2 $\pm$ 0.09 & \multicolumn{1}{c}{99.8 $\pm$ 0.01} \\
Mosque & \multicolumn{1}{c|}{73.8 $\pm$ 0.35} & 98.6 $\pm$ 0.29 & \multicolumn{1}{c|}{99.1 $\pm$ 0.26} & 98.9 $\pm$ 0.30 & \multicolumn{1}{c}{99.3 $\pm$ 0.15} \\
Nail & \multicolumn{1}{c|}{49.4 $\pm$ 1.05} & 92.8 $\pm$ 0.80 & \multicolumn{1}{c|}{94.0 $\pm$ 0.76} & 93.5 $\pm$ 0.32 & \multicolumn{1}{c}{94.5 $\pm$ 1.37} \\
Parking meter & \multicolumn{1}{c|}{46.3 $\pm$ 2.64} & 98.5 $\pm$ 0.29 & \multicolumn{1}{c|}{93.3 $\pm$ 1.64} & 99.3 $\pm$ 0.04 & \multicolumn{1}{c}{94.7 $\pm$ 0.76} \\
Pillow & \multicolumn{1}{c|}{40.8 $\pm$ 1.22} & 99.3 $\pm$ 0.14 & \multicolumn{1}{c|}{94.0 $\pm$ 0.47} & 99.2 $\pm$ 0.14 & \multicolumn{1}{c}{94.2 $\pm$ 0.42} \\
Revolver & \multicolumn{1}{c|}{50.6 $\pm$ 1.59} & 98.2 $\pm$ 0.30 & \multicolumn{1}{c|}{97.6 $\pm$ 0.25} & 98.6 $\pm$ 0.11 & \multicolumn{1}{c}{97.7 $\pm$ 0.68} \\
Rotary dial telephone & \multicolumn{1}{c|}{59.1 $\pm$ 1.68} & 90.4 $\pm$ 1.99 & \multicolumn{1}{c|}{97.7 $\pm$ 0.50} & 92.2 $\pm$ 0.33 & \multicolumn{1}{c}{98.3 $\pm$ 0.75} \\
Schooner & \multicolumn{1}{c|}{78.0 $\pm$ 1.94} & 99.1 $\pm$ 0.18 & \multicolumn{1}{c|}{99.2 $\pm$ 0.20} & 99.6 $\pm$ 0.02 & \multicolumn{1}{c}{99.1 $\pm$ 0.26} \\
Snowmobile & \multicolumn{1}{c|}{70.9 $\pm$ 0.79} & 97.7 $\pm$ 0.86 & \multicolumn{1}{c|}{99.0 $\pm$ 0.22} & 98.1 $\pm$ 0.15 & \multicolumn{1}{c}{99.1 $\pm$ 0.25} \\
Soccer ball & \multicolumn{1}{c|}{56.0 $\pm$ 1.68} & 97.3 $\pm$ 1.70 & \multicolumn{1}{c|}{92.9 $\pm$ 1.18} & 98.6 $\pm$ 0.13 & \multicolumn{1}{c}{93.6 $\pm$ 0.61} \\
Stingray & \multicolumn{1}{c|}{79.5 $\pm$ 0.98} & 99.3 $\pm$ 0.20 & \multicolumn{1}{c|}{99.1 $\pm$ 0.33} & 99.7 $\pm$ 0.04 & \multicolumn{1}{c}{99.2 $\pm$ 0.10} \\
Strawberry & \multicolumn{1}{c|}{65.0 $\pm$ 3.32} & 97.7 $\pm$ 0.64 & \multicolumn{1}{c|}{99.1 $\pm$ 0.20} & 99.1 $\pm$ 0.03 & \multicolumn{1}{c}{99.2 $\pm$ 0.22} \\
Tank & \multicolumn{1}{c|}{69.5 $\pm$ 1.83} & 97.3 $\pm$ 0.51 & \multicolumn{1}{c|}{98.6 $\pm$ 0.18} & 97.3 $\pm$ 0.47 & \multicolumn{1}{c}{98.9 $\pm$ 0.13} \\
Toaster & \multicolumn{1}{c|}{53.1 $\pm$ 0.42} & 97.7 $\pm$ 0.56 & \multicolumn{1}{c|}{92.2 $\pm$ 0.78} & 98.3 $\pm$ 0.05 & \multicolumn{1}{c}{92.2 $\pm$ 0.65} \\
Volcano & \multicolumn{1}{c|}{88.0 $\pm$ 3.27} & 89.6 $\pm$ 0.44 & \multicolumn{1}{c|}{99.5 $\pm$ 0.09} & 91.6 $\pm$ 0.90 & \multicolumn{1}{c}{99.4 $\pm$ 0.19} \\
\midrule
Mean & \multicolumn{1}{c|}{61.1 $\pm$ 10.61} & 96.7 $\pm$ 2.98 & \multicolumn{1}{c|}{97.3 $\pm$ 2.53} & 97.5 $\pm$ 2.43 & \multicolumn{1}{c}{97.7 $\pm$ 2.34} \\
\bottomrule
\end{tabular}
    
\end{table*}
%%%%%%%%%%%%%%%%%%%%%%%%%%%%%%%%%%%%%%%%%%%%%%%%%%%%%%%%%%%%
%%%%%%%%%%%%%%%%%%%%%%%%%%%%%%%%%%%%%%%%%%%%%%%%%%%%%%%%%%%%
\begin{table*}[th]
    \caption{Mean AUC detection performance in \% (over 10 seeds) for all individual classes for our end-to-end implementations on the Fashion-MNIST one vs.~rest benchmark with grayscale CIFAR-100 OE from Section \ref{sec:exp_sota_without_transfer}. }
    \label{tab:fmnist_one_vs_rest_wo_transfer}
    \vspace{0.5em}
    \centering\footnotesize
    
\begin{tabular}{lccccc}
\toprule
& \multicolumn{1}{c|}{Unsupervised} & \multicolumn{2}{c|}{Unsupervised OE} & \multicolumn{2}{c}{Supervised OE} \\
Class &   \multicolumn{1}{c|}{DSVDD} & DSAD & \multicolumn{1}{c|}{HSC} & Focal & \multicolumn{1}{c}{BCE} \\
\midrule
Top & \multicolumn{1}{c|}{82.8 $\pm$ 1.10} & 85.2 $\pm$ 2.00 & \multicolumn{1}{c|}{83.5 $\pm$ 4.13} & 78.6 $\pm$ 2.41 & \multicolumn{1}{c}{75.3 $\pm$ 2.12} \\
Trouser & \multicolumn{1}{c|}{94.8 $\pm$ 4.92} & 96.1 $\pm$ 1.90 & \multicolumn{1}{c|}{98.1 $\pm$ 0.18} & 95.9 $\pm$ 1.30 & \multicolumn{1}{c}{93.5 $\pm$ 0.81} \\
Pullover & \multicolumn{1}{c|}{79.3 $\pm$ 2.42} & 82.4 $\pm$ 0.95 & \multicolumn{1}{c|}{83.6 $\pm$ 2.59} & 88.9 $\pm$ 1.55 & \multicolumn{1}{c}{89.4 $\pm$ 1.37} \\
Dress & \multicolumn{1}{c|}{88.8 $\pm$ 2.22} & 88.1 $\pm$ 1.96 & \multicolumn{1}{c|}{88.0 $\pm$ 2.26} & 84.5 $\pm$ 2.04 & \multicolumn{1}{c}{82.6 $\pm$ 2.88} \\
Coat & \multicolumn{1}{c|}{82.9 $\pm$ 3.71} & 84.1 $\pm$ 2.04 & \multicolumn{1}{c|}{85.7 $\pm$ 1.71} & 86.2 $\pm$ 0.62 & \multicolumn{1}{c}{83.1 $\pm$ 2.20} \\
Sandal & \multicolumn{1}{c|}{75.1 $\pm$ 9.60} & 91.2 $\pm$ 1.18 & \multicolumn{1}{c|}{90.5 $\pm$ 0.76} & 92.7 $\pm$ 0.97 & \multicolumn{1}{c}{91.7 $\pm$ 2.09} \\
Shirt & \multicolumn{1}{c|}{75.5 $\pm$ 1.02} & 73.0 $\pm$ 2.40 & \multicolumn{1}{c|}{74.5 $\pm$ 3.16} & 78.4 $\pm$ 0.74 & \multicolumn{1}{c}{74.7 $\pm$ 1.58} \\
Sneaker & \multicolumn{1}{c|}{96.6 $\pm$ 0.53} & 96.0 $\pm$ 0.65 & \multicolumn{1}{c|}{97.1 $\pm$ 0.49} & 95.1 $\pm$ 0.69 & \multicolumn{1}{c}{95.0 $\pm$ 0.31} \\
Bag & \multicolumn{1}{c|}{88.9 $\pm$ 2.67} & 75.9 $\pm$ 5.94 & \multicolumn{1}{c|}{78.3 $\pm$ 3.86} & 88.0 $\pm$ 0.53 & \multicolumn{1}{c}{89.4 $\pm$ 2.01} \\
Ankle boot & \multicolumn{1}{c|}{98.2 $\pm$ 0.23} & 91.6 $\pm$ 1.91 & \multicolumn{1}{c|}{94.0 $\pm$ 1.36} & 88.9 $\pm$ 1.78 & \multicolumn{1}{c}{89.8 $\pm$ 1.79} \\
\midrule
Mean & \multicolumn{1}{c|}{86.3 $\pm$ 8.06} & 86.4 $\pm$ 7.44 & \multicolumn{1}{c|}{87.3 $\pm$ 7.38} & 87.7 $\pm$ 5.76 & \multicolumn{1}{c}{86.4 $\pm$ 6.84} \\
\bottomrule
\end{tabular}
    
\end{table*}
%%%%%%%%%%%%%%%%%%%%%%%%%%%%%%%%%%%%%%%%%%%%%%%%%%%%%%%%%%%%
%%%%%%%%%%%%%%%%%%%%%%%%%%%%%%%%%%%%%%%%%%%%%%%%%%%%%%%%%%%%
\begin{table*}[th]
    \caption{Mean AUC detection performance in \% (over 10 seeds) for all individual classes for our end-to-end implementations on the MNIST one vs.~rest benchmark with EMNIST OE from Section \ref{sec:exp_sota_without_transfer}. }
    \label{tab:mnist_one_vs_rest_wo_transfer}
    \vspace{0.5em}
    \centering\footnotesize
    
\begin{tabular}{lccccc}
\toprule
& \multicolumn{1}{c|}{Unsupervised} & \multicolumn{2}{c|}{Unsupervised OE} & \multicolumn{2}{c}{Supervised OE} \\
Class &   \multicolumn{1}{c|}{DSVDD} & DSAD & \multicolumn{1}{c|}{HSC} & Focal & \multicolumn{1}{c}{BCE} \\
\midrule
Zero & \multicolumn{1}{c|}{99.0 $\pm$ 0.27} & 99.3 $\pm$ 0.39 & \multicolumn{1}{c|}{99.6 $\pm$ 0.03} & 99.4 $\pm$ 0.34 & \multicolumn{1}{c}{99.4 $\pm$ 0.16} \\
One & \multicolumn{1}{c|}{99.8 $\pm$ 0.02} & 99.7 $\pm$ 0.12 & \multicolumn{1}{c|}{99.9 $\pm$ 0.02} & 99.7 $\pm$ 0.18 & \multicolumn{1}{c}{99.3 $\pm$ 0.34} \\
Two & \multicolumn{1}{c|}{92.3 $\pm$ 1.17} & 98.0 $\pm$ 0.12 & \multicolumn{1}{c|}{98.0 $\pm$ 0.49} & 97.6 $\pm$ 0.68 & \multicolumn{1}{c}{97.5 $\pm$ 0.23} \\
Three & \multicolumn{1}{c|}{94.9 $\pm$ 0.36} & 98.6 $\pm$ 0.35 & \multicolumn{1}{c|}{98.5 $\pm$ 0.94} & 99.5 $\pm$ 0.10 & \multicolumn{1}{c}{99.6 $\pm$ 0.13} \\
Four & \multicolumn{1}{c|}{94.4 $\pm$ 0.50} & 96.5 $\pm$ 1.02 & \multicolumn{1}{c|}{96.7 $\pm$ 0.74} & 97.5 $\pm$ 1.01 & \multicolumn{1}{c}{98.2 $\pm$ 0.52} \\
Five & \multicolumn{1}{c|}{94.2 $\pm$ 1.35} & 97.6 $\pm$ 0.24 & \multicolumn{1}{c|}{97.8 $\pm$ 0.90} & 97.9 $\pm$ 0.25 & \multicolumn{1}{c}{97.3 $\pm$ 0.69} \\
Six & \multicolumn{1}{c|}{99.0 $\pm$ 0.09} & 99.6 $\pm$ 0.08 & \multicolumn{1}{c|}{99.6 $\pm$ 0.08} & 99.7 $\pm$ 0.07 & \multicolumn{1}{c}{99.7 $\pm$ 0.07} \\
Seven & \multicolumn{1}{c|}{96.6 $\pm$ 0.57} & 97.1 $\pm$ 0.39 & \multicolumn{1}{c|}{97.4 $\pm$ 0.63} & 97.6 $\pm$ 0.65 & \multicolumn{1}{c}{97.7 $\pm$ 0.45} \\
Eight & \multicolumn{1}{c|}{90.9 $\pm$ 1.47} & 93.5 $\pm$ 2.49 & \multicolumn{1}{c|}{96.6 $\pm$ 0.81} & 97.3 $\pm$ 0.32 & \multicolumn{1}{c}{97.3 $\pm$ 0.57} \\
Nine & \multicolumn{1}{c|}{96.9 $\pm$ 0.36} & 96.3 $\pm$ 1.06 & \multicolumn{1}{c|}{97.6 $\pm$ 0.27} & 97.8 $\pm$ 0.31 & \multicolumn{1}{c}{97.7 $\pm$ 0.57} \\
\midrule
Mean & \multicolumn{1}{c|}{95.8 $\pm$ 2.83} & 97.6 $\pm$ 1.81 & \multicolumn{1}{c|}{98.2 $\pm$ 1.14} & 98.4 $\pm$ 0.97 & \multicolumn{1}{c}{98.4 $\pm$ 0.95} \\
\bottomrule
\end{tabular}
    
\end{table*}
%%%%%%%%%%%%%%%%%%%%%%%%%%%%%%%%%%%%%%%%%%%%%%%%%%%%%%%%%%%%
%%%%%%%%%%%%%%%%%%%%%%%%%%%%%%%%%%%%%%%%%%%%%%%%%%%%%%%%%%%%%%%%%%%%%%%%%%%%%%%%
%%%%%%%%%%%%%%%%%%%%%%%%%%%%%%%%%%%%%%%%%%%%%%%%%%%%%%%%%%%%%%%%%%%%%%%%%%%%%%%%
%%%%%%%%%%%%%%%%%%%%%%%%%%%%%%%%%%%%%%%%%%%%%%%%%%%%%%%%%%%%%%%%%%%%%%%%%%%%%%%%
%%%%%%%%%%%%%%%%%%%%%%%%%%%%%%%%%%%%%%%%%%%%%%%%%%%%%%%%%%%%%%%%%%%%%%%%%%%%%%%%

\clearpage
\subsection{Image AD with transfer learning-based methods on the one vs.~rest benchmark}
\begin{figure}[hbt] 
  \begin{center} 
      \includegraphics[width=0.99\textwidth]{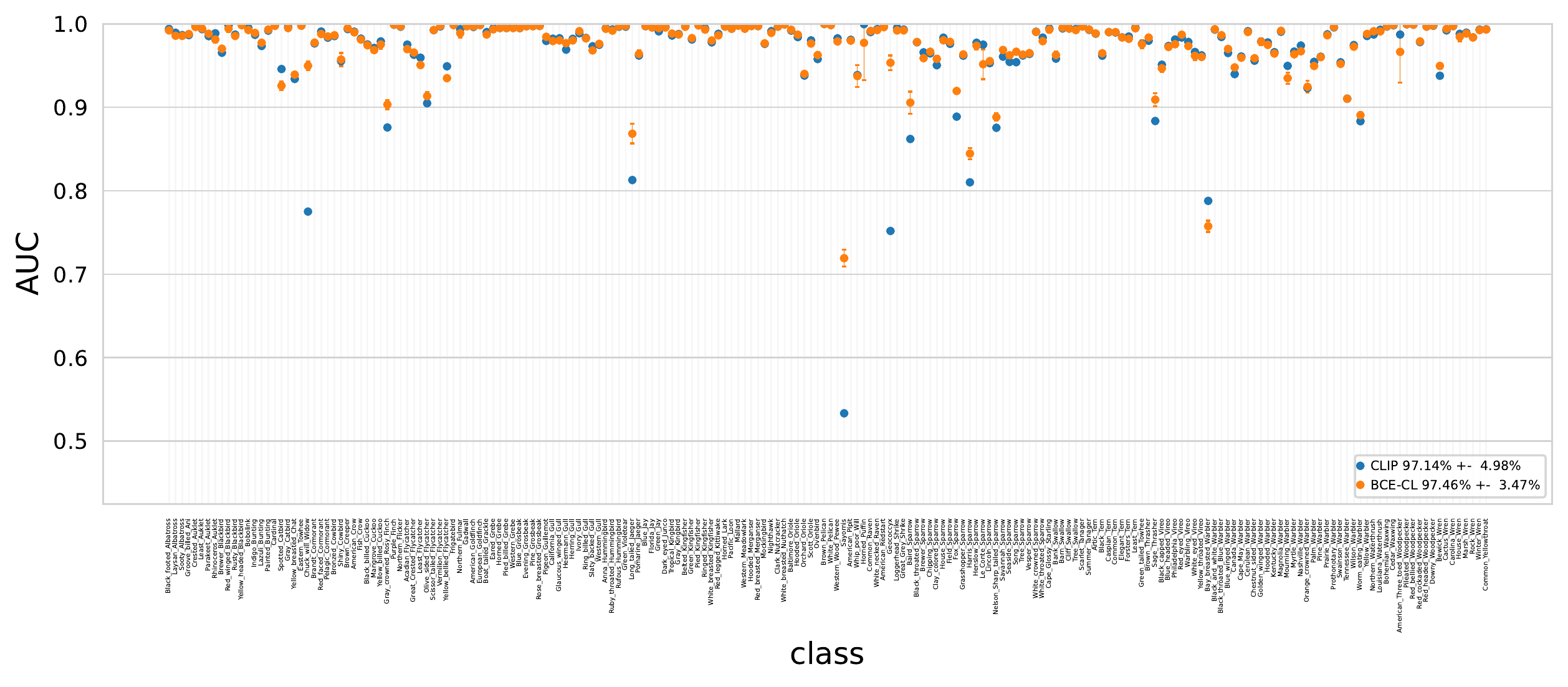}
      \caption{Mean AUC detection performance in \% (over 10 seeds) for all individual classes for our transfer learning-based implementations on the CUB one vs.~rest benchmark with ImageNet-22k OE from Section \ref{sec:exp_sota_with_transfer}.}
      \label{fig:auc_summary_cub_transfer}
      % \vspace{5em}
  \end{center}
\end{figure}

\begin{figure}[hbt] 
  \begin{center} 
      \includegraphics[width=0.99\textwidth]{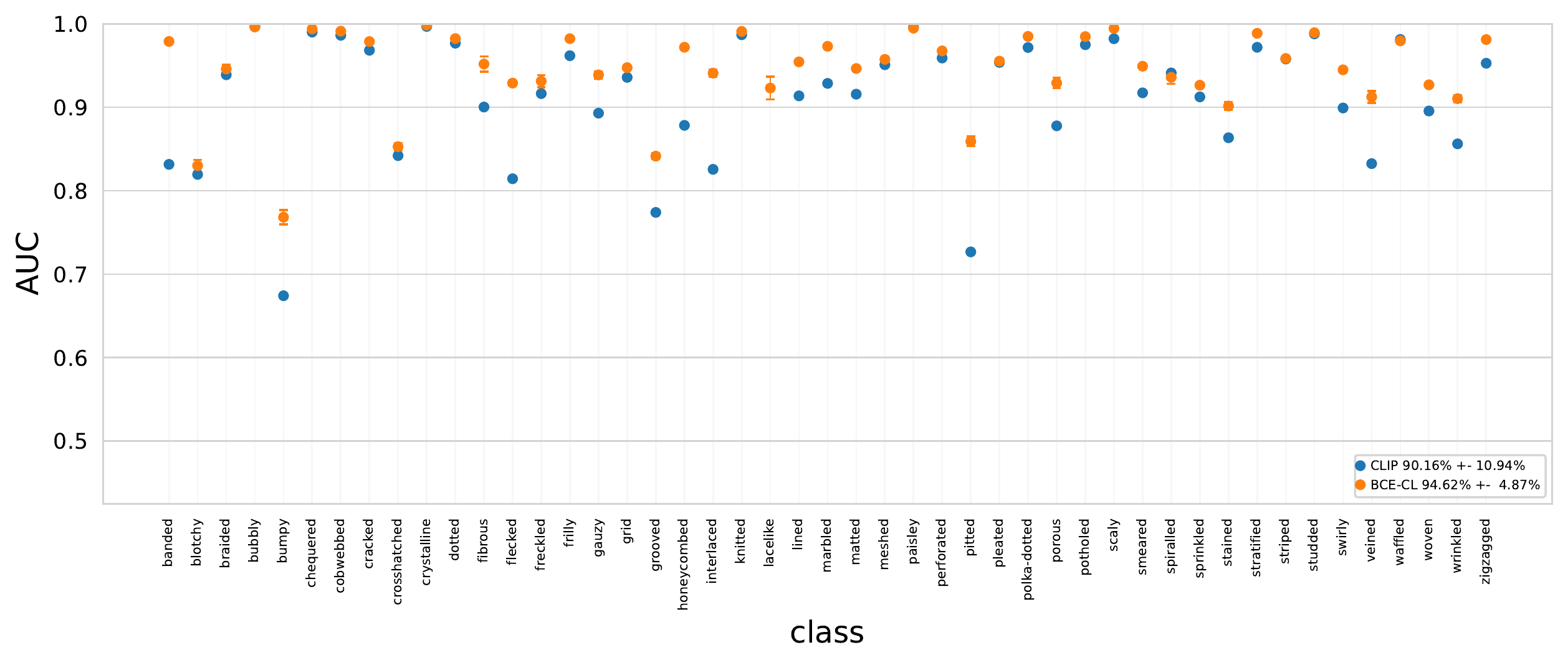}
      \caption{Mean AUC detection performance in \% (over 10 seeds) for all individual classes for our transfer learning-based implementations on the DTD one vs.~rest benchmark with ImageNet-22k OE from Section \ref{sec:exp_sota_with_transfer}.}
      \label{fig:auc_summary_dtd_transfer}
  \end{center}
\end{figure}

Here we report results for the one vs.~rest benchmark experiments from Section \ref{sec:exp_sota_with_transfer} for all individual classes and \emph{transfer learning-based} methods. 
For CIFAR-10, we report results in Table \ref{tab:cifar10_classes_transfer}, including results from the literature (left) and results with standard deviations for our implementation (right). 
For ImageNet-30, Fashion-MNIST, and MNIST, we report results with standard deviations for our implementations in Tables \ref{tab:imagenet_classes_transfer}, \ref{tab:fmnist_one_vs_rest_transfer}, and \ref{tab:mnist_one_vs_rest_transfer}, respectively. 
For CUB and DTD, we report class-wise results instead in form of a plot in Figures \ref{fig:auc_summary_cub_transfer} and \ref{fig:auc_summary_dtd_transfer}, as there are too many classes (47 for DTD and 200 for CUB) to report concisely in a Table. 

%%%%%%%%%%%%%%%%%%%%%%%%%%%%%%%%%%%%%%%%%%%%%%%%%%%%%%%%%%%%
\begin{table*}[th]
    \caption{Left: Mean AUC detection performance in \% (over 10 seeds) for all individual classes and transfer learning-based methods on the CIFAR-10 one vs.~rest benchmark with 80MTI OE from Section \ref{sec:exp_sota_with_transfer}. Right: Our transfer learning-based implementations for the same setup.}
    \label{tab:cifar10_classes_transfer}
    \vspace{0.5em}
    \centering
    \resizebox{!}{0.115\textheight}{% \begin{tabular}{lcccccc} 
\begin{tabular}{lcccccc} 
\toprule 
 & \multicolumn{3}{c|}{Unsupervised} & \multicolumn{3}{c}{Supervised OE} \\ 
Class & DN2* & PANDA* & \multicolumn{1}{c|}{CLIP} & PANDA* & ADIP* & \multicolumn{1}{c}{BCE-CL} \\ 
\midrule 
Airplane & 93.9 & $\times$ & \multicolumn{1}{c|}{99.4} & $\times$ & 99.2 & \multicolumn{1}{c}{\bf 99.7} \\ 
Automobile & 97.7 & $\times$ & \multicolumn{1}{c|}{99.4} & $\times$ & \bf 99.8 & \multicolumn{1}{c}{\bf 99.8} \\ 
Bird & 85.5 & $\times$ & \multicolumn{1}{c|}{97.4} & $\times$ & 98.6 & \multicolumn{1}{c}{\bf 99.2} \\ 
Cat & 85.5 & $\times$ & \multicolumn{1}{c|}{97.0} & $\times$ & 97.0 & \multicolumn{1}{c}{\bf 98.8} \\ 
Deer & 93.6 & $\times$ & \multicolumn{1}{c|}{98.1} & $\times$ & 99.3 & \multicolumn{1}{c}{\bf 99.6} \\ 
Dog & 91.3 & $\times$ & \multicolumn{1}{c|}{97.4} & $\times$ & 98.2 & \multicolumn{1}{c}{\bf 99.2} \\ 
Frog & 94.3 & $\times$ & \multicolumn{1}{c|}{98.1} & $\times$ & 99.6 & \multicolumn{1}{c}{\bf 99.9} \\ 
Horse & 93.6 & $\times$ & \multicolumn{1}{c|}{99.0} & $\times$ & \bf 99.8 & \multicolumn{1}{c}{\bf 99.8} \\ 
Ship & 95.1 & $\times$ & \multicolumn{1}{c|}{99.7} & $\times$ & 99.6 & \multicolumn{1}{c}{\bf 99.8} \\ 
Truck & 95.3 & $\times$ & \multicolumn{1}{c|}{99.3} & $\times$ & 99.5 & \multicolumn{1}{c}{\bf 99.9} \\ 
\midrule 
Mean AUC & 92.5 & 96.2 & \multicolumn{1}{c|}{98.5} & 98.9 & 99.1 & \multicolumn{1}{c}{\bf 99.6} \\ 
\bottomrule 
\end{tabular}
}
    \resizebox{!}{0.115\textheight}{\begin{tabular}{lcc} 
\toprule 
 & \multicolumn{1}{c|}{Unsupervised} & \multicolumn{1}{c}{Supervised OE} \\ 
Class & \multicolumn{1}{c|}{CLIP} & \multicolumn{1}{c}{BCE-CL} \\ 
\midrule 
Airplane & \multicolumn{1}{c|}{99.40 $\pm$ 0.00} & \multicolumn{1}{c}{99.74 $\pm$ 0.02} \\ 
Automobile & \multicolumn{1}{c|}{99.37 $\pm$ 0.00} & \multicolumn{1}{c}{99.82 $\pm$ 0.01} \\ 
Bird & \multicolumn{1}{c|}{97.37 $\pm$ 0.00} & \multicolumn{1}{c}{99.29 $\pm$ 0.04} \\ 
Cat & \multicolumn{1}{c|}{96.99 $\pm$ 0.00} & \multicolumn{1}{c}{98.86 $\pm$ 0.04} \\ 
Deer & \multicolumn{1}{c|}{98.05 $\pm$ 0.00} & \multicolumn{1}{c}{99.62 $\pm$ 0.02} \\ 
Dog & \multicolumn{1}{c|}{97.36 $\pm$ 0.00} & \multicolumn{1}{c}{99.26 $\pm$ 0.04} \\ 
Frog & \multicolumn{1}{c|}{98.05 $\pm$ 0.00} & \multicolumn{1}{c}{99.89 $\pm$ 0.01} \\ 
Horse & \multicolumn{1}{c|}{98.99 $\pm$ 0.00} & \multicolumn{1}{c}{99.84 $\pm$ 0.02} \\ 
Ship & \multicolumn{1}{c|}{99.71 $\pm$ 0.00} & \multicolumn{1}{c}{99.86 $\pm$ 0.01} \\ 
Truck & \multicolumn{1}{c|}{99.30 $\pm$ 0.00} & \multicolumn{1}{c}{99.90 $\pm$ 0.01} \\ 
\midrule 
Mean AUC & \multicolumn{1}{c|}{98.46 $\pm$ 0.96} & \multicolumn{1}{c}{99.61 $\pm$ 0.34} \\ 
\bottomrule 
\end{tabular} }
\end{table*}
%%%%%%%%%%%%%%%%%%%%%%%%%%%%%%%%%%%%%%%%%%%%%%%%%%%%%%%%%%%%
%%%%%%%%%%%%%%%%%%%%%%%%%%%%%%%%%%%%%%%%%%%%%%%%%%%%%%%%%%%%
\begin{table*}[th]
    \caption{Mean AUC detection performance in \% (over 10 seeds) for all individual classes for our transfer learning-based implementations on the Fashion-MNIST one vs.~rest benchmark with grayscale CIFAR-100 OE from Section \ref{sec:exp_sota_with_transfer}. }
    \label{tab:fmnist_one_vs_rest_transfer}
    \vspace{0.5em}
    \centering\small
    
\begin{tabular}{lcc}
\toprule
& \multicolumn{1}{c|}{Unsupervised} &   \multicolumn{1}{c}{Supervised OE} \\
Class &   \multicolumn{1}{c|}{CLIP} &     \multicolumn{1}{c}{BCE-CL} \\
\midrule
Top & \multicolumn{1}{c|}{82.9 $\pm$ 0.00} & \multicolumn{1}{c}{84.7 $\pm$ 0.34} \\
Trouser & \multicolumn{1}{c|}{87.7 $\pm$ 0.00} & \multicolumn{1}{c}{98.5 $\pm$ 0.03} \\
Pullover & \multicolumn{1}{c|}{91.6 $\pm$ 0.00} & \multicolumn{1}{c}{93.8 $\pm$ 0.06} \\
Dress & \multicolumn{1}{c|}{87.0 $\pm$ 0.00} & \multicolumn{1}{c}{97.0 $\pm$ 0.07} \\
Coat & \multicolumn{1}{c|}{91.8 $\pm$ 0.00} & \multicolumn{1}{c}{91.8 $\pm$ 0.34} \\
Sandal & \multicolumn{1}{c|}{78.7 $\pm$ 0.00} & \multicolumn{1}{c}{99.3 $\pm$ 0.02} \\
Shirt & \multicolumn{1}{c|}{81.3 $\pm$ 0.00} & \multicolumn{1}{c}{84.9 $\pm$ 0.22} \\
Sneaker & \multicolumn{1}{c|}{97.3 $\pm$ 0.00} & \multicolumn{1}{c}{97.8 $\pm$ 0.09} \\
Bag & \multicolumn{1}{c|}{92.4 $\pm$ 0.00} & \multicolumn{1}{c}{99.6 $\pm$ 0.02} \\
Ankle boot & \multicolumn{1}{c|}{99.1 $\pm$ 0.00} & \multicolumn{1}{c}{99.4 $\pm$ 0.04} \\
\midrule
Mean & \multicolumn{1}{c|}{89.0 $\pm$ 6.36} & \multicolumn{1}{c}{94.7 $\pm$ 5.49} \\
\bottomrule
\end{tabular}
    
\end{table*}
%%%%%%%%%%%%%%%%%%%%%%%%%%%%%%%%%%%%%%%%%%%%%%%%%%%%%%%%%%%%
%%%%%%%%%%%%%%%%%%%%%%%%%%%%%%%%%%%%%%%%%%%%%%%%%%%%%%%%%%%%
\begin{table*}[th]
    \caption{Mean AUC detection performance in \% (over 10 seeds) for all individual classes for our transfer learning-based implementations on the ImageNet-30 one vs.~rest benchmark with ImageNet-22K OE from Section \ref{sec:exp_sota_with_transfer}. }
    \label{tab:imagenet_classes_transfer}
    \vspace{0.5em}
    \centering\small
    \begin{tabular}{lcc} 
\toprule 
 & \multicolumn{1}{c|}{Unsupervised} & \multicolumn{1}{c}{Supervised OE} \\ 
Class & \multicolumn{1}{c|}{CLIP} & \multicolumn{1}{c}{BCE-CL} \\ 
\midrule 
Acorn & \multicolumn{1}{c|}{99.78 $\pm$ 0.00} & \multicolumn{1}{c}{99.96 $\pm$ 0.01} \\ 
Airliner & \multicolumn{1}{c|}{100.00 $\pm$ 0.00} & \multicolumn{1}{c}{100.00 $\pm$ 0.00} \\ 
Ambulance & \multicolumn{1}{c|}{100.00 $\pm$ 0.00} & \multicolumn{1}{c}{100.00 $\pm$ 0.00} \\ 
American alligator & \multicolumn{1}{c|}{99.98 $\pm$ 0.00} & \multicolumn{1}{c}{100.00 $\pm$ 0.00} \\ 
Banjo & \multicolumn{1}{c|}{100.00 $\pm$ 0.00} & \multicolumn{1}{c}{100.00 $\pm$ 0.00} \\ 
Barn & \multicolumn{1}{c|}{100.00 $\pm$ 0.00} & \multicolumn{1}{c}{100.00 $\pm$ 0.00} \\ 
Bikini & \multicolumn{1}{c|}{99.87 $\pm$ 0.00} & \multicolumn{1}{c}{100.00 $\pm$ 0.00} \\ 
Digital clock & \multicolumn{1}{c|}{99.69 $\pm$ 0.00} & \multicolumn{1}{c}{99.93 $\pm$ 0.02} \\ 
Dragonfly & \multicolumn{1}{c|}{100.00 $\pm$ 0.00} & \multicolumn{1}{c}{100.00 $\pm$ 0.00} \\ 
Dumbbell & \multicolumn{1}{c|}{99.91 $\pm$ 0.00} & \multicolumn{1}{c}{99.97 $\pm$ 0.02} \\ 
Forklift & \multicolumn{1}{c|}{100.00 $\pm$ 0.00} & \multicolumn{1}{c}{100.00 $\pm$ 0.00} \\ 
Goblet & \multicolumn{1}{c|}{99.29 $\pm$ 0.00} & \multicolumn{1}{c}{99.81 $\pm$ 0.04} \\ 
Grand piano & \multicolumn{1}{c|}{100.00 $\pm$ 0.00} & \multicolumn{1}{c}{98.36 $\pm$ 4.87} \\ 
Hotdog & \multicolumn{1}{c|}{99.99 $\pm$ 0.00} & \multicolumn{1}{c}{100.00 $\pm$ 0.00} \\ 
Hourglass & \multicolumn{1}{c|}{99.69 $\pm$ 0.00} & \multicolumn{1}{c}{99.97 $\pm$ 0.02} \\ 
Manhole cover & \multicolumn{1}{c|}{100.00 $\pm$ 0.00} & \multicolumn{1}{c}{100.00 $\pm$ 0.00} \\ 
Mosque & \multicolumn{1}{c|}{100.00 $\pm$ 0.00} & \multicolumn{1}{c}{100.00 $\pm$ 0.00} \\ 
Nail & \multicolumn{1}{c|}{99.61 $\pm$ 0.00} & \multicolumn{1}{c}{99.97 $\pm$ 0.01} \\ 
Parking meter & \multicolumn{1}{c|}{99.52 $\pm$ 0.00} & \multicolumn{1}{c}{99.97 $\pm$ 0.01} \\ 
Pillow & \multicolumn{1}{c|}{99.95 $\pm$ 0.00} & \multicolumn{1}{c}{100.00 $\pm$ 0.00} \\ 
Revolver & \multicolumn{1}{c|}{100.00 $\pm$ 0.00} & \multicolumn{1}{c}{100.00 $\pm$ 0.00} \\ 
Rotary dial telephone & \multicolumn{1}{c|}{99.84 $\pm$ 0.00} & \multicolumn{1}{c}{99.21 $\pm$ 1.75} \\ 
Schooner & \multicolumn{1}{c|}{100.00 $\pm$ 0.00} & \multicolumn{1}{c}{100.00 $\pm$ 0.00} \\ 
Snowmobile & \multicolumn{1}{c|}{99.99 $\pm$ 0.00} & \multicolumn{1}{c}{100.00 $\pm$ 0.00} \\ 
Soccer ball & \multicolumn{1}{c|}{99.97 $\pm$ 0.00} & \multicolumn{1}{c}{99.98 $\pm$ 0.05} \\ 
Stingray & \multicolumn{1}{c|}{100.00 $\pm$ 0.00} & \multicolumn{1}{c}{100.00 $\pm$ 0.00} \\ 
Strawberry & \multicolumn{1}{c|}{99.81 $\pm$ 0.00} & \multicolumn{1}{c}{99.97 $\pm$ 0.07} \\ 
Tank & \multicolumn{1}{c|}{100.00 $\pm$ 0.00} & \multicolumn{1}{c}{100.00 $\pm$ 0.00} \\ 
Toaster & \multicolumn{1}{c|}{99.44 $\pm$ 0.00} & \multicolumn{1}{c}{99.88 $\pm$ 0.01} \\ 
Volcano & \multicolumn{1}{c|}{100.00 $\pm$ 0.00} & \multicolumn{1}{c}{99.99 $\pm$ 0.04} \\ 
\midrule 
Mean AUC & \multicolumn{1}{c|}{99.88 $\pm$ 0.19} & \multicolumn{1}{c}{99.90 $\pm$ 0.32} \\ 
\bottomrule 
\end{tabular}

\end{table*}
%%%%%%%%%%%%%%%%%%%%%%%%%%%%%%%%%%%%%%%%%%%%%%%%%%%%%%%%%%%%
%%%%%%%%%%%%%%%%%%%%%%%%%%%%%%%%%%%%%%%%%%%%%%%%%%%%%%%%%%%%
\begin{table*}[th]
    \caption{Mean AUC detection performance in \% (over 10 seeds) for all individual classes for our transfer learning-based implementations on the MNIST one vs.~rest benchmark with EMNIST OE from Section \ref{sec:exp_sota_without_transfer}. }
    \label{tab:mnist_one_vs_rest_transfer}
    \vspace{0.5em}
    \centering\small
    
\begin{tabular}{lcc}
\toprule
& \multicolumn{1}{c|}{Unsupervised} &   \multicolumn{1}{c}{Supervised OE} \\
Class &   \multicolumn{1}{c|}{CLIP} &     \multicolumn{1}{c}{BCE-CL} \\
\midrule
Zero & \multicolumn{1}{c|}{73.8 $\pm$ 0.00} & \multicolumn{1}{c}{98.2 $\pm$ 2.22} \\
One & \multicolumn{1}{c|}{38.3 $\pm$ 0.00} & \multicolumn{1}{c}{99.6 $\pm$ 0.08} \\
Two & \multicolumn{1}{c|}{69.8 $\pm$ 0.00} & \multicolumn{1}{c}{89.6 $\pm$ 11.61} \\
Three & \multicolumn{1}{c|}{85.2 $\pm$ 0.00} & \multicolumn{1}{c}{99.4 $\pm$ 0.06} \\
Four & \multicolumn{1}{c|}{64.8 $\pm$ 0.00} & \multicolumn{1}{c}{98.4 $\pm$ 0.15} \\
Five & \multicolumn{1}{c|}{49.4 $\pm$ 0.00} & \multicolumn{1}{c}{96.6 $\pm$ 0.33} \\
Six & \multicolumn{1}{c|}{24.7 $\pm$ 0.00} & \multicolumn{1}{c}{88.5 $\pm$ 21.08} \\
Seven & \multicolumn{1}{c|}{77.9 $\pm$ 0.00} & \multicolumn{1}{c}{97.8 $\pm$ 0.06} \\
Eight & \multicolumn{1}{c|}{40.0 $\pm$ 0.00} & \multicolumn{1}{c}{95.6 $\pm$ 0.37} \\
Nine & \multicolumn{1}{c|}{66.1 $\pm$ 0.00} & \multicolumn{1}{c}{96.9 $\pm$ 0.43} \\
\midrule
Mean & \multicolumn{1}{c|}{59.0 $\pm$ 18.76} & \multicolumn{1}{c}{96.0 $\pm$ 3.70} \\
\bottomrule
\end{tabular}
    
\end{table*}
%%%%%%%%%%%%%%%%%%%%%%%%%%%%%%%%%%%%%%%%%%%%%%%%%%%%%%%%%%%%
%%%%%%%%%%%%%%%%%%%%%%%%%%%%%%%%%%%%%%%%%%%%%%%%%%%%%%%%%%%%%%%%%%%%%%%%%%%%%%%%
%%%%%%%%%%%%%%%%%%%%%%%%%%%%%%%%%%%%%%%%%%%%%%%%%%%%%%%%%%%%%%%%%%%%%%%%%%%%%%%%
%%%%%%%%%%%%%%%%%%%%%%%%%%%%%%%%%%%%%%%%%%%%%%%%%%%%%%%%%%%%%%%%%%%%%%%%%%%%%%%%
%%%%%%%%%%%%%%%%%%%%%%%%%%%%%%%%%%%%%%%%%%%%%%%%%%%%%%%%%%%%%%%%%%%%%%%%%%%%%%%%

\clearpage
\subsection{Image AD on the leave-one-class-out benchmark} 
\label{appx:section_loco_results}
In the main paper, we evaluate our methods on the ubiquitous one vs.~rest image-AD benchmark (see Section \ref{sec:experimental_setups}).
In Section \ref{sec:exp_var_oe_size} we also present results on the more uncommon leave-one-class-out benchmark when varying the OE training set size.
Here, we present results when using the full OE training set size, analogue to the experiments in Sections \ref{sec:exp_sota_without_transfer} and \ref{sec:exp_sota_with_transfer}.
Tables \ref{tab:appx_cifar_loco} and \ref{tab:appx_imagenet_loco} show results for leave-one-class-out CIFAR-10 and leave-one-class-out ImageNet-30, respectively.

Note that, to apply CLIP to the leave-one-class-out setting with $K$ normal classes, we use the text tuple $(v_1, \dots, v_{K+1}) =$ (``a photo of a \{NORMAL\_CLASS\_1\}'', ..., ``a photo of a \{NORMAL\_CLASS\_{K}\}'', ``a photo of something'').
For a test image $\bx$, we compute its anomaly score as
\begin{equation} \label{eq:appx_clip_ad_loco}
    s(\bx) = \frac{\exp(\left\langle f_u(\bx), f_v(v_{K+1}) \right\rangle \cdot 100)}{\sum_{k=1}^{K+1} \exp(\left\langle f_u(\bx), f_v(v_k) \right\rangle \cdot 100)}.
\end{equation}

%%%%%%%%%%%%%%%%%%%%%%%%%%%%%%%%%%%%%%%%%%%%%%%%%%%%%%%%%%%%
\begin{table*}[th]
    \caption{Mean AUC detection performance in \% (over 2 seeds) with standard deviations for all individual classes for our implementations on the ImageNet-30 leave-one-class-out AD benchmark with ImageNet-22k (with ImageNet-1k removed) OE. }
    \label{tab:appx_imagenet_loco}
    \vspace{0.5em}
    \centering\footnotesize
    
\begin{tabular}{lccccccc}
\toprule
& \multicolumn{2}{c|}{Unsupervised} & \multicolumn{2}{c|}{Unsupervised OE} & \multicolumn{3}{c}{Supervised OE} \\
Anomaly class & DSVDD & \multicolumn{1}{c|}{CLIP} & DSAD & \multicolumn{1}{c|}{HSC} & Focal & BCE & \multicolumn{1}{c}{BCE-CL} \\
\midrule
Acorn & 49.7 $\pm$ 1.02 & \multicolumn{1}{c|}{98.8 $\pm$ 0.00} & 92.0 $\pm$ 0.21 & \multicolumn{1}{c|}{92.6 $\pm$ 0.54} & 92.2 $\pm$ 0.51 & 90.4 $\pm$ 0.02 & \multicolumn{1}{c}{99.7 $\pm$ 0.05} \\
Airliner & 47.6 $\pm$ 3.67 & \multicolumn{1}{c|}{99.9 $\pm$ 0.00} & 93.8 $\pm$ 0.89 & \multicolumn{1}{c|}{92.9 $\pm$ 1.71} & 94.7 $\pm$ 0.22 & 93.9 $\pm$ 0.09 & \multicolumn{1}{c}{99.9 $\pm$ 0.01} \\
Ambulance & 50.5 $\pm$ 1.57 & \multicolumn{1}{c|}{99.7 $\pm$ 0.00} & 93.4 $\pm$ 0.97 & \multicolumn{1}{c|}{92.3 $\pm$ 0.92} & 93.1 $\pm$ 1.46 & 93.2 $\pm$ 1.66 & \multicolumn{1}{c}{99.9 $\pm$ 0.02} \\
American alligator & 53.4 $\pm$ 4.66 & \multicolumn{1}{c|}{98.6 $\pm$ 0.00} & 91.7 $\pm$ 0.89 & \multicolumn{1}{c|}{91.3 $\pm$ 0.12} & 91.5 $\pm$ 1.01 & 92.1 $\pm$ 0.01 & \multicolumn{1}{c}{99.7 $\pm$ 0.07} \\
Banjo & 50.3 $\pm$ 1.30 & \multicolumn{1}{c|}{99.4 $\pm$ 0.00} & 85.3 $\pm$ 0.98 & \multicolumn{1}{c|}{85.6 $\pm$ 0.97} & 85.3 $\pm$ 0.54 & 86.0 $\pm$ 0.04 & \multicolumn{1}{c}{99.9 $\pm$ 0.02} \\
Barn & 51.5 $\pm$ 0.96 & \multicolumn{1}{c|}{98.3 $\pm$ 0.00} & 92.0 $\pm$ 0.73 & \multicolumn{1}{c|}{90.6 $\pm$ 0.95} & 91.3 $\pm$ 0.17 & 89.3 $\pm$ 2.04 & \multicolumn{1}{c}{99.8 $\pm$ 0.05} \\
Bikini & 50.6 $\pm$ 1.03 & \multicolumn{1}{c|}{97.9 $\pm$ 0.00} & 89.8 $\pm$ 0.04 & \multicolumn{1}{c|}{90.0 $\pm$ 0.49} & 87.4 $\pm$ 0.72 & 90.6 $\pm$ 0.44 & \multicolumn{1}{c}{99.7 $\pm$ 0.03} \\
Digital clock & 49.1 $\pm$ 1.33 & \multicolumn{1}{c|}{94.9 $\pm$ 0.00} & 86.0 $\pm$ 0.42 & \multicolumn{1}{c|}{84.8 $\pm$ 1.80} & 84.7 $\pm$ 0.79 & 85.1 $\pm$ 0.83 & \multicolumn{1}{c}{98.5 $\pm$ 0.01} \\
Dragonfly & 51.2 $\pm$ 3.23 & \multicolumn{1}{c|}{99.8 $\pm$ 0.00} & 94.2 $\pm$ 0.01 & \multicolumn{1}{c|}{93.9 $\pm$ 0.24} & 93.8 $\pm$ 0.58 & 93.5 $\pm$ 0.72 & \multicolumn{1}{c}{99.9 $\pm$ 0.01} \\
Dumbbell & 46.1 $\pm$ 2.87 & \multicolumn{1}{c|}{97.7 $\pm$ 0.00} & 80.7 $\pm$ 1.12 & \multicolumn{1}{c|}{82.7 $\pm$ 0.78} & 81.7 $\pm$ 1.39 & 83.3 $\pm$ 0.71 & \multicolumn{1}{c}{99.4 $\pm$ 0.07} \\
Forklift & 50.6 $\pm$ 1.20 & \multicolumn{1}{c|}{96.8 $\pm$ 0.00} & 82.1 $\pm$ 0.37 & \multicolumn{1}{c|}{82.0 $\pm$ 0.42} & 82.9 $\pm$ 0.34 & 81.9 $\pm$ 0.71 & \multicolumn{1}{c}{99.6 $\pm$ 0.12} \\
Goblet & 46.8 $\pm$ 0.13 & \multicolumn{1}{c|}{96.3 $\pm$ 0.00} & 84.2 $\pm$ 2.64 & \multicolumn{1}{c|}{83.9 $\pm$ 0.15} & 84.4 $\pm$ 1.30 & 83.5 $\pm$ 0.99 & \multicolumn{1}{c}{98.8 $\pm$ 0.14} \\
Grand piano & 46.2 $\pm$ 2.79 & \multicolumn{1}{c|}{99.5 $\pm$ 0.00} & 82.2 $\pm$ 1.00 & \multicolumn{1}{c|}{82.4 $\pm$ 0.01} & 83.4 $\pm$ 0.71 & 82.6 $\pm$ 1.08 & \multicolumn{1}{c}{99.9 $\pm$ 0.02} \\
Hotdog & 50.4 $\pm$ 1.74 & \multicolumn{1}{c|}{99.7 $\pm$ 0.00} & 93.2 $\pm$ 0.62 & \multicolumn{1}{c|}{93.9 $\pm$ 0.49} & 94.3 $\pm$ 0.27 & 94.2 $\pm$ 0.23 & \multicolumn{1}{c}{99.7 $\pm$ 0.03} \\
Hourglass & 50.9 $\pm$ 2.30 & \multicolumn{1}{c|}{87.3 $\pm$ 0.00} & 87.1 $\pm$ 0.89 & \multicolumn{1}{c|}{85.9 $\pm$ 0.62} & 86.9 $\pm$ 0.71 & 85.5 $\pm$ 1.14 & \multicolumn{1}{c}{96.0 $\pm$ 0.29} \\
Manhole cover & 51.8 $\pm$ 5.59 & \multicolumn{1}{c|}{93.0 $\pm$ 0.00} & 84.7 $\pm$ 1.74 & \multicolumn{1}{c|}{84.9 $\pm$ 1.47} & 87.8 $\pm$ 0.48 & 88.3 $\pm$ 0.09 & \multicolumn{1}{c}{99.6 $\pm$ 0.04} \\
Mosque & 49.4 $\pm$ 6.93 & \multicolumn{1}{c|}{99.7 $\pm$ 0.00} & 91.7 $\pm$ 0.04 & \multicolumn{1}{c|}{89.4 $\pm$ 1.22} & 88.1 $\pm$ 0.84 & 89.1 $\pm$ 0.29 & \multicolumn{1}{c}{99.9 $\pm$ 0.06} \\
Nail & 49.3 $\pm$ 0.87 & \multicolumn{1}{c|}{97.1 $\pm$ 0.00} & 86.2 $\pm$ 0.44 & \multicolumn{1}{c|}{87.7 $\pm$ 0.74} & 86.6 $\pm$ 0.08 & 85.4 $\pm$ 1.49 & \multicolumn{1}{c}{97.0 $\pm$ 0.05} \\
Parking meter & 51.1 $\pm$ 2.91 & \multicolumn{1}{c|}{92.4 $\pm$ 0.00} & 85.1 $\pm$ 0.96 & \multicolumn{1}{c|}{82.1 $\pm$ 0.26} & 81.9 $\pm$ 0.87 & 81.3 $\pm$ 0.17 & \multicolumn{1}{c}{96.3 $\pm$ 0.55} \\
Pillow & 47.9 $\pm$ 0.37 & \multicolumn{1}{c|}{99.5 $\pm$ 0.00} & 91.6 $\pm$ 0.65 & \multicolumn{1}{c|}{90.6 $\pm$ 0.84} & 91.0 $\pm$ 0.65 & 90.6 $\pm$ 0.36 & \multicolumn{1}{c}{99.9 $\pm$ 0.02} \\
Revolver & 54.7 $\pm$ 0.47 & \multicolumn{1}{c|}{99.6 $\pm$ 0.00} & 91.3 $\pm$ 0.66 & \multicolumn{1}{c|}{88.9 $\pm$ 0.91} & 89.4 $\pm$ 0.15 & 90.5 $\pm$ 0.65 & \multicolumn{1}{c}{99.8 $\pm$ 0.04} \\
Rotary dial telephone & 50.0 $\pm$ 0.68 & \multicolumn{1}{c|}{97.3 $\pm$ 0.00} & 85.2 $\pm$ 1.38 & \multicolumn{1}{c|}{84.9 $\pm$ 1.09} & 82.9 $\pm$ 0.38 & 82.7 $\pm$ 0.41 & \multicolumn{1}{c}{99.3 $\pm$ 0.10} \\
Schooner & 42.0 $\pm$ 0.25 & \multicolumn{1}{c|}{99.9 $\pm$ 0.00} & 95.6 $\pm$ 0.12 & \multicolumn{1}{c|}{94.2 $\pm$ 0.65} & 95.6 $\pm$ 0.18 & 94.4 $\pm$ 0.61 & \multicolumn{1}{c}{99.9 $\pm$ 0.01} \\
Snowmobile & 48.2 $\pm$ 2.40 & \multicolumn{1}{c|}{99.5 $\pm$ 0.00} & 86.6 $\pm$ 0.01 & \multicolumn{1}{c|}{88.0 $\pm$ 0.12} & 87.5 $\pm$ 1.40 & 86.5 $\pm$ 0.16 & \multicolumn{1}{c}{99.9 $\pm$ 0.01} \\
Soccer ball & 51.1 $\pm$ 0.20 & \multicolumn{1}{c|}{99.4 $\pm$ 0.00} & 90.3 $\pm$ 0.46 & \multicolumn{1}{c|}{89.8 $\pm$ 0.67} & 90.5 $\pm$ 0.51 & 90.2 $\pm$ 0.92 & \multicolumn{1}{c}{99.8 $\pm$ 0.00} \\
Stingray & 48.6 $\pm$ 6.35 & \multicolumn{1}{c|}{97.1 $\pm$ 0.00} & 91.8 $\pm$ 0.17 & \multicolumn{1}{c|}{90.9 $\pm$ 1.21} & 90.8 $\pm$ 0.53 & 90.4 $\pm$ 0.39 & \multicolumn{1}{c}{99.8 $\pm$ 0.01} \\
Strawberry & 52.0 $\pm$ 2.28 & \multicolumn{1}{c|}{99.5 $\pm$ 0.00} & 95.5 $\pm$ 0.95 & \multicolumn{1}{c|}{96.4 $\pm$ 0.34} & 95.5 $\pm$ 0.23 & 95.0 $\pm$ 0.48 & \multicolumn{1}{c}{99.8 $\pm$ 0.01} \\
Tank & 51.8 $\pm$ 2.40 & \multicolumn{1}{c|}{98.4 $\pm$ 0.00} & 86.8 $\pm$ 1.76 & \multicolumn{1}{c|}{82.7 $\pm$ 0.72} & 84.9 $\pm$ 0.19 & 84.6 $\pm$ 1.32 & \multicolumn{1}{c}{99.8 $\pm$ 0.01} \\
Toaster & 48.3 $\pm$ 1.40 & \multicolumn{1}{c|}{97.2 $\pm$ 0.00} & 84.0 $\pm$ 0.08 & \multicolumn{1}{c|}{83.3 $\pm$ 1.03} & 86.1 $\pm$ 0.72 & 84.3 $\pm$ 0.32 & \multicolumn{1}{c}{98.0 $\pm$ 0.19} \\
Volcano & 51.3 $\pm$ 2.07 & \multicolumn{1}{c|}{99.6 $\pm$ 0.00} & 89.2 $\pm$ 1.10 & \multicolumn{1}{c|}{89.2 $\pm$ 0.77} & 88.5 $\pm$ 0.65 & 88.4 $\pm$ 0.94 & \multicolumn{1}{c}{99.9 $\pm$ 0.01} \\
\midrule
Mean & 49.7 $\pm$ 2.41 & \multicolumn{1}{c|}{97.8 $\pm$ 2.73} & 88.8 $\pm$ 4.20 & \multicolumn{1}{c|}{88.3 $\pm$ 4.16} & 88.5 $\pm$ 4.14 & 88.2 $\pm$ 4.12 & \multicolumn{1}{c}{99.3 $\pm$ 1.05} \\
\bottomrule
\end{tabular}
    
\end{table*}
%%%%%%%%%%%%%%%%%%%%%%%%%%%%%%%%%%%%%%%%%%%%%%%%%%%%%%%%%%%%

%%%%%%%%%%%%%%%%%%%%%%%%%%%%%%%%%%%%%%%%%%%%%%%%%%%%%%%%%%%%
\begin{table*}[htb]
    \caption{Mean AUC detection performance in \% (over 2 seeds) with standard deviations for all individual classes for our implementations on the CIFAR-10 leave-one-class-out AD benchmark with 80MTI OE. }
    \label{tab:appx_cifar_loco}
    \vspace{0.5em}
    \centering\small
    \resizebox{0.99\textwidth}{!}{
\begin{tabular}{lccccccc}
\toprule
& \multicolumn{2}{c|}{Unsupervised} & \multicolumn{2}{c|}{Unsupervised OE} & \multicolumn{3}{c}{Supervised OE} \\
Class & DSVDD & \multicolumn{1}{c|}{CLIP} & DSAD & \multicolumn{1}{c|}{HSC} & Focal & BCE & \multicolumn{1}{c}{BCE-CL} \\
\midrule
Airplane & 62.3 $\pm$ 8.90 & \multicolumn{1}{c|}{96.2 $\pm$ 0.00} & 85.5 $\pm$ 0.52 & \multicolumn{1}{c|}{87.0 $\pm$ 0.52} & 88.4 $\pm$ 0.28 & 87.9 $\pm$ 0.17 & \multicolumn{1}{c}{99.2 $\pm$ 0.03} \\
Automobile & 52.7 $\pm$ 0.69 & \multicolumn{1}{c|}{96.0 $\pm$ 0.00} & 85.6 $\pm$ 0.81 & \multicolumn{1}{c|}{87.2 $\pm$ 0.56} & 89.9 $\pm$ 0.71 & 90.3 $\pm$ 0.01 & \multicolumn{1}{c}{99.3 $\pm$ 0.02} \\
Bird & 54.2 $\pm$ 4.00 & \multicolumn{1}{c|}{93.5 $\pm$ 0.00} & 83.6 $\pm$ 0.04 & \multicolumn{1}{c|}{83.8 $\pm$ 0.03} & 86.2 $\pm$ 0.49 & 86.0 $\pm$ 0.45 & \multicolumn{1}{c}{97.9 $\pm$ 0.15} \\
Cat & 39.6 $\pm$ 2.98 & \multicolumn{1}{c|}{90.5 $\pm$ 0.00} & 82.4 $\pm$ 0.10 & \multicolumn{1}{c|}{82.4 $\pm$ 0.80} & 83.4 $\pm$ 0.52 & 84.1 $\pm$ 0.36 & \multicolumn{1}{c}{97.9 $\pm$ 0.01} \\
Deer & 56.6 $\pm$ 0.49 & \multicolumn{1}{c|}{79.5 $\pm$ 0.00} & 75.6 $\pm$ 0.11 & \multicolumn{1}{c|}{74.9 $\pm$ 0.06} & 76.9 $\pm$ 0.42 & 77.0 $\pm$ 0.13 & \multicolumn{1}{c}{96.6 $\pm$ 0.03} \\
Dog & 49.9 $\pm$ 4.06 & \multicolumn{1}{c|}{90.6 $\pm$ 0.00} & 82.7 $\pm$ 0.15 & \multicolumn{1}{c|}{82.8 $\pm$ 0.09} & 84.1 $\pm$ 0.75 & 84.6 $\pm$ 0.03 & \multicolumn{1}{c}{97.9 $\pm$ 0.00} \\
Frog & 53.2 $\pm$ 11.75 & \multicolumn{1}{c|}{94.0 $\pm$ 0.00} & 85.1 $\pm$ 0.10 & \multicolumn{1}{c|}{85.5 $\pm$ 0.35} & 86.1 $\pm$ 0.28 & 86.4 $\pm$ 0.31 & \multicolumn{1}{c}{98.3 $\pm$ 0.12} \\
Horse & 47.7 $\pm$ 0.69 & \multicolumn{1}{c|}{93.4 $\pm$ 0.00} & 85.2 $\pm$ 0.57 & \multicolumn{1}{c|}{85.9 $\pm$ 0.08} & 88.4 $\pm$ 0.04 & 88.2 $\pm$ 0.52 & \multicolumn{1}{c}{98.6 $\pm$ 0.09} \\
Ship & 54.7 $\pm$ 5.87 & \multicolumn{1}{c|}{97.8 $\pm$ 0.00} & 89.7 $\pm$ 0.12 & \multicolumn{1}{c|}{90.7 $\pm$ 0.06} & 91.3 $\pm$ 0.17 & 92.0 $\pm$ 0.35 & \multicolumn{1}{c}{99.4 $\pm$ 0.03} \\
Truck & 51.1 $\pm$ 1.86 & \multicolumn{1}{c|}{90.0 $\pm$ 0.00} & 86.7 $\pm$ 0.23 & \multicolumn{1}{c|}{88.0 $\pm$ 0.20} & 88.9 $\pm$ 0.41 & 89.2 $\pm$ 0.37 & \multicolumn{1}{c}{99.0 $\pm$ 0.01} \\
\midrule
Mean & 52.2 $\pm$ 5.66 & \multicolumn{1}{c|}{92.2 $\pm$ 4.89} & 84.2 $\pm$ 3.51 & \multicolumn{1}{c|}{84.8 $\pm$ 4.08} & 86.4 $\pm$ 3.94 & 86.6 $\pm$ 3.96 & \multicolumn{1}{c}{98.4 $\pm$ 0.83} \\
\bottomrule
\end{tabular}
    }
\end{table*}
%%%%%%%%%%%%%%%%%%%%%%%%%%%%%%%%%%%%%%%%%%%%%%%%%%%%%%%%%%%%

\FloatBarrier
\subsection{Image AD on the MVTec-AD benchmark}
\label{appx:section_mvtec_results}
As mentioned in Section \ref{sec:methods_background}, our paper focuses on natural images because random images from the web are likely not informative as OE for other data. 
In Section \ref{sec:exp_var_oe_size} we demonstrate this on the manufacturing dataset MVTec-AD, where anomalies are rather subtle, for varying OE set sizes.
Here we show results using the full ImageNet-22k dataset as OE in Table \ref{tab:appx_mvtec}.

%%%%%%%%%%%%%%%%%%%%%%%%%%%%%%%%%%%%%%%%%%%%%%%%%%%%%%%%%%%%
\begin{table*}[th]
    \caption{Mean AUC detection performance in \% (over 2 seeds) with standard deviations for all individual classes for our implementations on the MVTec-AD benchmark with ImageNet22k OE. }
    \label{tab:appx_mvtec}
    \vspace{0.5em}
    \centering\small
    \resizebox{0.99\textwidth}{!}{
\begin{tabular}{lccccccc}
\toprule
& \multicolumn{2}{c|}{Unsupervised} & \multicolumn{2}{c|}{Unsupervised OE} & \multicolumn{3}{c}{Supervised OE} \\
Class & DSVDD & \multicolumn{1}{c|}{CLIP} & DSAD & \multicolumn{1}{c|}{HSC} & Focal & BCE & \multicolumn{1}{c}{BCE-CL} \\
\midrule
Bottle & 83.0 $\pm$ 2.38 & \multicolumn{1}{c|}{34.8 $\pm$ 0.00} & 78.0 $\pm$ 6.07 & \multicolumn{1}{c|}{73.6 $\pm$ 0.95} & 66.9 $\pm$ 6.47 & 71.9 $\pm$ 5.28 & \multicolumn{1}{c}{76.0 $\pm$ 4.92} \\
Cable & 60.7 $\pm$ 3.02 & \multicolumn{1}{c|}{48.7 $\pm$ 0.00} & 71.3 $\pm$ 2.74 & \multicolumn{1}{c|}{69.1 $\pm$ 0.42} & 61.5 $\pm$ 3.47 & 54.8 $\pm$ 4.76 & \multicolumn{1}{c}{72.1 $\pm$ 4.82} \\
Capsule & 61.4 $\pm$ 4.09 & \multicolumn{1}{c|}{43.3 $\pm$ 0.00} & 53.2 $\pm$ 7.64 & \multicolumn{1}{c|}{59.7 $\pm$ 5.88} & 56.0 $\pm$ 11.81 & 57.5 $\pm$ 3.97 & \multicolumn{1}{c}{59.2 $\pm$ 1.07} \\
Carpet & 54.0 $\pm$ 1.89 & \multicolumn{1}{c|}{72.5 $\pm$ 0.00} & 55.6 $\pm$ 2.96 & \multicolumn{1}{c|}{63.5 $\pm$ 5.12} & 55.3 $\pm$ 0.22 & 57.1 $\pm$ 0.52 & \multicolumn{1}{c}{88.1 $\pm$ 0.63} \\
Grid & 47.0 $\pm$ 2.09 & \multicolumn{1}{c|}{60.6 $\pm$ 0.00} & 48.9 $\pm$ 16.75 & \multicolumn{1}{c|}{59.4 $\pm$ 1.46} & 46.0 $\pm$ 4.93 & 49.9 $\pm$ 3.63 & \multicolumn{1}{c}{72.7 $\pm$ 0.92} \\
Hazelnut & 80.5 $\pm$ 0.11 & \multicolumn{1}{c|}{48.1 $\pm$ 0.00} & 70.2 $\pm$ 6.56 & \multicolumn{1}{c|}{79.2 $\pm$ 3.38} & 73.2 $\pm$ 5.14 & 78.8 $\pm$ 2.82 & \multicolumn{1}{c}{62.8 $\pm$ 1.12} \\
Leather & 69.3 $\pm$ 3.21 & \multicolumn{1}{c|}{99.9 $\pm$ 0.00} & 80.1 $\pm$ 3.87 & \multicolumn{1}{c|}{84.6 $\pm$ 0.48} & 87.3 $\pm$ 4.52 & 87.3 $\pm$ 1.61 & \multicolumn{1}{c}{87.4 $\pm$ 12.50} \\
Metal nut & 55.7 $\pm$ 0.20 & \multicolumn{1}{c|}{45.0 $\pm$ 0.00} & 47.9 $\pm$ 5.33 & \multicolumn{1}{c|}{62.7 $\pm$ 5.76} & 61.1 $\pm$ 7.14 & 54.5 $\pm$ 4.42 & \multicolumn{1}{c}{75.3 $\pm$ 7.61} \\
Pill & 61.3 $\pm$ 2.17 & \multicolumn{1}{c|}{69.4 $\pm$ 0.00} & 58.8 $\pm$ 10.27 & \multicolumn{1}{c|}{55.8 $\pm$ 4.95} & 70.3 $\pm$ 2.96 & 53.8 $\pm$ 11.25 & \multicolumn{1}{c}{64.5 $\pm$ 1.20} \\
Screw & 49.5 $\pm$ 1.52 & \multicolumn{1}{c|}{64.4 $\pm$ 0.00} & 39.0 $\pm$ 1.30 & \multicolumn{1}{c|}{41.9 $\pm$ 1.84} & 59.5 $\pm$ 4.08 & 62.0 $\pm$ 7.20 & \multicolumn{1}{c}{76.6 $\pm$ 1.33} \\
Tile & 63.7 $\pm$ 0.05 & \multicolumn{1}{c|}{78.0 $\pm$ 0.00} & 95.4 $\pm$ 1.06 & \multicolumn{1}{c|}{95.3 $\pm$ 1.15} & 94.3 $\pm$ 2.09 & 95.5 $\pm$ 1.82 & \multicolumn{1}{c}{98.2 $\pm$ 0.20} \\
Toothbrush & 52.9 $\pm$ 6.53 & \multicolumn{1}{c|}{58.3 $\pm$ 0.00} & 70.3 $\pm$ 1.94 & \multicolumn{1}{c|}{88.3 $\pm$ 2.99} & 59.6 $\pm$ 1.81 & 66.9 $\pm$ 4.72 & \multicolumn{1}{c}{68.0 $\pm$ 6.76} \\
Transistor & 61.9 $\pm$ 6.73 & \multicolumn{1}{c|}{51.0 $\pm$ 0.00} & 70.4 $\pm$ 7.19 & \multicolumn{1}{c|}{64.9 $\pm$ 2.83} & 64.6 $\pm$ 3.87 & 66.9 $\pm$ 0.08 & \multicolumn{1}{c}{72.5 $\pm$ 6.65} \\
Wood & 87.7 $\pm$ 0.70 & \multicolumn{1}{c|}{35.5 $\pm$ 0.00} & 85.1 $\pm$ 11.27 & \multicolumn{1}{c|}{93.8 $\pm$ 1.67} & 73.6 $\pm$ 9.12 & 75.7 $\pm$ 20.57 & \multicolumn{1}{c}{94.3 $\pm$ 0.73} \\
Zipper & 70.8 $\pm$ 1.52 & \multicolumn{1}{c|}{36.5 $\pm$ 0.00} & 67.1 $\pm$ 6.22 & \multicolumn{1}{c|}{60.3 $\pm$ 10.50} & 51.8 $\pm$ 25.64 & 58.6 $\pm$ 0.54 & \multicolumn{1}{c}{74.5 $\pm$ 0.67} \\
\midrule
Mean & 64.0 $\pm$ 11.81 & \multicolumn{1}{c|}{56.4 $\pm$ 17.57} & 66.1 $\pm$ 14.89 & \multicolumn{1}{c|}{70.1 $\pm$ 14.78} & 65.4 $\pm$ 12.44 & 66.1 $\pm$ 12.93 & \multicolumn{1}{c}{76.2 $\pm$ 10.96} \\
\bottomrule
\end{tabular}
    }
\end{table*}
%%%%%%%%%%%%%%%%%%%%%%%%%%%%%%%%%%%%%%%%%%%%%%%%%%%%%%%%%%%%

\clearpage

\clearpage
\subsection{Varying the OE size}
\begin{figure*}[th]
\centering
\subfigure[Class: acorn]{\includegraphics[width=0.328\linewidth]{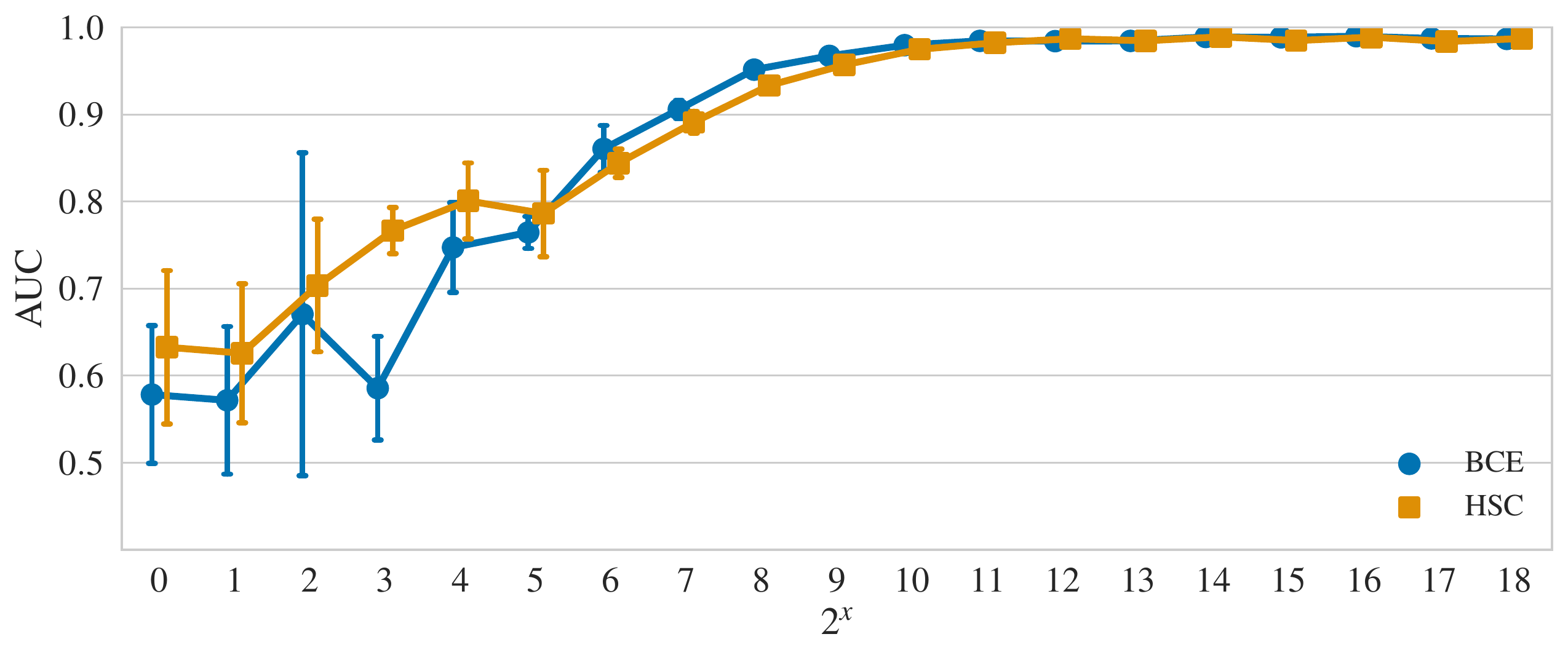}}
\subfigure[Class: airliner]{\includegraphics[width=0.328\linewidth]{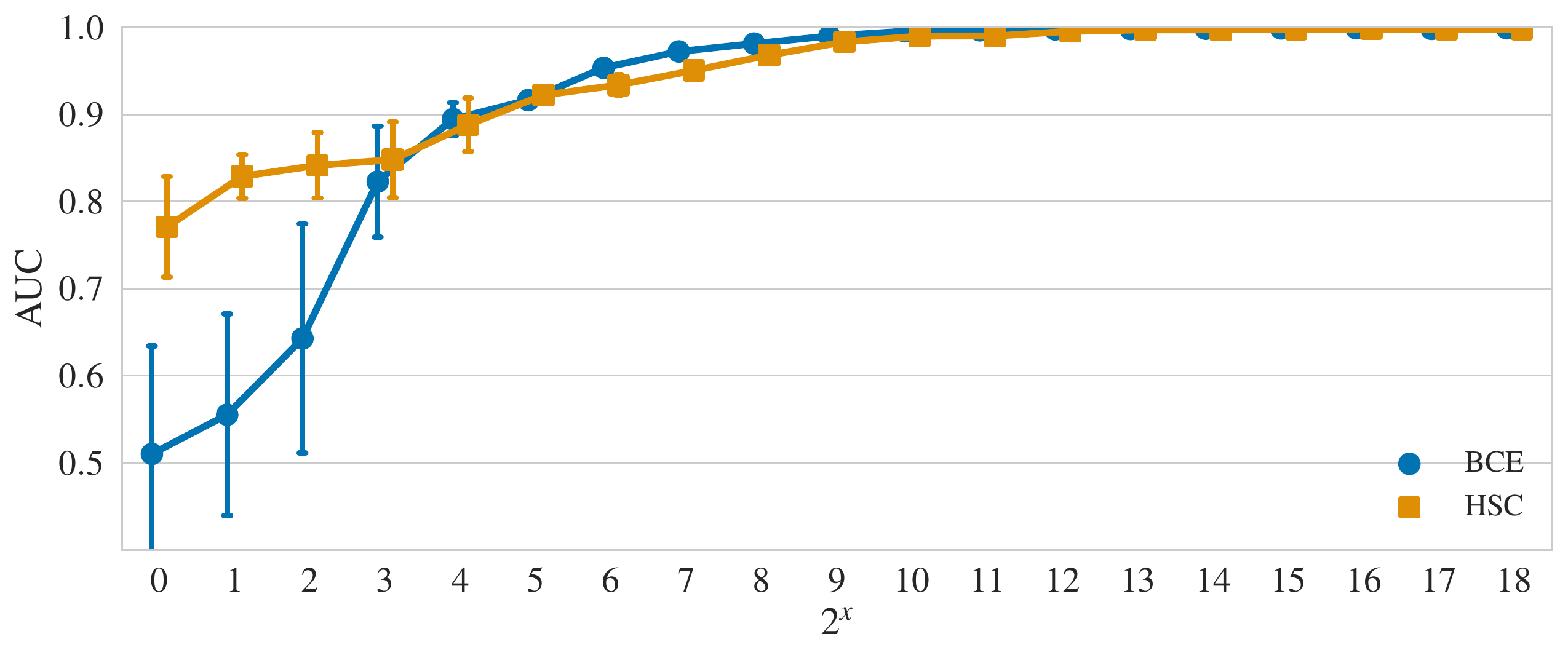}}
\subfigure[Class: ambulance]{\includegraphics[width=0.328\linewidth]{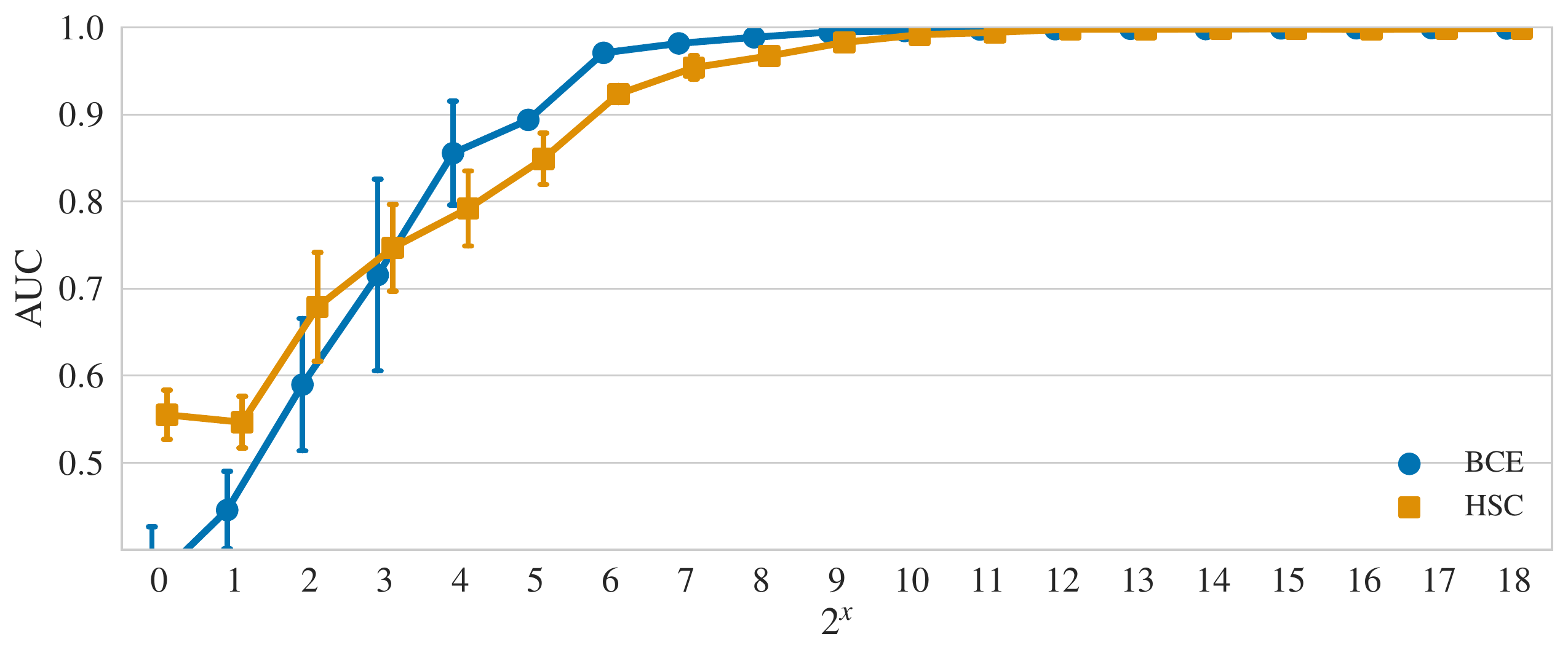}}
\subfigure[Class: american alligator]{\includegraphics[width=0.328\linewidth]{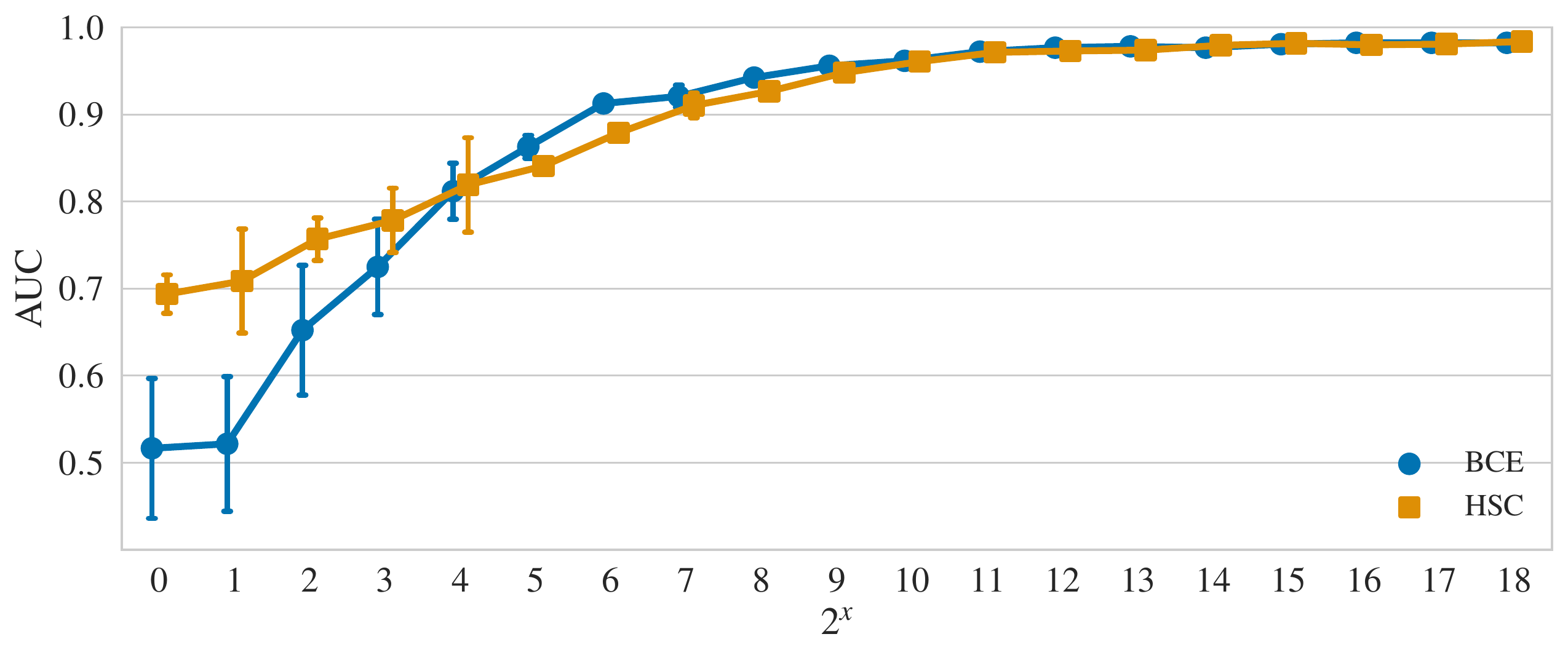}}
\subfigure[Class: banjo]{\includegraphics[width=0.328\linewidth]{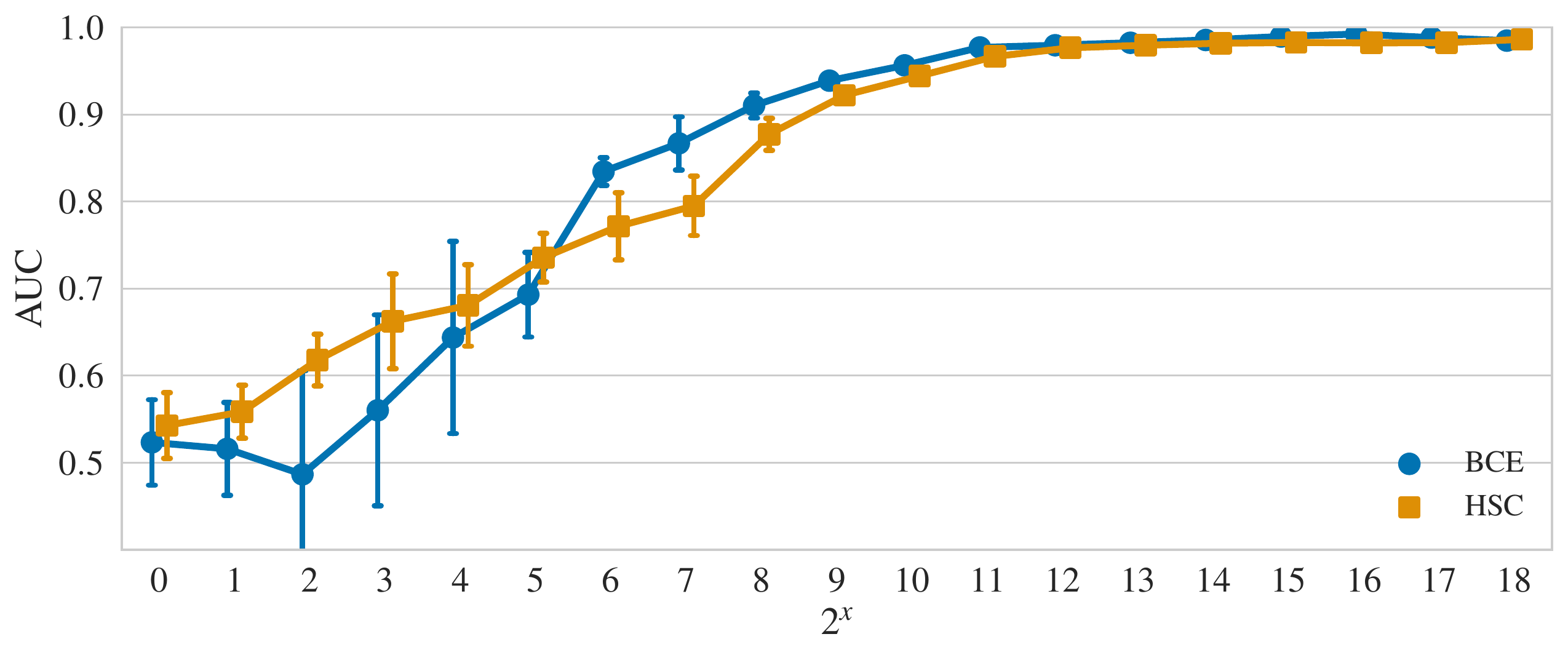}}
\subfigure[Class: barn]{\includegraphics[width=0.328\linewidth]{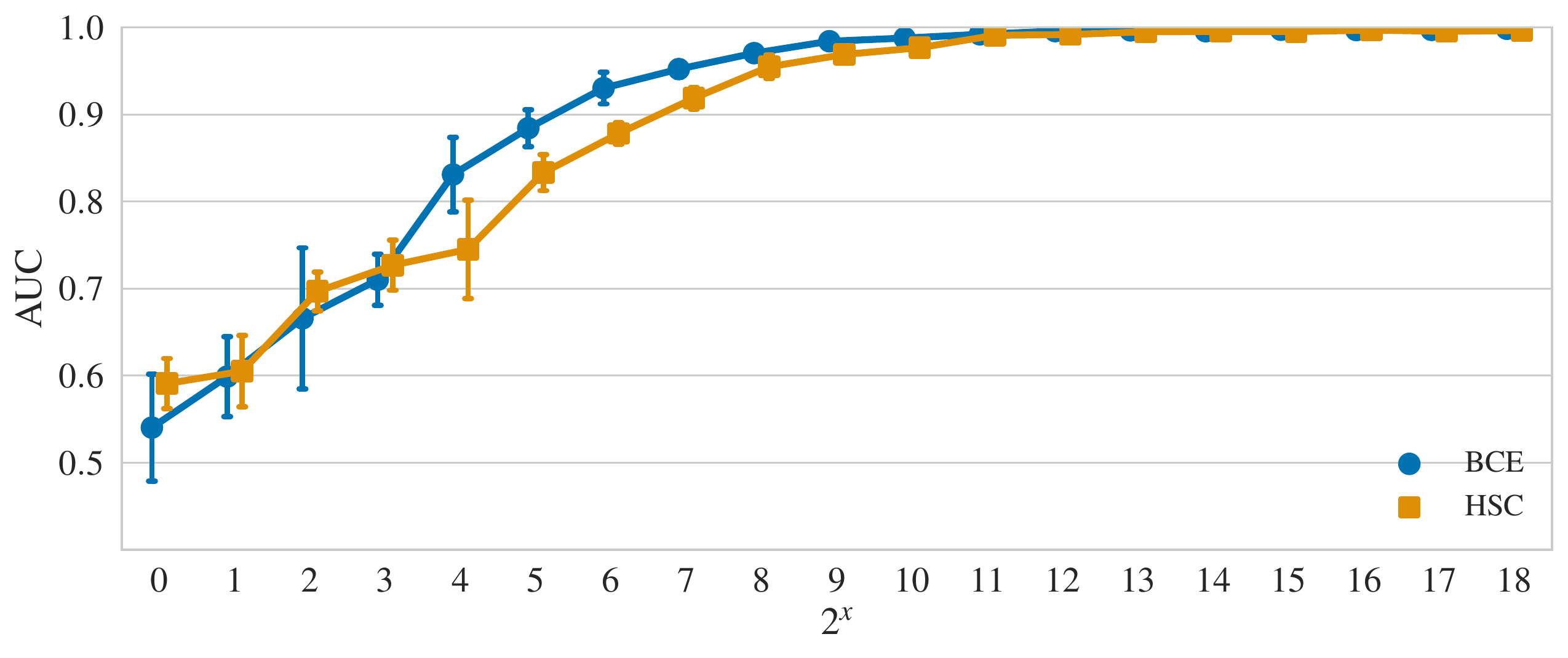}}
\subfigure[Class: bikini]{\includegraphics[width=0.328\linewidth]{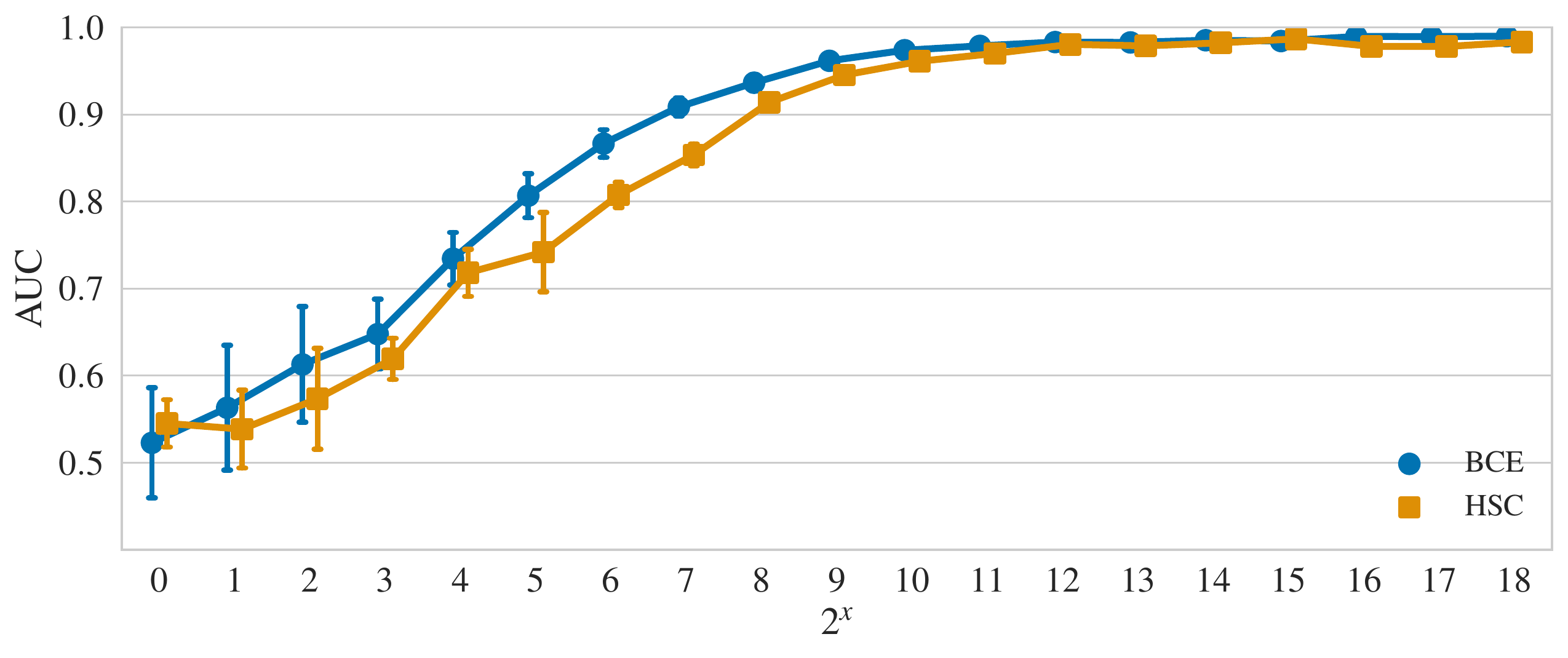}}
\subfigure[Class: digital clock]{\includegraphics[width=0.328\linewidth]{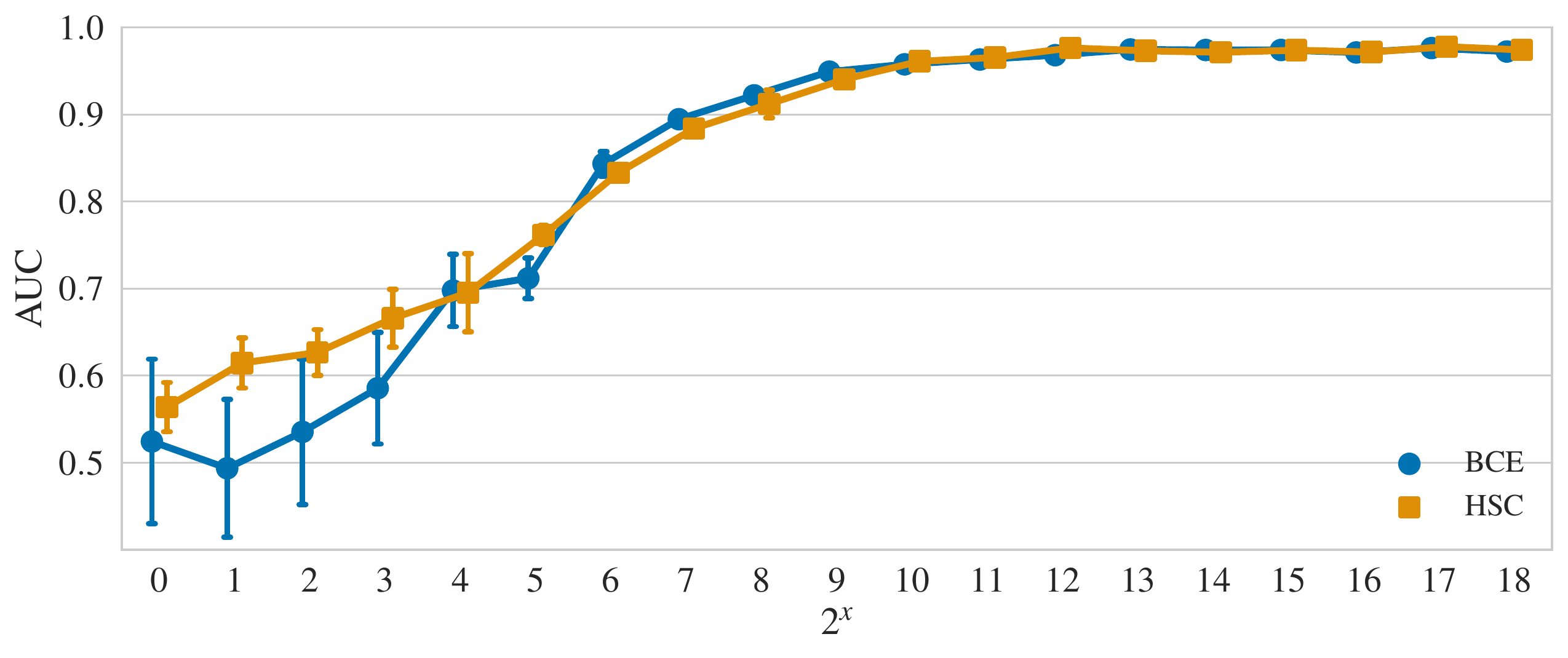}}
\subfigure[Class: dragonfly]{\includegraphics[width=0.328\linewidth]{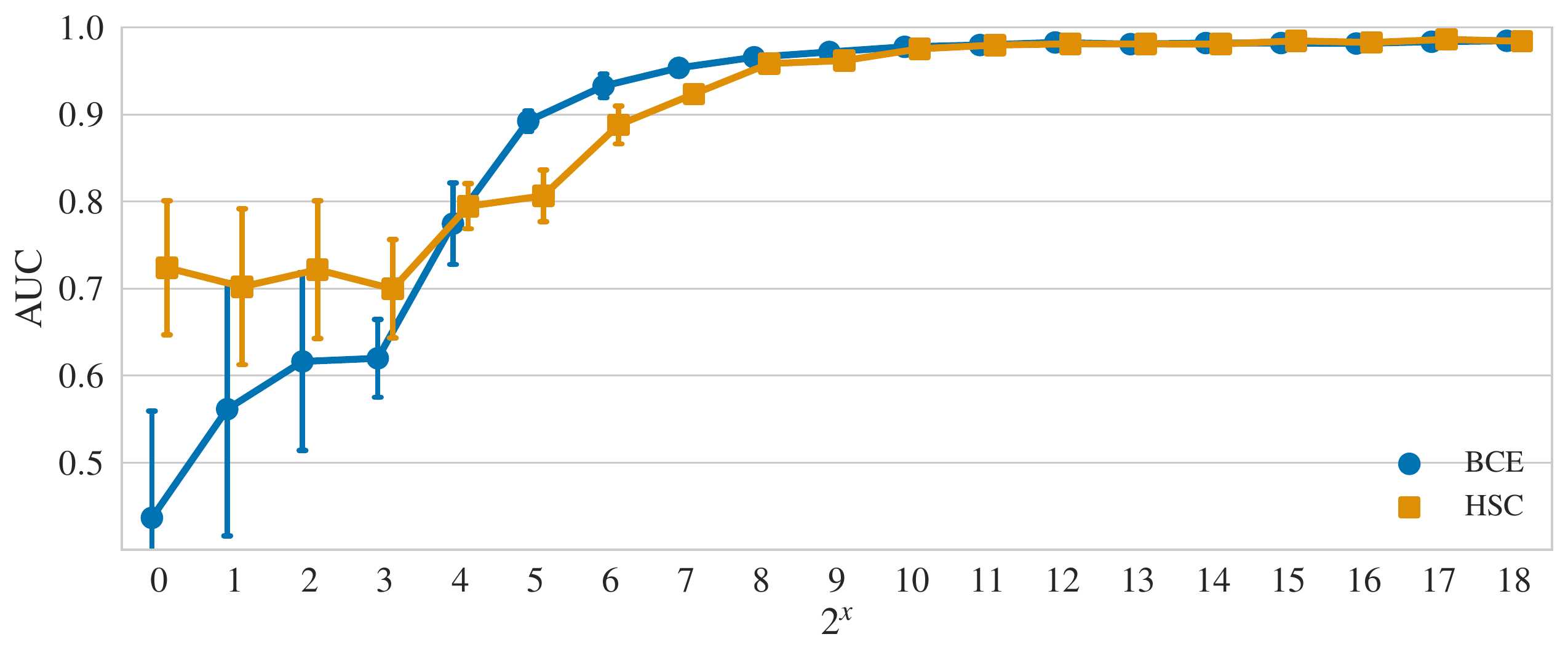}}
\subfigure[Class: dumbbell]{\includegraphics[width=0.328\linewidth]{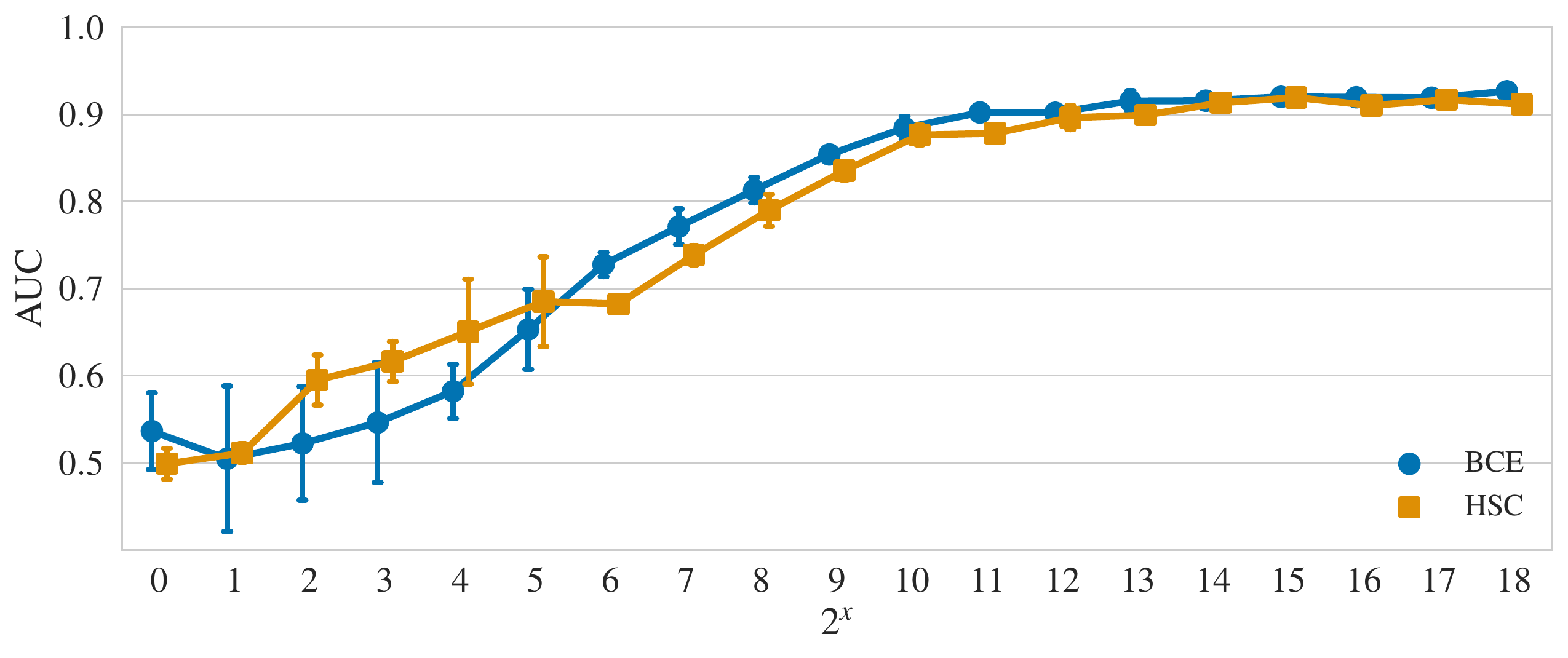}}
\subfigure[Class: forklift]{\includegraphics[width=0.328\linewidth]{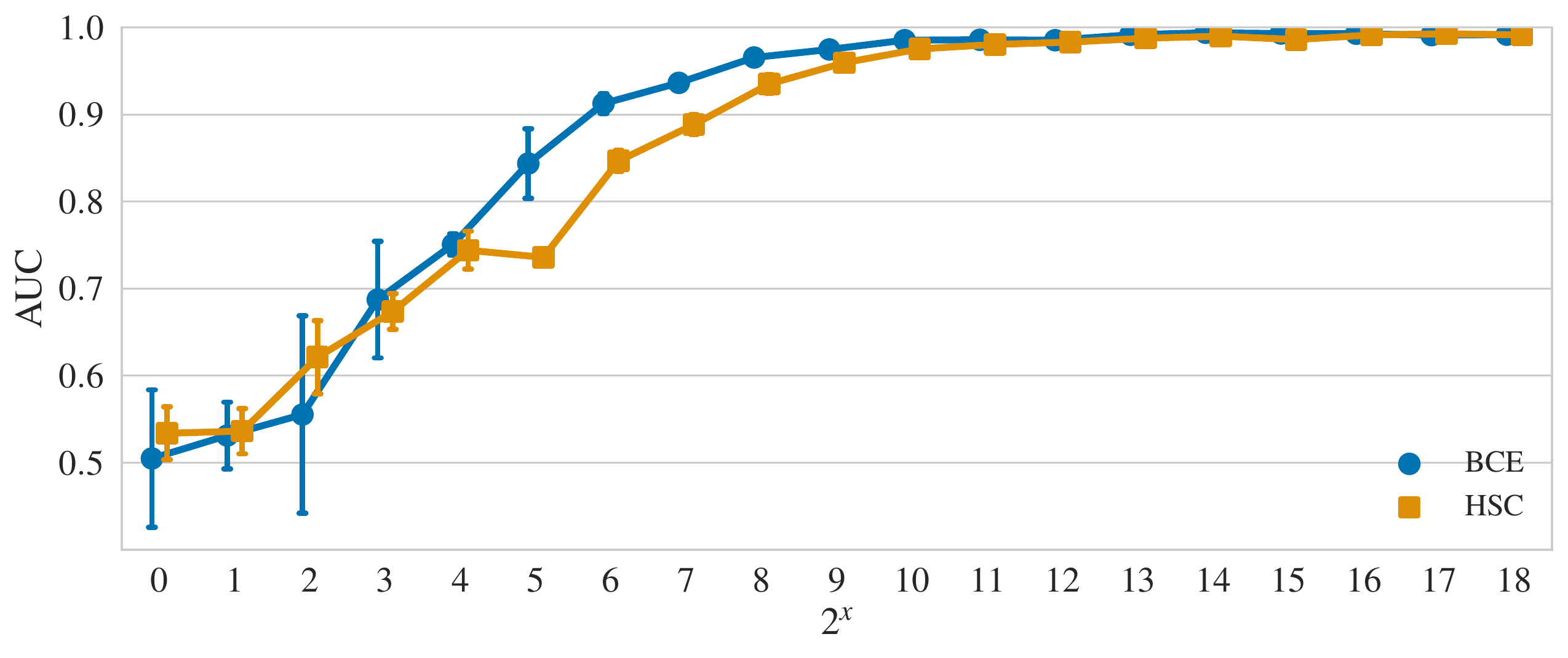}}
\subfigure[Class: goblet]{\includegraphics[width=0.328\linewidth]{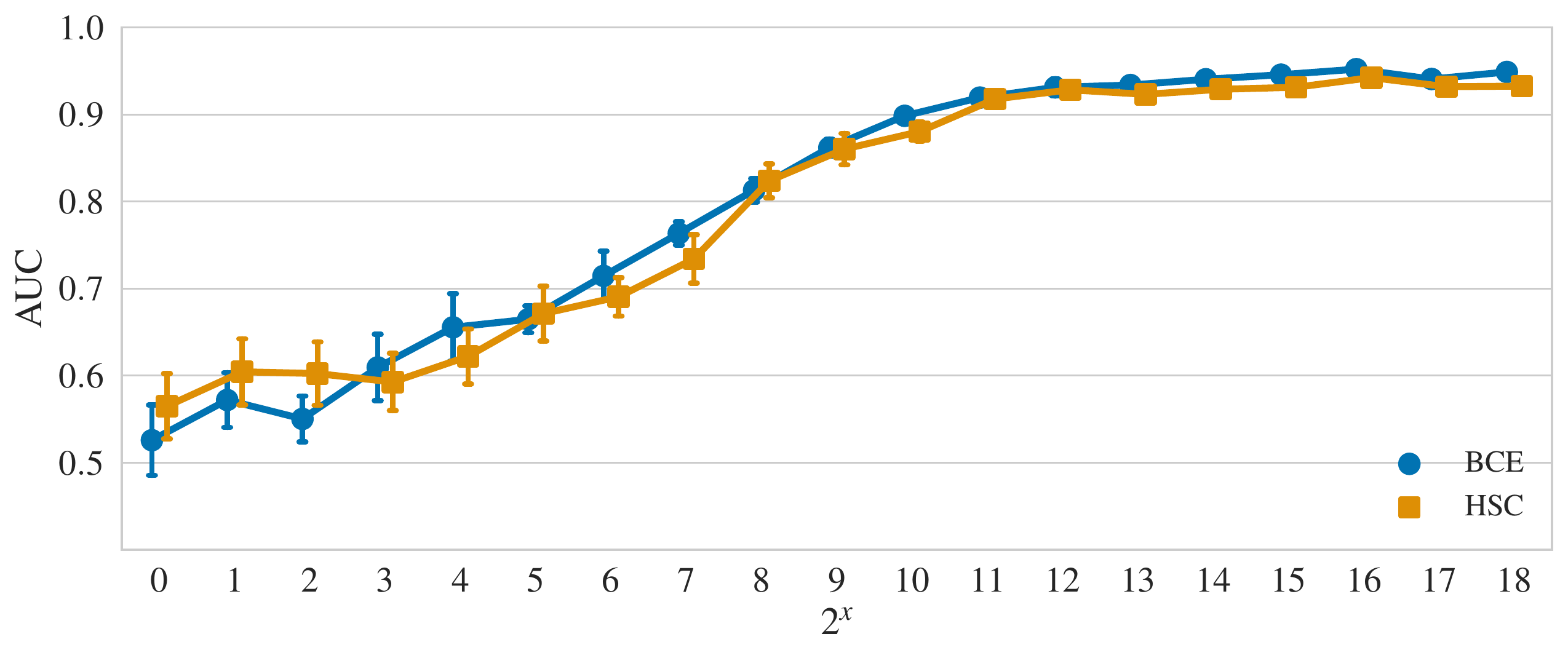}}
\subfigure[Class: grand piano]{\includegraphics[width=0.328\linewidth]{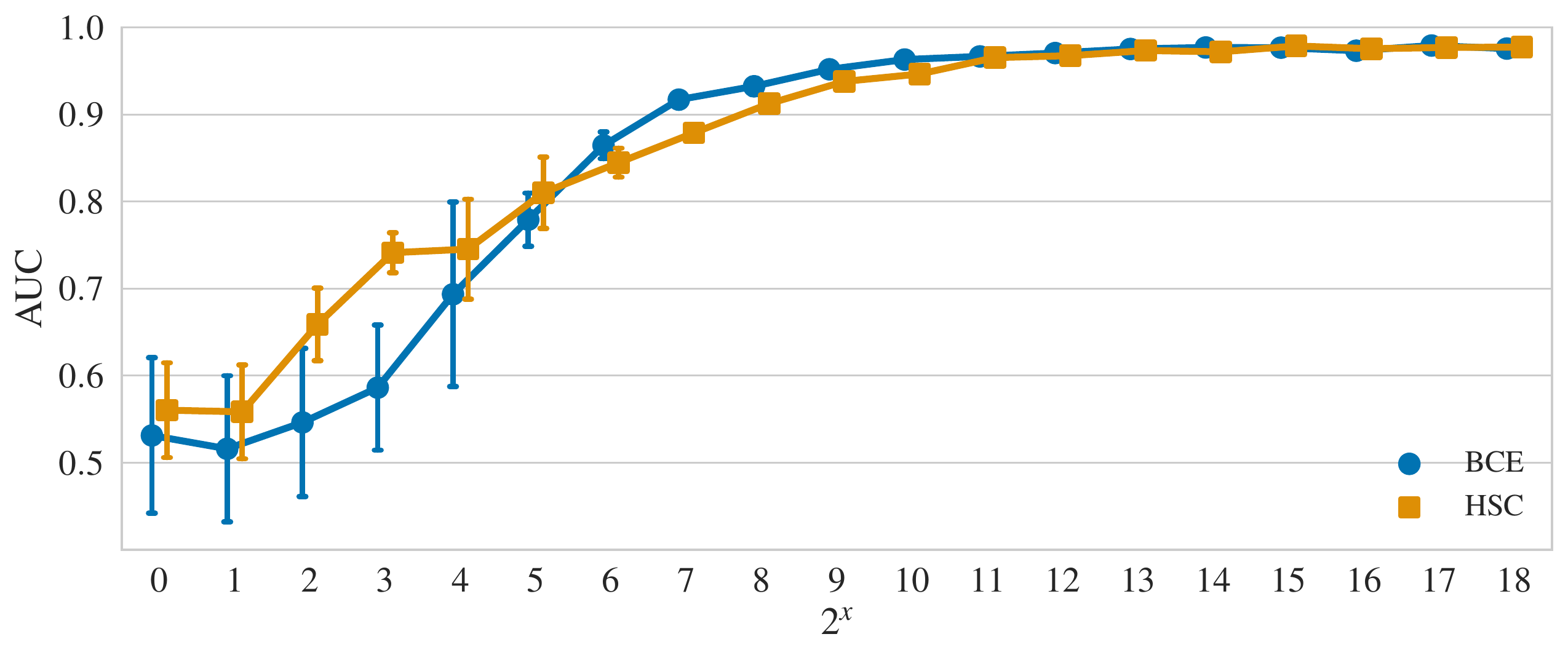}}
\subfigure[Class: hotdog]{\includegraphics[width=0.328\linewidth]{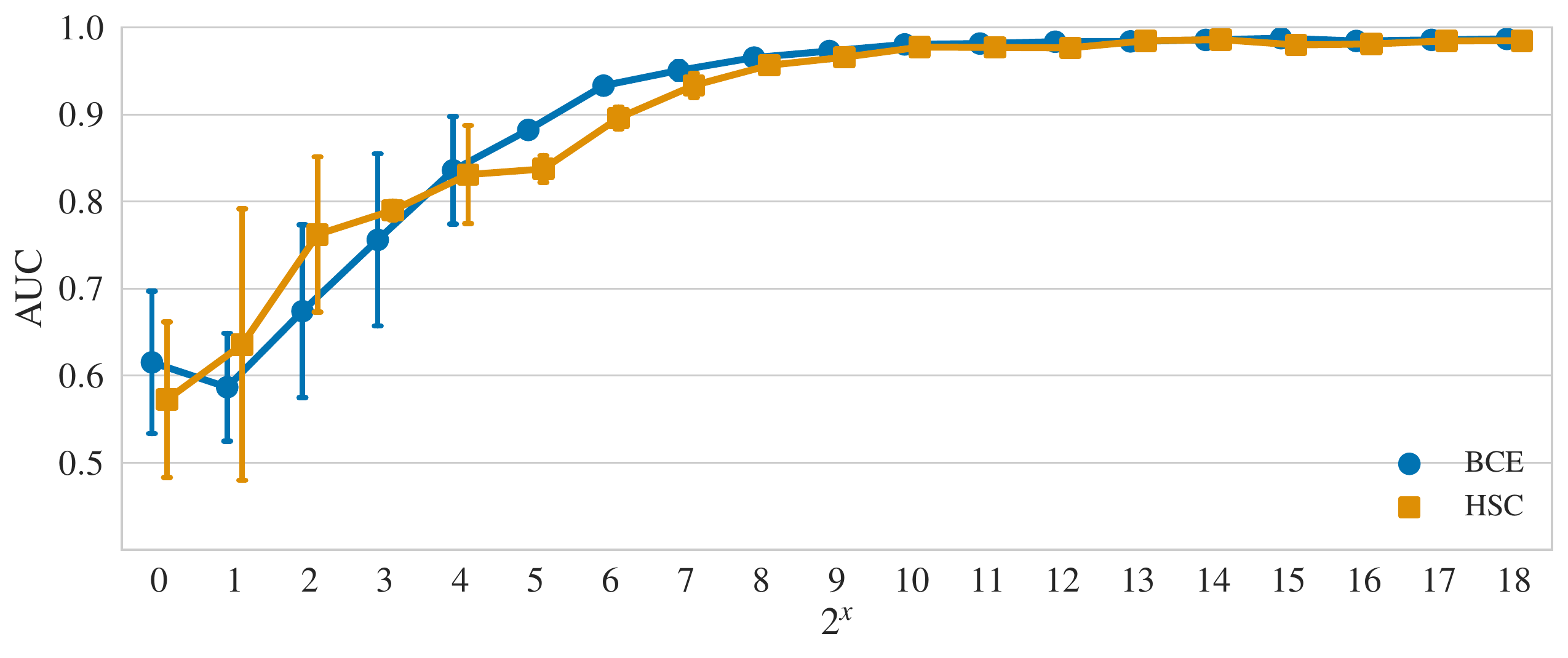}}
\subfigure[Class: hourglass]{\includegraphics[width=0.328\linewidth]{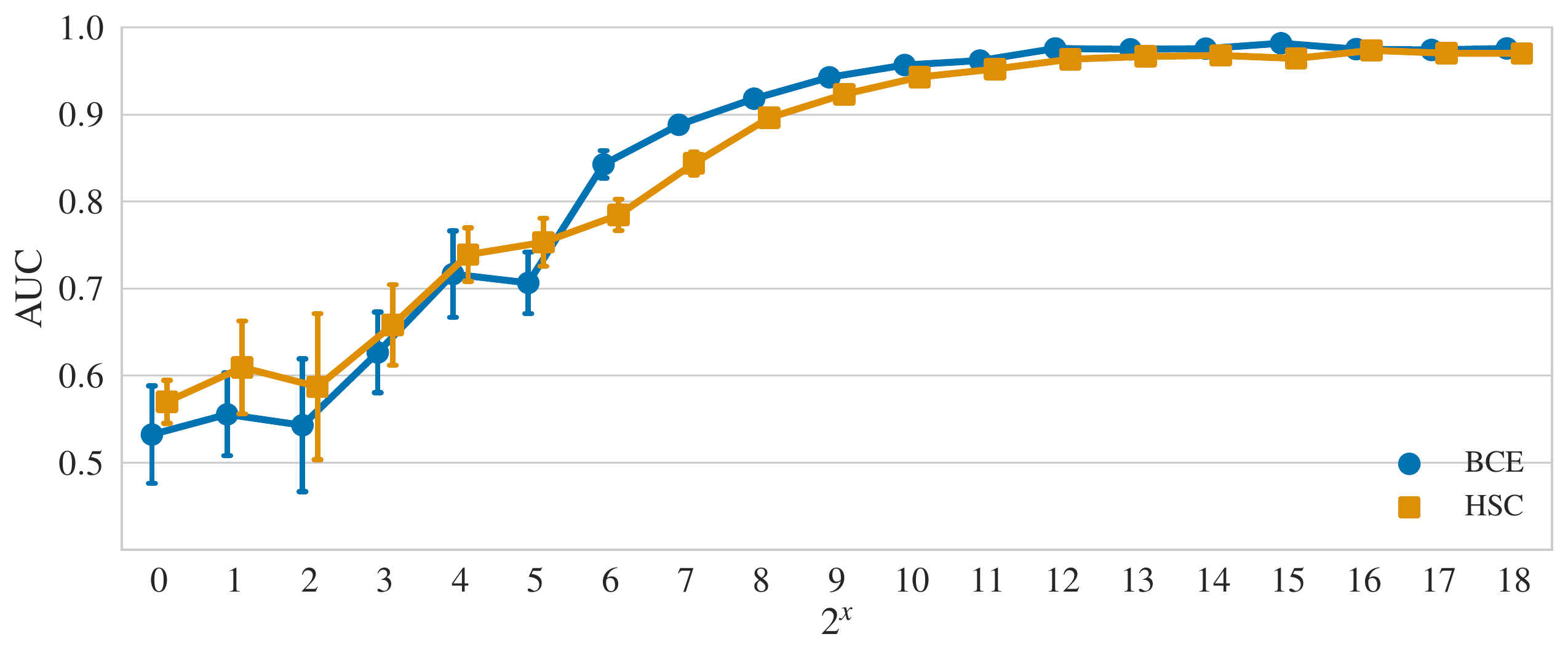}}
\caption{Mean AUC detection performance in \% (over 5 seeds) for all classes of the ImageNet-30 one vs.~rest benchmark from Section \ref{sec:exp_var_oe_size} when varying the number of ImageNet-22K OE samples. These plots correspond to Figure \ref{fig:imagenet1kvs22k}, but here we report the results for all individual classes (from class 1 (acorn) to class 15 (hourglass)).}
\label{fig:imagenet1kvs22k_classes1}
\end{figure*}

\begin{figure*}[th]
\centering
\subfigure[Class: manhole cover]{\includegraphics[width=0.328\linewidth]{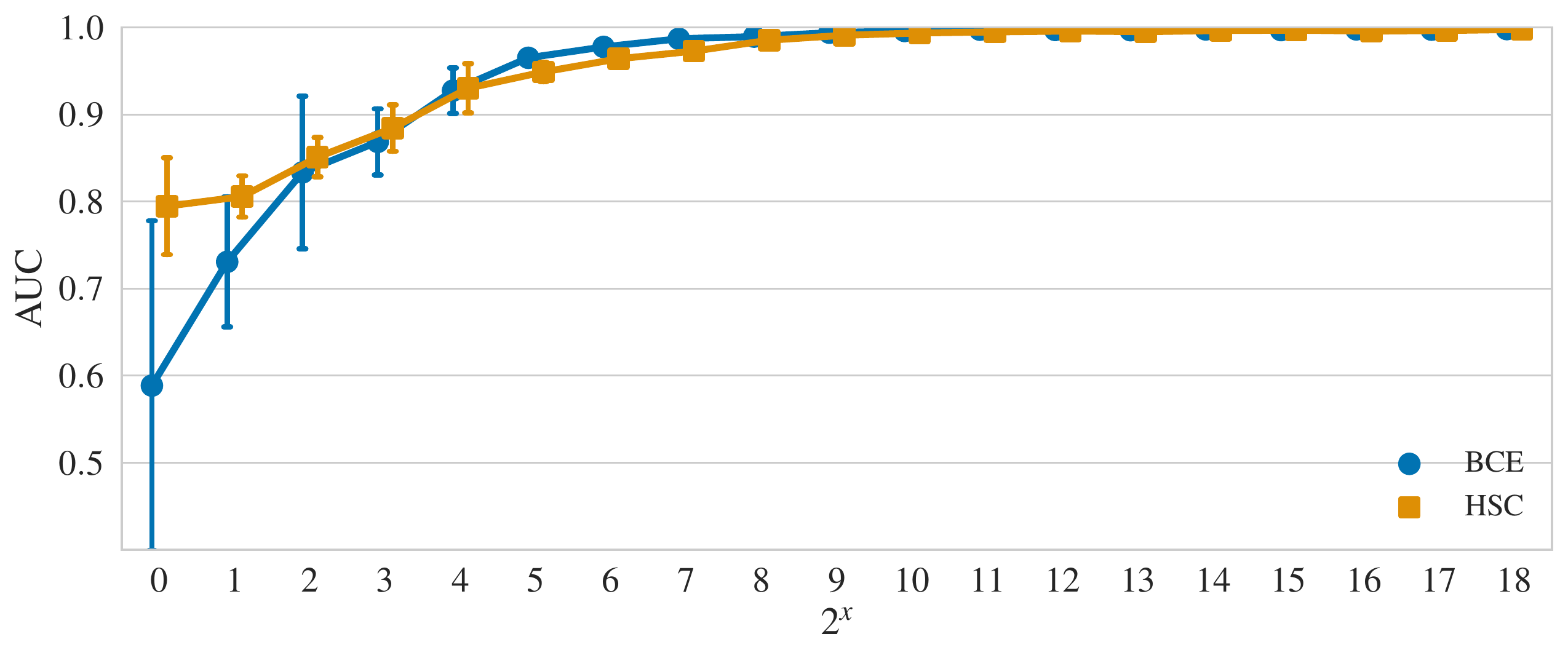}}
\subfigure[Class: mosque]{\includegraphics[width=0.328\linewidth]{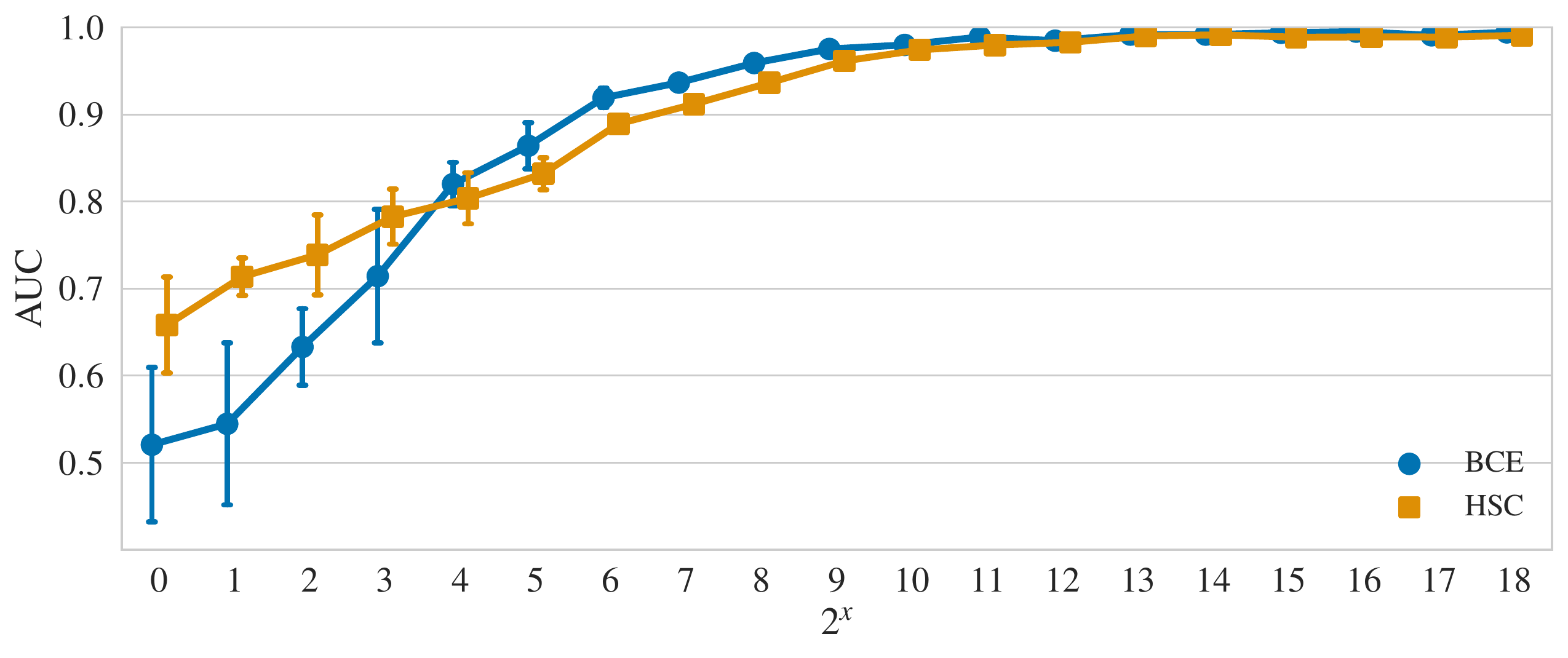}}
\subfigure[Class: nail]{\includegraphics[width=0.328\linewidth]{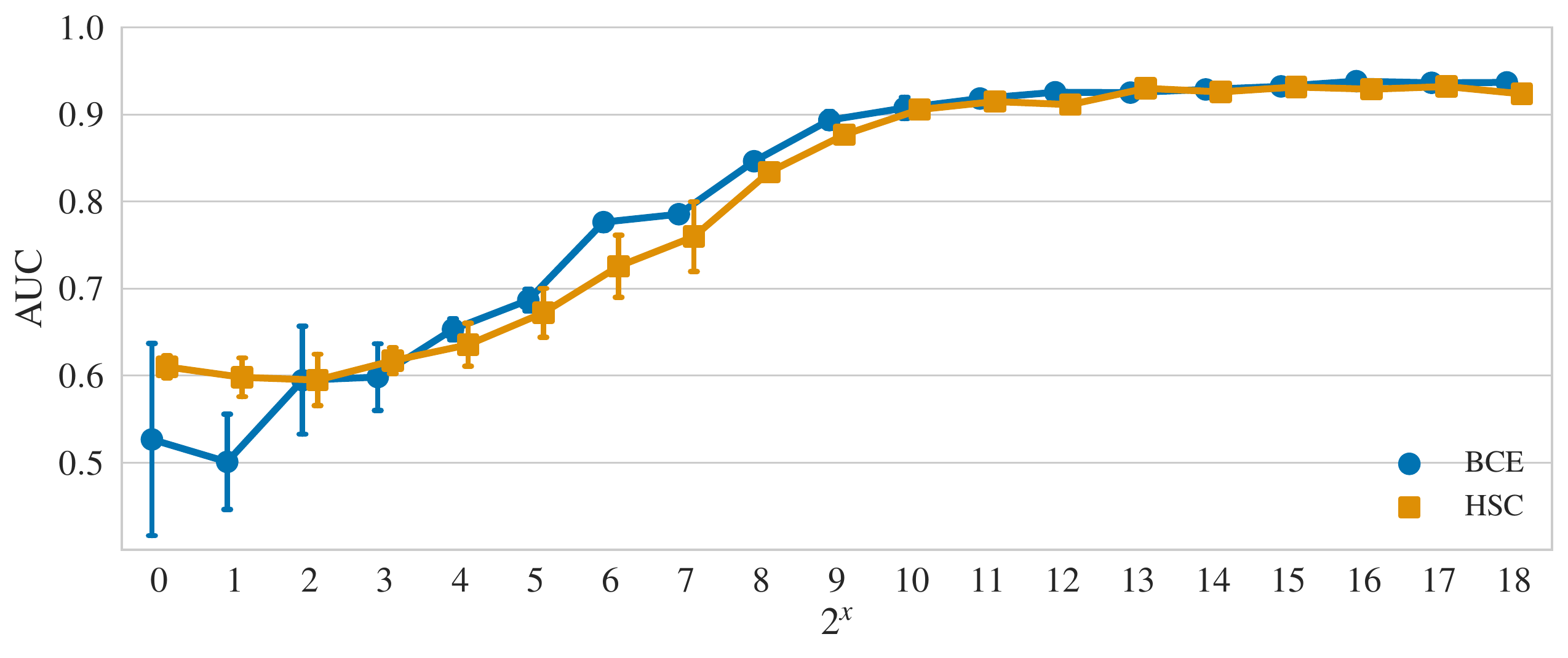}}
\subfigure[Class: parking meter]{\includegraphics[width=0.328\linewidth]{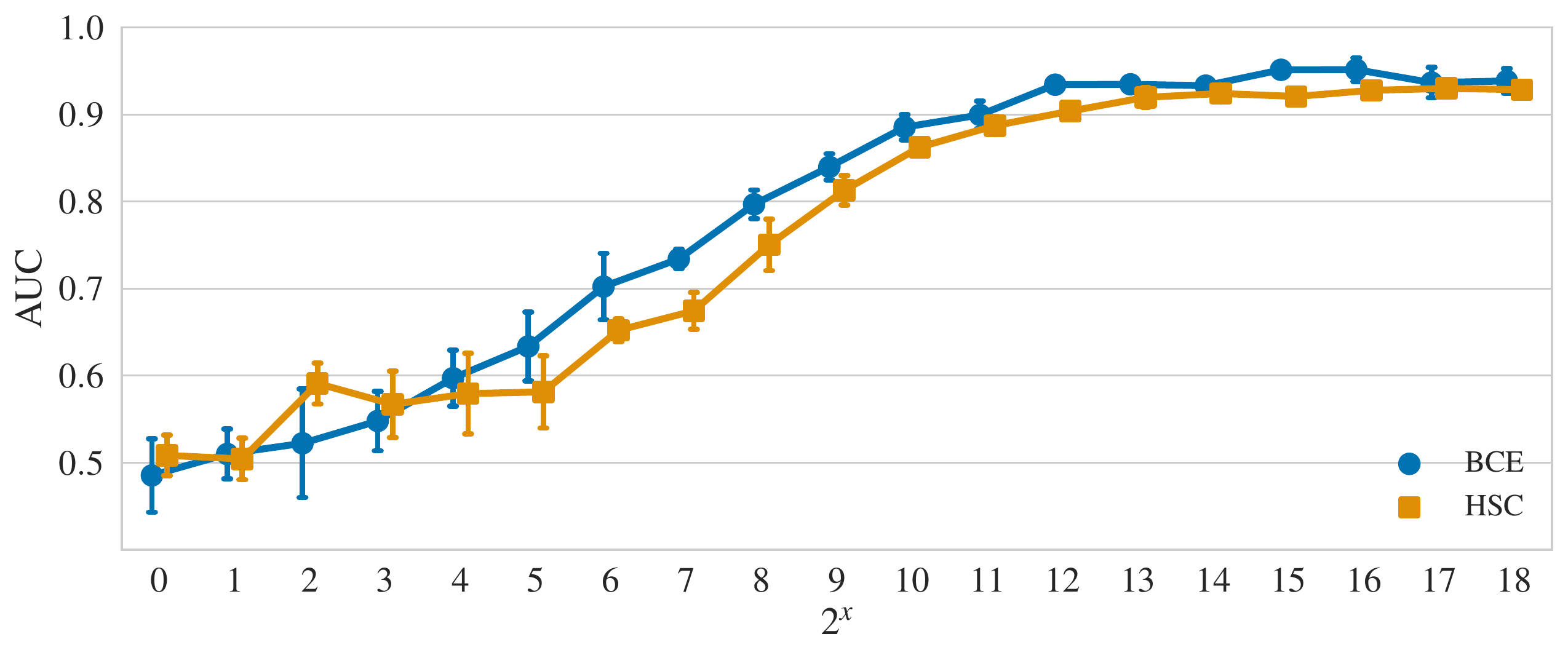}}
\subfigure[Class: pillow]{\includegraphics[width=0.328\linewidth]{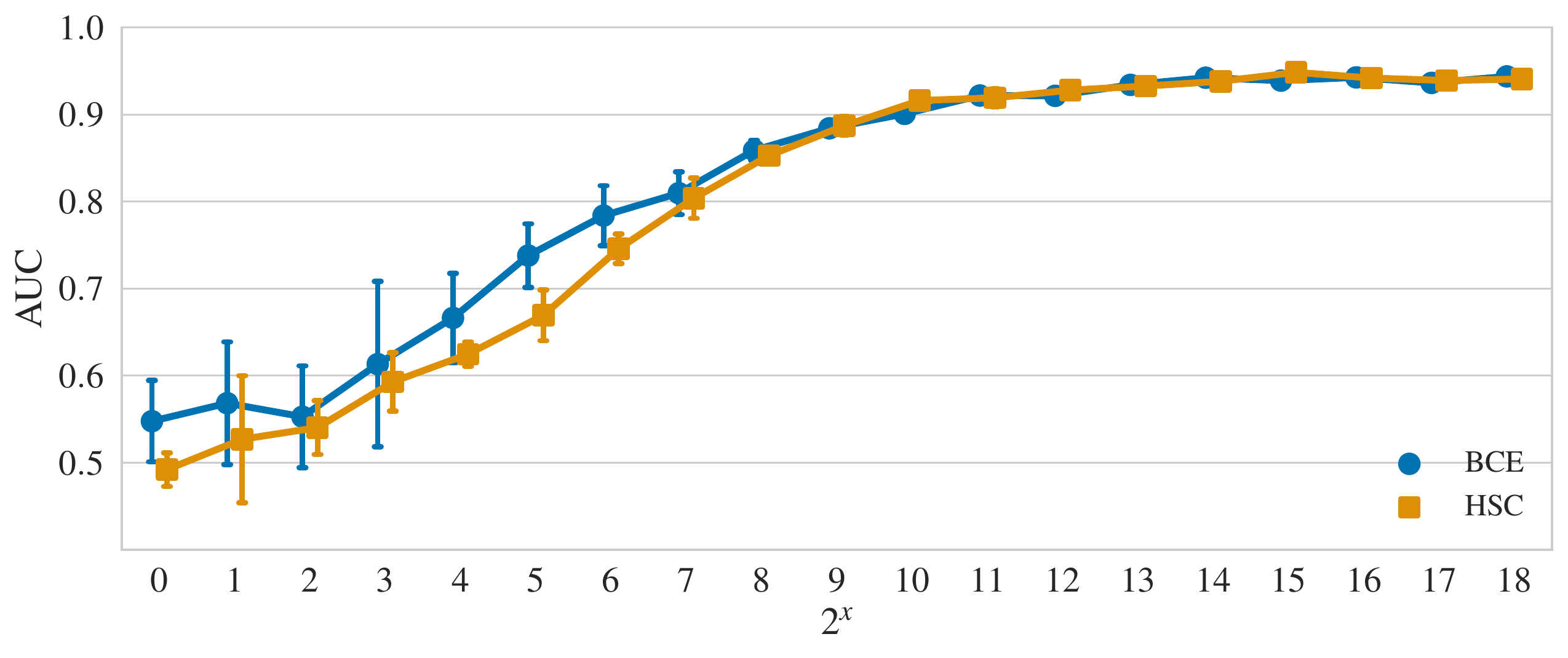}}
\subfigure[Class: revolver]{\includegraphics[width=0.328\linewidth]{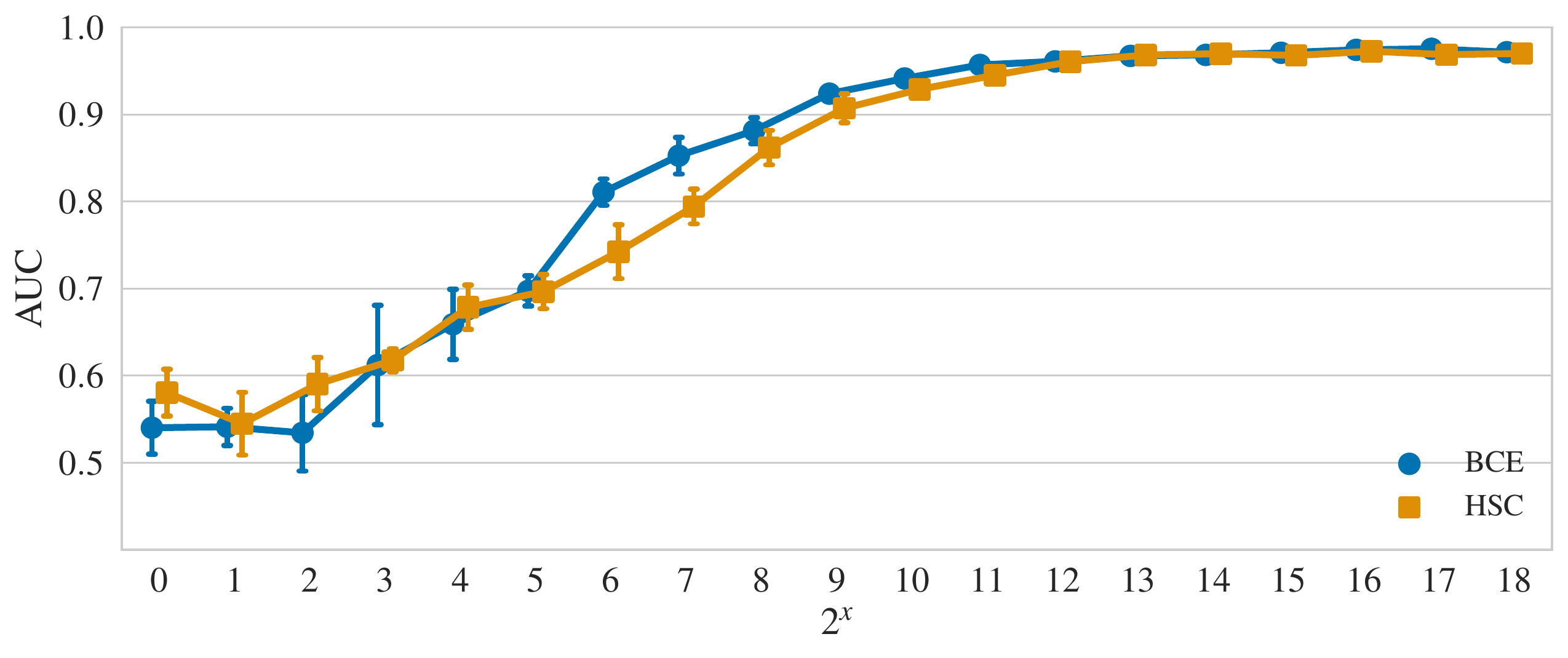}}
\subfigure[Class: rotary dial telephone]{\includegraphics[width=0.328\linewidth]{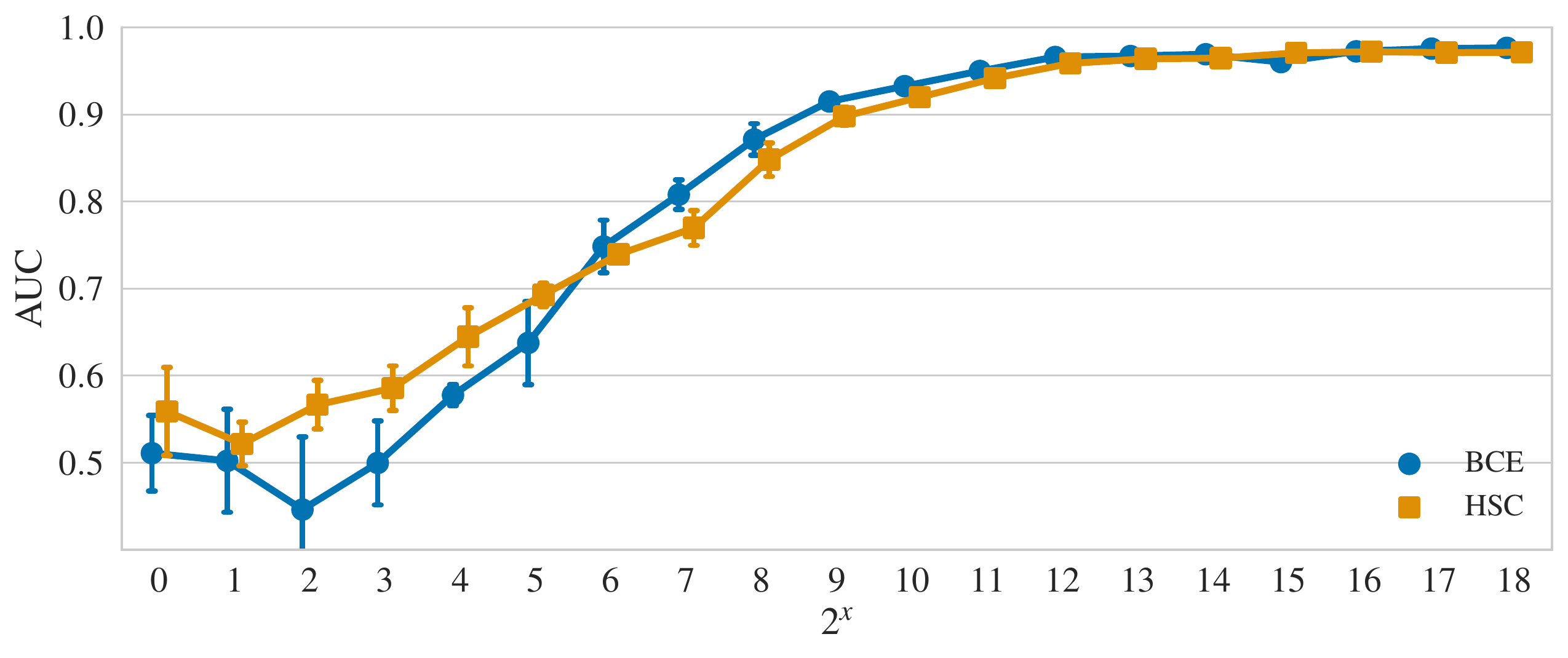}}
\subfigure[Class: schooner]{\includegraphics[width=0.328\linewidth]{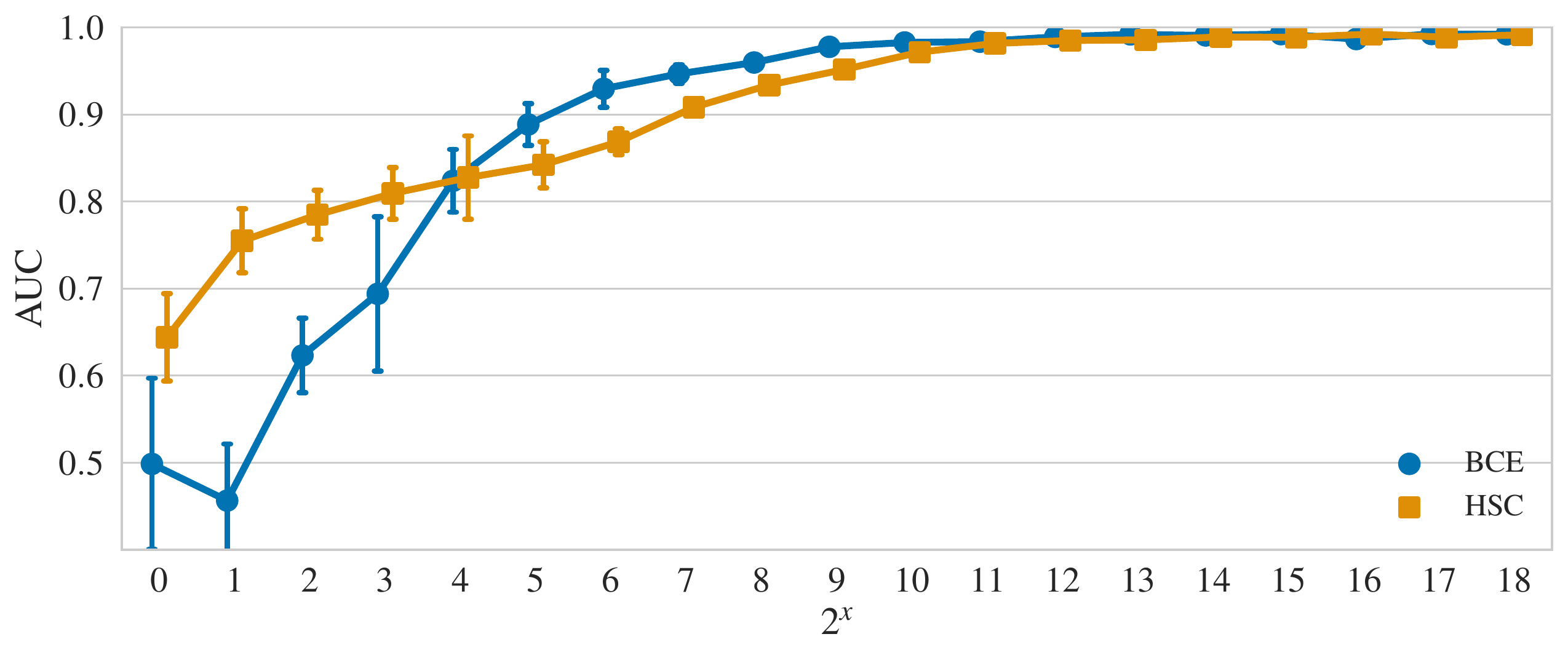}}
\subfigure[Class: snowmobile]{\includegraphics[width=0.328\linewidth]{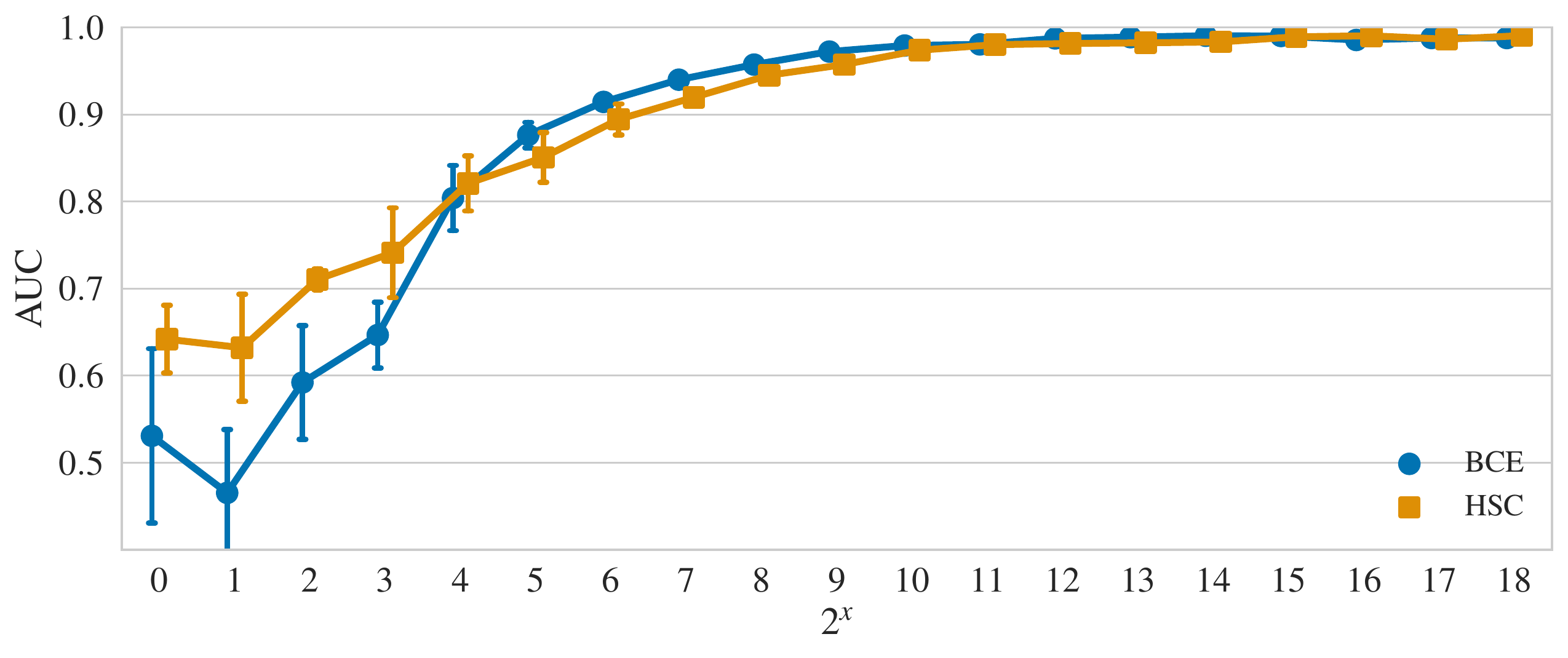}}
\subfigure[Class: soccer ball]{\includegraphics[width=0.328\linewidth]{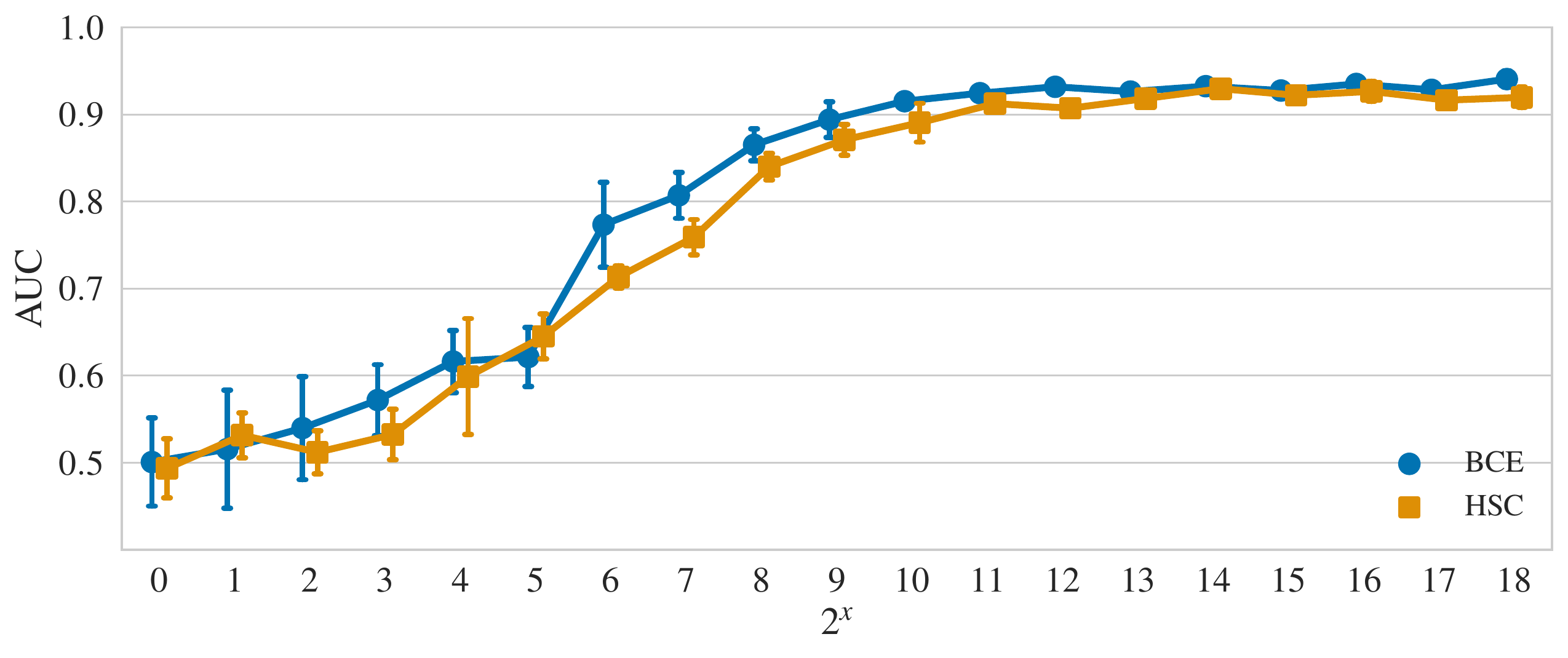}}
\subfigure[Class: stingray]{\includegraphics[width=0.328\linewidth]{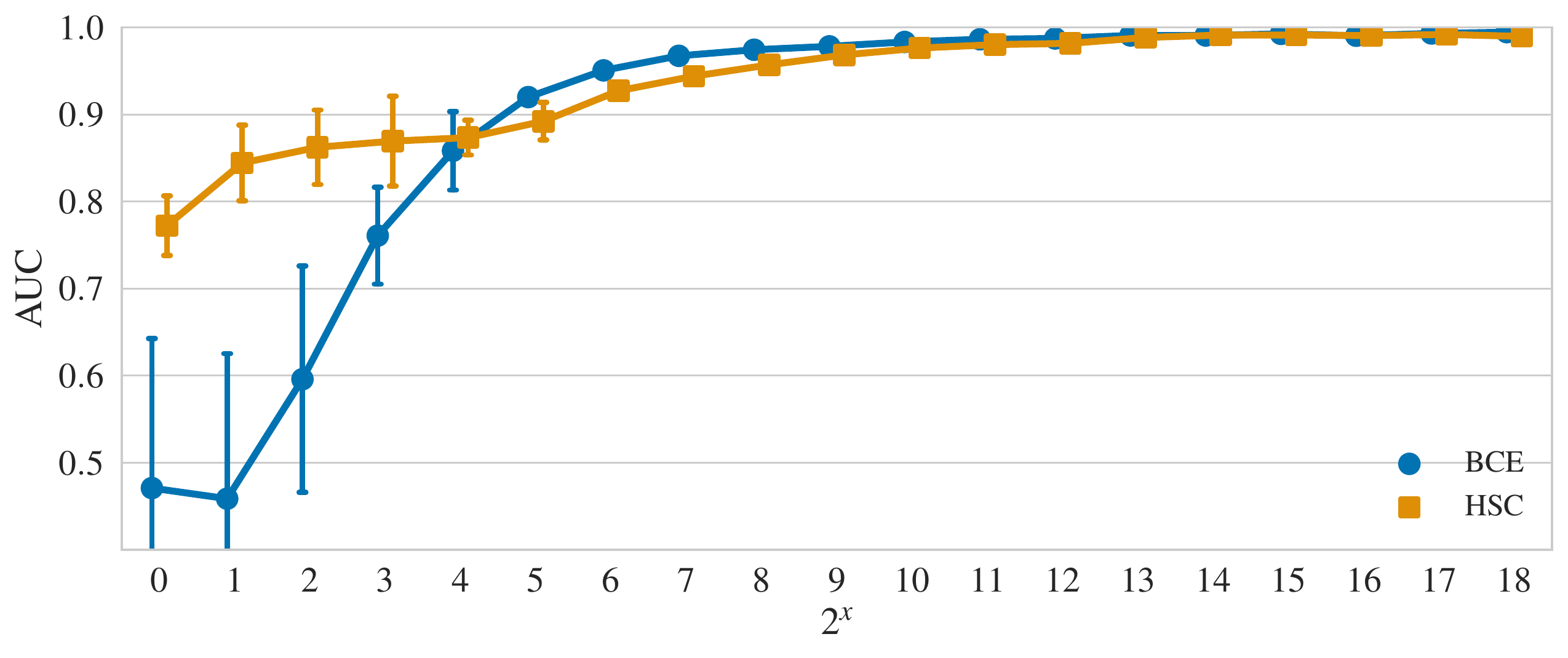}}
\subfigure[Class: strawberry]{\includegraphics[width=0.328\linewidth]{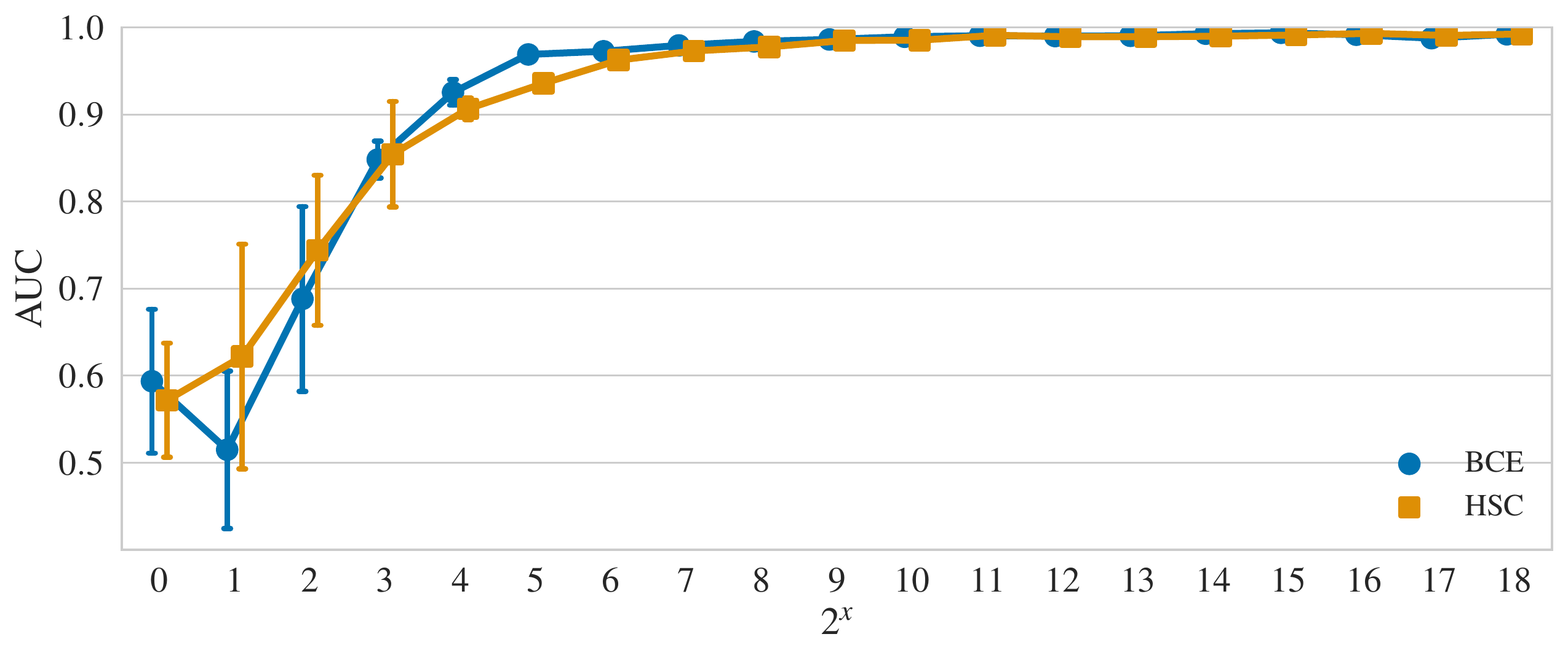}}
\subfigure[Class: tank]{\includegraphics[width=0.328\linewidth]{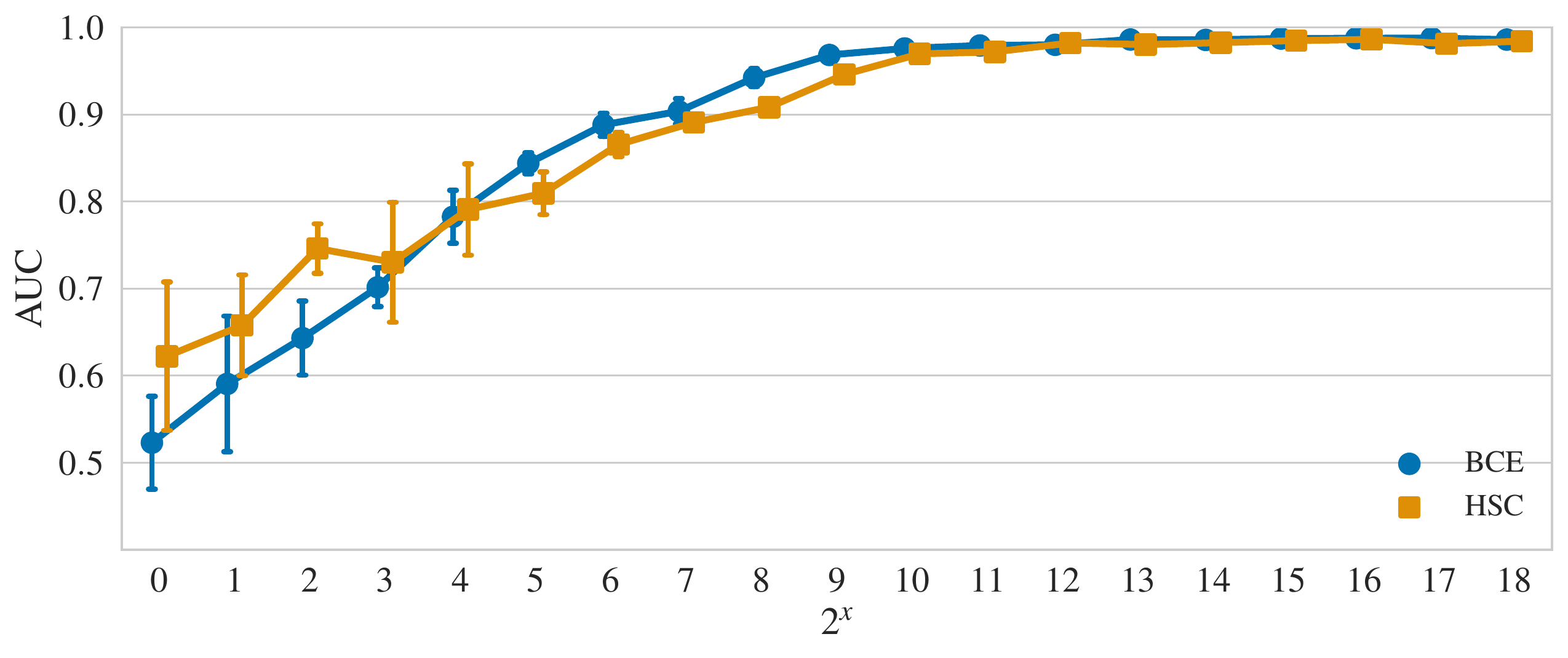}}
\subfigure[Class: toaster]{\includegraphics[width=0.328\linewidth]{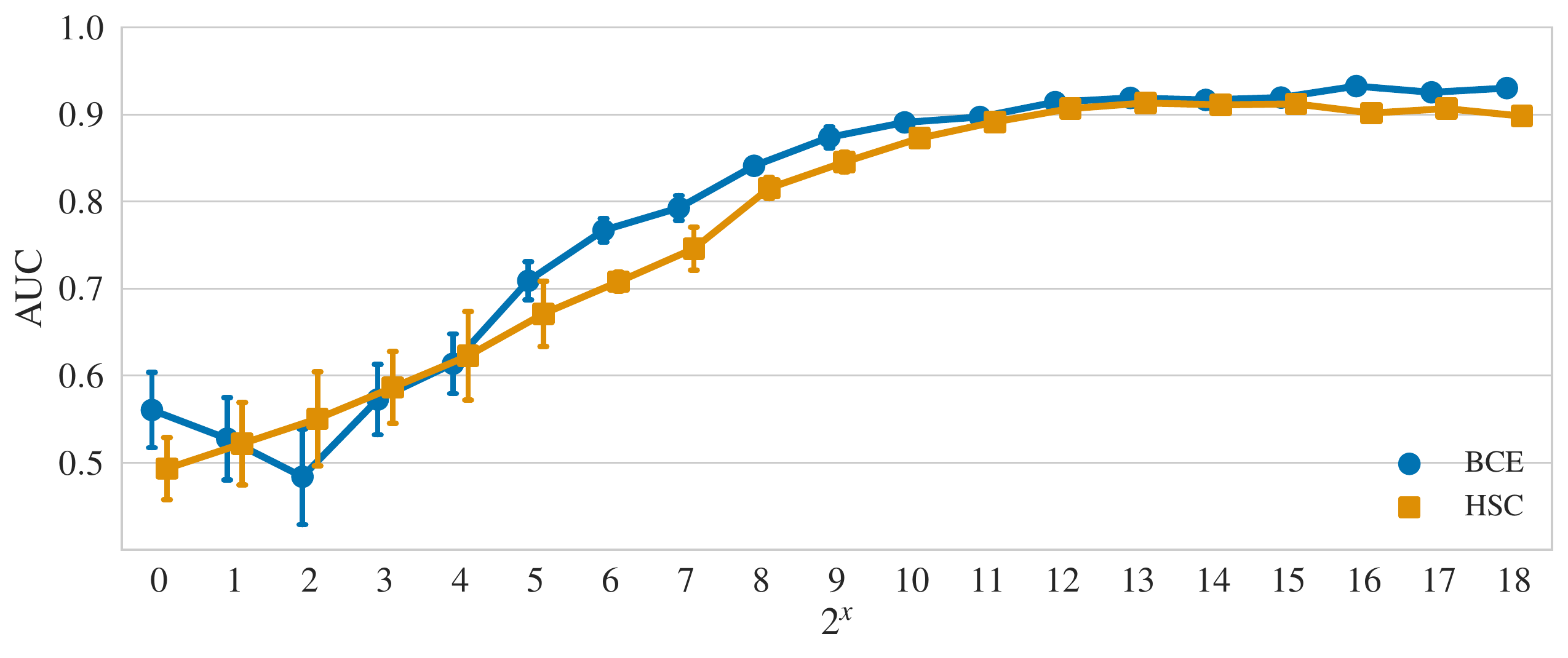}}
\subfigure[Class: volcano]{\includegraphics[width=0.328\linewidth]{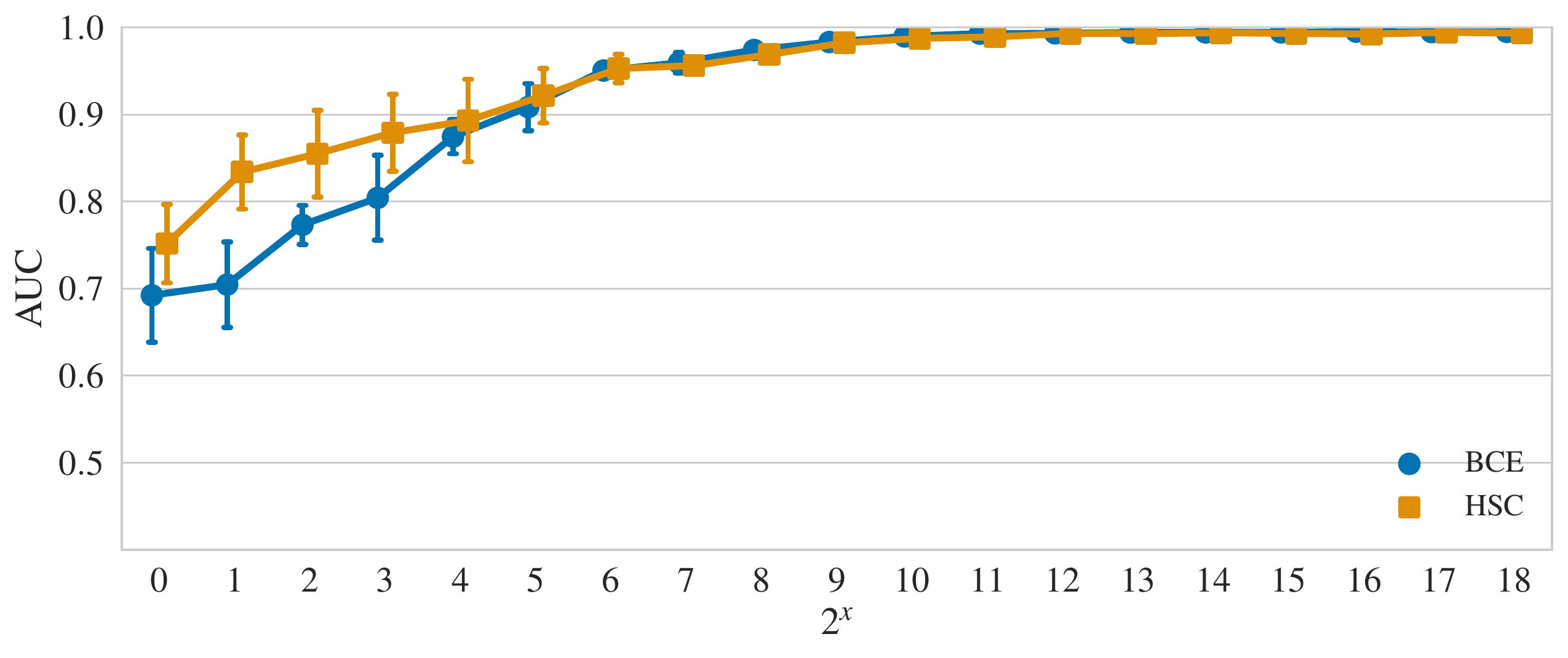}}
\caption{Mean AUC detection performance in \% (over 5 seeds) for all classes of the ImageNet-30 one vs.~rest benchmark from Section \ref{sec:exp_var_oe_size} when varying the number of ImageNet-22K OE samples. These plots correspond to Figure \ref{fig:imagenet1kvs22k}, but here we report the results for all individual classes (from class 16 (manhole cover) to class 30 (volcano)).}
\label{fig:imagenet1kvs22k_classes2}
\end{figure*}

\begin{figure*}[th]
\centering
\subfigure[Class: airplane]{\label{fig:cifarvstiny0}\includegraphics[width=0.49\linewidth]{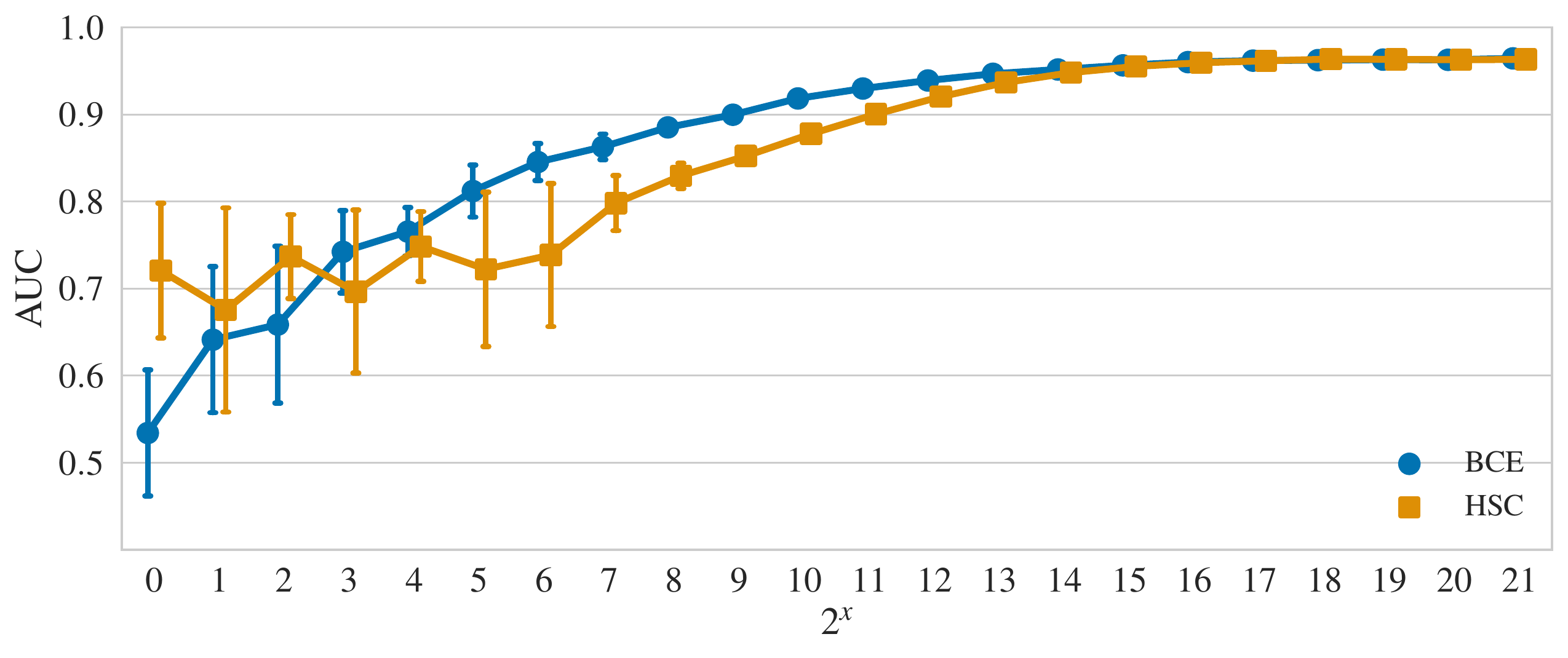}}
\subfigure[Class: automobile]{\label{fig:cifarvstiny1}\includegraphics[width=0.49\linewidth]{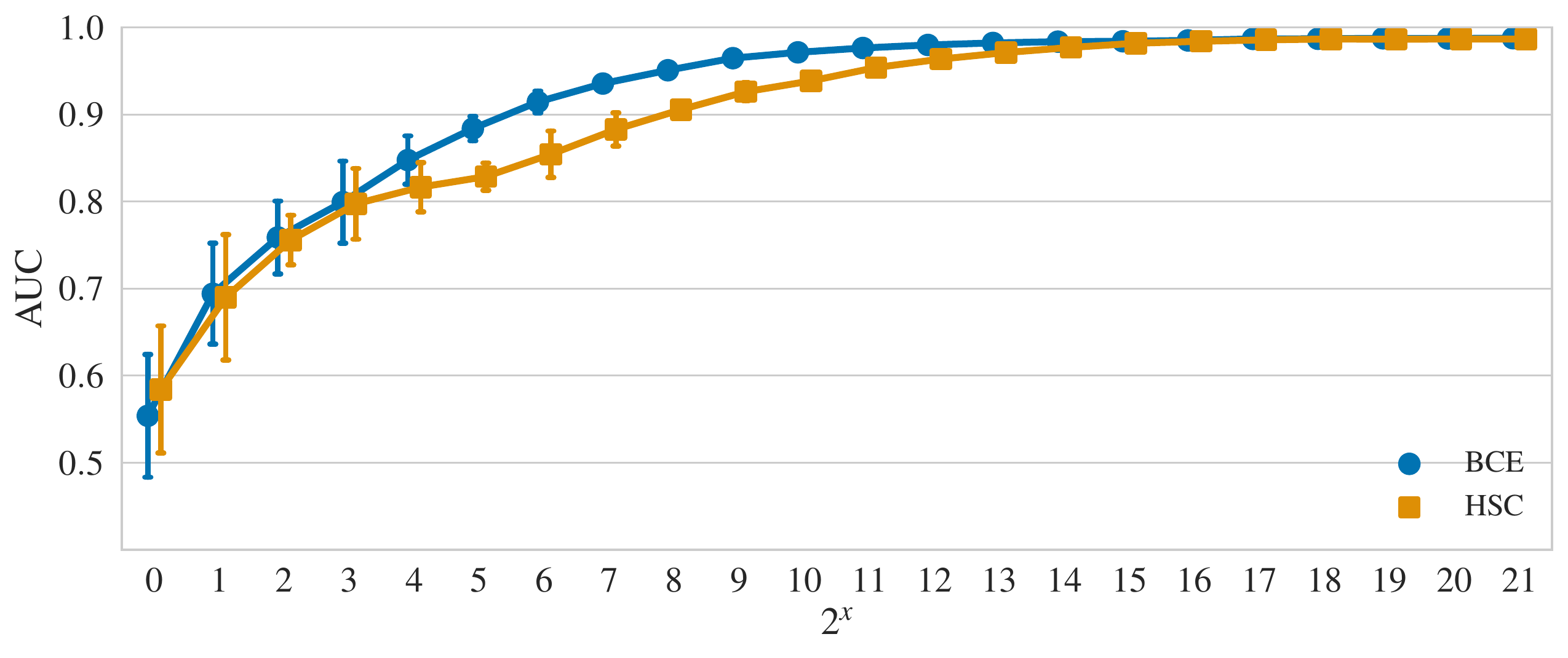}}
\subfigure[Class: bird]{\label{fig:cifarvstiny2}\includegraphics[width=0.49\linewidth]{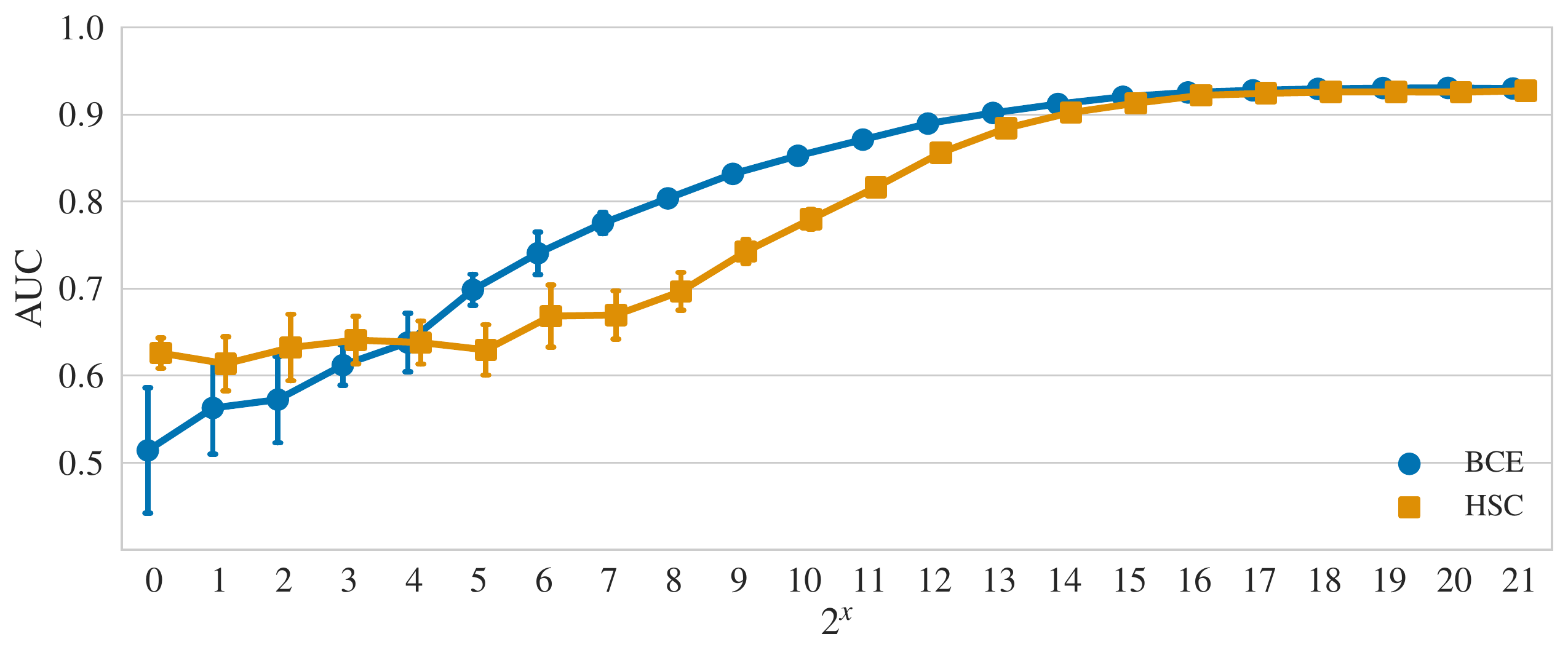}}
\subfigure[Class: cat]{\label{fig:cifarvstiny3}\includegraphics[width=0.49\linewidth]{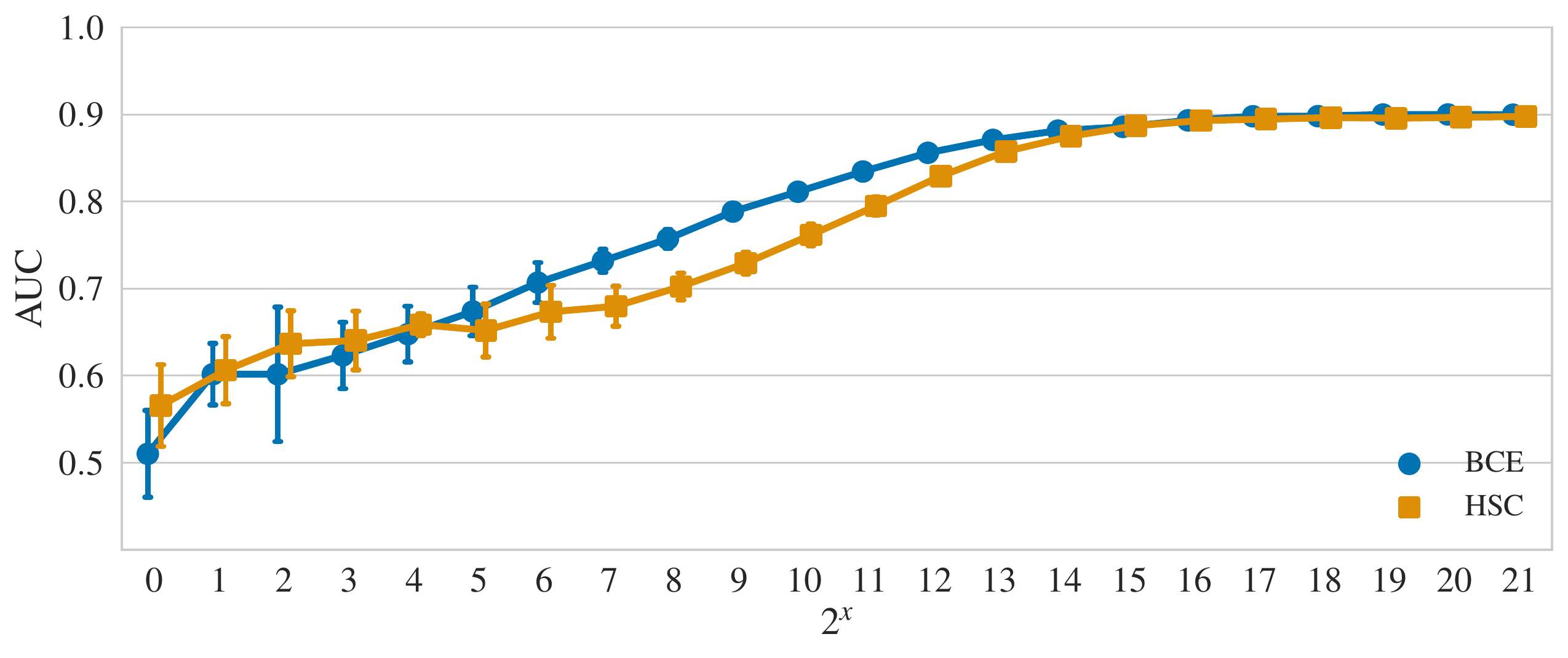}}
\subfigure[Class: deer]{\label{fig:cifarvstiny4}\includegraphics[width=0.49\linewidth]{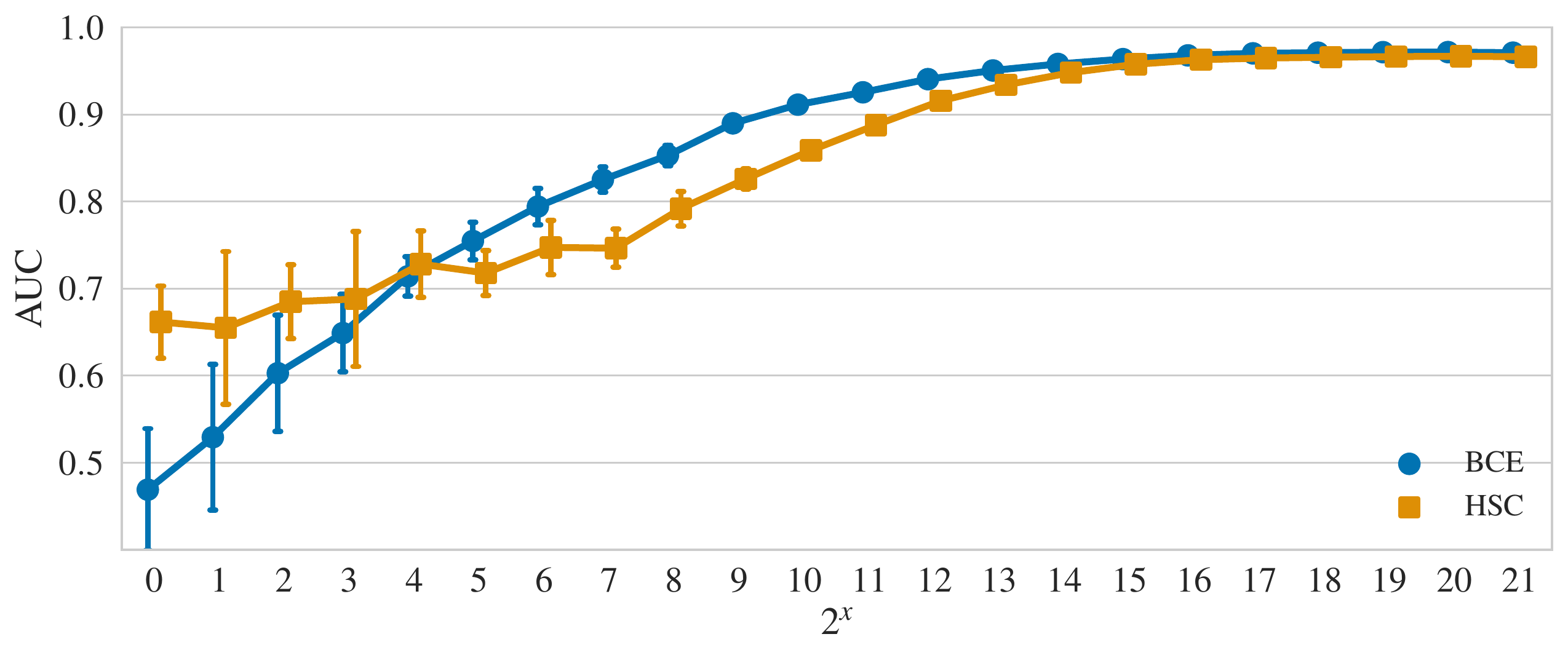}}
\subfigure[Class: dog]{\label{fig:cifarvstiny5}\includegraphics[width=0.49\linewidth]{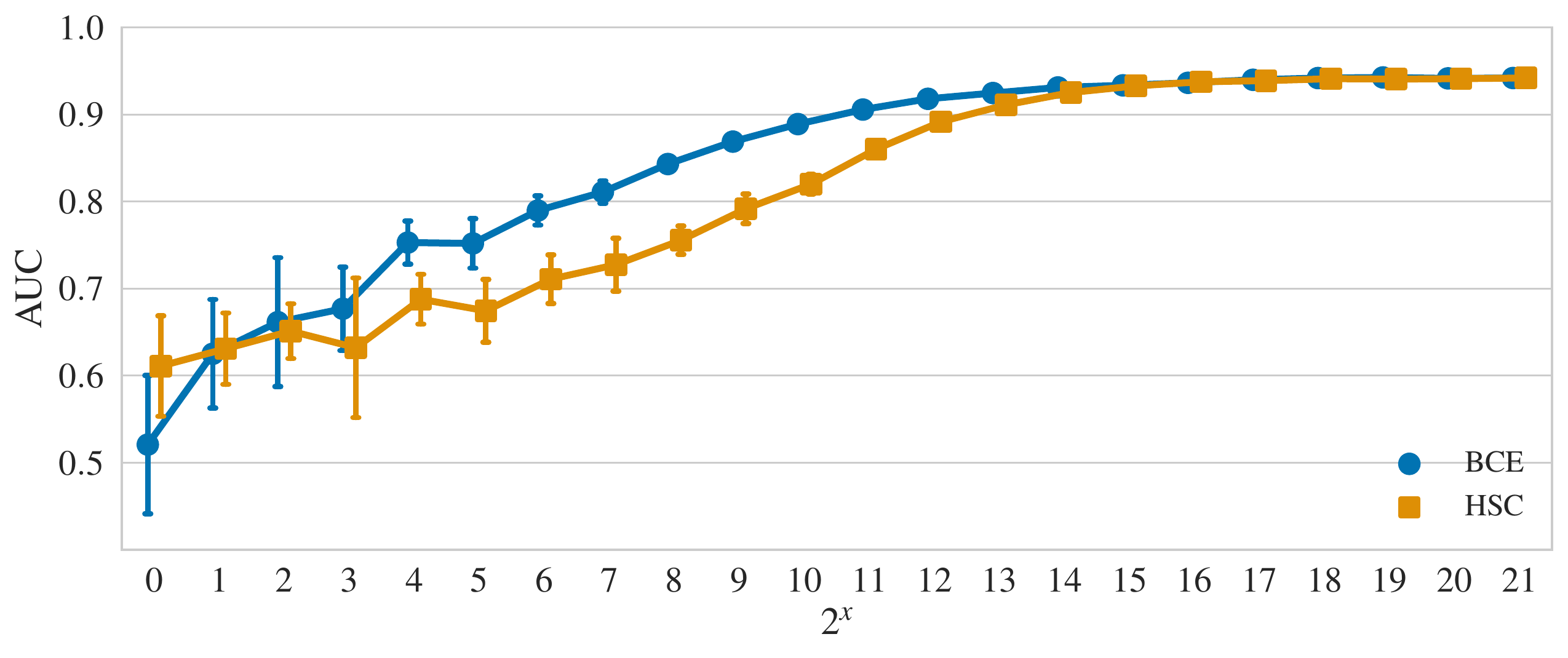}}
\subfigure[Class: frog]{\label{fig:cifarvstiny6}\includegraphics[width=0.49\linewidth]{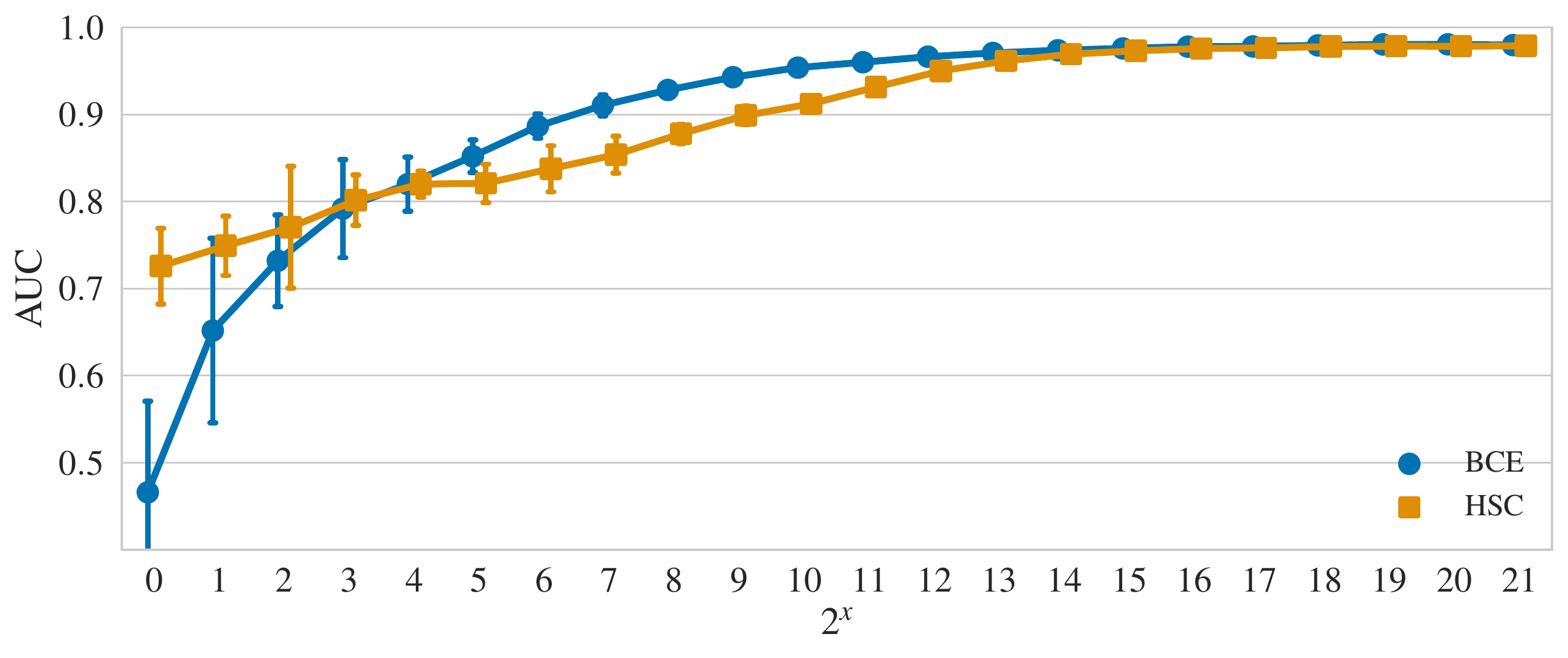}}
\subfigure[Class: horse]{\label{fig:cifarvstiny7}\includegraphics[width=0.49\linewidth]{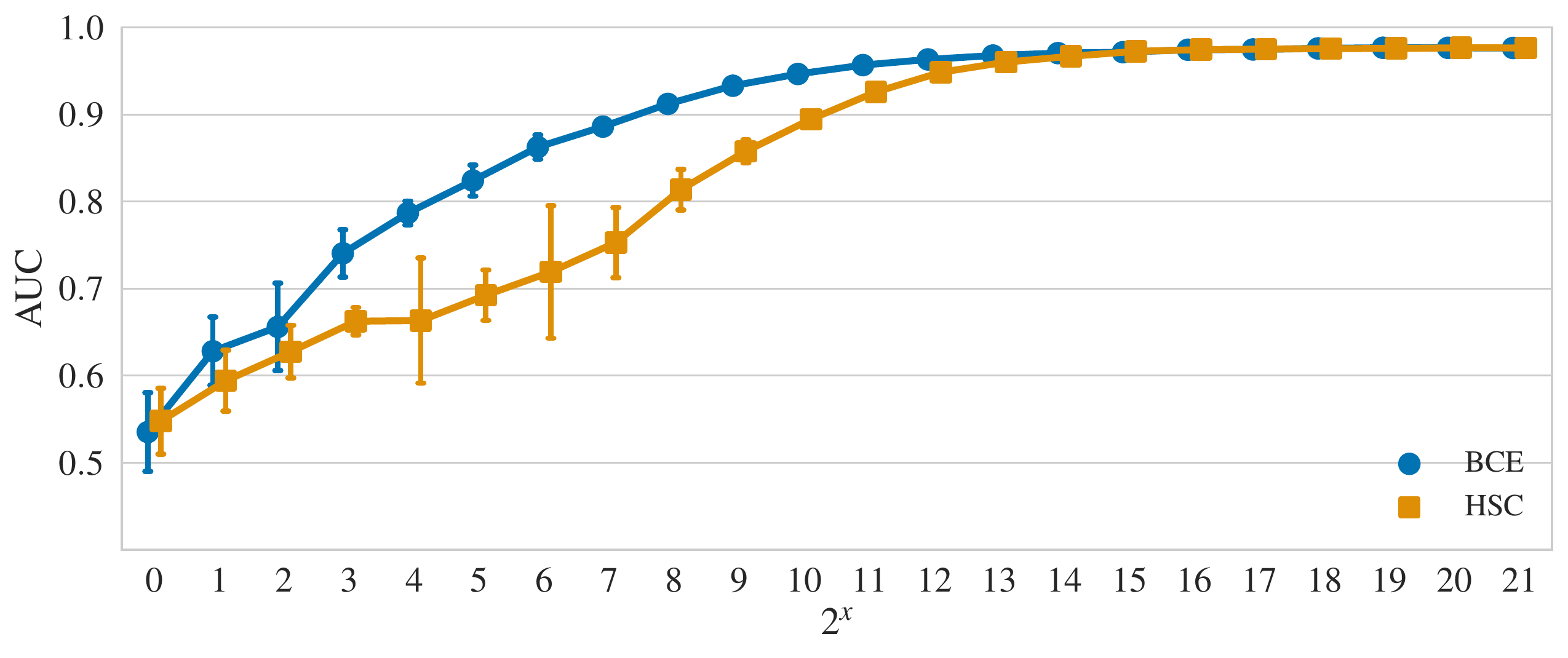}}
\subfigure[Class: ship]{\label{fig:cifarvstiny8}\includegraphics[width=0.49\linewidth]{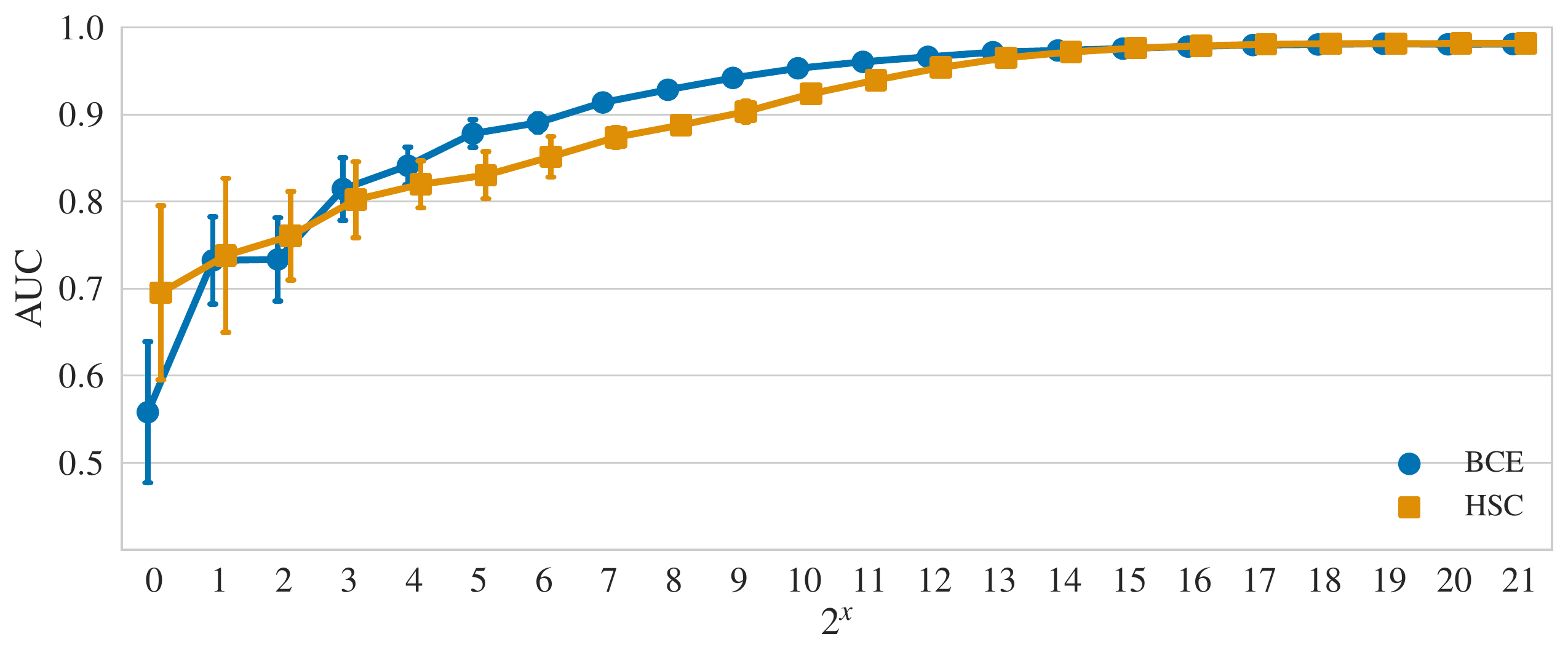}}
\subfigure[Class: truck]{\label{fig:cifarvstiny9}\includegraphics[width=0.49\linewidth]{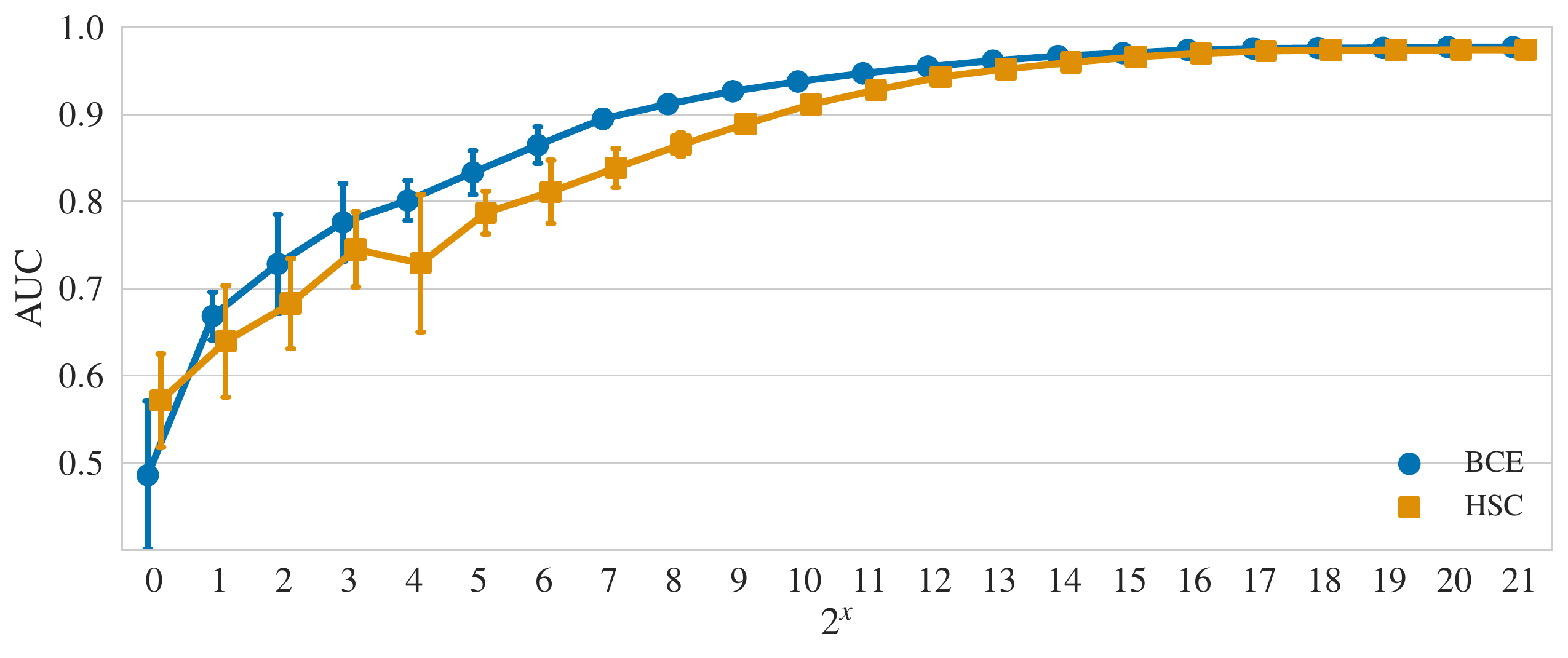}}
\caption{Mean AUC detection performance in \% (over 10 seeds) for all classes of the CIFAR-10 one vs.~rest benchmark from Section \ref{sec:exp_var_oe_size} when varying the number of 80MTI OE samples. These plots correspond to Figure \ref{fig:cifarvstiny}, but here we report the results for all individual classes.}
\label{fig:cifarvstiny_classes}
\end{figure*}

\begin{figure*}[th]
\centering
\subfigure[Class: airplane]{\label{fig:cifar10vscifar100_0}\includegraphics[width=0.328\linewidth]{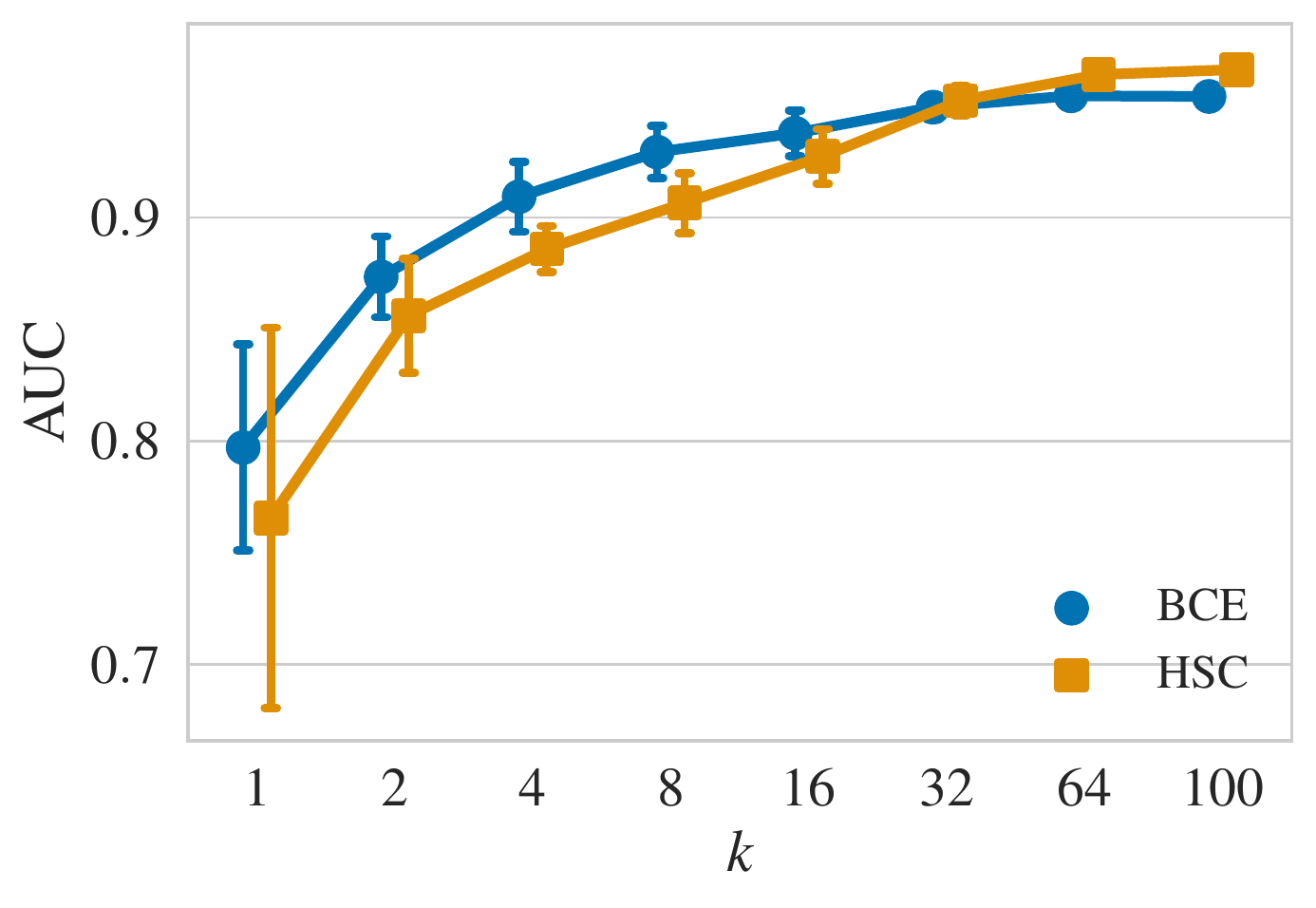}}
\subfigure[Class: automobile]{\label{fig:cifar10vscifar100_1}\includegraphics[width=0.328\linewidth]{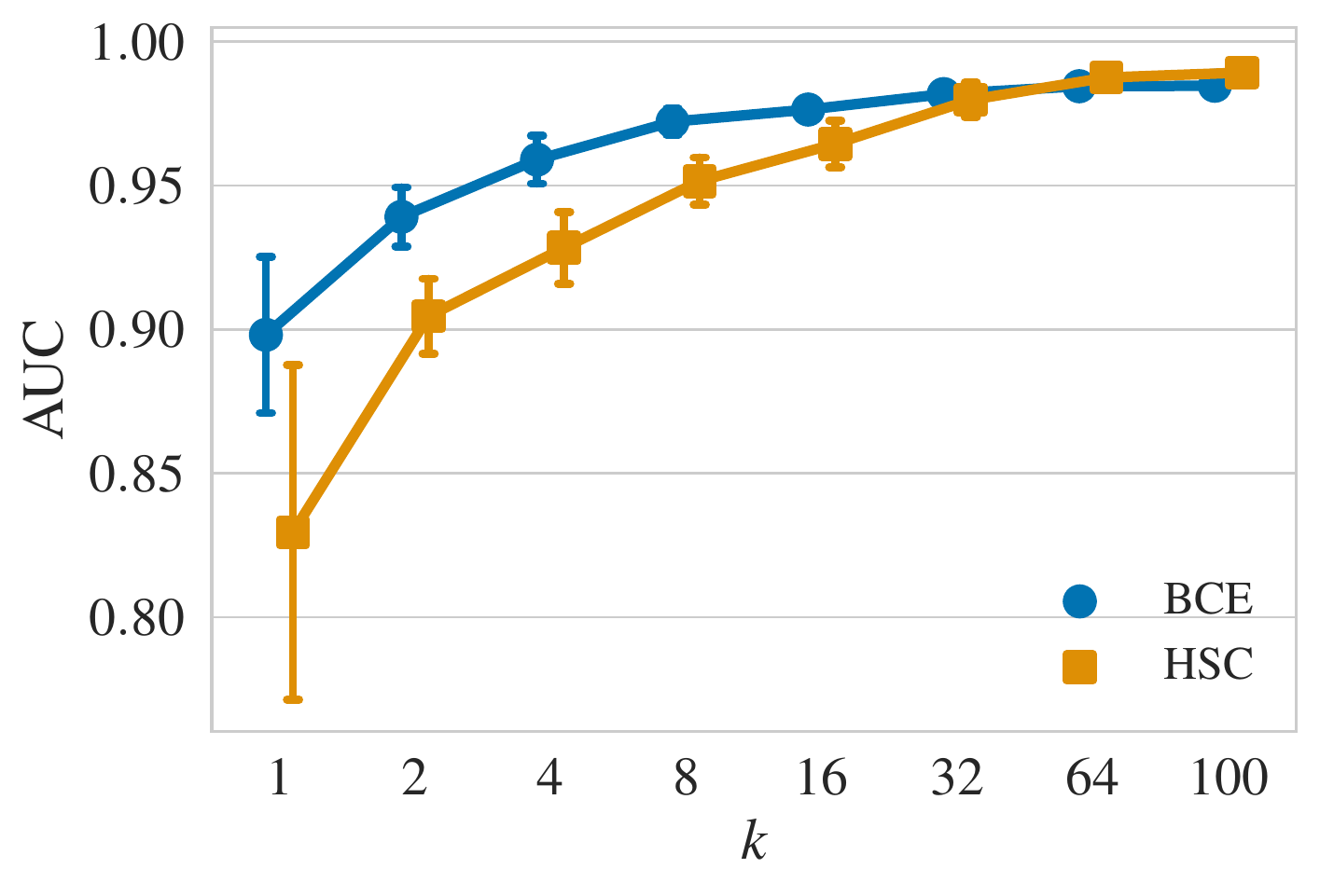}}
\subfigure[Class: bird]{\label{fig:cifar10vscifar100_2}\includegraphics[width=0.328\linewidth]{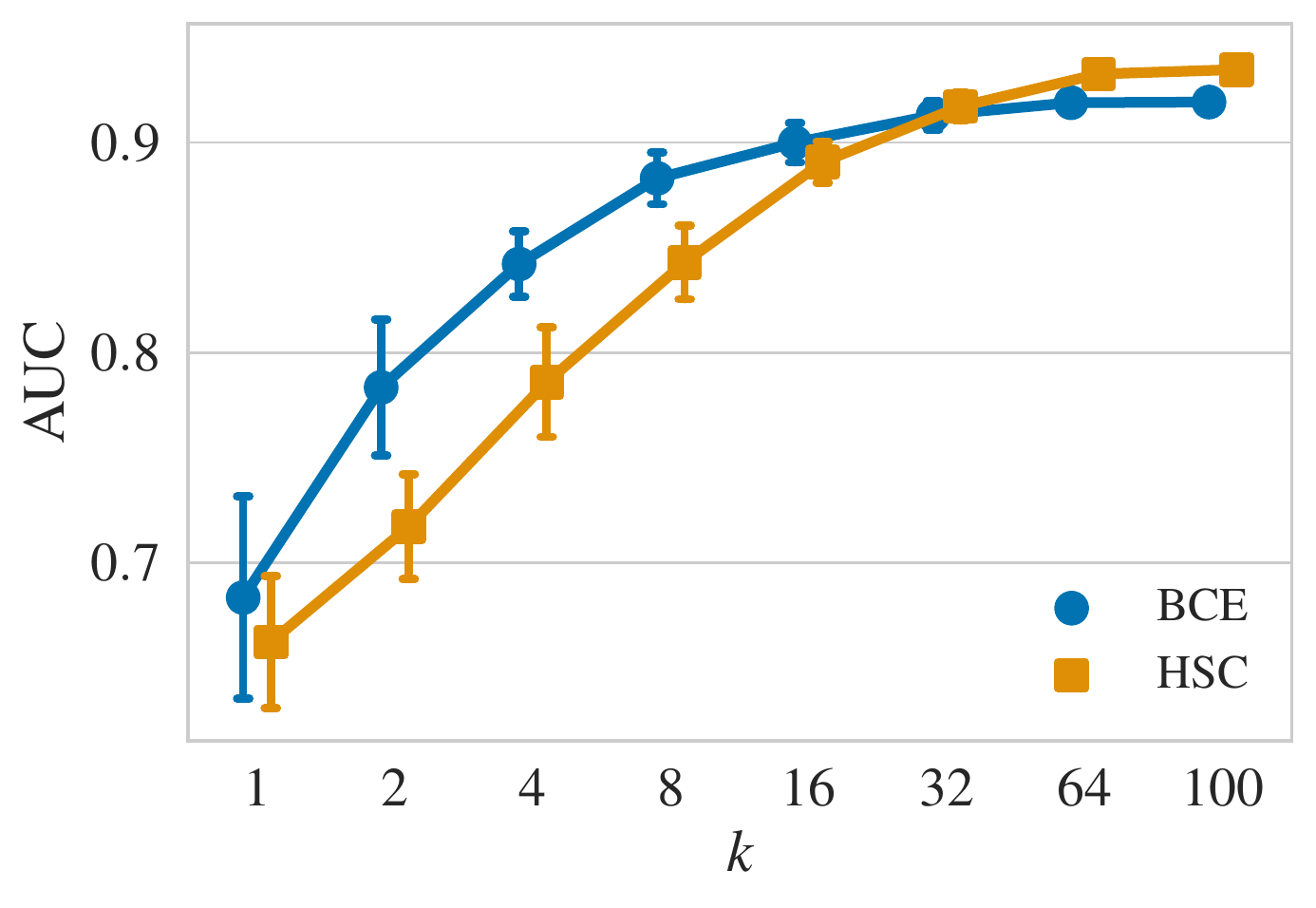}}
\subfigure[Class: cat]{\label{fig:cifar10vscifar100_3}\includegraphics[width=0.328\linewidth]{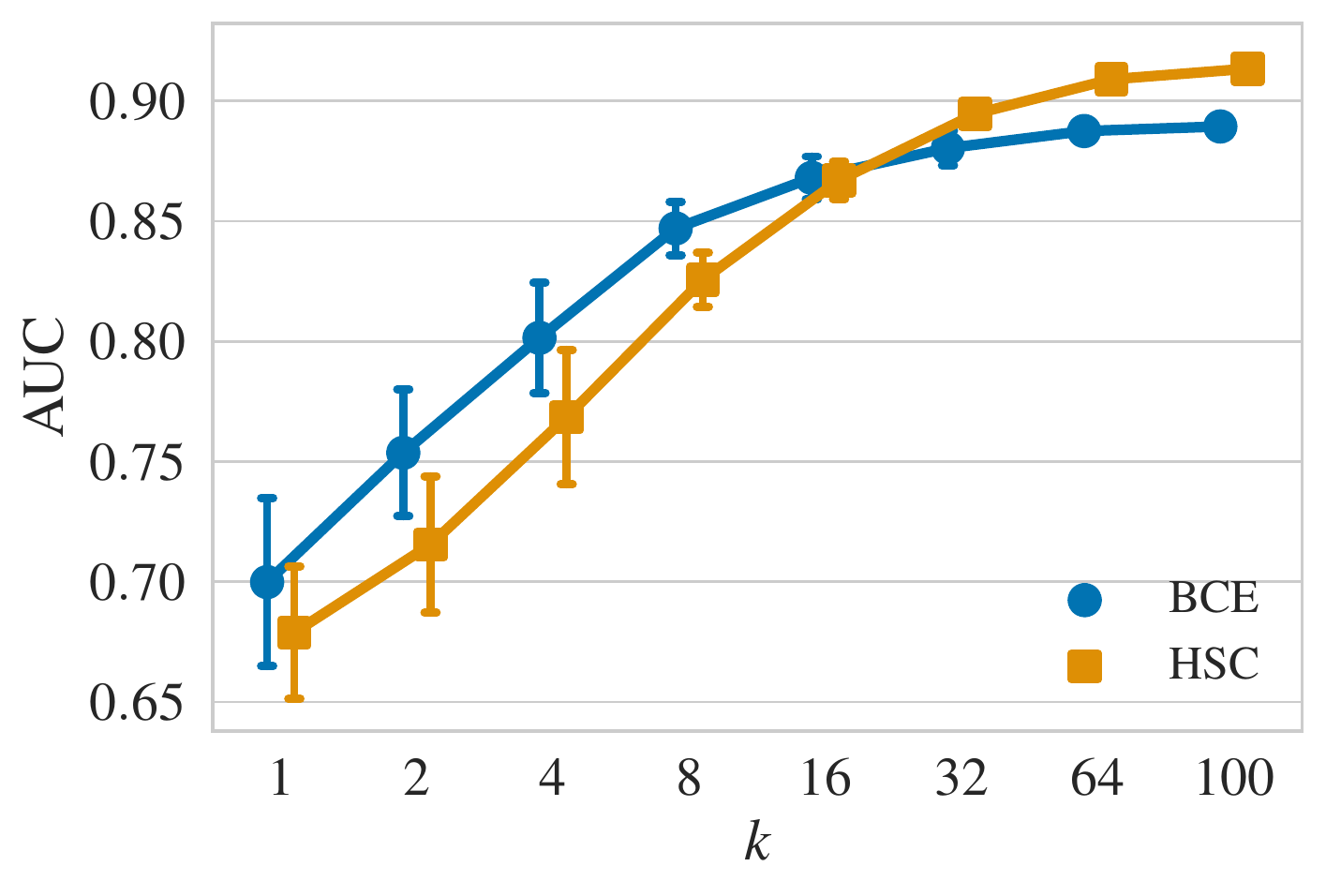}}
\subfigure[Class: deer]{\label{fig:cifar10vscifar100_4}\includegraphics[width=0.328\linewidth]{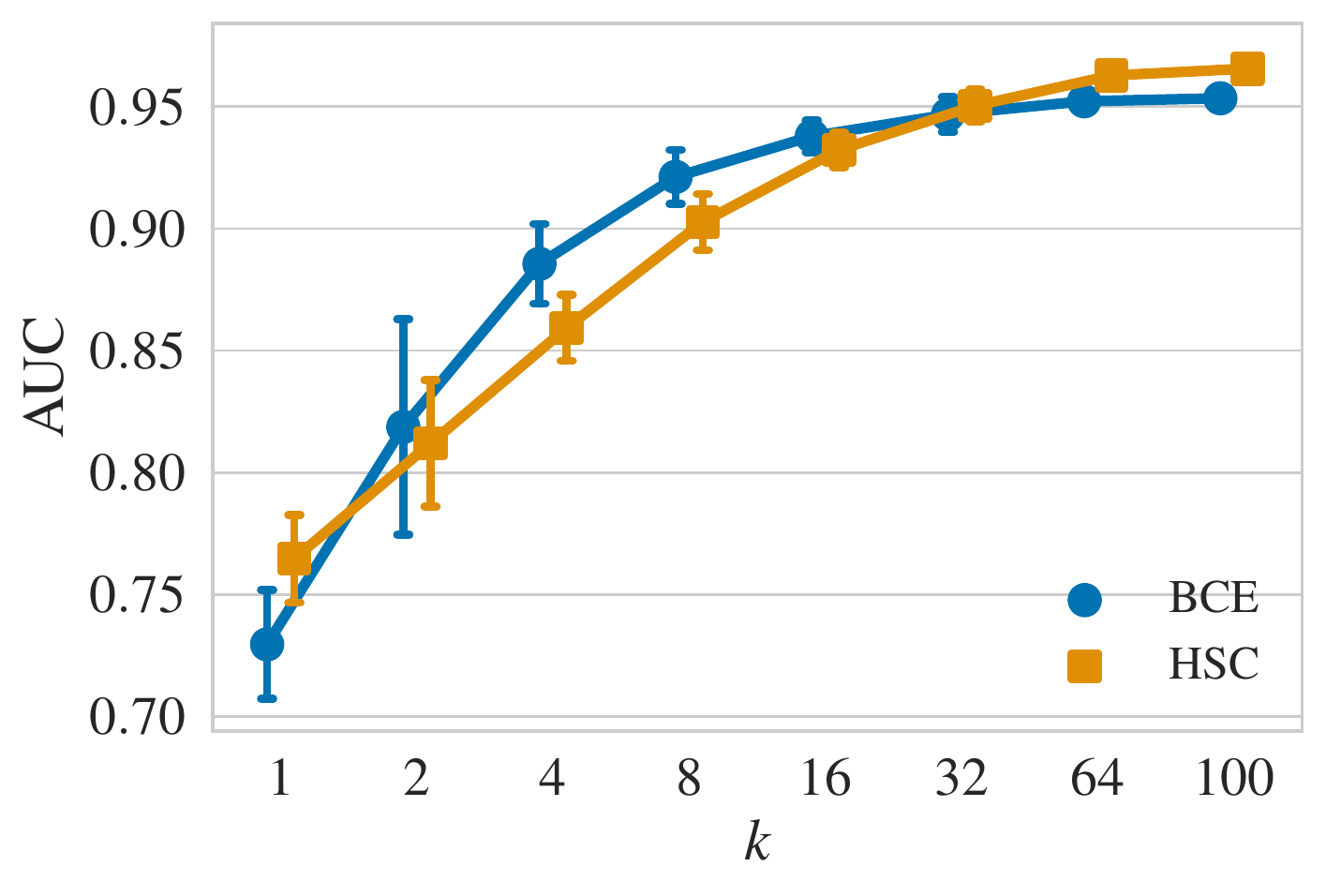}}
\subfigure[Class: dog]{\label{fig:cifar10vscifar100_5}\includegraphics[width=0.328\linewidth]{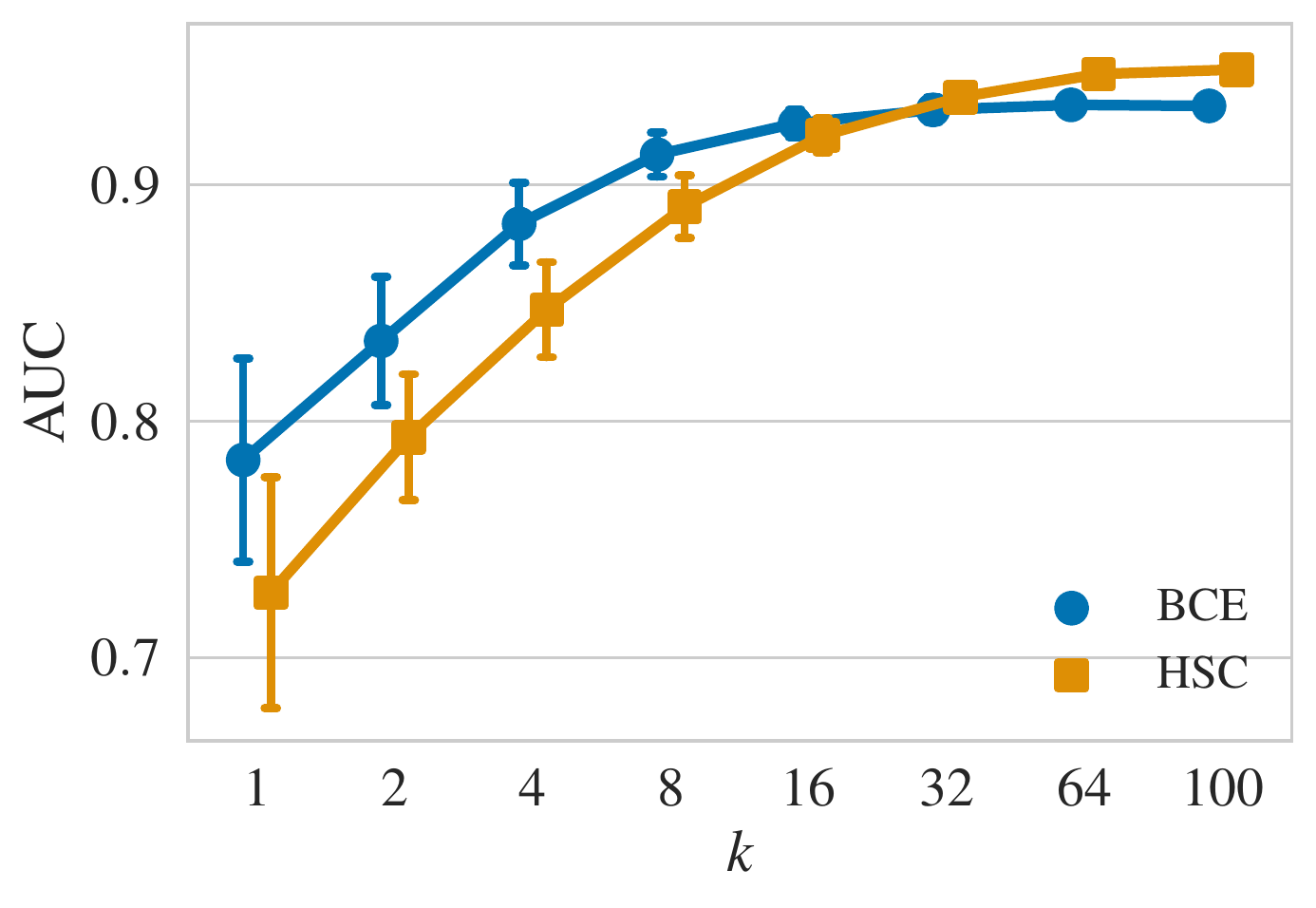}}
\subfigure[Class: frog]{\label{fig:cifar10vscifar100_6}\includegraphics[width=0.328\linewidth]{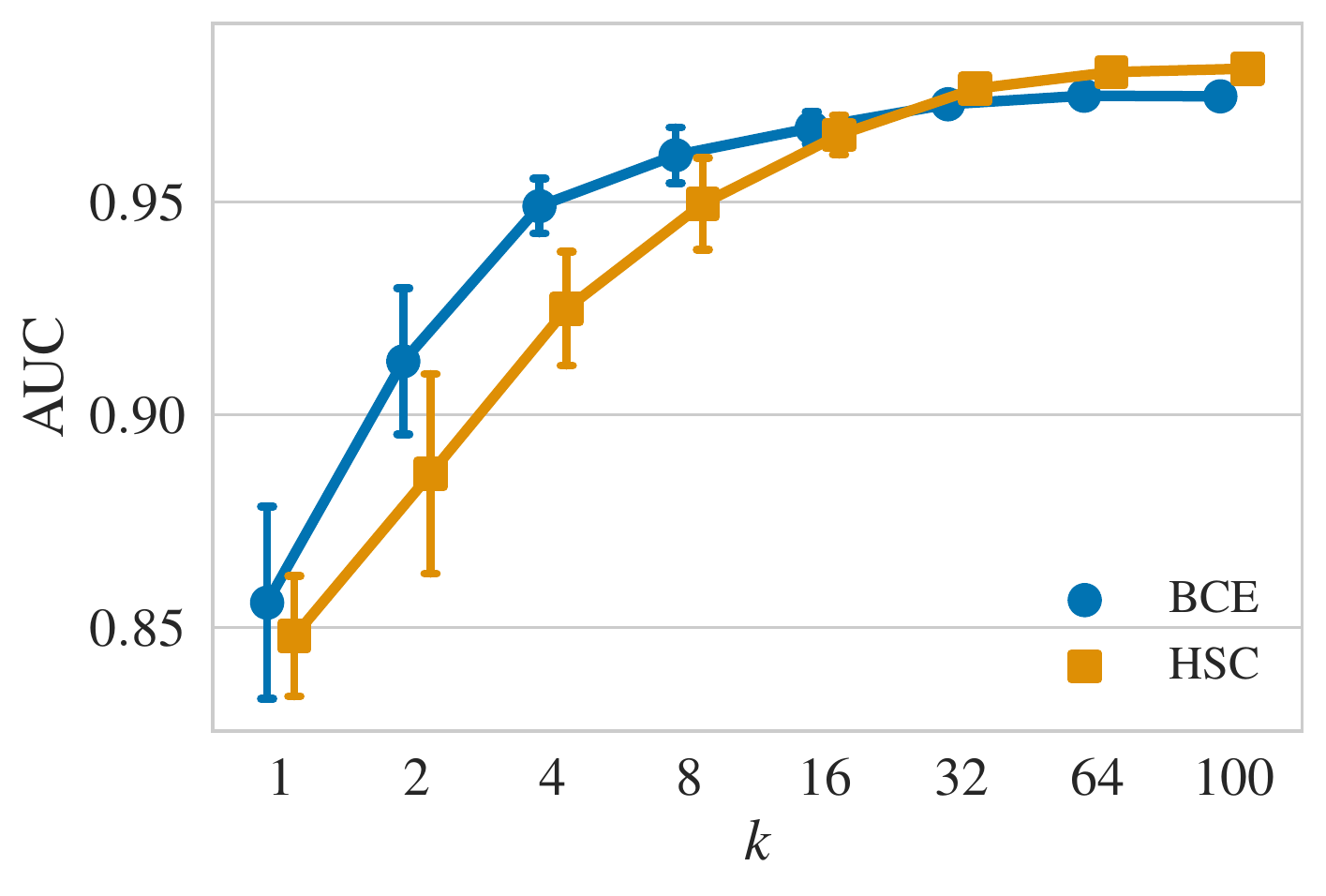}}
\subfigure[Class: horse]{\label{fig:cifar10vscifar100_7}\includegraphics[width=0.328\linewidth]{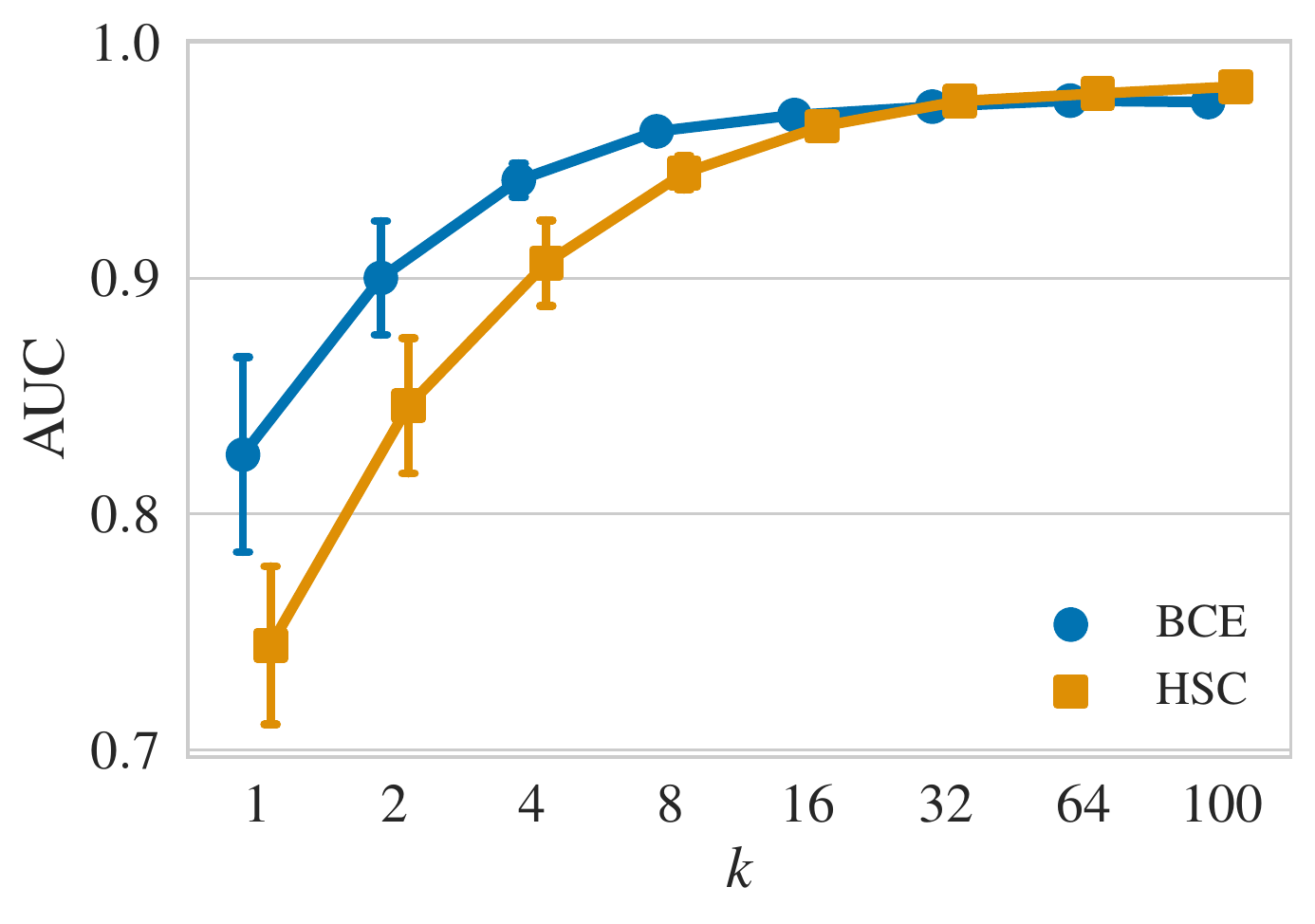}}
\subfigure[Class: ship]{\label{fig:cifar10vscifar100_8}\includegraphics[width=0.328\linewidth]{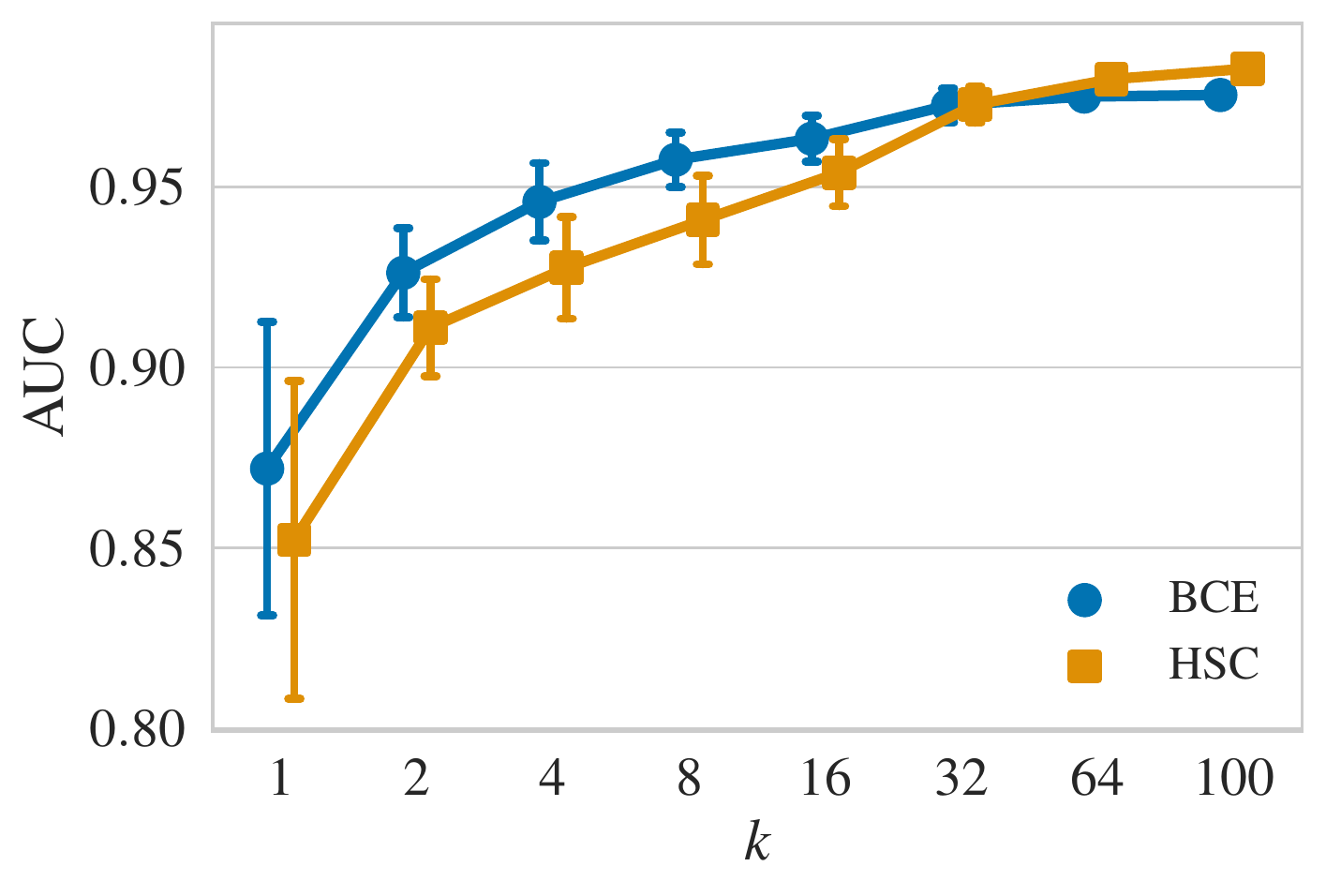}}
\subfigure[Class: truck]{\label{fig:cifar10vscifar100_9}\includegraphics[width=0.328\linewidth]{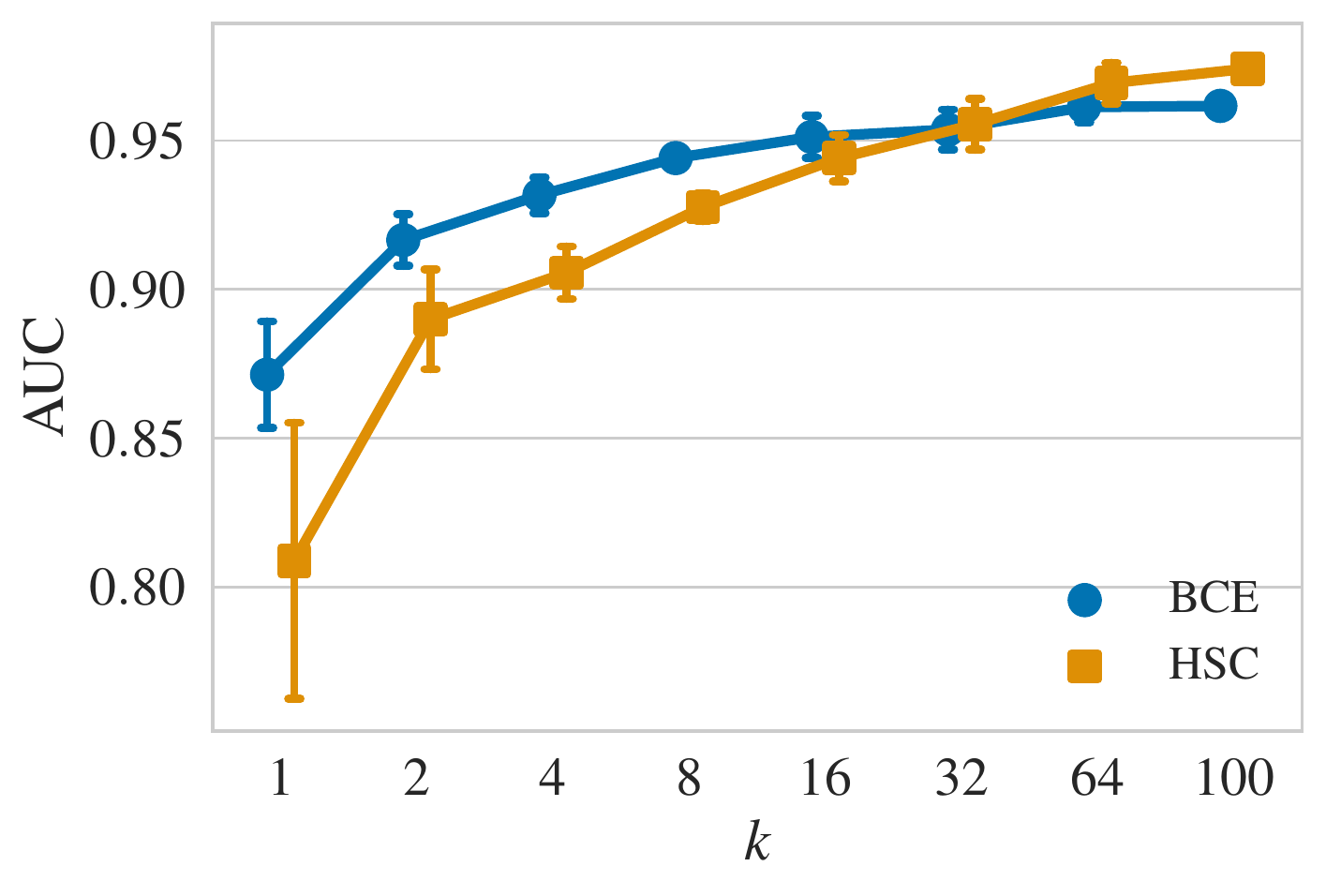}}
\caption{Mean AUC detection performance in \% (over 10 seeds) for all CIFAR-10 classes from the experiment in Appendix \ref{appx:oe_diversity} on varying the number of classes $k$ of the CIFAR-100 OE data. These plots correspond to Figure \ref{fig:oe_diversity}, but here we report the results for all individual classes.}
\label{fig:cifar10vscifar100}
\end{figure*}

For the experiments on varying the number of OE samples (Section \ref{sec:exp_var_oe_size}), we include plots for all individual classes in Figure \ref{fig:cifarvstiny_classes} for CIFAR-10 and in Figures \ref{fig:imagenet1kvs22k_classes1} and \ref{fig:imagenet1kvs22k_classes2} for ImageNet-30, respectively.
Additionally, for the experiments on varying the diversity of OE data on CIFAR-10 with CIFAR-100 OE, we added the plots for all individual classes in Figure \ref{fig:cifar10vscifar100}.

%%%%%%%%%%%%%%%%%%%%%%%%%%%%%%%%%%%%%%%%%%%%%%%%%%%%%%%%%%%%
%%%%%%%%%%%%%%%%%%%%%%%%%%%%%%%%%%%%%%%%%%%%%%%%%%%%%%%%%%%%
%%%%%%%%%%%%%%%%%%%%%%%%%%%%%%%%%%%%%%%%%%%%%%%%%%%%%%%%%%%%
%%%%%%%%%%%%%%%%%%%%%%%%%%%%%%%%%%%%%%%%%%%%%%%%%%%%%%%%%%%%
%%%%%%%%%%%%%%%%%%%%%%%%%%%%%%%%%%%%%%%%%%%%%%%%%%%%%%%%%%%%
%%%%%%%%%%%%%%%%%%%%%%%%%%%%%%%%%%%%%%%%%%%%%%%%%%%%%%%%%%%%
%%%%%%%%%%%%%%%%%%%%%%%%%%%%%%%%%%%%%%%%%%%%%%%%%%%%%%%%%%%%
%%%%%%%%%%%%%%%%%%%%%%%%%%%%%%%%%%%%%%%%%%%%%%%%%%%%%%%%%%%%

\clearpage
\subsection{Robustness towards choice of OE samples}
\begin{figure}[htb] 
\centering \small 
\subfigure[ 
``airplane'' is normal 
]{ 
\includegraphics[width=0.24\textwidth]{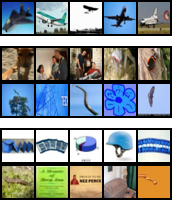} 
} 
\subfigure[ 
``automobile'' is normal 
]{ 
\includegraphics[width=0.24\textwidth]{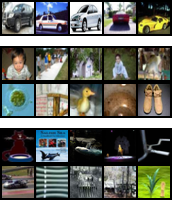} 
} 
\subfigure[ 
``bird'' is normal 
]{ 
\includegraphics[width=0.24\textwidth]{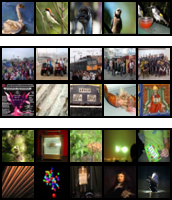} 
} 
\subfigure[ 
``cat'' is normal 
]{ 
\includegraphics[width=0.24\textwidth]{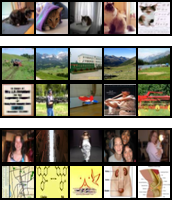} 
} 
\subfigure[ 
``deer'' is normal 
]{ 
\includegraphics[width=0.24\textwidth]{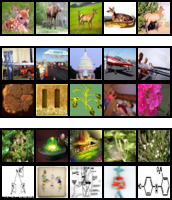} 
} 
\subfigure[ 
``dog'' is normal 
]{ 
\includegraphics[width=0.24\textwidth]{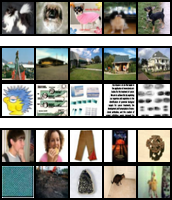} 
} 
\subfigure[ 
``frog'' is normal 
]{ 
\includegraphics[width=0.24\textwidth]{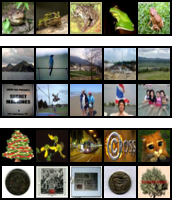} 
} 
\subfigure[ 
``horse'' is normal 
]{ 
\includegraphics[width=0.24\textwidth]{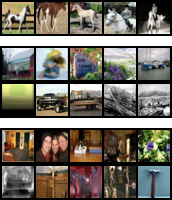} 
} 
\subfigure[ 
``ship'' is normal 
]{ 
\includegraphics[width=0.24\textwidth]{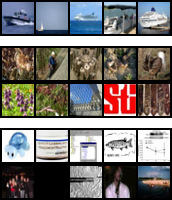} 
} 
\subfigure[ 
``truck'' is normal 
]{ 
\includegraphics[width=0.24\textwidth]{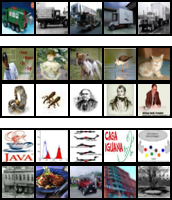} 
} 
\vspace{-1em} 
\caption{ 
Optimal OE samples for CIFAR-10 with 80MTI as OE. The first row shows normal samples, the next two rows the best samples found via HSC (top) and BCE (bottom), and the last two rows the worst samples found via HSC (top) and BCE (bottom).  
} 
\label{fig:evolve_cifar_plain} 
\end{figure}

\begin{figure}[htb] 
\centering \small 
\subfigure[ 
``airplane'' is normal 
]{ 
\includegraphics[width=0.24\textwidth]{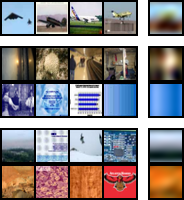} 
} 
\subfigure[ 
``automobile'' is normal 
]{ 
\includegraphics[width=0.24\textwidth]{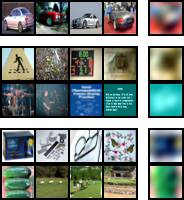} 
} 
\subfigure[ 
``bird'' is normal 
]{ 
\includegraphics[width=0.24\textwidth]{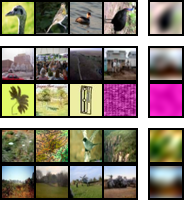} 
} 
\subfigure[ 
``cat'' is normal 
]{ 
\includegraphics[width=0.24\textwidth]{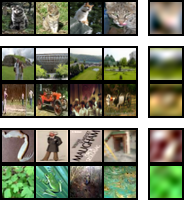} 
} 
\subfigure[ 
``deer'' is normal 
]{ 
\includegraphics[width=0.24\textwidth]{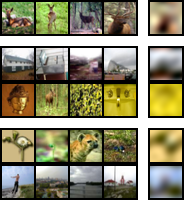} 
} 
\subfigure[ 
``dog'' is normal 
]{ 
\includegraphics[width=0.24\textwidth]{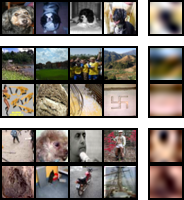} 
} 
\subfigure[ 
``frog'' is normal 
]{ 
\includegraphics[width=0.24\textwidth]{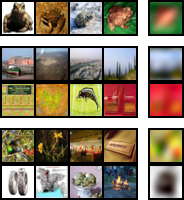} 
} 
\subfigure[ 
``horse'' is normal 
]{ 
\includegraphics[width=0.24\textwidth]{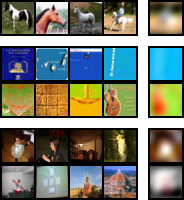} 
} 
\subfigure[ 
``ship'' is normal 
]{ 
\includegraphics[width=0.24\textwidth]{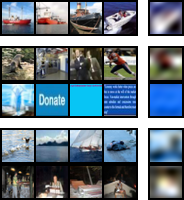} 
} 
\subfigure[ 
``truck'' is normal 
]{ 
\includegraphics[width=0.24\textwidth]{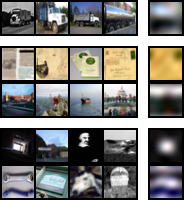} 
} 
\vspace{-1em} 
\caption{ 
Optimal OE samples for low-pass-filtered CIFAR-10 with low-pass-filtered 80MTI as OE. The first row shows normal samples, the next two rows the best samples found via HSC (top) and BCE (bottom), and the last two rows the worst samples found via HSC (top) and BCE (bottom). The last column shows the low-pass-filtered version of the images, which is what the network sees during training and testing.  
} 
\label{fig:evolve_cifar_lpf} 
\end{figure}

\begin{figure}[htb] 
\centering \small 
\subfigure[ 
``airplane'' is normal 
]{ 
\includegraphics[width=0.24\textwidth]{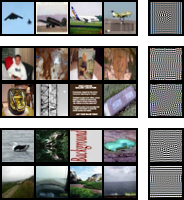} 
} 
\subfigure[ 
``automobile'' is normal 
]{ 
\includegraphics[width=0.24\textwidth]{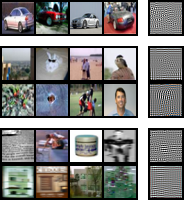} 
} 
\subfigure[ 
``bird'' is normal 
]{ 
\includegraphics[width=0.24\textwidth]{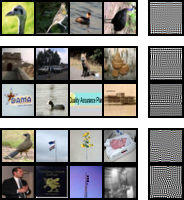} 
} 
\subfigure[ 
``cat'' is normal 
]{ 
\includegraphics[width=0.24\textwidth]{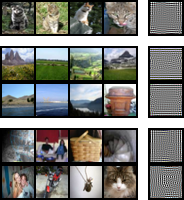} 
} 
\subfigure[ 
``deer'' is normal 
]{ 
\includegraphics[width=0.24\textwidth]{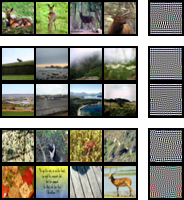} 
} 
\subfigure[ 
``dog'' is normal 
]{ 
\includegraphics[width=0.24\textwidth]{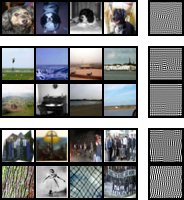} 
} 
\subfigure[ 
``frog'' is normal 
]{ 
\includegraphics[width=0.24\textwidth]{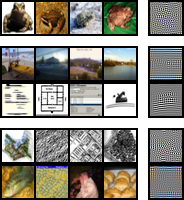} 
} 
\subfigure[ 
``horse'' is normal 
]{ 
\includegraphics[width=0.24\textwidth]{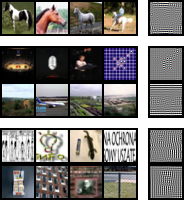} 
} 
\subfigure[ 
``ship'' is normal 
]{ 
\includegraphics[width=0.24\textwidth]{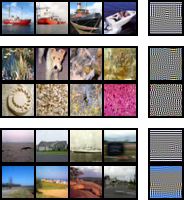} 
} 
\subfigure[ 
``truck'' is normal 
]{ 
\includegraphics[width=0.24\textwidth]{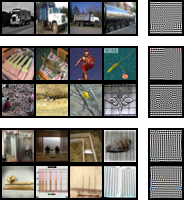} 
} 
\vspace{-1em} 
\caption{ 
Optimal OE samples for high-pass-filtered CIFAR-10 with high-pass-filtered 80MTI as OE. The first row shows normal samples, the next two rows the best samples found via HSC (top) and BCE (bottom), and the last two rows the worst samples found via HSC (top) and BCE (bottom). The last column shows the high-pass-filtered version of the images, which is what the network sees during training and testing.  
} 
\label{fig:evolve_cifar_hpf} 
\end{figure}

\begin{figure}[htb] 
\centering \small 
\subfigure[ 
``acorn'' is normal 
]{ 
\includegraphics[width=0.24\textwidth]{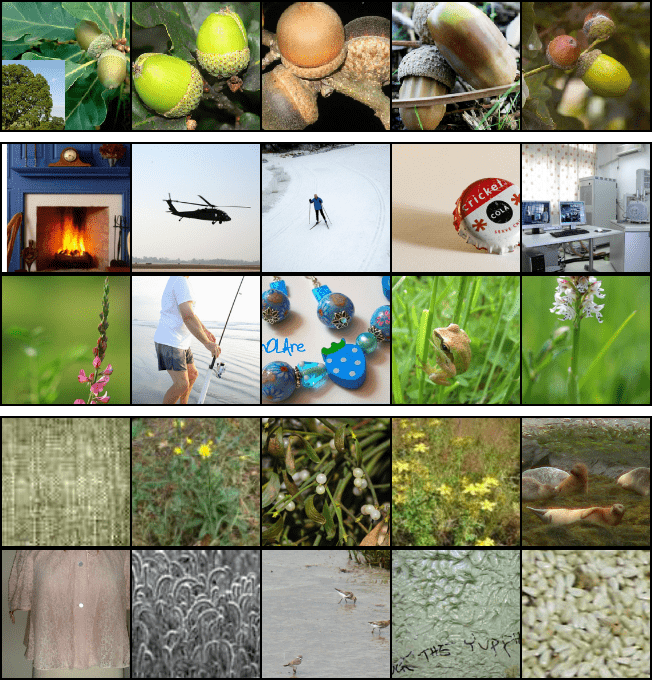} 
} 
\subfigure[ 
``airliner'' is normal 
]{ 
\includegraphics[width=0.24\textwidth]{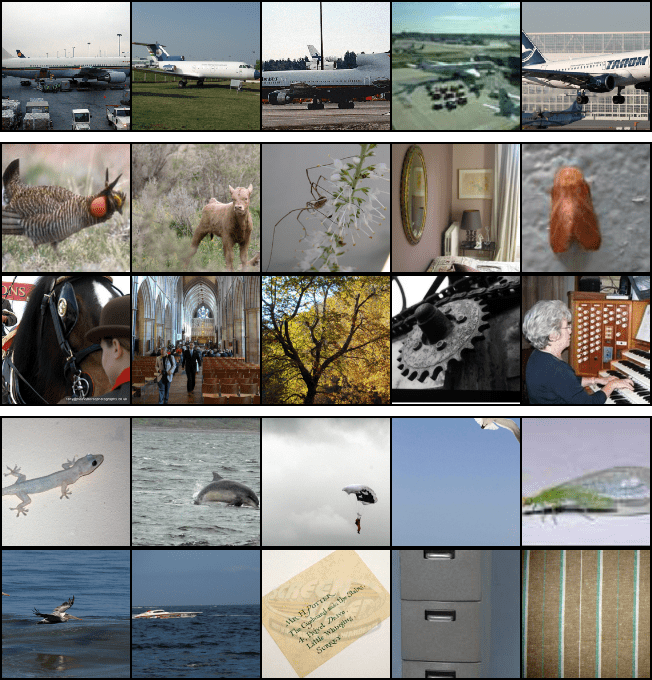} 
} 
\subfigure[ 
``ambulance'' is normal 
]{ 
\includegraphics[width=0.24\textwidth]{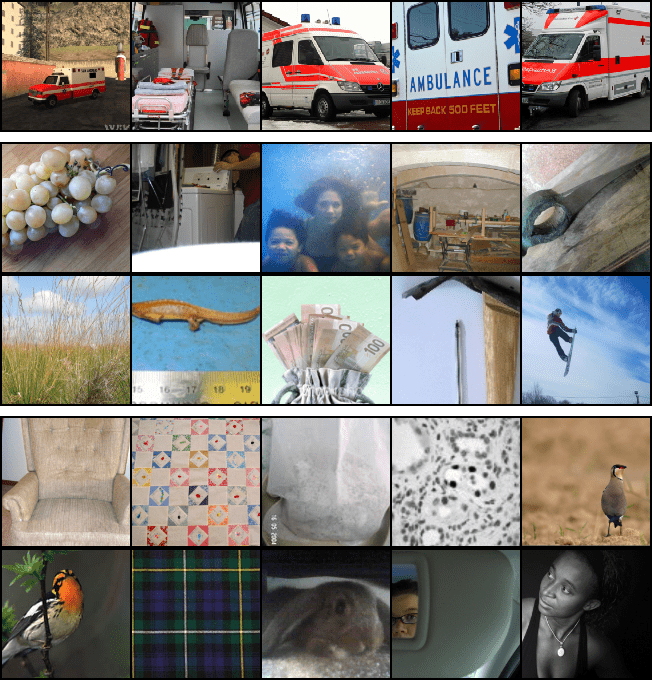} 
} 
\subfigure[ 
``american alligator'' is normal 
]{ 
\includegraphics[width=0.24\textwidth]{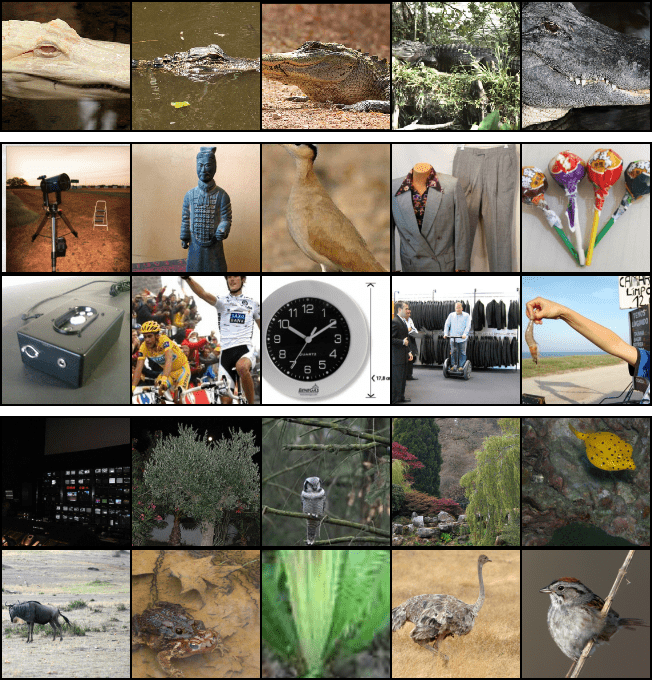} 
} 
\subfigure[ 
``banjo'' is normal 
]{ 
\includegraphics[width=0.24\textwidth]{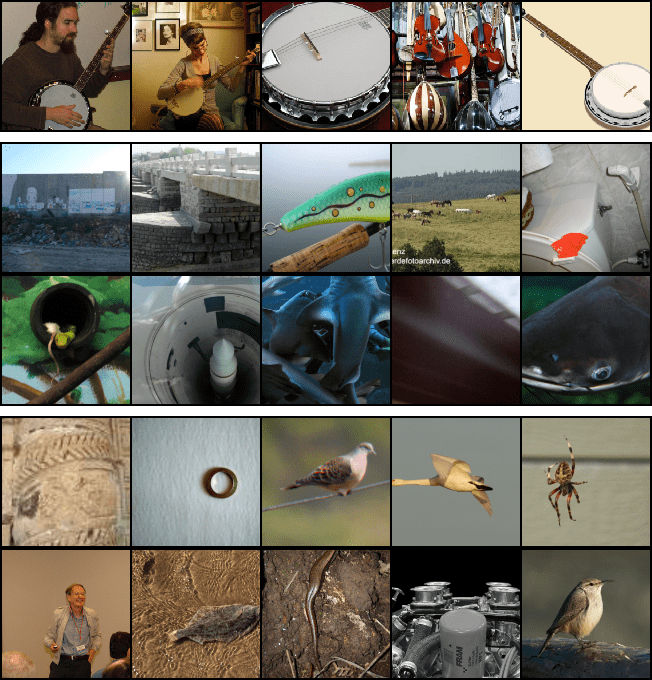} 
} 
\subfigure[ 
``barn'' is normal 
]{ 
\includegraphics[width=0.24\textwidth]{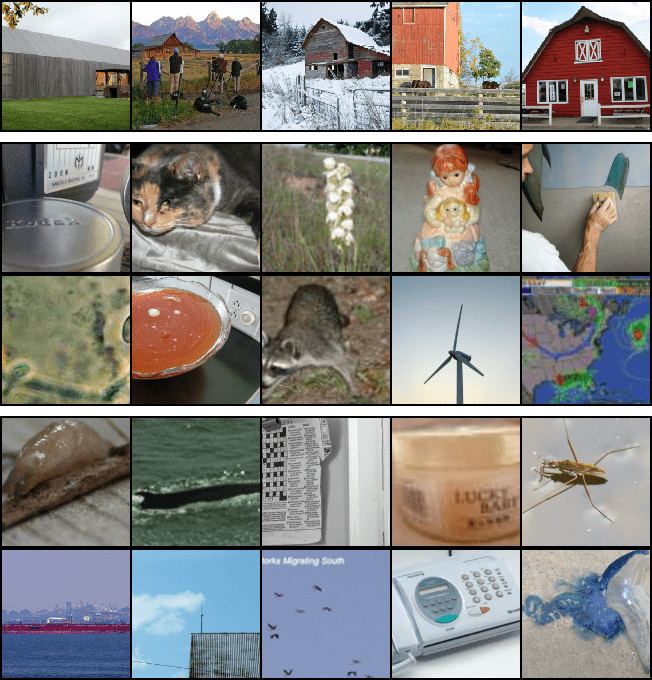} 
} 
\subfigure[ 
``bikini'' is normal 
]{ 
\includegraphics[width=0.24\textwidth]{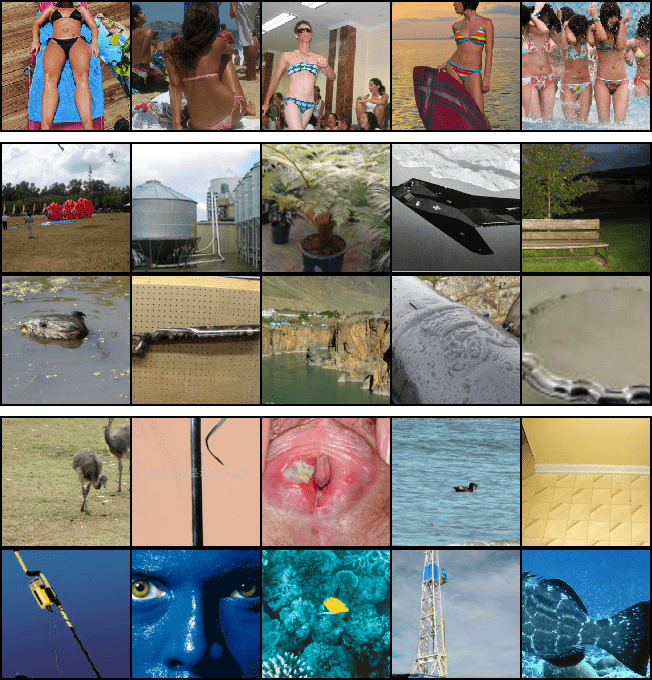} 
} 
\subfigure[ 
``digital clock'' is normal 
]{ 
\includegraphics[width=0.24\textwidth]{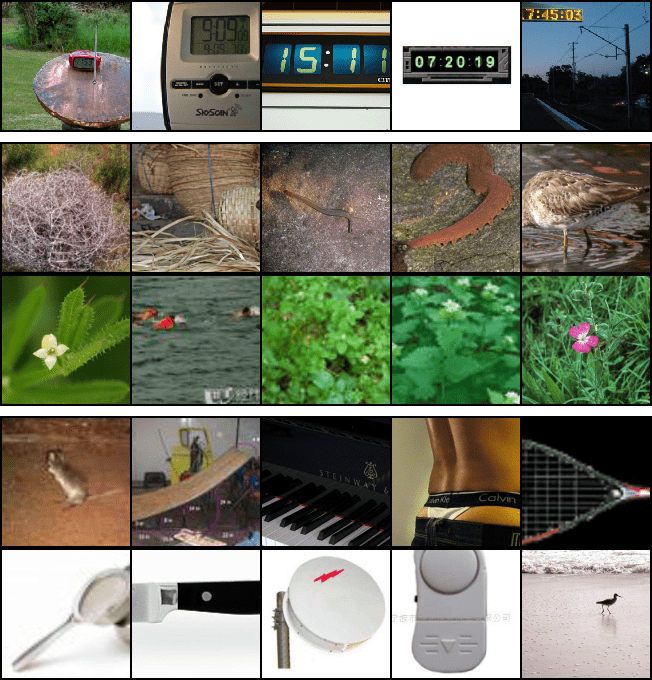} 
} 
\subfigure[ 
``dragonfly'' is normal 
]{ 
\includegraphics[width=0.24\textwidth]{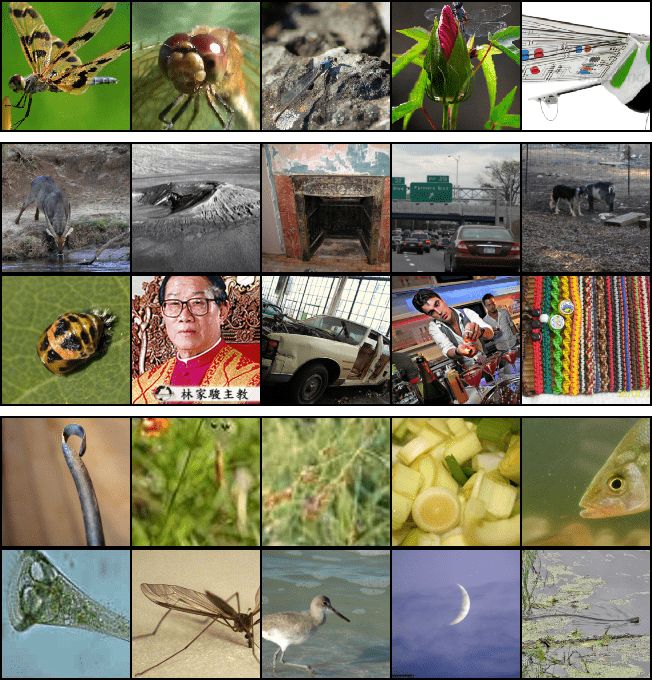} 
} 
\subfigure[ 
``dumbbell'' is normal 
]{ 
\includegraphics[width=0.24\textwidth]{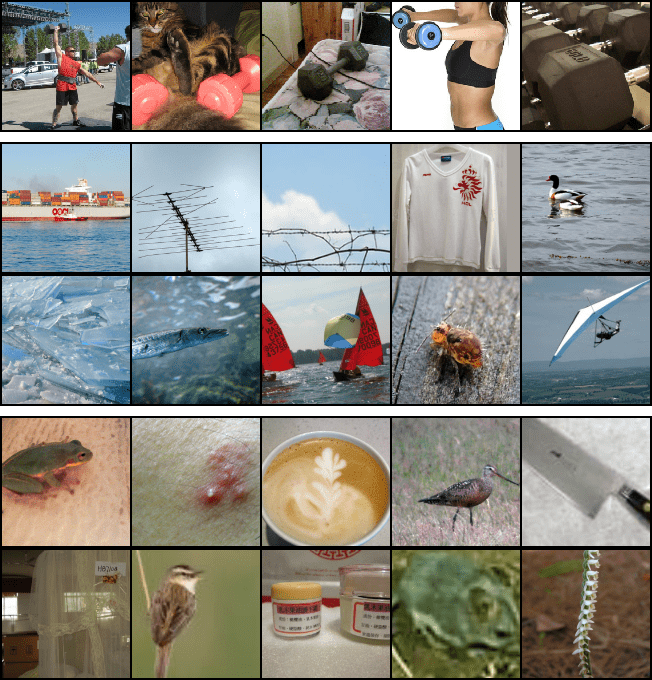} 
} 
\vspace{-1em} 
\caption{ 
Optimal OE samples for ImageNet1k with ImageNet22k (with the 1K classes removed) as OE. The first row shows normal samples, the next two rows the best samples found via HSC (top) and BCE (bottom), and the last two rows the worst samples found via HSC (top) and BCE (bottom).  
} 
\label{fig:evolve_imagenet_plain} 
\end{figure}

\begin{figure}[htb] 
\centering \small 
\subfigure[ 
``acorn'' is normal 
]{ 
\includegraphics[width=0.24\textwidth]{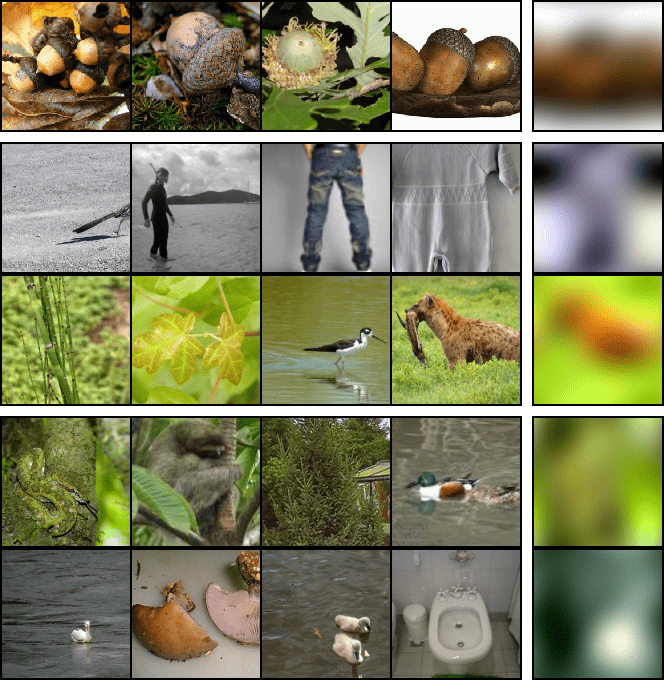} 
} 
\subfigure[ 
``airliner'' is normal 
]{ 
\includegraphics[width=0.24\textwidth]{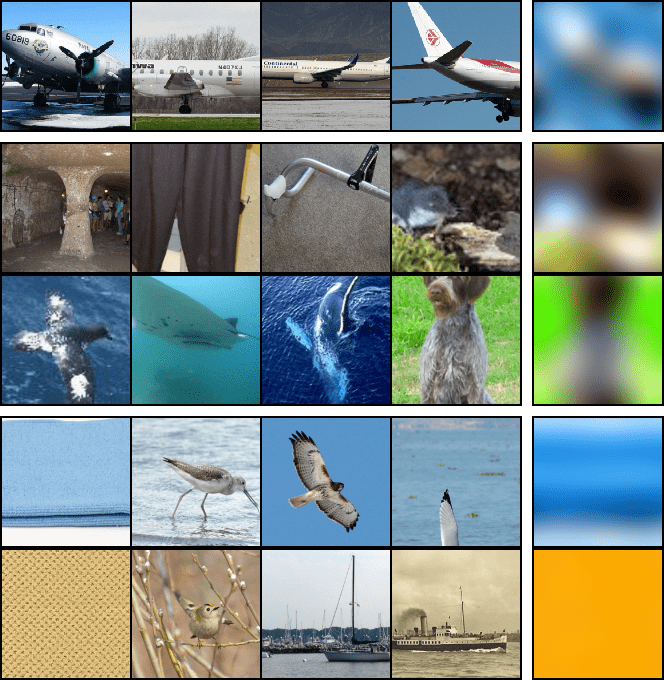} 
} 
\subfigure[ 
``ambulance'' is normal 
]{ 
\includegraphics[width=0.24\textwidth]{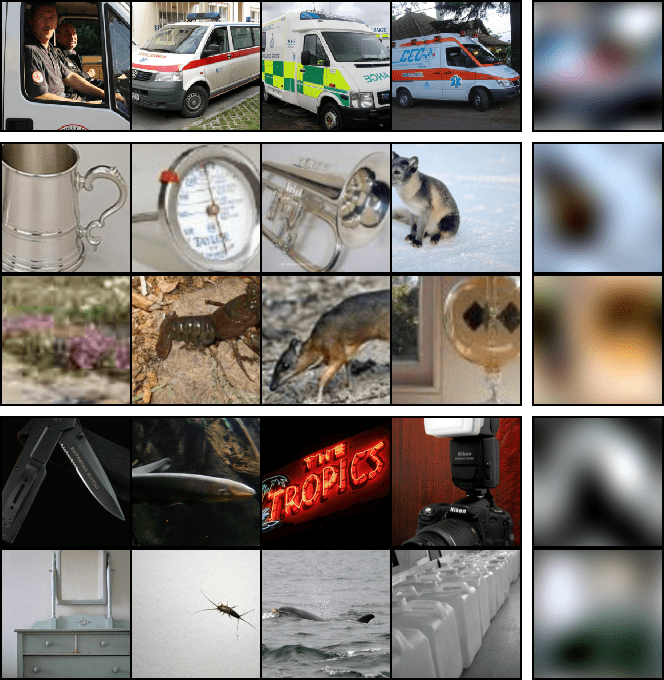} 
} 
\subfigure[ 
``american alligator'' is normal 
]{ 
\includegraphics[width=0.24\textwidth]{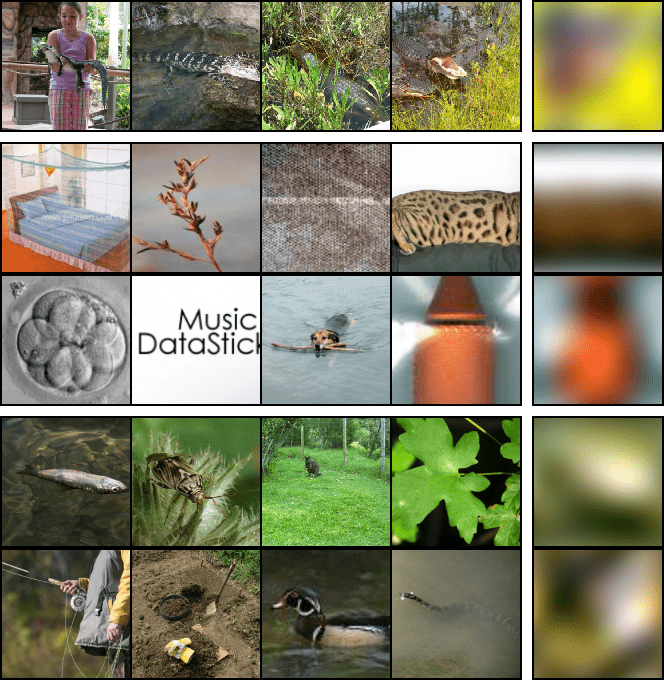} 
} 
\subfigure[ 
``banjo'' is normal 
]{ 
\includegraphics[width=0.24\textwidth]{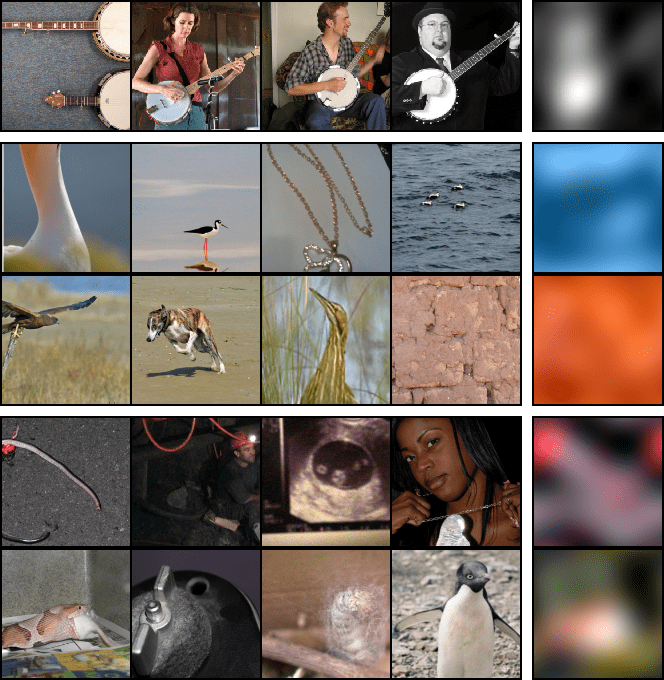} 
} 
\subfigure[ 
``barn'' is normal 
]{ 
\includegraphics[width=0.24\textwidth]{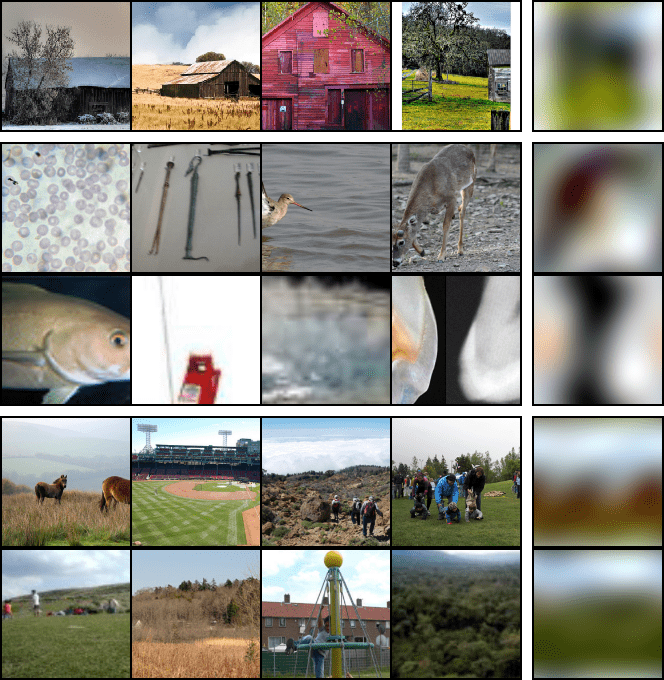} 
} 
\subfigure[ 
``bikini'' is normal 
]{ 
\includegraphics[width=0.24\textwidth]{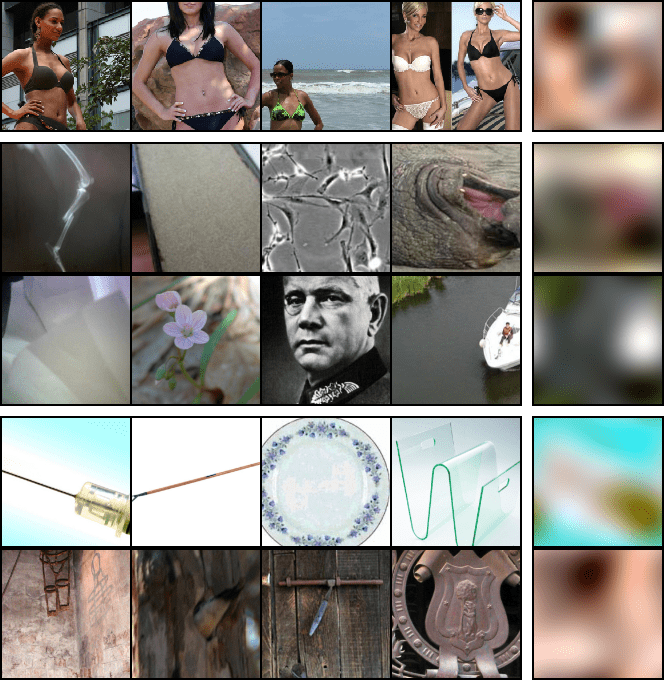} 
} 
\subfigure[ 
``digital clock'' is normal 
]{ 
\includegraphics[width=0.24\textwidth]{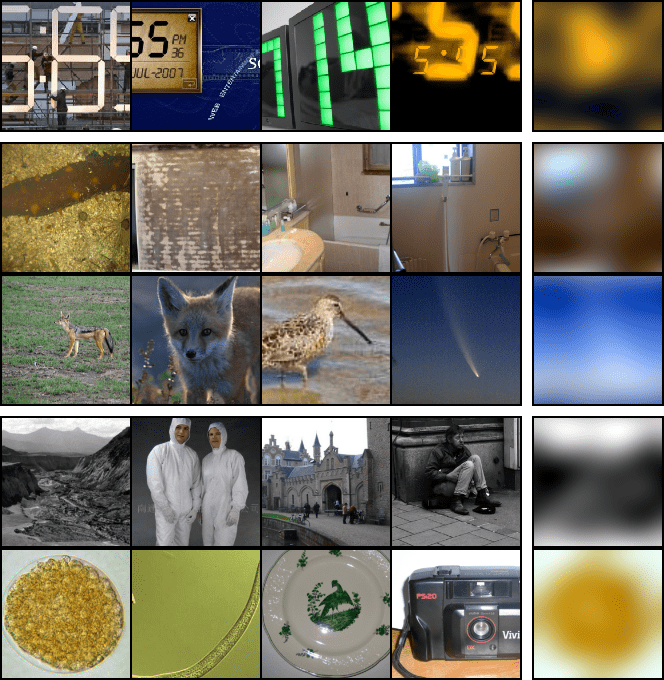} 
} 
\subfigure[ 
``dragonfly'' is normal 
]{ 
\includegraphics[width=0.24\textwidth]{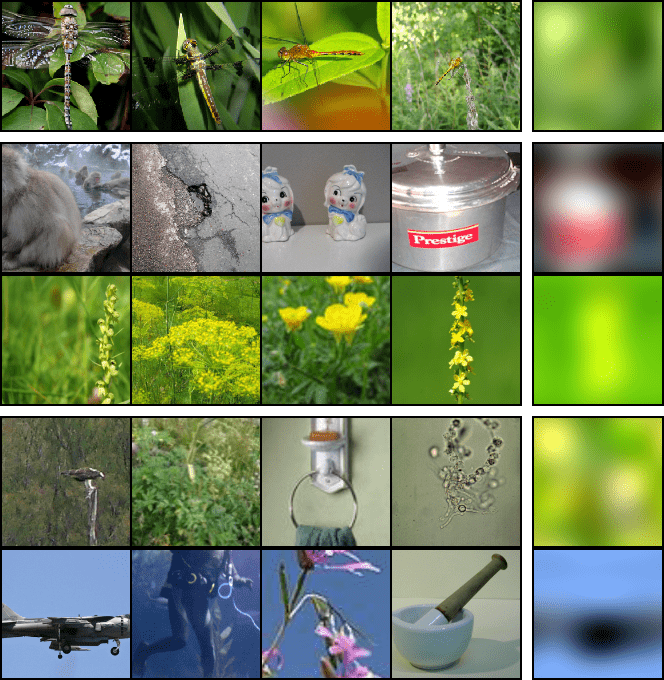} 
} 
\subfigure[ 
``dumbbell'' is normal 
]{ 
\includegraphics[width=0.24\textwidth]{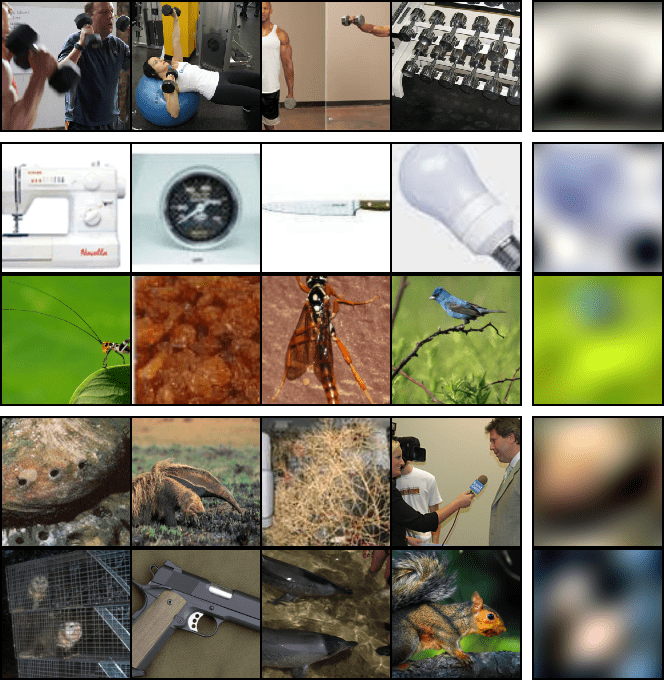} 
} 
\vspace{-1em} 
\caption{ 
Optimal OE samples for low-pass-filtered ImageNet1k with low-pass-filtered ImageNet22k (with the 1K classes removed) as OE. The first row shows normal samples, the next two rows the best samples found via HSC (top) and BCE (bottom), and the last two rows the worst samples found via HSC (top) and BCE (bottom). The last column shows the low-pass-filtered version of the images, which is what the network sees during training and testing.  
} 
\label{fig:evolve_imagenet_lpf} 
\end{figure}

\begin{figure}[htb] 
\centering \small 
\subfigure[ 
``acorn'' is normal 
]{ 
\includegraphics[width=0.24\textwidth]{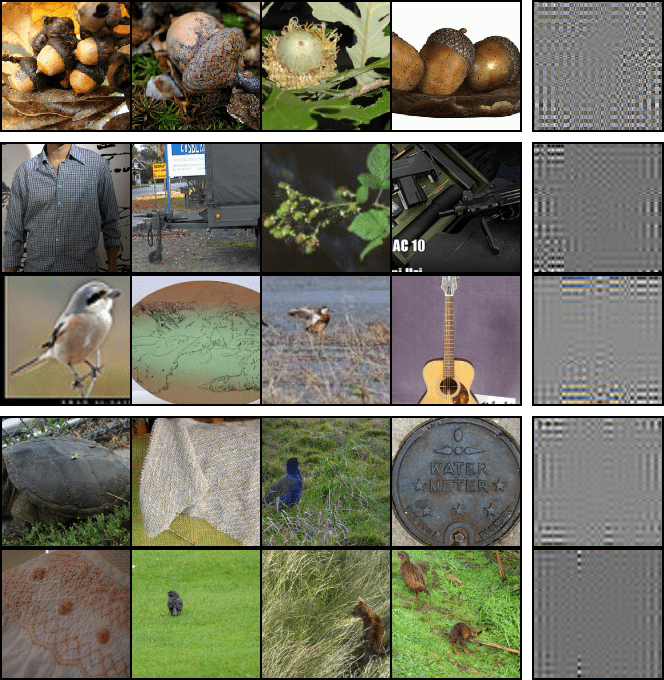} 
} 
\subfigure[ 
``airliner'' is normal 
]{ 
\includegraphics[width=0.24\textwidth]{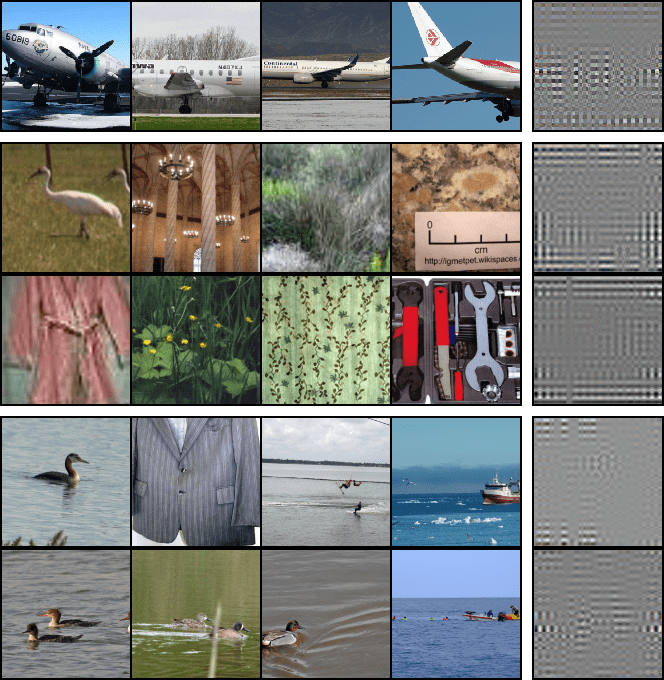} 
} 
\subfigure[ 
``ambulance'' is normal 
]{ 
\includegraphics[width=0.24\textwidth]{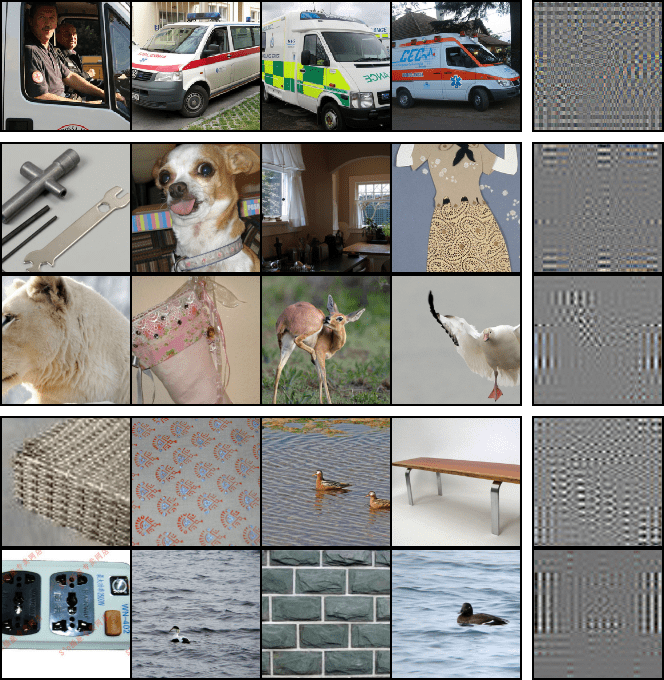} 
} 
\subfigure[ 
``american alligator'' is normal 
]{ 
\includegraphics[width=0.24\textwidth]{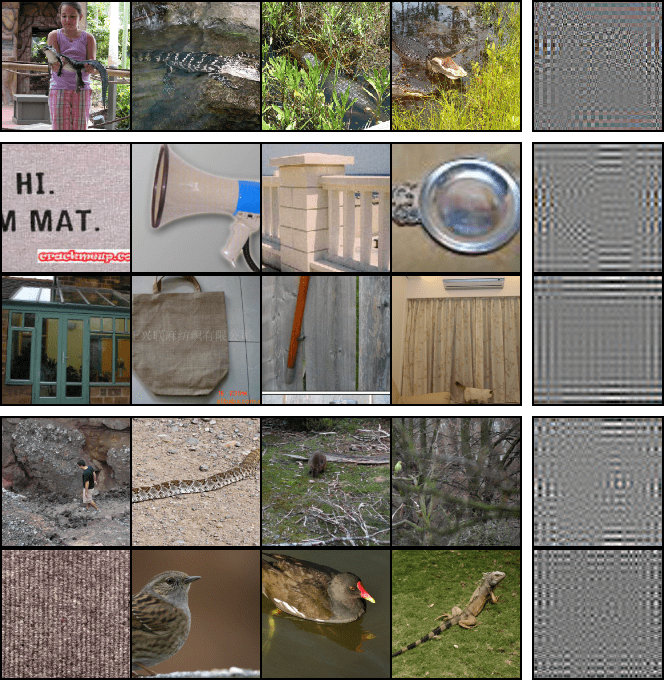} 
} 
\subfigure[ 
``banjo'' is normal 
]{ 
\includegraphics[width=0.24\textwidth]{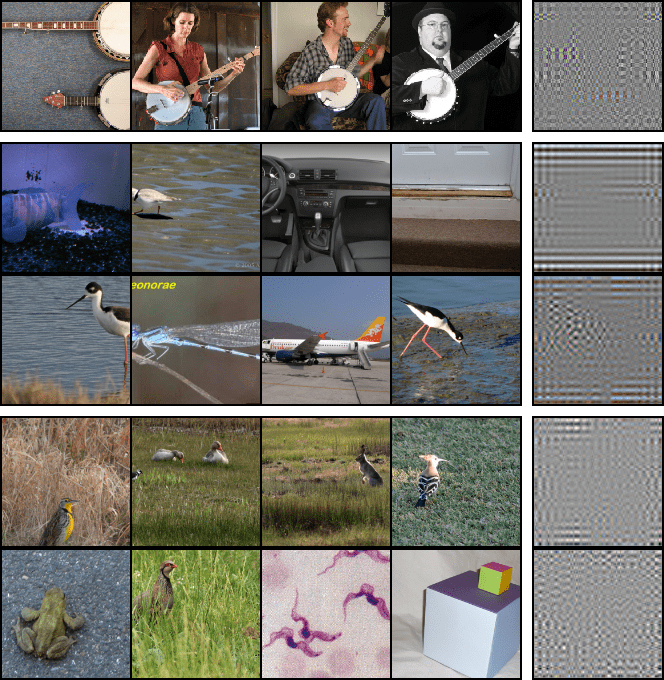} 
} 
\subfigure[ 
``barn'' is normal 
]{ 
\includegraphics[width=0.24\textwidth]{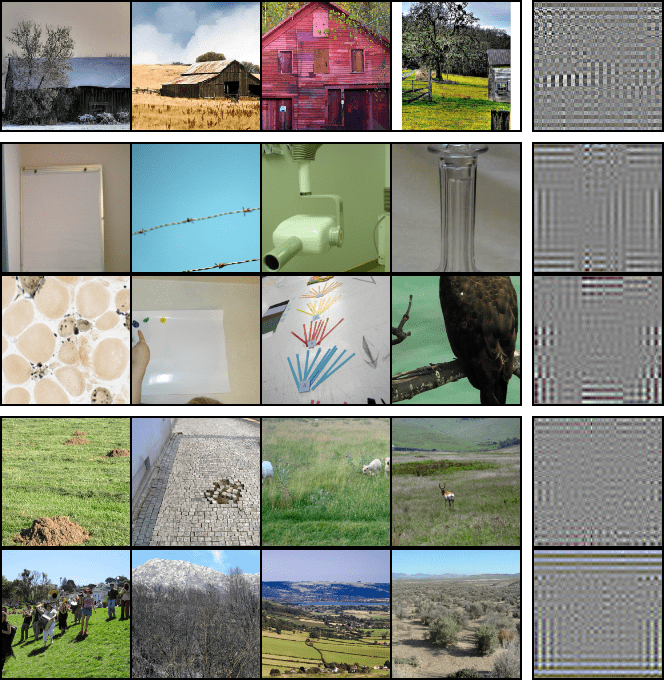} 
} 
\subfigure[ 
``bikini'' is normal 
]{ 
\includegraphics[width=0.24\textwidth]{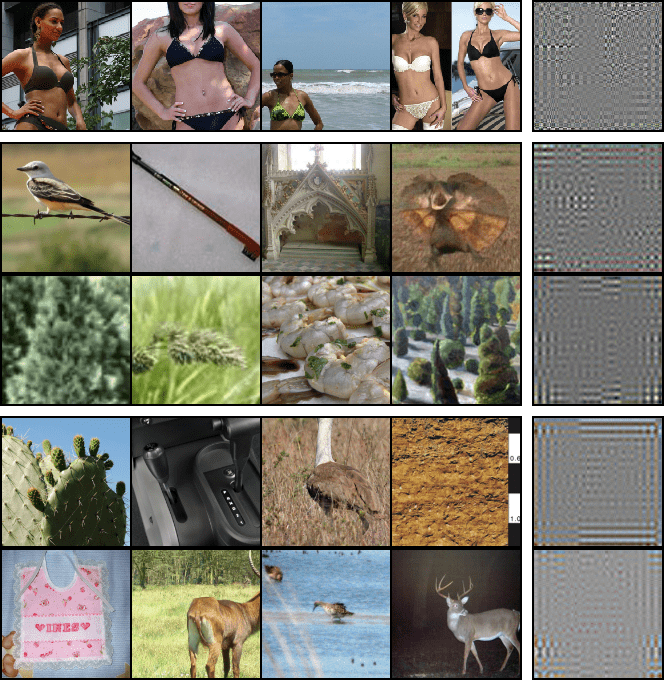} 
} 
\subfigure[ 
``digital clock'' is normal 
]{ 
\includegraphics[width=0.24\textwidth]{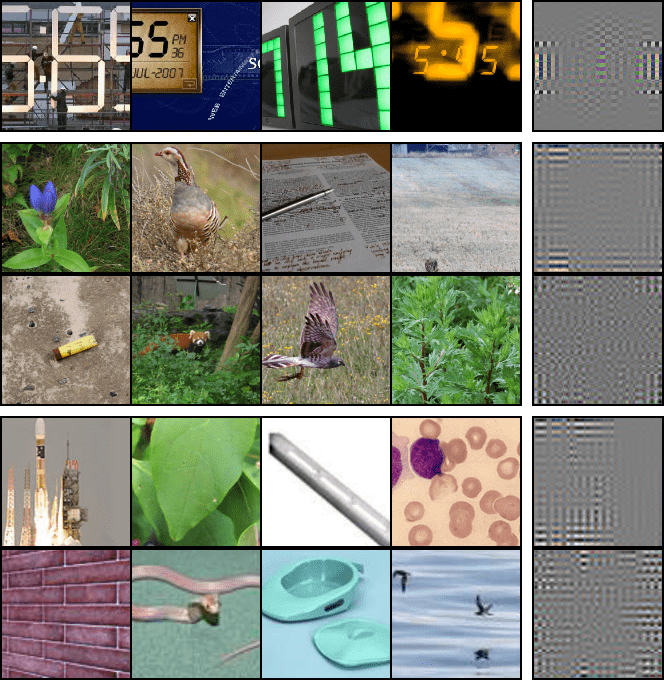} 
} 
\subfigure[ 
``dragonfly'' is normal 
]{ 
\includegraphics[width=0.24\textwidth]{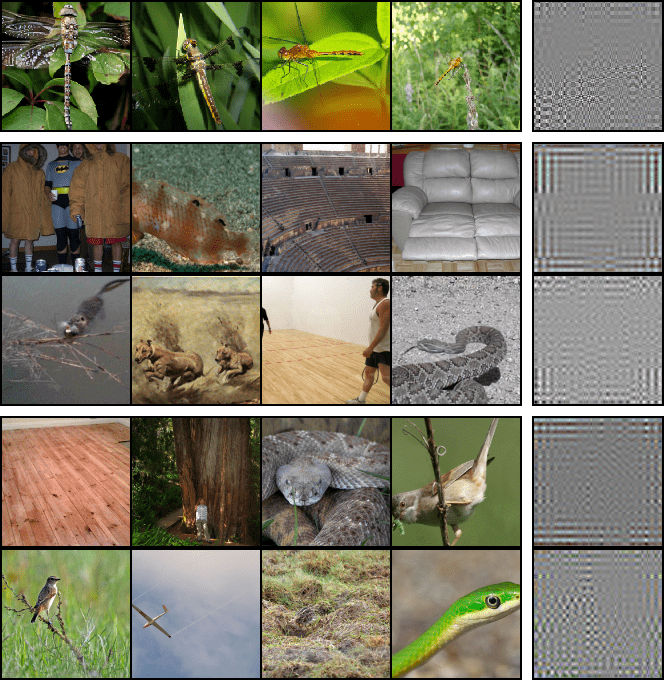} 
} 
\subfigure[ 
``dumbbell'' is normal 
]{ 
\includegraphics[width=0.24\textwidth]{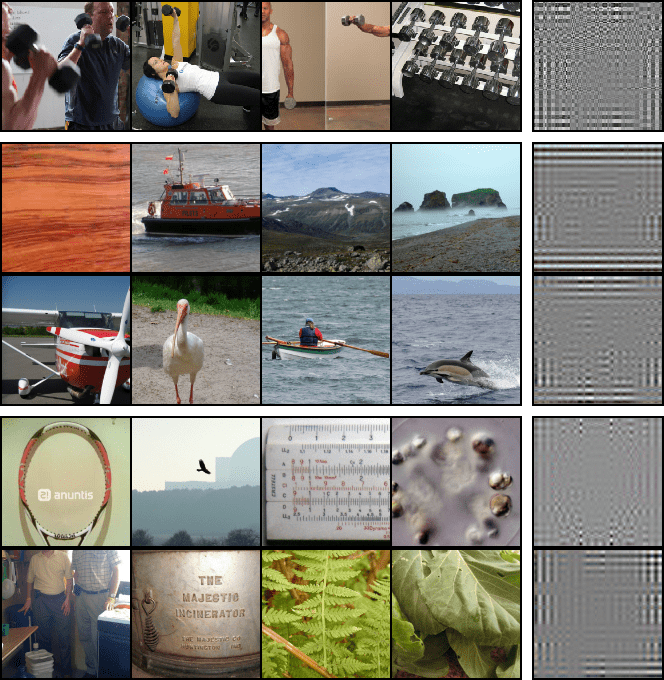} 
} 
\vspace{-1em} 
\caption{ 
Optimal OE samples for high-pass-filtered ImageNet1k with high-pass-filtered ImageNet22k (with the 1K classes removed) as OE. The first row shows normal samples, the next two rows the best samples found via HSC (top) and BCE (bottom), and the last two rows the worst samples found via HSC (top) and BCE (bottom). The last column shows the high-pass-filtered version of the images, which is what the network sees during training and testing.  
} 
\label{fig:evolve_imagenet_hpf} 
\end{figure}

Here we provide class-wise results for the best and worst single OE samples found via an evolutionary algorithm (Section \ref{sec:exp_robustness}).
On ImageNet, we only consider the first 10 classes of the ones used by \citep{hendrycks2019using}: ``Acorn'', ``Airliner'', ``Ambulance'', ``American alligator'', ``Banjo'', ``Barn'', ``Bikini'', ``Digital clock'', ``Dragonfly'', ``Dumbbell''. 
However, the full 30-class one vs.~rest benchmark was used to evaluate for each of these ten classes.
Tables \ref{tab:evolve_classwise_cifar_plain} and \ref{tab:evolve_classwise_cifar_freq} show the performance of the best or worst single OE sample (from 80MTI) for each class in CIFAR-10, while the former shows results for unfiltered images and the latter for low-pass-filtered or high-pass-filtered images.
Tables \ref{tab:evolve_classwise_imagenet_plain} and \ref{tab:evolve_classwise_imagenet_freq} show the same for ImageNet-10 with ImageNet-22K as OE.
Figures \ref{fig:evolve_cifar_plain}, \ref{fig:evolve_cifar_lpf}, \ref{fig:evolve_cifar_hpf}, \ref{fig:evolve_imagenet_plain}, \ref{fig:evolve_imagenet_lpf}, and \ref{fig:evolve_imagenet_hpf} show the 4-5 best and worst samples for HSC and BCE, on CIFAR-10 or ImageNet-10, for unfiltered or filtered images, respectively. 

%%%%%%%%%%%%%%%%%%%%%%%%%%%%%%%%%%%%%%%%%%%%%%%%%%%%%%%%%%%%
\begin{table*}[th]
    \caption{Class-wise AUC detection performance in \% for the best and worst single OE samples found via an evolutionary algorithm (Section \ref{sec:exp_robustness}) on the CIFAR-10 one vs.~rest AD benchmark using 80MTI as OE.}
    \label{tab:evolve_classwise_cifar_plain}
    \vspace{0.5em}
    \centering\small
    \begin{tabular}{lcccc} 
\toprule 
 & \multicolumn{2}{c|}{Best OE} & \multicolumn{2}{c}{Worst OE} \\ 
Class & HSC & \multicolumn{1}{c|}{BCE} & HSC & \multicolumn{1}{c}{BCE} \\ 
\midrule 
Airplane & 85.2 & \multicolumn{1}{c|}{76.1} & 37.4 & \multicolumn{1}{c}{29.9} \\ 
Automobile & 78.1 & \multicolumn{1}{c|}{71.5} & 36.4 & \multicolumn{1}{c}{32.9} \\ 
Bird & 74.0 & \multicolumn{1}{c|}{67.9} & 47.2 & \multicolumn{1}{c}{35.8} \\ 
Cat & 72.5 & \multicolumn{1}{c|}{64.9} & 38.7 & \multicolumn{1}{c}{35.1} \\ 
Deer & 79.9 & \multicolumn{1}{c|}{68.2} & 55.1 & \multicolumn{1}{c}{29.8} \\ 
Dog & 71.6 & \multicolumn{1}{c|}{70.6} & 41.0 & \multicolumn{1}{c}{35.0} \\ 
Frog & 84.2 & \multicolumn{1}{c|}{71.7} & 47.7 & \multicolumn{1}{c}{24.8} \\ 
Horse & 66.2 & \multicolumn{1}{c|}{66.2} & 41.9 & \multicolumn{1}{c}{38.7} \\ 
Ship & 86.5 & \multicolumn{1}{c|}{72.8} & 46.2 & \multicolumn{1}{c}{27.1} \\ 
Truck & 78.8 & \multicolumn{1}{c|}{69.1} & 41.2 & \multicolumn{1}{c}{26.9} \\ 
\midrule 
Mean AUC & 77.7 & \multicolumn{1}{c|}{69.9} & 43.3 & \multicolumn{1}{c}{31.6} \\ 
\bottomrule 
\end{tabular}
\end{table*}
%%%%%%%%%%%%%%%%%%%%%%%%%%%%%%%%%%%%%%%%%%%%%%%%%%%%%%%%%%%%
%%%%%%%%%%%%%%%%%%%%%%%%%%%%%%%%%%%%%%%%%%%%%%%%%%%%%%%%%%%%
\begin{table*}[th]
    \caption{Class-wise AUC detection performance in \% for the best and worst single OE samples found via an evolutionary algorithm (Section \ref{sec:exp_robustness}) on the CIFAR-10 one vs.~rest AD benchmark using 80MTI as OE. All images are either low-pass-filtered (LPF) or high-pass-filtered (HPF), both during training and testing. }
    \label{tab:evolve_classwise_cifar_freq}
    \vspace{0.5em}
    \centering\small
    \resizebox{0.99\textwidth}{!}{\begin{tabular}{lcccccccc} 
\toprule 
 & \multicolumn{4}{c|}{Best OE} & \multicolumn{4}{c}{Worst OE} \\ 
Class & HSC LPF & BCE LPF & HSC HPF & \multicolumn{1}{c|}{BCE HPF} & HSC LPF & BCE LPF & HSC HPF & \multicolumn{1}{c}{BCE HPF} \\ 
\midrule 
Airplane & 83.0 & 68.5 & 68.5 & \multicolumn{1}{c|}{64.6} & 36.0 & 22.1 & 40.7 & \multicolumn{1}{c}{32.5} \\ 
Automobile & 75.0 & 69.5 & 60.1 & \multicolumn{1}{c|}{69.0} & 47.5 & 32.6 & 40.2 & \multicolumn{1}{c}{40.3} \\ 
Bird & 71.1 & 63.5 & 67.9 & \multicolumn{1}{c|}{61.6} & 40.3 & 39.1 & 46.4 & \multicolumn{1}{c}{39.2} \\ 
Cat & 73.6 & 63.2 & 65.5 & \multicolumn{1}{c|}{63.3} & 44.0 & 34.2 & 48.6 & \multicolumn{1}{c}{39.4} \\ 
Deer & 76.9 & 69.2 & 68.5 & \multicolumn{1}{c|}{63.8} & 38.0 & 31.7 & 47.2 & \multicolumn{1}{c}{38.9} \\ 
Dog & 74.2 & 64.3 & 69.4 & \multicolumn{1}{c|}{67.1} & 43.0 & 34.1 & 53.3 & \multicolumn{1}{c}{37.3} \\ 
Frog & 83.5 & 73.1 & 78.2 & \multicolumn{1}{c|}{70.3} & 40.1 & 28.2 & 38.0 & \multicolumn{1}{c}{37.0} \\ 
Horse & 72.5 & 66.5 & 66.7 & \multicolumn{1}{c|}{63.0} & 50.9 & 35.5 & 49.0 & \multicolumn{1}{c}{40.3} \\ 
Ship & 85.7 & 75.8 & 74.8 & \multicolumn{1}{c|}{72.2} & 47.8 & 23.0 & 35.5 & \multicolumn{1}{c}{35.8} \\ 
Truck & 79.9 & 71.0 & 67.9 & \multicolumn{1}{c|}{69.2} & 53.0 & 30.1 & 37.3 & \multicolumn{1}{c}{39.7} \\ 
\midrule 
Mean AUC & 77.5 & 68.5 & 68.8 & \multicolumn{1}{c|}{66.4} & 44.1 & 31.1 & 43.6 & \multicolumn{1}{c}{38.0} \\ 
\bottomrule 
\end{tabular} }
\end{table*}
%%%%%%%%%%%%%%%%%%%%%%%%%%%%%%%%%%%%%%%%%%%%%%%%%%%%%%%%%%%%
%%%%%%%%%%%%%%%%%%%%%%%%%%%%%%%%%%%%%%%%%%%%%%%%%%%%%%%%%%%%
\begin{table*}[th]
    \caption{Class-wise AUC detection performance in \% for the best and worst single OE samples found via an evolutionary algorithm (Section \ref{sec:exp_robustness}) on the ImageNet-30 one vs.~rest AD benchmark using ImageNet-22k (with the 1K classes removed) as OE.  } 
    \label{tab:evolve_classwise_imagenet_plain}
    \vspace{0.5em}
    \centering\small
    \begin{tabular}{lcccc} 
\toprule 
 & \multicolumn{2}{c|}{Best OE} & \multicolumn{2}{c}{Worst OE} \\ 
Class & HSC & \multicolumn{1}{c|}{BCE} & HSC & \multicolumn{1}{c}{BCE} \\ 
\midrule 
Acorn & 84.0 & \multicolumn{1}{c|}{76.6} & 43.6 & \multicolumn{1}{c}{22.3} \\ 
Airliner & 91.1 & \multicolumn{1}{c|}{80.8} & 41.3 & \multicolumn{1}{c}{15.7} \\ 
Ambulance & 84.0 & \multicolumn{1}{c|}{86.4} & 24.0 & \multicolumn{1}{c}{22.9} \\ 
American alligator & 81.2 & \multicolumn{1}{c|}{75.7} & 54.6 & \multicolumn{1}{c}{30.1} \\ 
Banjo & 81.3 & \multicolumn{1}{c|}{75.8} & 37.2 & \multicolumn{1}{c}{26.8} \\ 
Barn & 78.4 & \multicolumn{1}{c|}{68.0} & 40.0 & \multicolumn{1}{c}{29.0} \\ 
Bikini & 66.7 & \multicolumn{1}{c|}{67.0} & 41.5 & \multicolumn{1}{c}{38.9} \\ 
Digital clock & 67.7 & \multicolumn{1}{c|}{73.2} & 32.5 & \multicolumn{1}{c}{28.1} \\ 
Dragonfly & 86.6 & \multicolumn{1}{c|}{84.8} & 36.1 & \multicolumn{1}{c}{13.6} \\ 
Dumbbell & 71.8 & \multicolumn{1}{c|}{66.5} & 40.8 & \multicolumn{1}{c}{35.9} \\ 
\midrule 
Mean AUC & 79.3 & \multicolumn{1}{c|}{75.5} & 39.2 & \multicolumn{1}{c}{26.3} \\ 
\bottomrule 
\end{tabular} 
\end{table*}
%%%%%%%%%%%%%%%%%%%%%%%%%%%%%%%%%%%%%%%%%%%%%%%%%%%%%%%%%%%%
%%%%%%%%%%%%%%%%%%%%%%%%%%%%%%%%%%%%%%%%%%%%%%%%%%%%%%%%%%%%
\begin{table*}[th]
    \caption{Class-wise AUC detection performance in \% for the best and worst single OE samples found via an evolutionary algorithm (Section \ref{sec:exp_robustness}) on the ImageNet-30 one vs.~rest AD benchmark using ImageNet-22k (with the 1K classes removed) as OE. All images are either low-pass-filtered (LPF) or high-pass-filtered (HPF), both during training and testing. } 
    \label{tab:evolve_classwise_imagenet_freq}
    \vspace{0.5em}
    \centering\small
    \resizebox{0.99\textwidth}{!}{\begin{tabular}{lcccccccc} 
\toprule 
 & \multicolumn{4}{c|}{Best OE} & \multicolumn{4}{c}{Worst OE} \\ 
Class & HSC LPF & BCE LPF & HSC HPF & \multicolumn{1}{c|}{BCE HPF} & HSC LPF & BCE LPF & HSC HPF & \multicolumn{1}{c}{BCE HPF} \\ 
\midrule 
Acorn & 82.8 & 83.8 & 80.4 & \multicolumn{1}{c|}{75.9} & 41.2 & 20.0 & 45.8 & \multicolumn{1}{c}{31.0} \\ 
Airliner & 78.2 & 77.3 & 83.6 & \multicolumn{1}{c|}{86.3} & 44.6 & 21.7 & 52.5 & \multicolumn{1}{c}{18.0} \\ 
Ambulance & 86.7 & 76.3 & 79.1 & \multicolumn{1}{c|}{83.3} & 53.6 & 26.1 & 44.4 & \multicolumn{1}{c}{29.2} \\ 
American alligator & 69.3 & 67.4 & 78.3 & \multicolumn{1}{c|}{80.1} & 43.4 & 32.3 & 42.2 & \multicolumn{1}{c}{29.0} \\ 
Banjo & 83.0 & 74.8 & 71.3 & \multicolumn{1}{c|}{75.9} & 48.0 & 21.5 & 39.1 & \multicolumn{1}{c}{31.9} \\ 
Barn & 79.6 & 71.5 & 73.0 & \multicolumn{1}{c|}{77.6} & 45.2 & 25.6 & 44.4 & \multicolumn{1}{c}{26.2} \\ 
Bikini & 70.4 & 68.0 & 68.6 & \multicolumn{1}{c|}{65.9} & 47.4 & 35.6 & 45.9 & \multicolumn{1}{c}{34.9} \\ 
Digital clock & 63.3 & 65.7 & 68.4 & \multicolumn{1}{c|}{76.2} & 41.2 & 35.9 & 34.9 & \multicolumn{1}{c}{26.0} \\ 
Dragonfly & 85.0 & 85.6 & 83.7 & \multicolumn{1}{c|}{84.6} & 33.0 & 12.9 & 51.4 & \multicolumn{1}{c}{20.2} \\ 
Dumbbell & 73.2 & 68.1 & 63.8 & \multicolumn{1}{c|}{67.2} & 48.1 & 29.1 & 40.0 & \multicolumn{1}{c}{32.8} \\ 
\midrule 
Mean AUC & 77.2 & 73.8 & 75.0 & \multicolumn{1}{c|}{77.3} & 44.6 & 26.1 & 44.1 & \multicolumn{1}{c}{27.9} \\ 
\bottomrule 
\end{tabular} }
\end{table*}
%%%%%%%%%%%%%%%%%%%%%%%%%%%%%%%%%%%%%%%%%%%%%%%%%%%%%%%%%%%%

%%%%%%%%%%%%%%%%%%%%%%%%%%%%%%%%%%%%%%%%%%%%%%%%%%%%%%%%%%%%

%%%%%%%%%%%%%%%%%%%%%%%%%%%%%%%%%%%%%%%%%%%%%%%%%%%%%%%%%%%%

\end{document}